\documentclass[3p]{elsarticle} % DO NOT CHANGE THIS
\usepackage{longtable} 

\newcounter{mylabelcounter}

\makeatletter
\newcommand{\labelText}[2]{%
#1\refstepcounter{mylabelcounter}%
\immediate\write\@auxout{%
  \string\newlabel{#2}{{1}{\thepage}{{\unexpanded{#1}}}{mylabelcounter.\number\value{mylabelcounter}}{}}%
}%
}
\makeatother
\usepackage{tabu} 
\usepackage{footmisc} 
\usepackage{tabulary}
\usepackage{booktabs}
\usepackage{prftree}    
\usepackage{ntheorem}  
\usepackage[linesnumbered,ruled]{algorithm2e} 

\SetCommentSty{mycommfont}

\makeatletter
\renewtheoremstyle{plain}% Adds automatic line break, if heading is too long
  {\item{\theorem@headerfont ##1\ ##2\theorem@separator}~}
  {\item{\theorem@headerfont ##1\ ##2\ (##3)\theorem@separator}~}
\makeatother

\usepackage{wrapfig} 
\usepackage{booktabs} \newcommand{\ra}[1]{\renewcommand{\arraystretch}{#1}}
\usepackage{mathtools}

\usepackage{amsmath}    
\usepackage{multicol} 
\usepackage{amssymb} 
\usepackage{extarrows}    
\usepackage[normalem]{ulem}
\usepackage{bcprules}
\usepackage{enumitem}  
\usepackage{amsfonts} 
\usepackage{stmaryrd} 
\usepackage{tikz}  
\usepackage{tikz-cd}  
\usetikzlibrary{matrix,fit,decorations.pathmorphing,intersections,arrows,positioning,shapes.misc}  
\usepackage{ifthen}

\usepackage[hyphens]{url}  % DO NOT CHANGE THIS
\usepackage{graphicx} % DO NOT CHANGE THIS 
\usepackage{xcolor} 
\usepackage{natbib}  % DO NOT CHANGE THIS AND DO NOT ADD ANY OPTIONS TO IT 
\setcitestyle{authoryear,open={(},close={)}}
\usepackage{caption} % DO NOT CHANGE THIS AND DO NOT ADD ANY OPTIONS TO IT
\newcommand{\hide}[1]{}

\newcommand{\wrtM}{with respect to}

\newcommand\utimes{\mathbin{\ooalign{$\cup$\cr%
   \hfil\raise0.42ex\hbox{$\scriptscriptstyle\times$}\hfil\cr}}}

\newcommand{\Sat}[1]{\mbox{\text{\mbox{Sat:}}{\text{\pmb{#1}}}}}
\newcommand{\SatOne}[1]{\mbox{\text{p-\mbox{Sat:}}{\text{\pmb{#1}}}}}

\newcommand{\modell}{\ensuremath{\llparenthesis G, \Pi\rrparenthesis}}  
\newcommand{\taam}{\ensuremath{\langle \modell, \mathcal{D}, \mathcal{I} \rangle}}

\usepackage{bm}
\DeclareMathAlphabet{\mathsfit}{T1}{\sfdefault}{\mddefault}{\sldefault}
\SetMathAlphabet{\mathsfit}{bold}{T1}{\sfdefault}{\bfdefault}{\sldefault}

\newcommand{\range}{\mathsf{range}}
 
\newcommand{\finsubseteq}{\mathbin{\small \overset{\makebox[0pt]{\tiny{fin}}}{\subseteq}}}

\newtheorem{definition}{Definition} 
\newtheorem{innerdefinition}{Inner-Definition} 

\newtheorem{example}{Example}  
\newtheorem{lemma}{Lemma}  
\newtheorem{innerlemma}{Inner-Lemma}
\newtheorem{observation}{Observation} 
\newtheorem{proposition}{Proposition}  
 
\newtheorem{note}{Remark} 
\newtheorem{theorem}{Theorem}   
\newtheorem{innertheorem}{Inner-Theorem} 
\newtheorem{corollary}{Corollary}   
 
\newproof{proof}{Proof} 

\usepackage[backref=page]{hyperref}
\usepackage{blindtext}
\usepackage{nameref}

\title{Theme Aspect Argumentation Model for Handling Fallacies
}
\author[1]{Ryuta Arisaka\corref{cor1}%
\fnref{fn1}}
\ead{ryutaarisaka@gmail.com}

\author[1]{Ryoma Nakai}
\ead{nakai.ryoma.34a@st.kyoto-u.ac.jp}

\author[2]{Yusuke Kawamoto}
\ead{yusuke.kawamoto@aist.go.jp}

\author[1]{Takayuki Ito}
\ead{ito@i.kyoto-u.ac.jp} 

\cortext[cor1]{Corresponding author. %The authors are ordered by contributions. 
The research idea was conceived and formulated by the 1st author alone. The 
literature survey was done 
by the 1st author alone, with 2 references provided by the 2nd author
and 1 reference by the 4th author. 
Formal proofs 
were produced and checked by the 1st (90\%) and the 2nd authors alone. 
Implementation was done by the 2nd author alone. 
The manuscript was produced by the 1st (98\%) and the 2nd authors alone.  
It was edited by the 1st ($\approx$ 100\%) 
and the 2nd authors alone, while the 3rd author 
suggested to bring formal proofs to the appendix. 
All the authors reflected/commented 
on the presentation. % ($\sim$ 100\% by the 1st author; 
}

\affiliation[1]{organization={Kyoto University}, addressline={Yoshida Honmachi},
postcode={606-8501}, city={Kyoto}, country={Japan}}

\affiliation[2]{organization={AIST}, addressline={Koto City 2-3-26}, postcode={135-0064}, city={Tokyo}, country={Japan}}

\begin{document}

\begin{abstract}         
	From daily discussions to marketing ads to political 
	statements, information manipulation is rife. 
	It is increasingly more important that we have the right 
	set of tools to defend ourselves from manipulative rhetoric, 
	or fallacies. Suitable techniques to automatically 
	identify fallacies are being investigated in natural language 
	processing research in this regard. However, a fallacy in one 
	context may not be a fallacy in another context, so there is 
	also a need to explain how and why it has come to be judged 
	a fallacy. For the explainable fallacy identification, 
	we present a novel approach to characterising fallacies 
	through formal constraints, as a viable alternative to more 
	traditional fallacy classifications by informal criteria. 
	To achieve this objective, we introduce a novel 
	context-aware argumentation model, the theme aspect 
	argumentation model, which can do both: the modelling of a given 
	argumentation as it is expressed (rhetorical modelling); and 
	a deeper semantic analysis of the rhetorical argumentation model.
	By identifying fallacies with formal constraints, 
	it becomes possible to 
	tell whether a fallacy lurks in the modelled rhetoric with a 
	formal rigour. We present core formal constraints for 
	the theme aspect argumentation model and then more formal 
	constraints that improve its fallacy identification capability.
	We show and prove the consequences of these formal 
	constraints. We then analyse the computational 
	complexities of 
	deciding the satisfiability of the constraints.

\end{abstract}
\maketitle
\section{Introduction} \label{sec_introduction}   
%\hypersetup{
%  colorlinks   = true, %Colours links instead of ugly boxes
%  urlcolor     = blue, %Colour for external hyperlinks
%  linkcolor    = blue, %Colour of internal links
%  citecolor   = red, %Colour of citations 
%  linkbordercolor = {white}
%%  hidelinks = true
%}
%\makeatletter
%\DeclareRobustCommand*{\nameref}{%
%\color{blue}%
%        \@ifstar\T@nameref\T@nameref
%        }%
%\makeatother
%\makeatletter
%\DeclareRobustCommand*{\ref}{%
%\color{blue}%
%        \@ifstar\T@ref\T@ref
%        }%
%\makeatother
From daily discussions to marketing ads to political statements, information manipulation is rife. 
It is increasingly more important that 
we have the right set of tools to defend ourselves from 
manipulative rhetoric, or fallacies. 
Recently within natural language processing research, propagandist message classifications were reported for news articles and news 
outlets \citep{Barron19,Martino19}. It was then reported that argumentative features 
added as extra cues 
further improved the quality of automatic propaganda and fallacy detection \citep{Vorakitphan21,Goffredo22}.  
Given many of the text-based propagandist messages covered in \citep{Barron19,Martino19} are fallacies, and fallacies are 
argumentations, this is an assuring empirical result confirming the actual relevance of argumentation to fallacy detection. 
\subsection{The big picture} \label{subsec_big_picture} 
%	The high-level research problem: fallacy identification} \label{subsec_high_level_research_problem} 
However, fallacy classification typically suffers from significant context-dependability of a fallacy in addition to  
classifiers' presuppositions. Even for the long-known {\it ad hominem} fallacy, 
telling exactly what properties count as its constituents is an ordeal \citep{Habernal18} (and given the 
innate ambiguity in natural language expressions, the exact properties are likely unidentifiable). 
Indeed, many fallacies ae fallacies only circumstantially \citep{Walton08}. 
Entailed is the following nettlesome problem: one's argumentation can be classified fallacious 
when it is actually not, resulting in a dilemma, that is to say, the fallacy classification may itself be fallacious. 
This can pose a setback to fallacy classification efforts in the long run, and may even 
trigger legal consequences in the case of false accusations. Thus, ultimately, we need to be able 
to explain very precisely why and how one's argumentation has come to be judged a fallacy, should 
the explanation be demanded. This is not an easy task at present, however. 
%The number of informally classified fallacies 
%surpassing 50 \citep{Walton08}  
%also makes it difficult to see which ones are potentially 
%overlapping, where the overlaps (if any) may be 
%and whether they actually cover all that can be fallacies.  
%It is also 
% unclear whether they cover 
%the literature has covered all the fallacies. 

%The sheer number of fallacies surpassing 50 \citep{Walton08}  
%makes it hard to see 
%how some of them overlap  
%further complicates the analysis on understanding 
% 
%whether they may overlap 
%and whether they cover  
%Added to this is the fact that there are many a number of 
%fallacies with likely overlaps. 

%and despite the number, it is not 
%clear whether 
%they cover  

%\subsection{A solution to the high-level research problem: formal fallacy identification} \label{subsec_solution_high_level_research_problem}
In light of the setback, we initiate formal fallacy identification (to be 
read as ``formal `fallacy identification'''; not as 
```formal fallacy' identification''), 
viewing a fallacy as a failure to satisfy formal constraints, as a viable alternative 
to the traditional fallacy classification through informal criteria.   
%In a little more detail, in the traditional fallacy classification, 
%we would be identifying {\it ad hominem}, {\it fear appeals}, 
%{\it slippery slope}, and so on, based on the informal criteria. 
This shift transforms the question  
from whether {\it ad hominem} and so on 
are present - the traditional way - to whether formal constraints 
have been met. %are satisfied or not satisfied. 
It seems to us this is a reasonable way out of  
the fallacious fallacy judgement dilemma, since, as stated above, informal properties are bound to be ambiguous and can fluctuate 
%subject to
%interpretations, and 
without widespread agreement. Through the proposed formal fallacy identification,  
the process of fallacy judgement becomes explainable in detail: 
from how an argumentation subjected to fallacy judgement is being modelled; 
to under which context(s) the judgement is being made; and, if deemed 
fallacious, which formal constraints 
have not been met. Furthermore, this method allows for 
a formal analysis of the interactions 
among these constraints. %can be formally analysed.   

%All these explanations are formally rigorous. 

\subsection{The high-level research problem: 
the argumentation modelling concept for fallacy identification} \label{subsec_mid_level_research_problem} 
\begin{figure*}[!t]   
%	\hspace{-2.3cm}  
	\vspace*{-3cm} 
	\includegraphics[width=\columnwidth]{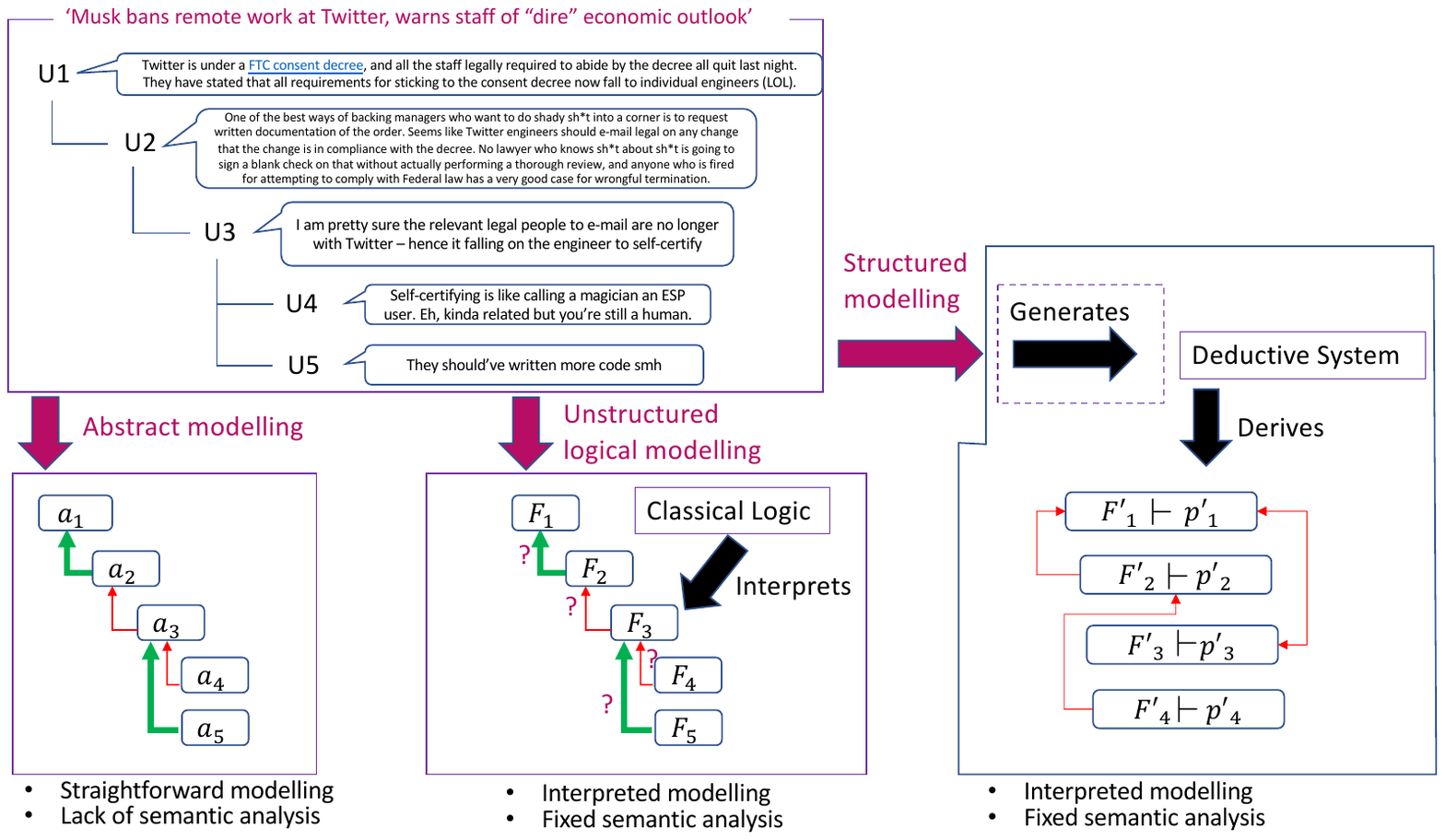} 
	 \vspace*{-15mm}
	\caption{%(\textbf{Top half}) 
	Different types of modelling of argumentation. The thick (green) arrows indicate supports and the thin (red) arrows 
	indicate attacks. In  abstract modelling, 
	the given argumentation may be straightforwardly modelled 
	by allocating a node to each comment. In unstructured logical modelling, 
	each comment is logically interpreted into a logical formula. 
	Conflicts 
	between two nodes are often judged by logical contradiction in the two formulas 
	put together. As such, the indicated arrows may or may not actually exist in 
	the model. 
	In the like manner, there may be other arrows that are not indicated here.
	The structured 
	modelling derives an argumentation graph from some deductive system as a knowledge base. 
%	which may be an existing one or a newly generated one from the given argumentation. 
	In the derived argumentation graph, every node 
	represents an argument in a specific sense comprising  
	premises ($F'_x$ in the figure) and the conclusion ($p'_x$ in the figure).   
	In general, the degree of applied interpretation increases from the left to the right.  
	%Depending on how the knowledge base is constructed, it could be the most heavily interpreted 
	%modelling of the three modellings.  \\ 
%	(\textbf{Bottom half}) An example of hybrid modelling comprising the abstract modelling part and 
%	the semantics as two separate components. The abstract modelling 
%	is divided further into uninterpreted graphical modelling (specifically no arrows 
%	presupposing semantic attacks, supports and possibly others) 
%	and a type assignment which gives syntactic types to the graph nodes and edges. 
%	Contrary to the abstract modelling in the top half where 
%	the syntax and the semantics of attacks and supports are not well-insulated, `$attack$', `$support$' and 
%	any other types here are only syntactic. Meanwhile, the semantics consists of 
%	an algebraic structure and an interpretation which takes  
%	node types and a node and which  outputs elements of the underlying set of 
%	the algebraic structure as the semantics of the nodes. The semantics of `$attack$' and `$support$' 
%	is not defined in the model: they are meant to be defined by formal constraints 
%	on this model.  
	}
	\label{fig_big_picture_smaller}  
\end{figure*}

 But this solution is much easier described than materialised. While formal 
studies on {\it abstract argumentation} \citep{Dung95} have been conducted for a long time, 
it is a formal framework 
primarily for modelling rhetorical argumentations where any argument can attack any argument. %In this paper, we 
%call them rhetoric argumentations. 
There, so long as a speaker intends an attack in his/her argument against another argument, 
there can be 
an attack, for ``{\it one who speaks last speaks loudest (one who laughs last laughs best)}" 
is a guiding principle \citep{Dung95} for winning a rhetorical 
argumentation. The abstract argumentation model  
excels at capturing this `{\it attack as an intention to attack}'
through straightforward modelling of a given argumentation 
as it is expressed, which is particularly convenient 
for modelling argumentations on SNS (such as YouTube, 
Twitter and Reddit) and dialogues in general. 
We may simply assign a node  
to each post or utterance and an edge to 
a post-to-post or utterance-to-utterance 
relation, as depicted on the bottom left-hand side of Figure 
\ref{fig_big_picture_smaller}. %giving us the subject  
%against which to conduct fallacy identification. 
%needs to be conducted and the derived rhetoric model needs to be kept. 
%The abstract modelling takes care of this part. 
However, the abstract argumentation model does not 
provide any semantics of arguments. 
There is no semantics of attacks or any other relations as could be inferred from 
that of arguments, either, trivially because the arguments do not have any semantics. 
%Hence, even though it gives us the subject, we do not have the component 
%to conduct semantic analysis on it. 
The {\it acceptability semantics} of the abstract argumentation model 
are high-level criteria to determine   
the conclusion of rhetorical argumentation, {\it i.e.} 
which set(s) of arguments can be accepted with respect to the criteria, which is 
too high-level for fallacy detection in general. 
\begin{note}\label{note_1} \rm 
	Indeed, suppose an  argumentation where Alice states: ``{\it We should 
buy fire insurance.}'' and Bob states: ``{\it I don't think so. 
	You are overcautious and see danger in everything.}'', 
Bob certainly intends an attack on Alice's statement, and, 
	with the abstract modelling, it is fair
to model this argumentation by modelling their statements as arguments and to put 
an attack relation from Bob's argument to Alice's argument. A majority of the acceptability semantics  
in the abstract argumentation model judge Bob's argument as acceptable and 
	Alice's argument as not acceptable. While 
	this judgement may be reasonable in light of the above-described  
	guiding principle, Bob's statement here can be {\it ad hominem}, 
	as he criticises Alice's personal trait rather than pointing 
	out a direct issue of buying fire insurance.  
	%Unfortunately, after deriving the rhetoric model, 
	%we are no longer able to check the appropriateness of the claimed attack 
	%relation. 
	\hfill$\clubsuit$
\end{note} 

%\begin{figure*}[!t]   
%	\includegraphics[scale=0.55]{Overview2.pdf} 
%	\vspace*{-10mm} 
%	\caption{(\textbf{Top left}) Argumentation on Reddit (expletives and inappropriate words have 
%	been edited). 
%	(\textbf{Bottom left}) An abstract modelling.  
%	(\textbf{Bottom centre}) Logical modelling with no structuralisation. Each node 
%	is 
%	The relations
%	among the nodes are determined by the semantics 
%	(logically 
%	(bottom centre), and logically and structurally 
%	(bottom right).}
%	\label{fig_big_picture} 
%\end{figure*} 

Structured argumentation formalisms \citep{Prakken01,Besnard01,Garcia04,Dung09,Modgil13} construct an argumentation graph 
from a given knowledge base, 
based on premise-conclusion-oriented argument construction rules as well as on the notion of 
conflict that derives from logical contradiction and its proxies.  
As shown on the bottom right hand side of Figure \ref{fig_big_picture_smaller},  
some deductive system (for example, 
a proof theory of classical logic such as \textsf{G3c} \cite{351148,Kleene52}) 
is posited. The knowledge base 
is then a set of members (often propositions) 
in the deductive system and each node of the obtained argumentation graph 
represents an argument in a specific sense comprising  
a subset of the knowledge base as the premises 
($F'_1, \ldots, F'_4$ in the figure) and a single member 
of the knowledge base 
derivable from the premises as its conclusion ($p'_1, \ldots, p'_4$ in the figure). 
Conflicts between two nodes are judged by logical contradiction and 
its proxies typically between
the conclusion of one node and the premises of another node. 
Hence, within the modelled argumentation 
it is possible to determine the semantics of an attack.  

However, structured argumentation formalisms do not always retain 
the rhetorical intention of a given argumentation, for they fix a 
specific form 
that an argument must take on and the specific condition 
of a conflict, too, as to when an argument 
can be considered attacking an argument at all.  
%As exemplified on the bottom right-hand side of Figure \ref{fig_big_picture_smaller}, 
%for a given argumentation,  
%structured argumentation derives its own account of argumentation graph 
%for it from a deductive system (as a knowledge base) which may be an existing one 
%or a newly generated one from the given argumentation. Every 
%node represents an argument in a specific sense 
%comprising premises ($F'_x$ in Figure \ref{fig_big_picture_smaller}) 
%and the conclusion ($p'_x$ in Figure \ref{fig_big_picture_smaller}). % \vdash p'_j$ 
%represents an argument in the premise ($F'_i$)-conclusion ($p'_j$) form. 
More restrictions on the form of an argument and of a conflict 
mean that an argumentation to be modelled in structured argumentation 
formalisms 
can go through a heavy pre-screening leading to any of the following. 
\begin{itemize} 
	\item elimination of expressed attacks if they do not conform to the logical contradiction and 
its  proxies recognised as a conflict in structured argumentation 
		formalisms. 
	\item addition of unexpressed attacks if they are inferred from the knowledge base. 
	\item addition of premises \citep{Hunter22} and/or conclusion if a statement in the argumentation to be modelled is not already 
in the specified premise-conclusion form.
\end{itemize} 
This is rather problematic:  
first for agile modelling 
of rhetorical argumentations, say those on SNS (such as \mbox{YouTube}, \mbox{Twitter} and \mbox{Reddit}), 
since it is time-consuming to apply the structured argumentation 
convention to mould the shape 
of raw argumentations on SNS; second for the difficulty of reacting to changes to the knowledge base, since the formal construction of 
argumentation is coupled with it and modelling of arguments 
must naturally be done anew at each such update; 
and then for ensuring the impartiality of pre-screening \citep{Hunter07}: once again the entire argumentation modelling must be done 
anew in the case it is found to be biased. 
%needs to be carried out for the given , 
%the pre-screening can lead to the loss of the very thing 
%({\it i.e.} the rhetoric 
%of the given argumentation) to subject to fallacy 
%identification. It also makes it harder to react to changes 
%to 
%itself is rather problematic: 
%first for agile modelling 
%of rhetorical argumentations, say those on SNS (such as \mbox{YouTube}, \mbox{Twitter} and \mbox{Reddit}), 
%since it is time-consuming to apply the structured argumentation 
%convention to mould the shape 
%of raw argumentations on SNS; second for the difficulty of reacting to changes to 
%the knowledge base, since the formal construction of 
%an argumentation graph is coupled with it and 
%modelling of arguments must naturally be done anew at each such update. 
%Ensuring the impartiality of pre-screening \citep{Hunter07} 
%is another difficulty: once again the entire argumentation modelling 
%must be done anew in the case it is found to be biased.     

%Fragile formal grounds were also noted \citep{Amgoud12}. %Apart from structured argumentation, 
%other logic-based argumentations model an argument as a logical formula, but identify a conflict 
%again in logical contradiction and proxies.  

\begin{note}\label{note_2} \rm   
	Observe the argumentation 
	in Remark \ref{note_1} where Alice states: 
	``{\it We should buy fire insurance.}'' 
	and Bob states: ``{\it I don't think so. You 
	are overcautious and see danger in everything.}'' 
	Since Alice's argument is not in the specific premise-conclusion 
	form, we pre-screen it into: ``{\it We should buy fire insurance 
	because X.}'' where 
	 X can be (but has to be) any 
	 that deductively derives the conclusion. 
	 If we presume an enthymeme \citep{Hunter07,Hunter22}, it 
	 can be 
	 ``{\it we use gas often, and 
	 if we use gas often, then we should buy fire insurance.}''  
	Bob's argument, too, may not be in the specific premise-conclusion 
	form, depending on whether Alice's being overcautious and seeing 
	danger in everything deductively derives that they should not buy fire 
	insurance. %Such derivation is well-nigh 
	%impossible if the deductive system is   
	If the deductive system is a proof theory 
	of a formal logic, the said deductive derivability 
	is well-nigh unattainable.\footnote{For the conclusion 
	to be deductively derivable from the premises, 
	if the premises are satisfiable, the conclusion must also be  
	satisfiable not just in some interpretation {\it but in every interpretation} 
	of the sentences occurring in them. 
	Denote ``{\it Alice is overcautious and sees danger in everything.}'' 
	by $p$ and ``{\it Bob and Alice should buy fire insurance.}'' by $q$, 
	then it is possible to find some model of $p \Rightarrow \neg q$ ($\equiv 
	\neg p \vee \neg q$); however, 
	$p \Rightarrow \neg q$ 
	is obviously not a universally valid proposition.} 
	 Hence,  as with 
	Alice's argument, one may pre-screen it into: 
	``{\it I don't think so because you are overcautious and see danger in everything and if you are overcautious and see danger in everything, then I don't think so.}''
	Within this new argumentation, the conclusion 
	of Bob's pre-screened argument (= it is not the case that they should buy fire 
	insurance) and the premises of Alice's pre-screened argument
	(which materially imply 
	that they should buy fire insurance) 
	entail logical contradiction. At the same time, 
		the conclusion of Alice's pre-screened 
			argument (= that they should buy 
			fire insurance) 
			and the premises 
			of Bob's pre-screened argument 
			(which materially imply 
			that they should not buy fire insurance) 
			entail logical contradiction. 
			Hence, Alice's pre-screened 
			argument and Bob's pre-screened argument 
			are regarded as attacking each other.  
			The rhetorical 
			intention of the original argumentation 
			is not preserved here. 
			As for the arguments 
			themselves, while both of the 
			pre-screened 
			arguments are in the premise-conclusion form,
			they are not as they were originally
			expressed.\footnote{Let us take Alice's original 
			statement, if there must be 
			some reason for the statement, 
			the closest nuance 
			would be: ``{\it We should buy fire insurance 
			because of whatever reason I may have that I 
			do not need to disclose to you.}''  
			The `{\it whatever reason}' being a 
			disjunction of 
			many possibilities, this reason part 
			is closer to a tautology than 
			`{\it We should buy fire insurance}' is. 
			The reason part is therefore a logical 
			consequence of ``{\it We should buy fire 
			insurance}''. But this cannot be expressed 
			in deductive argumentation owing 
			to the required relationship between 
			the presmies and the conclusion.\label{fnlabel}}
			The argumentation 
			as it is expressed (the rhetoric
			model of the argumentation), 
			for which we need to reason about  
			fallacies, does not 
			remain here. %In other words, we do not 
			%have the very target 
			%of fallacy identification in 
	\hfill$\clubsuit$ 
\end{note}  
Further, for a given argumentation, while a structured argumentation 
formalism identifies it with one particular interpretation of it, 
normally there is not one correct interpretation. How it should be interpreted 
depends on the context and how much of its detail one needs to 
model. For the explainable fallacy identification therefore, 
(1) the rhetorical intention of a given argumentation (the target 
of fallacy identification) 
needs to be available within the formal argumentation model, 
as stated above, and (2) it needs to be 
checkable 
against a context-dependent interpretation of it, again within 
the formal argumentation model. (3) The relationship 
between the given argumentation and its interpretation 
is 1-to-many. These 3 requirements  
are not embodied in the design of structured 
argumentation formalisms. % by design. 
%do not embed these features within. 

There are logic-based argumentation models that are unstructured, 
{\it e.g.} \citep{Gabbay09,Arisaka16b} and in particular \cite{Arisaka16b}, 
expressing an argument as 
a classical logic formula without much structuralisation.
In the unstructured logic-based modelling 
as is illustrated in \mbox{Figure \ref{fig_big_picture_smaller}} at the bottom 
centre,   
each comment may be assigned a node 
just as in the abstract argumentation 
modelling covered earlier. 
However, the unstructured logic-based argumentation models 
also seek out a conflict in contradiction, entailing similar issues 
experienced in structured argumentation formalisms.\\ %as 
%structured 
%similar issues experienced in structured 
%argumentation formalisms. \\ %are 
%experienced in the unstructured logic-based argumentation modelling, too. 
%elimination of expressed attacks and addition of unexpressed attacks in particular. 

To recapitulate, the current situation is there is either very little semantics of the modelled argumentation itself (in the case of 
abstract argumentation models), 
or too much dependence of argumentation modelling on a knowledge  
base as well as often too narrow a definition of an attack limited to contradiction and its proxies (in the case of 
more logic-oriented argumentation models, structured or unstructured). 
%and 
%often too narrow a definition of an attack limited to logical contradiction and its close proxies (in case of structured argumentation 
%and other logic-based argumentations \citep{Arisaka16b,Gabbay09,Amgoud09}).  
Abstract or structured, both modelling approaches 
have their own advantages and shares,
but, taken individually, neither seems 
fit for the task of formal fallacy identification.  

\subsection{A solution to the high-level research problem: 
a hybrid argumentation model and 
constraint-based fallacy identification} 
\label{subsec_mid_level_research_solution} 
We propose to go the middle way of the two approaches with what we describe as the hybrid argumentation model. It is a general argumentation 
modelling concept,  modelling an argumentation in 3 components.

\begin{itemize} 
	\item 
The first component is some abstract argumentation model (there are plenty of abstract argumentation models proposed to this day 
to choose from) for modelling the rhetorical intention of a given argumentation, 
enabling the agile modelling of an argumentation as it is expressed. 
	Especially for argumentations on SNS, 
when an SNS user posts a statement (that is, an argument, but to avoid any confusion 
with the `argument' in the specific sense in structured argumentation
		formalisms, we 
will use this term in the rest of this paper),  
to which statement the user is responding to is often obvious, too, due to the `reply' 
		functionality, and the stance of the user: for; against; or both, is often also clear (sometimes 
		there is a functionality to select the stance). All this makes 
agile modelling within reach. 
		Moreover, the modelling is automatable in 
		argumentation mining \citep{Habernal17,Lawrence20} 
		given 
the availability of relevant large corpora, {\it e.g.} \citep{Abbott16}. 
		With the advent of ChatGPT,  
		the extraction has become even simpler and generally more accurate 
		for a number of natural language processing tasks. 
%		simply in ChatGPT.  
%for selected discussion topics, given 
%the availability of relevant large corpora, {\it e.g.} \citep{Abbott16}. 
%Moreover, by the advent of ChatGPT, 
%this extraction process has become very accessible 
%to many regardless of expertise in natural language processing. 
%with typically a better result than average humans. 
\item The second component is a semantic structure for the semantics of statements as the basis 
	of the semantics of `$attack$'s and `$support$'s. 

\item The third component is an interpretation function that does the actual mapping, 
mapping statements to elements of the semantic structure. 
\end{itemize} 
The formal constraints described earlier (see section \ref{subsec_big_picture}) then correspond to formal constraints on these 3 components determining how each of the components
should behave. Fallacy identification becomes very intuitive this way. For instance, a suitable set of constraints on  
the 2nd and the 3rd components determine under what conditions a statement can semantically attack or support 
another statement, which verifies whether 
the `$attack$'s and the `$support$'s in the 1st component are really attacks and supports (if not, we judge that there is a fallacy). 
The mismatch of the rhetoric `$attack$' and the semantic 
attack is a key contributor to many fallacies as 
we shall see. 
%in examples. 
%which  
%affirms the need of such formal constraints. 
%therefore needs to be expressible with formal constraints. 
\subsection{The low-level research problem: a specific argumentation model} \label{subsec_low_level_research_problem} 
However, the prospect and the effectiveness of this program depend hugely on 
the choice of a specific hybrid argumentation model.     
\begin{itemize} 
	\item 
For the 1st component, the typical 
abstract argumentation models defining only `$attack$'s and `$support$'s \citep{Cayrol13} 
are just too weak to serve the purpose, for instance. Recall the context-dependability of a fallacy being one major difficulty of 
fallacy identification; it is a must for the 1st component to be able to express  which context(s) (discussion threads,  
discussion users and groups, discussion topics, and so on) each statement belongs to. 
Another desired feature is {\it referencing}, allowing a statement in a context 
		to make a reference to another context or even to some statement in a context. There are indeed numerous referencings 
in argumentations on SNS. There are URL links and hyperlinks within a discussion thread to another discussion thread intending 
		to refer to 
a summary of the other discussion thread. 
		%There are ``Luca 1:30-33 in the Bible", ``Bob's birthday party two years ago", 
%and so on, referring to some specific part, with the Bible as its context, and to Bob's birthday party, with the stated 
		%time in the past as its context \citep{Barringer12}. 
		Some types of fallacies such as %{\it false flags} or 
		{\it equivocation} do manipulate references. 
\item For the 2nd and the 3rd components, it is impractical to try to map every part of each statement in the 1st component 
	to the 2nd component, for computational and linguistic reasons. We must restrict ourselves to 
		mapping representative features of
		a statement only. Moreover, we need at least some ordered semantic structure suitable 
for conducting comparisons between the semantics of statements, so the semantics 
of `$attack$'s and `$support$'s can be always characterised from them. %as is envisioned in logic-based 
%		argumentation models (structured or unstructured). 
				%In general, we need to even 
		%consider the possibility where a statement 
		%is not entirely textual. 
		%What then? If we recognise any message in 
		%a visual cue, we may need to interpret it 
		%textually, but the particular interpretation  
		%cannot be identical with the original statement containing the visual cue. 
\end{itemize}  
\subsection{Our solution to the low-level research problem: the theme aspect argumentation model} \label{subsec_low_level_research_solution} 
As the solution to this low-level research problem, we deliver the {\it theme aspect argumentation model}. 
\begin{figure*}[!h]   
%	\hspace{-2.3cm} 
	\includegraphics[width=\columnwidth]{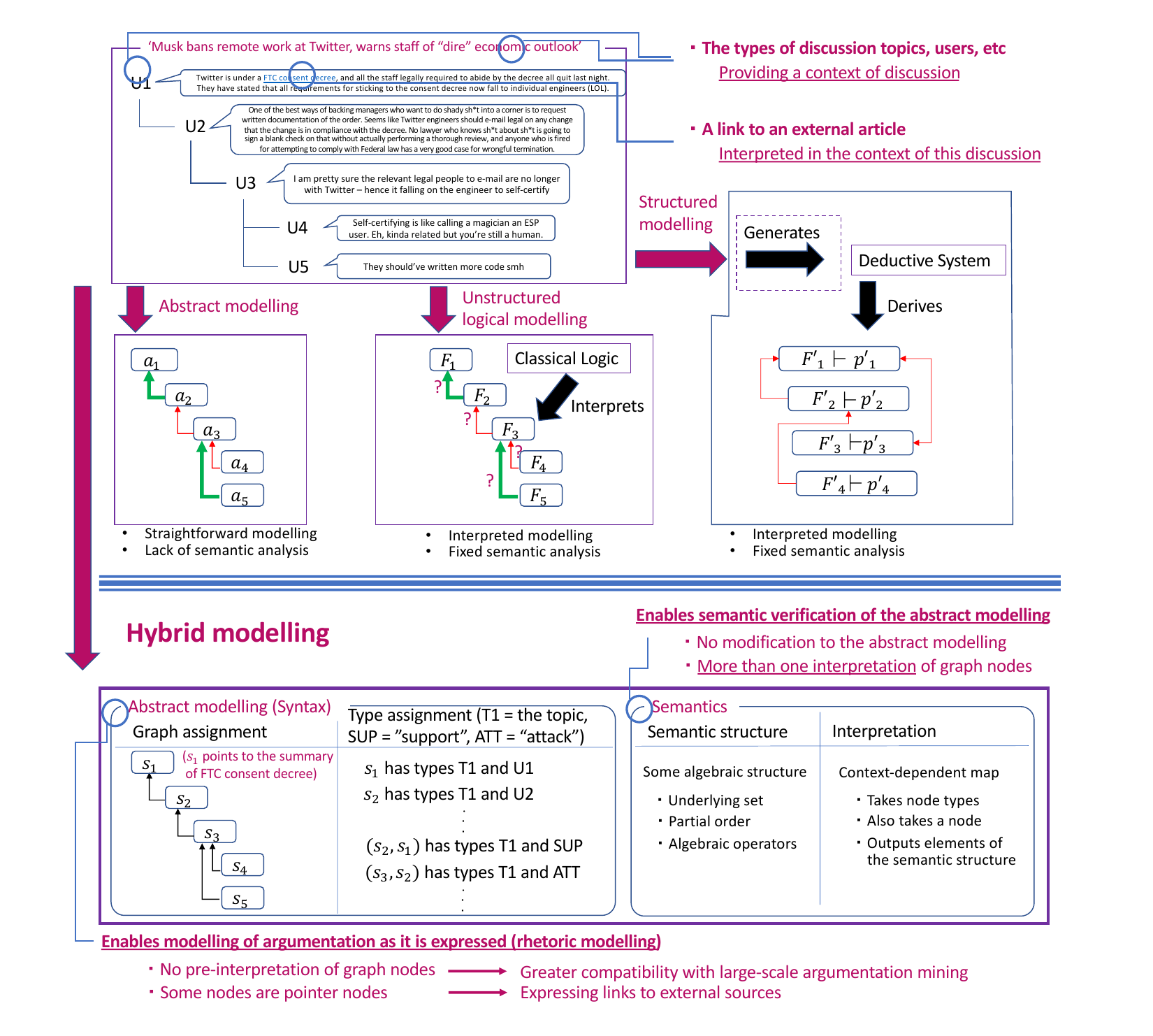} 
	% \vspace*{-10mm}
	\caption{(\textbf{Top half})  
	From Figure \ref{fig_big_picture_smaller}.  
	The contextual information and a link to an external article 
	in the given argumentation are highlighted. 
%	Different types of modellings of argumentation. The thick (green) arrows indicate supports and the thin (red) arrows 
%	indicate attacks. In  abstract modelling, the given argumentation may be straightforwardly modelled 
%	by allocating a node to each comment. In unstructured logical modelling, 
%	each comment is logically interpreted into a logical formula. Conflicts 
%	between two nodes are often judged by logical contradiction in the two formulas 
%	put together. As such, the indicated arrows may or may not actually exist in the modelling. 
%	In the like manner, there may be other arrows that are not indicated here. The structured 
%	modelling derives an argumentation graph from some deductive system as a knowledge base 
%	which may be an existing one or a newly generated one from the given argumentation. 
%	Every node represents an argument in a specific sense comprising  
%	premises ($F'_x$ in the figure) and the conclusion ($p'_x$ in the figure).   
%	In general, the degree of applied interpretation increases from the left to the right.  
%	%Depending on how the knowledge base is constructed, it could be the most heavily interpreted 
	%modelling of the three modellings.  \\ 
	(\textbf{Bottom half}) An example of hybrid modelling comprising the abstract modelling part and 
	the semantics as two separate components. The abstract modelling 
	is divided further into uninterpreted graphical modelling (specifically no arrows 
	presupposing semantic attacks, supports and possibly others) 
	and a type assignment which gives syntactic types to the graph nodes and edges. 
%	Contrary to the abstract modelling in the top half where 
%	the syntax and the semantics of attacks and supports are not well-insulated, `$attack$', `$support$' and 
%	any other types here are only syntactic. Meanwhile, the semantics consists of 
%	an algebraic structure and an interpretation which takes  
%	node types and a node and which  outputs elements of the underlying set of 
%	the algebraic structure as the semantics of the nodes. The semantics of `$attack$' and `$support$' 
%	is not defined in the model: they are meant to be defined by formal constraints 
%	on this model.  
	}
	\label{fig_big_picture}  
\end{figure*} 

\begin{itemize} 
	\item Its 1st component is an abstract argumentation model which has 
		two types of statement-to-statement relations `$attack$' and `$support$', 
		and a finite number of context types, each of which we call a theme type in this paper.
		Every relation is typed with at least one of the relation types, so it is possible 
		that some relation is typed with both. 
		Every statement and every relation is typed with at least one theme 
		type.  
		This is illustrated on the bottom left of Figure \ref{fig_big_picture} 
		which extends Figure \ref{fig_big_picture_smaller} 
		with the hybrid modelling approach. 
		Figure \ref{fig_big_picture} also 
		highlights %in the given argumentation 
		the contextual information such as the types 
of discussion topics, users and so on  
as well as a link referring to an external article, both of which are 
explicitly taken into account in the theme aspect argumentation model.   
		Any statement and any relation typed with the same theme type 
		is in the theme. In case any of them are typed with multiple theme types, 
		they belong to more than one theme. 
		There are two disjoint sets of statements. 
		One is the set of {\it ordinary} statements that do not refer to 
		anything. 
		The other set of {\it pointer} statements always refer 
		either to a theme or to an ordinary statement of some theme. %In this paper, reference to a theme signifies reference to 
		%a summary of the theme. 
		As is clear, 
		this abstract modelling captures contexts and referencing. 
	\item Its 2nd component is a complete Boolean algebra, the familiar structure 
		for propositional logic. While it is relatively simple, 
		it is equipped with algebraic operations of complement, conjunction and disjunction, 
		which we can make use of for the semantic characterisation of statements, `$attack$'s and `$support$'s fairly 
		generally. 
	\item For its 3rd component, we presume that a statement is mapped to a set of elements of the Boolean algebra 
		and that the elements represent aspects of the statement. This is in response to 
		the need to extract the representative features of the statement. Since aspects can be coarse or finer, 
		the granularity of information can be easily tuned to, for example, 
		the size of the 1st component (when it is very large, more abstraction 
		is needed for keeping the computational time small, 
		but when it is relatively small, more detailed aspects can be assigned for greater accuracy). 
		Note here that the 1st component 
		is not going to be altered. %, Cf. Figure \ref{fig_big_picture}.  
		The 3rd component (the interpretation function) needs to 
		map each 
		statement to the 2nd component, but in such a way that 
		each statement may be interpreted  differently according 
		to 
		the provided context. 
\end{itemize} 
%Figure \ref{fig_big_picture} 
%extends Figure \ref{fig_big_picture_smaller} 
%with the hybrid modelling approach. %\begin{note} \rm 
In contrast to the abstract modelling (shown in the top half of  
	Figure \ref{fig_big_picture}) where 
	the syntax and the semantics of attacks and supports are not well-insulated, `$attack$', `$support$' and 
	any other types in the 1st component of a hybrid argumentation model 
	(in particular, the theme aspect argumentation 
	model) are only syntactic. The rhetorical intention 
	of argumentation is modelled in this component 
	which 
	remains unaffected by the other components.  
	The semantics (the 2nd and the 3rd components) 
	consist of 
	an algebraic structure (in particular, a Boolean algebra) and an 
	interpretation function which takes  
	node types and a node and which  outputs elements of the underlying set of 
	the algebraic structure as the semantics of the statements. Then,   
	as per our \hyperref[subsec_mid_level_research_solution]{approach to the high-level research problem},  formal constraints on this model determine 
	the semantics of `$attack$' and `$support$'.  %\hfill$\clubsuit

\subsection{Contributions and the structure of the paper}  
We address both the low-level and the high-level research problems to 
mark a significant 
advance in this novel research direction of formal fallacy identification. 
We also analyse the computational complexities of deciding the 
satisfiability of constraints. 
%Overall, our contributions 
%are geared towards theory, unsurprisingly, 
%since this work is on the theoretical 
%foundation of formal fallacy identification, 
%but a number of examples 
%in the paper should help readers consolidate intuition.  
%In addition to the main achievements, we 
%report evaluation of our implementation 
% and carry out a detailed theoretical relation to 
% deductive argumentation 
%\citep{Besnard01} and its recent variations \citep{Hunter22,Amgound21} 
%to see .  
Below, we state our key contributions and where they are found. 
\begin{itemize} 
	\item We present a comprehensive survey. \mbox{(Section \ref{sec_related_work})}
	\item  We formalise the theme aspect argumentation model  
		as our hybrid argumentation model as a response to the \hyperref[subsec_low_level_research_problem]{low-level research problem}.
%This is a novel context-aware argumentation model that can cater for 
%		the modelling of a given argumentation as it is 
%		expressed (rhetorical modelling) and for a deeper 
%		semantic analysis 
%		of the rhetorical argumentation model. 
		\mbox{(Section \ref{sec_theme_aspect_argumentation_model})} 
	\item  We analyse the semantics of statements, `$attack$'s and `$support$'s in the theme aspect argumentation model, 
		to identify many types of rhetorically viable attacks and supports, which goes well beyond 
		the normal understanding of a conflict (and a support) within typical logic-based 
		argumentation models, structured or unstructured. %\citep{Modgil13,Dung09,Garcia04,Besnard01} and 
		%unstructured {\it e.g.} \citep{Arisaka16b} logic-based argumentations. 
		This is for the low-to-high-level.
		The closest to this contribution  
		is the contribution by \citep{Heyninck20} who 
		recognise 
		a conflict in conditionally contradictory 
		statements. We take up the  
		recent research activity and extend 
		the semantics of a conflict, 
		facilitating 
		rhetorical-grade attacks. Further, we identify 
		the semantics of rhetorical-grade supports. 
		 \mbox{(Section \ref{sec_understanding_the_semanics})} 
	\item  We formulate {\it core formal constraints}, some on 
		the 1st component, and others on 
		the semantics (the 2nd and the 3rd components) and, through them, define 
		conditions that the theme aspect argumentation model should satisfy. 
		This is for the \hyperref[subsec_mid_level_research_problem]{high-level research problem}.
		We establish a bidirectional theoretical correspondence between 
		the constraints on the 1st component (which are graphic/syntactic) and 
		those on its semantics along with 
		many other theoretical results on the consequences of the constraints. Examples that illustrate the role of individual 
		constraints 
		in formal fallacy identification 
	are given.  
		(Section \ref{sec_core_constraints_of_theme_aspect_argumentation_model}) %\pagebreak
	\item {\it More constraints} are formed for both the 1st component 
		and for its semantics, for detecting well-schemed fallacies that cannot be 
		identified in the core constraints alone. This is for the \hyperref[subsec_mid_level_research_problem]{high-level research problem}. We explain why they should not be core 
		constraints. 
		Theoretical results on their consequences are established. 
		Again, examples are given for illustrating their 
		roles.   
		We also formulate the notion of {\it logico-rhetorical 
		conclusion} which derives from a rhetorical conclusion ({\it i.e.} the conclusion of a 
		rhetorical argumentation) 
		its logical conclusion, allowing detection of more logical fallacies. 
		\mbox{(Section \ref{sec_more_constraints})} 
%	\item  {\it We demonstrate formal fallacy identification with the theme aspect argumentation model}, achieving  
%		our goal to respond to the \hyperref[subsec_high_level_research_problem]{high-level research problem}. 
%		The mid-level description of this achievement is more technical. 
%		We are introducing {\it normal forms} of theme aspect argumentation model to detect 
%		fallacies of relevance, false flags, straw man, question-begging, vagueness of verbal classification and others; and 
%		introducing the notion of {\it logico-rhetorical 
%		conclusion} which derives from a rhetorical conclusion ({\it i.e.} the conclusion of a 
%		rhetorical argumentation) 
%		its logical conclusion, allowing detection of more logical fallacies. 
%		\mbox{(Section \ref{sec_application_fallacy_detection_and_prevention})}  
	\item  We prove that the {\it computational complexity} of 
		deciding the satisfiability of the constraints 
		is: \linebreak 
		\mbox{co-NP(N}P)-complete ($\mathsf{\Pi}_2$-complete)
		for core constraints; and  
		\mbox{co-NP(N}P)-hard ($\mathsf{\Pi}_2$-hard) 
		and in\linebreak \mbox{co-NP(NP(NP))} 
		($\mathsf{\Pi}_3$) 
		for all the constraints.
		For core constraints, we further show that 
		the bidirectional theoretical correspondence 
		between the constraints on the 1st component 
		and those on its semantics 
		as described above allows for polynomial-time 
		verification of the existence of the 2nd and the 3rd components 
		given the 1st component. 
		\mbox{(Section \ref{sec_computational_complexities})} %Some of 
\end{itemize}
The first version of this paper was made public in May 2022 on arXiv \citep{Arisaka22}.    
This paper amends errors, %around the definitions of the width-/depth-statements-sets and the definition of redundancy 
%used in two formal constraints, 
makes clearer 
some of the concepts, 
enriches the literature survey (Section \ref{sec_related_work}), and 
adds a section for computational complexities. 
%On the other hand, preliminary experimental analysis has been 
%excluded from the main contents. 
%gearing towards application from the main sections. %are excluded from this paper 
%which we will discuss in another paper. We focus on theoretical 
%matters. 
%and 
%implementation. 
Extensive checks on the correctness of the proofs of formal results
were conducted both manually and through computational runs.  
While we will focus on theoretical matters in this paper, 
implementation of the theme 
aspect argumentation 
model and all the constraints as well as the data we gathered 
are in \citep{Nakai22}. %Nevertheless, 
%we will focus on theoretical matters in this paper. 
A detailed experimental evaluation will be in another paper
gearing towards application. 
%Another paper will 
%handle a detailed 
%experimental evaluation. 
%which also contains all the data we generated. %See  
%{\it \hyperref[app_implementation_detail]{Appendix: implementation, 
%gathered data 
%and run-time evaluation}} for more detail. 

%Many fallacy examples in this paper are generated in ChatGPT (GPT-4). 
%Proof-reading is also done in ChatGPT (GPT-4). 
%Fallacy examples in this paper are generated in ChatGPT (GPT-4). 
%The main authors 
%of this paper (the 1st and the 2nd authors) have over 98\% confidence 
%on the correctness of all the theoretical results in this paper 
%and 100\% confidence on that of most of those that were 
 %implements the theme aspect argumentation model 
%and all the formal constraints.  
\section{Related Work: A Survey} \label{sec_related_work} 
\subsection{Misleading deeds and fallacies}\label{subsec_misleading_deeds} Within formal argumentation,  
it is vastly more common to study misleadings in the context of persuasion 
and dialogue games, {\it e.g.} 
\citep{Kakas05,Parsons05,Rahwan09,Sakama12,Thimm14,ArisakaSatoh18,Hunter18} 
and negotiation (surveys \citep{Rahwan03,Dimopoulos14} and concurrent 
negotiation \citep{arisaka2020b}). Among them, 
deceptive argumentation \citep{Sakama12} studies 
dishonest arguments in abstract argumentation 
and how they can be detected. Issues 
of the proposed deception detection method around belief were identified to exist but were 
addressed in \citep{AHT19}. 
Other kinds of manipulation were also 
explored such as the exploitation of 
agents' inference capacity \citep{Kuipers10} 
and lie-hiding estimation \citep{Kontarinis15}.
Availability of these few studies notwithstanding, they 
cannot really detect any fallacious argumentations properly, since no semantics (note, not the 
acceptability semantics) 
are given to statements for checking the semantics of `$attack$' (and `$support$' if  it is available). 

To look at fallacies in particular, \citep{Walton08} compile numerous arguments that are potentially fallacies, list up
critical questions to see if they indeed are, create argumentation schemes, and also point out a research direction for 
formalising the schemes which was done by themselves 
but was recently also followed up by \citep{Bisquert19} who showed encoding 
of some of the schemes as rules into 
a structured argumentation formalism.  
Corpora are obtained for 
some of the argumentation schemes. However, the informal classification is troubled by 
vague boundaries of a fallacy or not a fallacy, and 
the classifications were found to be painstaking even 
for experts \citep{Lawrence20}.  
%\citep{Bisquert19} propose formalisation    
%of argumentation schemes as a deductive system 
%(see also section \ref{subsec_logic_based_argumentation}). However,  
%the deductive system focuses on recognising pattern matchings. 
%\ryuta{Detail this later.} 

Propagandist messages classification was reported 
for news articles and news outlets \citep{Barron19,Martino19}. Recently,  
it was reported \citep{Vorakitphan21,Vorakitphan21b,Jin22,Goffredo22} that consideration for argumentative features led to 
improvement  
of the classification quality. 
However, 
they, too, encounter the context-dependability of a fallacy 
as well as from classifiers' presuppositions.
The theme aspect argumentation model 
that we present in this paper accommodates the formal classification of fallacies based on 
the unsatisfaction of constraints defined on its components, addressing 
the issue of ambiguity, and achieving explainability. It also makes fallacy identification simpler. In fact, 
annotators will have little need to directly annotate fallacies, since 
the actual fallacy identification is dealt with within the theme aspect 
argumentation model. They can instead focus on 
classifying discussion themes for statements, classifying reference links, and choosing a suitable 
aspect ontology, one burden less for classifiers. There are already 
several datasets for topic classification 
({\it e.g.} Wikipedia \citep{Lehmann15}, Reddit \citep{Wallace14}, PubMed \citep{Lu11} and EUR-lex \citep{Mencia08}) developed before the arrival of ChatGPT.  
ChatGPT and subsequent large language models then 
largely supersede a lot of the natural language processing tasks 
in terms of accuracy.  
%many natural language processing tasks
%have been superceded 
%and there are also prominent machine-learning-based tools such as GPT-3 
%\citep{Brown20} 
%for large-scale knowledge, 
They could be utilised for these 3 tasks. As such, 
our proposal is timely for furthering the neuro-symbolic research which 
is seeing progress in question answering \citep{Andreas16,Iyyer17,Singhal22},  
math problem solving \citep{Schlag20} and grounded text generation \citep{Gao20} 
among many others.

%Propagandist messages classification  was reported 
%for news articles and news outlets \citep{Barron19,Martino19}. Recently,  
%it was reported that consideration for argumentative features led to 
%improvement  
%of the classification quality \citep{Vorakitphan21,Vorakitphan21b}. However, 
%they, too, suffer from significant context-dependability of a fallacy 
%as well as from classifiers' presuppositions. and there is not very strong explainability. 
%We present a formal alternative which we believe will assist deepen our understanding of a fallacy further. \\

\subsection{Contexts and referencing}  
A context is a broad term, but the specific sense of a context 
assumed in this paper is a conceptual boundary. As such, 
any previous work on multi-agent argumentation defining the scope of an agent  
has this feature. 
%To name but a few, 
In dialogue games, two agents/groups {\it e.g.} \citep{Kakas05,Oren09,Bonzon11,Sakama12,Hadjinikolis13,Hadoux17}, 
2 agents/groups plus one observer \citep{Grossi13}, 
or more recently 3+ agents/groups \citep{ArisakaSatoh18,AHT19,Tarle22}, are posited. 
Many of them presume a global argumentation and presume that each agent has its own  
model of it. %which is the agent's subjective representation of the (objective) global argumentation. 
In other words, the global argumentation is viewed in the context of each agent. 
Others include argumentation context systems \citep{Brewka09},  modal and temporal argumentation networks \citep{Barringer12} 
and the block argumentation model \citep{ArisakaSantini19}. 

Very few of them, however, characterise the 
referencing we described earlier. Modal and temporal argumentation 
networks \citep{Barringer12} with its conference publication in 2005 
treat each context as though a possible world. It can be expressed 
that some statement in a context is in all connected contexts 
or in some connected context, like in modal logic \citep{Garson13}. 
However, what we need in the referencing 
is the capability of pointing to a specific context itself 
or a statement in the context, for which 
modal/temporal operators are insufficient. 
Gabbay adapts his Fibring \citep{Gabbay99} to argumentation \citep{Gabbay09c}, illustrating  
the replacement of a statement with an argumentation graph with 
bridge rules to ensure integration \citep{Gabbay09b}.  
What we require, however, is not substitution but 
a pointer; these two are not the same. 
The block argumentation model \citep{ArisakaSantini19} 
advances the point of view: a statement is an argumentation and vice 
versa, giving them two roles, leading to a fractal-like argumentation 
model. 
Then, since a statement representing an argumentation can be referred to 
as the context 
of the argumentation,  
and can at the same time act as a statement, 
the block argumentation model can in fact express 
the theme aspect argumentation model's 
referencing to a theme. It is, however, difficult to clearly distinguish 
it from referencing to a statement of a theme.

\subsection{Hybrid argumentation models} 
As far as we are aware, 
there is a hybrid argumentation model proposed in \citep{ArisakaDauphin18,Arisaka21}. 
%\footnote{An 
%ArXiv technical report; Arisaka and Dauphin, 2018.}. 
It comes with an abstract argumentation graph and 
a complete lattice as the semantic structure 
representing an ontology. An interpretation 
function maps each statement to an element 
in the lattice assigning it its semantic meaning. 
The semantic mapping facilitates abstraction (over-approximation \citep{Cousot77}) of loops 
which 
in abstract argumentation graphs typically represent inconclusive 
argumentations.  
The abstraction via the semantic structure was reported to better define the acceptance statuses of 
the statements in the loops than those that are obtained purely 
argumentation-graphically.  
The idea of checking the appropriateness of pure-argumentation-graph acceptability judgement 
against the semantics of statements (note, not their 
acceptability semantics) was exhibited. % identifying that some of the well-known acceptability semantics   
%for dealing with loops  can judge a statement in loop that should not be 
%accepted as acceptable. 
However, it was assumed that  
directional edges signify attacks semantically. In other words,  
the syntax-semantics insulation required for 
checking whether the syntactic `$attack$'s are semantic attacks  
was not carved out. Thus, the potential application was limited to the high-level 
verification of acceptability semantics. By contrast,   
we make clear the separation, enabling formal fallacy identification.  

\subsection{Compositional argumentation models}  
There are 
many works that promote the separation of formal components 
for characterising different aspects of argumentations.  
They typically attach some components $X$ to an argumentation graph.  

$X$ may include functions that assign 
statements and statement-to-statement relations 
attributes (types) such as values 
\citep{Bench-Capon03}, probabilistic confidence \citep{Hunter13}, quantities of goods \citep{arisaka2020b}, 
statements' epistemic/doxastic ownership in multi-agent argumentation (see the references given earlier), 
and trustworthiness \citep{Costa11} including epistemic trustworthiness \citep{AHT19}.   
As \citep{Dung95} originally recognised, which set of statements   
may be acceptable largely depends on which set of constraints, 
synonymously which set of rational principles underlying them \citep{Baroni07}, 
are adopted. 
By keeping  
an argumentation graph separated from the attributing functions, 
%the argumentation graph separated from the constraints, 
it becomes easier to choose and enforce a suitable subset of constraints.  
The literature has seen a few compilations of constraints on acceptability ranks \citep{Amgoud13,Bonzon16,Baroni19} and weights 
\citep{Amgoud22}. %and agency \citep{Yu21}. 

$X$ may specify constraints on the relationships 
among acceptability statuses of statements \citep{Brewka13,Sakama20}, on state transitions in multi-agent persuasion \citep{ArisakaSatoh18,arisaka2020b,Yu20} 
and so on. 
%\footnote{Of course, $X$ may include both attributing functions and constraints together, 
%which is not atypical in multi-agent argumentation (see the references given earlier).}  
In the context of preference-based argumentation models \citep{Amgoud98,Kaci08,Amgoud14},  
the traditional approach is to derive a preference-adjusted argumentation graph  
and obtain its acceptability semantics. 
A recent work \citep{Sakama20} shows a non-invasive characterisation of the acceptability semantics 
of preference-based argumentation models by keeping the underlying 
argumentation graph separated from  
a preference constraint, claiming the benefit of insulating objective and subjective perspectives. 
%\citep{Sakama20} with their epistemic constraints show  
%a non-invasive characterisation of acceptability semantics with preference (see 
%\citep{Amgoud98,Amgoud14,Kaci08} for preference-based argumentation). 
%Acceptability semantics is given for a given argumentation graph with no preference as an objective 
%judgement. The epistemic constraint is then used to retain members of the global acceptability semantics  
%that satisfy it, which forms agent's subjective view. The separation allows them to be compared against.  
Outside of the preference-based argumentation models, it is typical in 
multi-agent argumentation to distinguish a 
global perspective from local perspectives as we stated earlier. %This we stated earlier.   
There are also some works with closely related ideas \citep{Rienstra11,Baroni14}. 
The importance of having multiple reasoning modes in multi-agent 
argumentation 
has been noted \citep{Kakas05,AHT19}. 

%which hence belongs 
%to the former type of component-based argumentation model.  
To compare, most of these works deals with 
the high-level acceptability semantics, not 
the semantics of statements. They do not look at the low-level issue  
of whether the `$attack$' and other relation types in an argumentation graph 
really have matching semantics. We do. In \citep{ArisakaSatoh18,arisaka2020b,Yu20}, 
availability of statements (or structured argumentation arguments) and 
statement-to-statement relations in a state is made conditional to X. 
They, too, however,
do not concern the low-level issue. 
%X is used to determine which  available in a state in dynamic setting. 
% in light of the semantics of the statements.   
Fortunately, the hybrid argumentation model in our proposal is mostly backwards-compatible with 
the compositional argumentation models in the first type having attributing function(s). 
%the first type of compositional argumentation model with attributing function(s).   
Since the attributes can be regarded as syntactic types, they can be used as the 1st component of the hybrid 
argumentation model. %by regarding  
%the attributes as syntactic types. 
The interpretation (the 3rd component) 
can then be made to depend on the additional types which will contextualise the semantic mapping 
further.

\subsection{Logic-based argumentation models} \label{subsec_logic_based_argumentation}
We already mentioned structured and unstructured logic-based 
argumentation models earlier in the text. 
The logic-based argumentation models before the advent of 
Dung's abstract argumentation model are compiled in \citep{Prakken01}. 
%Logic-based argumentation may be 
%structured \citep{Besnard01,Garcia04,Dung09,Modgil13,Prakken01,Gabbay09} 
%or unstructured, {\it e.g.} \citep{Arisaka16b}.     
The unstructured logic-based argumentation model \citep{Arisaka16b} 
which instantiates an idea intuitively described 
first in \citep{Gabbay09} supposes formally processable classical logic 
propositions as statements. In other words, 
they are encoded into a particular formal expression 
in classical logic. But 
there is no trace of a rhetoric model, 
and we lose access to the original argumentation. 
Since there is no guarantee that the natural language statements coincide 
in intention perfectly 
with their formal encoding, so the encoding is lossy. 
A conflict is a logical contradiction in classical logic 
and the conflicts are resolved by the AGM belief contraction 
\citep{Makinson85} minimally 
removing conflicting propositions off. 
As we already stated and shall expound upon in Section \ref{sec_understanding_the_semanics}, however, logical contradiction is not the only  
form of conflict in rhetorical argumentation. 
Any other types of conflicts thrown away, 
the already lossy encoding becomes even lossier. In our proposal, 
a given argumentation on SNS can be modelled as it is presented, which minimises  
the encoding loss during the rhetorical modelling. 
%keeping the modelling loss in rhetorical modelling at minimum if not nil. 
Further, the interpretation function (the 3rd component) 
does not modify the rhetorical model of the argumentation. The interpretation 
is context-dependent. %any statement in the 1st component may have some semantic meaning in 
%a context, but the same statement may have a different semantic meaning in a different context.  
The semantics of `$attack$' covers a wide array of conflicts. 

All structured argumentation formalisms derive an argumentation model from a knowledge base, 
and a structured argumentation argument is specified to be a pair of 
premise(s) of the structured argumentation argument and the conclusion of the structured argumentation argument.  
The knowledge base may be pre-existing or may be generated from
a given argumentation. Among them, 
\citep{Besnard01} propose a classical-logic-based structured argumentation 
formalism 
called deductive argumentation, of which Remark \ref{note_2} had a glimpse. 
As we described earlier, its knowledge base 
is a set of classical logic propositions.  
The premise of any deductive argumentation argument 
(which is called support in \citep{Besnard01}) 
is a subset of the knowledge base 
%of any deductive argumentation argument 
and  
the conclusion of the deductive argumentation argument 
is a single proposition that is proof-theoretically 
derivable (say, in a classical logic sequent calculus \textsf{G3c} \citep{Kleene52,351148}) 
from the premise. The premise does not proof-theoretically derive 
inconsistency and a strict subset of the premise does not proof-theoretically 
derive the conclusion (the minimality condition 
on the premise). %\footnote{At least 
%by how it is defined in the paper, even though most of the examples in the paper appear to concern only one 
%specific model.} 
As for the conflicts in deductive argumentation, any two deductive 
argumentation arguments $A$ and $B$ 
are in conflict when %either the negation of $A$'s conclusion is 
%proof-theoretically derivable from  
%$B$'s conclusion (rebuttal) or 
the negation of the conjunction of $A$'s premise is proof-theoretically 
derivable from $B$'s conclusion (defeater).\footnote{There are other types, rebuttal and undercut, but every undercut 
is a defeater and the presence of rebuttal materially implies the presence of a defeater \citep{Besnard01}.} %In either way, 
The conclusion of $B$ is a contrary  
%either 
of the conjunction of $A$'s premise. 
%or of $A$'s conclusion. 
The contrariness here is in the proper Aristotelian sense, {\it i.e.} as in {\it Organon}.\footnote{It is freely available 
		at \url{http://classics.mit.edu/Browse/browse-Aristotle.html}} 
		Contrary propositions are propositions that cannot both hold true simultaneously.\footnote{ 
	An introductory discussion on the Aristotelian Square of Opposition is found in \citep{Horn89} 
	and a correct interpretation of it is in \citep{Hart51}.} 
As for the derivation of an argumentation model, when a natural language argumentation is given, it is first 
encoded into the specific premise-conclusion form through the knowledge base, which 
produces all the deductive argumentation arguments, 
and then conflicts among them are identified, which produces an 
argumentation model of the given argumentation. However, the premise-conclusion formulation 
is lossy for the same reason as with the unstructured logic-based 
argumentation model and can also introduce artefacts (see \mbox{Remark \ref{note_2}}). %\citep{Hunter22}.  
Expecting enthymemes \citep{Hunter22} is very useful and supportive 
when only the honest are involved in argumentation and when it is only premises 
that can be missing. However, neither of the conditions can be taken for granted to hold in practice.
Both self-interested and malicious participants exploit
such a complementation facility by deliberately saying less and letting 
the honest and unprepared fill the hole with some `missing information' cogent to them.
(Notorious examples are criminal interrogation techniques \citep{Inbau13} and 
marketing techniques  \citep{Pratkanis92}.) %This technique is exploited 
It may be the conclusion 
that is missing. Also, given a natural language statement,    
it is not always possible to clearly designate a portion of it 
as the premise(s) and a portion of it as the conclusion \citep{Arisaka16b}. 
Since this paper's objective 
is to handle fallacies, 
we thus cannot have these two presumptions. As for the conflicts available to deductive argumentation,  
they are limited to contraries.  
%The available conflicts are limited to  

\citep{Dung09} and \citep{Modgil13} propose structured argumentation formalisms 
that are not necessarily based 
on classical logic or any mainstream logic. Any deductive system may be used instead. Both of them presume 
a contrary function which does not need to respect the Aristotelian sense of contrariness. 
Unfortunately, the more unconstrained a deductive system is, the more loosely logical it becomes \citep{Amgoud12,Heyninck20}.    
For achieving formal fallacy identification, we do require some ontological reasoning and the ability to identify a unique aspect 
as the opposite aspect 
of a given aspect, %the opposite 
%unique aspect of 
%a given aspect, 
and so it is not desirable to have the flexibility of permitting just any structure or potentially designating 
a non-contrary thing as a contrary.

\section{Theme Aspect Argumentation Model}\label{sec_theme_aspect_argumentation_model} 
\subsection{Technical Preliminaries} \label{subsec_technical_preliminaries} 
We first go through preliminaries on argumentation structures, lattices and 
Boolean algebras. 
A comprehensive list 
of notations as well as where they first appear in this paper is in 
\hyperref[app_table_of_notations]{\it \labelText{Appendix: table of notations}{text_table_of_notations}}. 

\subsubsection{Argumentation structures} 
$\pmb{A}$ denotes an infinite set of 
statements (as abstract entities) in argumentation.  
To represent an argumentation formally, 
we 
first have a finite graph $(A, R)$ 
for argumentation, where $A \finsubseteq \pmb{A}$ 
represents statements and 
$R \subseteq A \times A$ directional relations 
among them. We denote the set of all finite argumentation graphs 
by $\mathcal{G}$.  
Let $G$ denote some $(A, R) \in \mathcal{G}$, we denote by $Stmt[G]$ the set of all statements in $G$, {\it i.e.} $A$, 
and by $Rel[G]$ the directional relations $R$. 
Although the literature often keeps it implicit, the formal 
structure of an argumentation is generally a finite argumentation 
graph 
adjoined with 
a type function typing: a relation with `$attack$', `$support$' {\it etc} 
and/or with numerical values (weights); and a statement with  
numerical values and attributes (such as which agents have it). Specifically, 
suppose $\pmb{Types}$ is some countable set of types, then the 
formal structure of argumentation is some 
$(G, \Pi)$ for $G \in \mathcal{G}$ and $\Pi:  \pmb{A} \cup (\pmb{A} \times \pmb{A}) \rightarrow \mathfrak{p}(\pmb{Types})$, where 
$\mathfrak{p}(...)$ denotes the power set of ... .  
When we state `$\Pi$ types a statement or a relation $x \in Stmt[G] \cup 
Rel[G]$ with $type \in \pmb{Types}$', we specifically mean $type \in \Pi(x)$.  
%When we model a natural language argumentation in abstract argumentation,  
%every natural language statement as a string 
%natural language statements are often too long to write, 
%so they are given a placeholder, often a single letter, and the correspondence 
%is recorded in a dictionary data structure.  
%It is conventional to say that $s_1$ attacks $s_2$  
%iff $(s_1, s_2) \in Rel[G]$ and $`attack' \in \Pi((s_1, s_2))$ 
%both hold 
%and that $s_1$ supports $s_2$ 
%iff $(s_1, s_2) \in Rel[G]$ and $`support' \in \Pi((s_1, s_2))$ both hold.  

\subsubsection{Lattices and Boolean algebra based on \citep{Davey02}} 
\label{subsec_lattices_boolean_algebra} 
Given a non-empty partially ordered set $D$, $(D, \vee, \wedge)$ 
is a {\it lattice} iff (if and only if) the following two conditions hold for every $x, y \in D$: 
$x \vee y \in D$ and $x \wedge y \in D$ where $x \vee y$ denotes the supremum of $\{x, y\}$, 
and $x \wedge y$ denotes the infimum of $\{x, y\}$. $D' \subseteq D$ is a {\it down set} of $D$ iff the following holds 
for every $x \in D'$ and every $y \in D$: if $y$ is smaller than or equal to $x$ 
in the partial order, $y$ is in $D'$. Dually, $D' \subseteq D$ is an {\it up set} of $D$ iff the following holds for every $x \in D'$ and every $y \in D$: 
if $y$ is greater than or equal to $x$ in the partial order, then 
$y$ is in $D'$. Given $D' \subseteq D$, 
$\downarrow\! D'$ denotes the smallest down set of $D$ such that $D' \subseteq {\downarrow\! D'}$. 
Dually, $\uparrow\! D'$ denotes the smallest up set of $D$ such that $D' \subseteq {\uparrow\! D'}$. 

%A down set $D'$ of $D$ is an {\it ideal} iff, for every $x, y \in D'$, $x \vee y$ is also in $D'$. 
%Dually, an up set $D'$ of $D$ is a {\it filter} iff, for every $x, y \in D'$, $x \wedge y$ is also in $D'$. 

A lattice $(D, \vee, \wedge)$ is {\it distributive} 
iff it satisfies 
the following distributivity for every $x, y, z \in D$:  
$x \wedge (y \vee z) = (x \wedge y) \vee (x \wedge z)$.    
Let $(D, \wedge, \vee)$ be a lattice with $\pmb{1}$ denoting the greatest element in $D$ and $\pmb{0}$ denoting the least element in $D$, then for any $x \in D$, 
$y \in D$ is said to be a {\it complement} of $x$  
iff $x \wedge y = \pmb{0}$ and $x \vee y = \pmb{1}$ both hold. 
A {\it Boolean algebra} is then a structure $(D, \pmb{1}, \pmb{0}, \neg, \vee, \wedge)$ 
such that: $(D, \vee, \wedge)$ is a distributive lattice 
with $\pmb{0}, \pmb{1} \in D$, and that, for every $x \in D$, $\neg x$ 
is a (necessarily unique) complement of $x$. 
A {\it complete lattice} $(D, \vee, \wedge)$ is a lattice which contains a supremum of every subset $D' \subseteq D$. 
A {\it complete Boolean algebra} is then a Boolean algebra that is complete. 

Any finite Boolean algebra is 
complete. It is moreover isomorphic to a power set lattice.  
To be more specific, let $\mathcal{L}$ 
be a lattice. 
	For any $x \in L$ ($L$ is the underlying set of 
	$\mathcal{L}$), $x$ is an {\it atom} iff 
	 the following hold: (1) $\pmb{0} < x$; and (2) for any $y \in L$, $y \leq x$ materially implies  
	 either $y = x$ or $y = \pmb{0}$. 
	Given a lattice $\mathcal{L}$, 
	 $\pmb{Atom}(\mathcal{L})$ denotes 
	 all the atoms 
	 of $\mathcal{L}$. %by $\pmb{Atom}(\mathcal{L})$.   
%When $\mathcal{L}$ is a finite Boolean algebra, any member of (the underlying set of) a finite Boolean algebra is characterised 
%by a distinct set of atoms. %This is a well-known correspondence between a Boolean algebra  
%and a power-set algebra. 
%\begin{lemma}[Isomorphism and representation; 5.5 in \citep{Davey02}]\label{lem_isomorphism}   
	Given a finite Boolean algebra $\mathcal{D}$, $\eta_D: x \mapsto \{y \in \pmb{Atom}(\mathcal{D}) \mid y \leq x\}$ is an isomorphism 
	of $D$ (which is the underlying set of $\mathcal{D}$) 
	onto $\mathfrak{p}(\pmb{Atom}(\mathcal{D}))$. 
	%Given some set $X$, a Boolean algebra $(\mathfrak{p}(X), X, \emptyset, - , \cap, \cup)$ 
	%where $-$ is a unary operation satisfying $- x = X \backslash x$ for every $x \in \mathfrak{p}(X)$ 
	%is called a powerset algebra. 
	%Let $\pmb{Atom}(\mathfrak{p}({X}))$ be 
	%$\{x \in \mathfrak{p}(X) \mid \emptyset \subset x \text{ and }   
	%  \forall y \in \mathfrak{p}(X).\emptyset \subset 
	%  y \subseteq x 
	%\}$ 
%\end{lemma}  

%\begin{lemma}\label{lem_x}  Every finite Boolean algebra is isomorphic to a lattice of power set  
%	ordered by set inclusion together with . 
%\end{lemma} 

%\subsection{Properties of finite Boolean algebras} 

%as a response to the low-level research problem. 
%\subsection{Technical preliminaries}\label{subsec_technical_preliminaries} 
%\ryuta{Move this to section 4, but make sure to check that nothing 
%goes wrong in the transfer.} 
\subsection{Argumentation structure for theme-aspect argumentation models} 
We now formalise the theme aspect argumentation model  
		as our hybrid argumentation model.  
Let $\pmb{Themes}$ be an at most countable set of theme 
constants ($t_1, t_2, \ldots$). 
For the argumentation structure of our {\it theme aspect argumentation model}, 
we set $\pmb{Types} = \pmb{Themes} \cup \{`attack$'$\} \cup \{`support$'$\}$. 
We shall presume the following {\it well-formedness} in every 
$(G, \Pi)$: that $\Pi$ types 
all the statements in $Stmt[G]$ with some theme(s) and never with `$attack$' 
or `$support$'; that $\Pi$ types 
all the relations $(s_1, s_2) \in Rel[G]$ with `$attack$' and/or `$support$' and with 
some theme(s) in $\Pi(s_1) \cup \Pi(s_2)$; and that 
$\Pi$ does not type any other statements and any other relations with anything ({\it i.e.} it types 
every $s \in \pmb{A} \backslash Stmt[G]$ and every $rel \in (\pmb{A} \times \pmb{A}) \backslash Rel[G]$ 
with $\emptyset$). 
We shall write $\modell$ to indicate the well-formedness of $(G, \Pi)$.  

To shape the argumentation structure further, we identify 
pointer statements and ordinary ({\it i.e.} non-pointer) 
statements in $Stmt[G]$. Specifically, we define $PStmt[G]$ 
and $OStmt[G]$ 
as the set of pointer statements and that of ordinary statements. 

\begin{definition}[Pointer and ordinary statements]\label{def_pointer_and_ordinary_statements} \rm     
	Let $\pmb{A}^{pnt} (\not= \emptyset)$ and $\pmb{A}^{ord} (\not= \emptyset)$ be such that 
	$\pmb{A} = \pmb{A}^{pnt} \cup \pmb{A}^{ord}$ and that 
	$\pmb{A}^{pnt} \cap \pmb{A}^{ord} = \emptyset$.   
	Given  a typed argumentation graph $\modell$,   
	let $OStmt[G]$ be $Stmt[G] \cap \pmb{A}^{ord}$ 
	and $PStmt[G]$ be $Stmt[G] \cap \pmb{A}^{pnt}$.  
	We define that 
	each member of $PStmt[G]$ is either a $t.\mathcal{C}$ 
	for $t \in \pmb{Themes}$ and a special symbol $\mathcal{C}$ 
	or $t.a$ for $t \in \pmb{Themes}$ and $a \in OStmt[G]$.  
	Given $t.x \in PStmt[G]$,  
	we say {\it $t.x$ refers to an ordinary statement} 
	iff $x \in OStmt[G]$, and we say {\it $t.x$ refers to 
	\pmb{s}ummary of a theme} 
	iff $x = \mathcal{C}$.  \hfill$\spadesuit$
\end{definition}  
\begin{note} \rm It is possible to delegate this statement differentiation 
	to $\Pi$ by letting it type a relation with `$points$-$to$'. However, 
	$t.\mathcal{C}$ refers to \pmb{s}ummary 
	of an argumentation structure $t$ denotes. Such reference is best 
	regarded as a reference to the theme $t$. Unfortunately, it 
	is a burden on visualisation, since 
	we would need to draw a theme node in the graph while 
	showing which statement node is in the theme. This factor weighed in, our choice 
	is to specialise $Stmt[G]$ itself. \hfill$\clubsuit$
\end{note} 
\begin{note} \rm A link to a link to a summary of a theme
	or to a statement of theme is already uncommon in 
	typical discussion on SNS.  
	Treatment of a fully general recursive referencing (a pointer to a pointer) is out of the scope, 
	which we leave to future work. 
%	in association 
%	with block argumentation \citep{ArisakaSantini19}.
	\hfill$\clubsuit$
\end{note}
%\ryuta{
%\begin{example}[Pointer and ordinary statements] 
%         
%\end{example} 
%}
%\begin{definition}[Referred \pmb{s}ummary and referred ordinary statements] \rm 
%	Given $\modell$, we say that a member $s$ of $PStmt[G]$  
%	points to: a \pmb{s}ummary of a theme $t$ if $s \equiv t.\mathcal{C}$; 
%	and an ordinary statement of a theme $t$ if there is some $a$ 
%	such that $s \equiv t.a$. 
%	\ryuta{remove?	For any $s \in PStmt[G]$, we define 
%	$Ref[s]$ to denote: $\mathcal{C}$ if $s$ points to conclusion of a theme; 
%	and $a \in OStmt[G]$ if $s$ points to $a$ of a theme.} 
%\end{definition} 
%
\subsection{Theme aspect argumentation models} 
In order to make the fallacy judgement formally, we must have certain criteria  
with which 
to evaluate statements and relations in the argumentation structure. 
In particular, 
we need to tell whether a statement or a relation is adequate 
against such criteria, for which we need to have a semantic  
representation of our argumentation structure.  
First, for the semantic structure, we presume a complete Boolean algebra 
and regard every element as an aspect to discuss. 
The reason for this choice is, despite being relatively simple, it 
easily accommodates ontological features of aspects, 
conjunctions and disjunctions, 
and also positive and negative aspects. %Moreover, with a finite 
%domain of discourse, the universal quantification is just conjunction 
%and the existential quantification is just disjunction  

\begin{definition}[Complete Boolean algebra aspects]\label{def_complete_boolean_algebra_aspects} \rm 
	We define $\mathcal{D}$ as a complete Boolean algebra\linebreak $(D, \pmb{1}, \pmb{0}, \neg, 
	\wedge, \vee)$. We call each member of $D$ {\it an aspect}. 
	We call $\neg x \in D$ {\it a negative aspect} of $x \in D$. \hfill$\spadesuit$
%	Given $x \in D$, we call $y \in {\downarrow\! \{x\}}$ 
%	a sub-aspect of $x$.
	\end{definition}      
We regard \pmb{0} as  an {\it inconsistent aspect}, and \pmb{1} as  a  {\it trivial aspect}.  
%\begin{lemma}[Negative sub-aspects]\label{lem_negative_sub_aspects} {\ }\\
%	Let $y$ be a sub-aspect of $x$, then $\neg y \in {\uparrow\!\{\neg x\}}$.  
%\end{lemma} 
Next, we presume an interpretation function 
$\mathcal{I}: \mathfrak{p}(\pmb{Themes}) \times (\pmb{A} \cup \{\omega\})
\rightarrow \mathfrak{p}(D)$, which, given theme(s), 
associates every statement to aspects in $D$, as a subset of $D$.  
Intuitively, $\mathcal{I}(T, s)$ denotes $s$'s aspects for $T$ (or with respect to $T$, 
or under the context of $T$). 
Meanwhile, $\omega$ is a special symbol and 
$\mathcal{I}(T, \omega)$ denotes the aspects of $T$.

\begin{definition}[Theme aspect argumentation model] \label{def_theme_aspect_argumentation_model} \rm  
	We define {\it a theme aspect argumentation model} as  
	a tuple $\langle \modell, \mathcal{D}, \mathcal{I} \rangle$. 
	\hfill$\spadesuit$
\end{definition}   
\begin{example}[On the semantic granularity]\label{ex_illustration_semantic_granularity}  \rm       
	As we mentioned in section \ref{subsec_low_level_research_solution},  
	statements may be interpreted with a different level of granularity. 
	 Let us suppose an argumentation of two 
		statements.
		\begin{itemize} 
			\item $s_1$: ``{\it Alice should not buy 
	fire insurance for her summer house. 
	It costs her a lot.}'' 
			\item $s_2$: 
	``{\it She should buy property insurance for the house. 
	Bob did not buy it for his house, 
	and the decision cost him millions in the end.}'' 
		\end{itemize} %for  
	%the theme ``{\it Should Alice buy property insurance for her summer house?}''   
	The rhetorical intention of this argumentation is such that 
	$s_2$ intends to attack $s_1$. 
	Accordingly, the typed argumentation graph contains two nodes 
	$s_1$ and $s_2$ and 
	one edge from $s_2$ to $s_1$ %$s_2 \rightarrow s_1$ 
	 having 
	 `$attack$' as its type. %We graphically represent 
	% this by
	% \begin{tikzcd}[row sep=tiny]   
	%	 s_2 \arrow{r}[above]{\text{`}attack\text{'}} & s_1 . 
%		aCostH \arrow[-,ur]\arrow[-,dr] & 	  & \neg aCostH \arrow[-,ul]\arrow[-,dl] & bCostH \arrow[-,ur]\arrow[-,dr] & & \neg bCostH \arrow[-,ul]\arrow[-,dl]\\
%		& \pmb{0} & & & \pmb{0} \\ 
%                & \pmb{1} &   & & \pmb{1} \\ 
%		aP \arrow[-,ur]\arrow[-,dr] & 	  & \neg aP \arrow[-,ul]\arrow[-,dl] & bP \arrow[-,ur]\arrow[-,dr] & & \neg bP 
%		\arrow[-,ul]\arrow[-,dl]\\
%		& \pmb{0} & & & \pmb{0} 
	%\end{tikzcd}
	%For the given theme, the following are, among any others, 
	%reasonable aspects. 
		Among others, let us suppose the following aspects. 
	\begin{itemize} 
		\item $x_1$:`{\it purchasing property insurance}' 
			\begin{itemize} 
				\item $x'_1$:`{\it purchasing fire insurance}', 
					$x'_2$:`{\it purchasing homeowners insurance}' 
					(and any other 
					more specific types of property insurance) 
			\end{itemize}  
		\item $x_2$:`{\it  Alice}' and $x_3$:`{\it Bob}' (as the subjects 
			of the dialogue)
%			\begin{itemize} 
	%			\item $x'_3$:``{\it Chris}'' and 
	%				$x'_4$:``{\it David}''  
	%				(as the utterers)   
	%		\end{itemize} 
		\item $x_4$:`{\it house}' 
			\begin{itemize} 
				\item $x'_3$:`{\it summer house}' (a more specific
					type of house) 
			\end{itemize} 
		\item $x_5$:`{\it costing a lot}' 
			\begin{itemize} 
				\item $x'_4$:`{\it costing millions}' 
				%	$x'_5$:``{\it costing dearly}'' 
					(and any other more specific types
					of `{\it costing a lot}') 
			\end{itemize} 
%	a reasonable aspect to discuss. 
%	 Since both fire insurance and homeowners insurance are property insurance, 
%	 ``{\it purchasing fire insurance}'' and 
%	 ``{\it purchasing homeowners insurance}'' are 
%	 also reasonable aspects to discuss. 
	\end{itemize}  
	In a complete Boolean algebra $\mathcal{D}$ that contains all 
	the above aspects,  
	the supremum and the infimum of any set of them 
	are also an aspect. The negative aspect of any aspect, too, is 
	an aspect. For example, ($x_2 \wedge x_4$), 
	($\neg x'_3 \wedge \neg x'_4$),  
	($x_1 \vee \neg x_1$) % {\it purchasing property insurance} $\wedge$ 
%	$\neg$ {\it purchasing property insurannce}) 
	are all in $\mathcal{D}$. 
	 Now, suppose this argumentation  
	is for the theme $t$:`{\it Should Alice buy fire insurance for her 
	summer house?}', 
	the interpretation function $\mathcal{I}$  
	can be such that 
	\begin{itemize} 
		\item $\mathcal{I}(\{t\}, s_1) = \{(x_2 \wedge x'_3) \Rightarrow \neg (x_2 \wedge x'_1),
	(x_2 \wedge x'_1) \Rightarrow (x_2 \wedge x_5)\}$.  
		\item $\mathcal{I}(\{t\}, s_2) = \{(x_2 \wedge x'_3) \Rightarrow 
	(x_2 \wedge x_1), (x_3 \wedge x_4 \wedge \neg x_1) \wedge 
	(x_3 \wedge x'_4)\}$. 
	\end{itemize} 
%	such that $\{(x_2 \wedge x'_3) \Rightarrow \neg (x_2 \wedge x_1),
%	(x_2 \wedge x_1) \Rightarrow (x_2 \wedge x'_5)\} 
%	\subseteq \mathcal{I}_1(\{t\}, s_1)$, 
%	and that $\{   \} \subseteq \mathcal{I}_1(\{t\}, s_2)$. 
	
	Here and elsewhere, $x \Rightarrow y$ abbreviates $\neg x \vee y$. (The dialogue does not make it clear if Alice has 
	a house, so the first sentence of $s_1$ 
	is weakly interpreted $(x_2 \wedge x'_3) \Rightarrow 
	\neg (x_2 \wedge x_1')$. If she has it, this interpretation  
	can be strengthened to $x_2 \wedge x'_3 \wedge \neg x'_1$.  
	Similarly for the others.) 
	But suppose 
	the given argumentation is only a small part of
	a much larger argumentation with many other themes concerning 
	Alice and Bob running alongside. In that case, this semantic 
	mapping can be too detailed and the computational overhead 
	can be too much. 
	We may need to trade 
	precision for scalability by reducing 
	the amount of information captured in 
	the complete Boolean algebra (see sections \ref{subsec_low_level_research_problem} and \ref{subsec_low_level_research_solution}). 
	For example, we may abstract away the fine details of 
	a house, regarding a summer house as a house. 
	We may abstract away the fine detail of insurance, 
	regarding fire insurance as property insurance. 
	We may abstract away the fine detail of cost, 
	regarding millions of costs as a lot of cost. 
%	is a more abstract complete Boolean algebra, say $\mathcal{D}_1$, 
%	in which no $x'_i$ ($1 \leq i \leq 4$) occur.  
%	With $\mathcal{D}_1$, 
	The result is a more abstract interpretation function $\mathcal{I}_1$ which can be: 
	\begin{itemize} 
		\item $\mathcal{I}_1(\{t\}, s_1) = 
			\{(x_2 \wedge x_4) \Rightarrow \neg (x_2 \wedge x_1), 
			(x_2 \wedge x_1) \Rightarrow (x_2 \wedge x_5)\}$. 
		\item $\mathcal{I}_1(\{t\}, s_2) = 
			\{(x_2 \wedge x_4) \Rightarrow 
			(x_2 \wedge x_1), (x_3 \wedge x_4 \wedge \neg x_1) \wedge 
			(x_3 \wedge x_5)\}$.  
	\end{itemize}   
	While we do not have to change $\mathcal{D}$, it can also be abstracted 
	into $\mathcal{D}_1$ in which no $x'_i$ ($1 \leq i \leq 4$) occur.  

	The semantic interpretation with $\mathcal{D}$ and $\mathcal{I}$ 
	is clearly more precise than with $\mathcal{D}$ and $\mathcal{I}_1$ or
	with $\mathcal{D}_1$ and $\mathcal{I}_1$. 
	As for the computational cost, $\mathcal{D}_1$ and $\mathcal{I}_1$ contain 
	a fewer number of symbols. So, working with $\mathcal{D}_1$ and $\mathcal{I}_1$ 
	is computationally cheaper than with $\mathcal{D}$ and $\mathcal{I}$. 
	%So there is indeed a trade-off.  
%	It is also possible 
%	to further concretise $\mathcal{D}$ with aspects 
%	denoting situations or possible worlds that 
%	concretise the deontic and other modalities. Any modality 
%	can be made explicit by including  
%	specific situations in the complete Boolean algebra.    
%	With the new complete Boolean algebra, it is possible 
%	to intrepret both $s_1$ and $s_2$ more precisely.  
%	Boolean algebra. 
	%making explicit in which situation(s) 
	 
	\underline{\textbf{But the key point is}} that 
	the typed argumentation graph (the rhetoric model) is  
	unaffected by the choice. Whichever complete Boolean algebra 
	and whichever interpretation function are used, 
	it is explainable 
	{\it within the theme aspect argumentation model}  
	what the rhetorical intention of a given argumentation (expressed 
	in the typed argumentation graph) is, 
	what the semantic structure (the complete Boolean algebra) is, 
	and how the statements in the rhetoric model are mapped to 
	the semantic structure (the interpretation function). 
	How  fallacy judgement is being made 
	(in particular, with respect 
	to which semantic structure and which interpretation 
	on the rhetoric model) is thus transparent. %\footnote{The appropriateness
%	of interpretation is not guaranteed. But in no formal 
%	logic is there any such guarantee. The point is 
%	the appropriateness of the interpretation is readily analysable
%	thanks to the straightforward } 
	Cf. section \ref{subsec_big_picture}. 
	 \hfill$\clubsuit$ 
\end{example}

\section{Understanding the Semantics of `$attack$'s and `$support$'s}\label{sec_understanding_the_semanics} 

At this stage, the theme aspect argumentation model 
$\langle \modell, \mathcal{D}, \mathcal{I} \rangle$ is unrestricted. 
Some constraints are thus helpful. However, before going into a list of constraints, 
we ought to have a good understanding of  
the semantics of `$attack$' and `$support$', for the syntactic `$attack$' should be semantically an attack and 
the syntactic `$support$' 
should be semantically a support.  

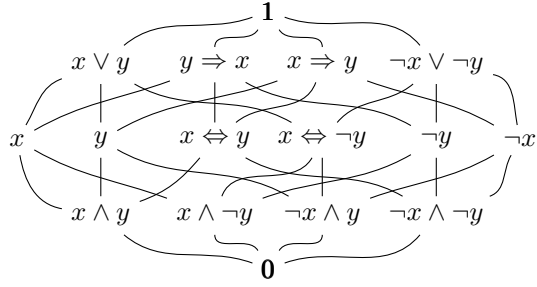
\begin{figure}[!t]  
	\centering 
	\begin{tikzpicture} %lxy    x means the level (down the greater), y means the width (right the greater) 
		\node (top) at (0, 0) {$\pmb{1}$}; 
		%%%%%%%l1 starts 
		\node [below left of=top] (l12) {$y \Rightarrow x$}; 
		\node [below right of=top] (l13) {$x \Rightarrow y$}; 
		\node [left of=l12,xshift=-0.5cm] (l11) {$x \vee y$}; 
		\node [right of=l13,xshift=0.5cm] (l14) {$\neg x \vee \neg y$};  
		\draw [out=-120,in=90] (top) to (l12);
		\draw [out=-60,in=90] (top) to (l13);
		\draw [out=-160,in=40] (top) to (l11);
		\draw [out=-20,in=130] (top) to (l14);
		%%%%%%%%l2 starts
		\node [below of=l11] (l22) {$y$}; 
		\node [left of=l22,xshift=-0.1cm] (l21) {$x$};
		\node [below of=l12] (l23) {$x \Leftrightarrow y$};  
		\node [below of=l13] (l24) {$x \Leftrightarrow \neg y$}; 
		\node [below of=l14] (l25) {$\neg y$}; 
		\node [right of=l25,xshift=0.1cm] (l26) {$\neg x$};   
		\draw (l11) to (l22);
		\draw [out=-160,in=30] (l13) to (l22);
		\draw (l12) to (l23);
		\draw [out=-110,in=40] (l13) to (l23); 
		\draw (l14) to (l25);
		\draw [out=-160,in=60] (l11) to (l21);
		\draw [out=-160,in=30] (l12) to (l21);
		\draw [out=-30,in=150] (l12) to (l25);
		\draw [out=-10,in=100] (l14) to (l26);
		\draw [out=-20,in=150] (l13) to (l26);
		\draw [out=-30,in=150] (l11) to (l24);
		\draw [out=-140,in=50] (l14) to (l24);
		%%%%%%%l3 starts  
		\node [below of=l22] (l31) {$x \wedge y$}; 
		\node [below of=l23] (l32) {$x \wedge \neg y$};
		\node [below of=l24] (l33) {$\neg x \wedge y$};  
		\node [below of=l25] (l34) {$\neg x \wedge \neg y$};   
		\draw (l25) to (l34);
		\draw [out=20,in=-150] (l32) to (l25);
		\draw (l24) to (l33);
		\draw [out=70,in=-120] (l32) to (l24); 
		\draw (l31) to (l22);
		\draw [out=20,in=-120] (l34) to (l26);
		\draw [out=20,in=-150] (l33) to (l26);
		\draw [out=150,in=-30] (l33) to (l22);
		\draw [out=170,in=-80] (l31) to (l21);
		\draw [out=160,in=-30] (l32) to (l21);
		\draw [out=150,in=-30] (l34) to (l23);
		\draw [out=20,in=-130] (l31) to (l23);
		%%%%%%%l4 starts 
		\node [below left of=l33] (l41) {$\pmb{0}$};  
		\draw [out=120,in=-90] (l41) to (l32);
		\draw [out=60,in=-90] (l41) to (l33);
		\draw [out=160,in=-40] (l41) to (l31);
		\draw [out=20,in=-130] (l41) to (l34);
%		\node [left of=l12,xshift=-0.5cm] (l11) {$x \vee y$}; 
%		\node [right of=l13,xshift=0.5cm] (l14) {$\neg x \vee \neg y$}; 
	\end{tikzpicture}  
	\caption{A complete Boolean algebra. 
	%$y \Rightarrow x$ abbreviates $\neg y \vee x$ and $x \Leftrightarrow y$ abbreviates 
	%$(x \Rightarrow y) \wedge (y \Rightarrow x)$. Similarly for the 
	%rest.
	} 
	\label{fig_boolean_algebra} 
\end{figure}

Identifying what can be an attack or a support in the context 
of the theme aspect argumentation model, however, is far from trivial, requiring attention to detail: we ought to observe the relationships carefully through 
a concrete complete Boolean algebra $\mathcal{D}$. Thus, we refer to Figure \ref{fig_boolean_algebra} for reasoning about them, %assuming 
%that this is $\mathcal{I}(\{t\}, \omega)$ and 
assuming that $x$ is an aspect 
``{\it purchasing fire insurance}'' and $y$ is an aspect ``{\it purchasing homeowners insurance}''. 
Assume $D$ is the underlying set of $\mathcal{D}$, 
for a compact description, we abbreviate $\{\neg x \in D \mid x \in D'\}$ by $\neg D'$ for any $D' \subseteq D$. 
For any $D' \subseteq D$, we also denote by $\bigwedge D'$ an infimum of $D'$ (which always exists in $D$; Cf. section \ref{subsec_technical_preliminaries}). 
\subsection{Understanding aspects of a statement}  
The difference of ordinary and pointer statements 
can wait. For now, we define specifically how we understand $\mathcal{I}(T, s)$ for 
$T \subseteq \pmb{Themes}$ and $s \in Stmt[G]$, given 
a theme aspect argumentation model $\langle \modell, \mathcal{D}, \mathcal{I} \rangle$. 
Stating it at the front, we view members of $\mathcal{I}(T, s)$ as aspects that are conjunctively expressed. 
\begin{definition}[Effective aspect of a statement]\label{def_effective_aspect_of_statement} \rm 
	Given a theme aspect argumentation model $\langle \modell, \mathcal{D}, \mathcal{I} \rangle$, $T \subseteq \pmb{Themes}$ and $s \in Stmt[G]$, we say 
	$\bigwedge \mathcal{I}(T, s)$ is {\it the effective aspect} of $\mathcal{I}(T, s)$. We also say 
	$s \in Stmt[G]$ effectively touches upon $\bigwedge \mathcal{I}(T, s)$.  
	We define that
	the effective aspect of $\mathcal{I}(T, s)$ is $\emptyset$ in case $\mathcal{I}(T, s) = \emptyset$. \hfill$\spadesuit$
\end{definition}   
Now, strictly speaking $\bigwedge \mathcal{I}(T, s)$ does not exist when $\mathcal{I}(T, s)$ is $\emptyset$. As in the given definition above, 
a rigorous description is: the effective aspect of $\mathcal{I}(T, s)$ is $\emptyset$ if $\mathcal{I}(T, s) = \emptyset$; 
$\bigwedge \mathcal{I}(T, s)$, otherwise. Nonetheless, for the sake of a compact description, we adopt a slight abuse 
and write $\bigwedge \mathcal{I}(T, s)$ for either of the cases. 

Let us assume that $\mathcal{D}$ is the complete
Boolean algebra shown above. Let us also assume that every $s \in Stmt[G]$ is typed with a theme $t$: `{\it Should we buy  property insurance?}'
\begin{description} 
	\item[Case $|\mathcal{I}(\{t\}, s)| = 1$.] Suppose $s$: 
		``{\it We should buy fire insurance or homeowners insurance.}'' is in $Stmt[G]$. 
		It touches upon 
		just $x \vee y$ in $D$ and nothing else explicitly. Thus, it certainly makes sense 
		to consider that $x \vee y$ is the effective aspect of $\mathcal{I}(\{t\}, s)$. Generally, 
		when $|\mathcal{I}(\{t\}, s)| = 1$, it is justified to regard $\bigwedge \mathcal{I}(\{t\}, s)$ as the effective aspect of 
		$\mathcal{I}(\{t\}, s)$. 
	\item[Case $|\mathcal{I}(\{t\}, s')| > 1$.] 
		Suppose $s'$: ``{\it We should buy fire insurance or homeowners insurance. We should not buy homeowners insurance.}'', 
		then $s'$ is touching upon $(x \vee y)$ and $\neg y$. 
		Thus, the effective aspect of 
		 $\mathcal{I}(\{t\}, s')$ is $(x \vee y) \wedge \neg y = x \wedge \neg y$ of $D$. Generally again, 
		 even when $|\mathcal{I}(\{t\}, s)| > 1$, it is justified to regard $\bigwedge \mathcal{I}(\{t\}, s)$ 
		 as the effective aspect of $\mathcal{I}(\{t\}, s)$. 
\end{description} 
%\begin{note}[Possible extension to deductive argumentation convention]\rm    
%%	Now that we have shaped the understanding of $\mathcal{I}(T, s)$, 
%	it is good to contemplate an embedding to deductive argumentation's 
%	convention. Since it holds 
%	for a single $T \subseteq \pmb{Themes}$, 
%	we may try to embed the semantics of a statement $s$ into 
%	a premise-conclusion tuple of $(\mathcal{I}(T, s), \bigwedge \mathcal{I}(T, s))$. 
%	It is clearly redundant, however. 
%	%as the conclusion.  
%\end{note} 
%
\begin{note}\label{note_effective} \rm   
%	In many of the logic-based structured argumentations \citep{Prakken01,Besnard01},  
%	see section \ref{subsec_logic_based_argumentation}, 
%	the conjunctive understanding is simply assumed for premises. There 
%	is normally only one proposition for the conclusion, so the conjunctive understanding 
%	for the conclusion is vacuous. 
%	However, ultimately, there is no premise-conclusion 
%	separation in $\mathcal{D}, \mathcal{I}$.  
	%The effective aspect of a statement 
	One naive way of simulating    
	the effective aspect of a statement in 
	deductive argumentation (a logic-based structured argumentation
	formalism) \citep{Besnard01} 
	is by treating $\mathcal{I}(T, s)$ as 
	the premise(s) when $s$ is interpreted in $T$ and 
	$\bigwedge \mathcal{I}(T, s)$ 
	as the conclusion when $s$ is interpreted in $T$. 
%	For the example in Example \ref{ex_illustration_semantic_granularity},   
	However, %more than that this viewpoint 
%	admits useless formal redundancy,
	the viewpoint neglects the fact that $\mathcal{I}(T, s)$ may 
	contain 
	aspects from both premises and conclusion (provided that the clear-cut premise-conclusion differentiation is possible 
	at all 
	for the statement). Also, the effective aspect of the strict subset(s) of $\mathcal{I}(T, s)$ 
	may still be $\bigwedge \mathcal{I}(T, s)$, not in accordance with the premise's minimality condition prescribed 
	for deductive argumentation arguments (see section \ref{subsec_logic_based_argumentation}). %These points make the simulation 
	%naive from 
	At the level of argumentation graph generation, too, 
	this simulation 
	may not be as preferred in deductive argumentation as to other 
	alternatives.
	For example, in the argumentation in 
	Example \ref{ex_illustration_semantic_granularity}, 
	there was a statement $s_1$: ``{\it Alice should not buy fire insurance for 
	her summer house. It costs her a lot."}. 
	%and $t$:`{\it Should Alice buy insurance for her summer house?}', 
	The premise-conclusion form requirement may incentivise 
	one to view 
	the first sentence in $s_1$ as the conclusion 
	and the second sentence in $s_1$ as the premise of $s_1$. 
	So the structuralist interpretation of 
	$s_1$ may be: 
	\begin{itemize} 
		\item Premise(s): $\{${\it ``Fire insurance costs Alice a lot.''},  
			{\it ``If Fire insurance costs Alice a lot, then 
			Alice should not buy fire insurance for her 
			summer house.''}$\}$  
		\item Conclusion: {\it ``Alice should not buy fire insurance 
			for her summer house.''} 
	\end{itemize} 
        The addition of an extra sentence to the premise 
	is due to the requirement of deductive derivability. Cf. Remark \ref{note_2}. This interpretation is clearly different from  
	the above-described simulation. 

	All these indicate that the suggested simulation of the 
	effective aspect 
	of a statement is indeed naive. To be better able to take the structuralist stance, 
	a more proper way is to 
	add one more input parameter to $\mathcal{I}$ so that 
	when `premise' is given to the input parameter, 
	$\mathcal{I}$ gives back aspects of a statement for its premises 
	and when `conclusion' is instead given to the input parameter $\mathcal{I}$ gives back aspects of a statement for its conclusion. 
	The structuralisation, however, is not the main point of this paper. 
	More relevantly, the premise-conclusion divisibility cannot be 
	presumed in this paper; Cf.  
	section \ref{subsec_logic_based_argumentation}. 
	\hfill$\clubsuit$ 
\end{note} 

\subsection{Understanding attacks} 
The view that a contradiction or a contrary in the Aristotelian sense forms a controversy in argumentation is dominant in
logic-based argumentation models, 
unstructured or 
structured. As we see below, in the theme aspect argumentation model, 
contraries cover only a part of what can be an attack. 

\begin{description} 
	\item[Contraries.] For each aspect, an obvious aspect that is in direct conflict with it is its complement. $\neg x$ is in conflict with $x$.  
Suppose $s_1$: ``{\it We should buy fire insurance}'', it touches upon $x$. $s_2$: ``{\it We should not buy fire insurance}'' that touches upon $\neg x$ 
clearly is against $s_1$, which can thus be considered an attack. A possibility of attack can be also found in a contrary 
aspect which in the proper sense of contrariness is any alternative to $x$ that cannot hold simultaneously with it. 
		They are $\neg x$, \mbox{$\neg x \wedge \neg y$}, \mbox{$\neg x \wedge y$} and \pmb{0}, {\it i.e.} a member of 
		\mbox{$\downarrow\! \{\neg x\}$}. 
		Generalising this reasoning, we obtain that 
		a statement $s'$ may attack $s$ for $t$ when $\bigwedge \mathcal{I}(\{t\}, s')$ is a member of 
		$\downarrow \{\neg (\bigwedge \mathcal{I}(\{t\}, s))\}$. 
	\item[Weakening.]  However, if we state $s_3$: ``{\it We should buy fire insurance or homeowners insurance.}'' touching upon $x \vee y$ and thereby 
weakening $x$, that may very well be an attack directed at $s_1$, even though $x$ and $x \vee y$ can hold simultaneously. 
This is because $s_3$ no longer insists the purchase of fire insurance which you perhaps see necessary. It is only 
when we had stated $s_1'$: ``{\it We should buy fire insurance or we should buy homeowners insurance.}'' 
touching upon $x \vee y$ that it would become absurd to recognise a possibility of attack in $s_3$ against $s_1'$.  
	Generalising this reasoning, we obtain that a statement $s'$ may attack $s$ 
		when $\bigwedge \mathcal{I}(\{t\}, s') \in\ \uparrow 
		(\{\bigwedge \mathcal{I}(\{t\}, s)\}$ 
		and $\bigwedge \mathcal{I}(\{t\}, s') \not= 
		\bigwedge \mathcal{I}(\{t\}, s)$.  
	\item[Strengthening.] Furthermore, if we instead state $s_4$: 
		``{\it We should buy fire insurance. We should buy homeowners insurance.}'' touching upon 
$x \wedge y$ and thereby strengthening $x$, that, too, may very well be an attack directed at $s_1$, for
$s_4$ asks for an additional burden which you perhaps do not see necessary. 
It is only when we had stated $s_1''$: ``{\it We should buy fire insurance. We should buy homeowners insurance.}'' touching 
upon $x \wedge y$ that it would become absurd to recognise a possibility of attack in $s_4$ against $s_1''$. 
		Generalising this reasoning, we obtain that a statement  
		$s'$ may attack $s$ when 
		$\bigwedge 	\mathcal{I}(\{t\}, s') \in 
		{\downarrow(\{\bigwedge \mathcal{I}(\{t\}, s)\})}$ and $\bigwedge \mathcal{I}(\{t\}, s') \not= 
		\bigwedge \mathcal{I}(\{t\}, s)$.  
	\item[Weakened contradiction \citep{Heyninck20}.] If we instead 
		state $s_5$: ``{\it We should either not buy 
		fire insurance or not buy homeowners insurance.}'' touching upon $\neg x \vee \neg y \in 
		({\uparrow \{\neg \bigwedge \mathcal{I}(\{t\}, s_1)\}}) \backslash \linebreak
		\{\bigwedge \mathcal{I}(\{t\}, s_1)\}$, 
		that may too be an attack. It may help 
		to note that 
		$\neg x \vee \neg y$ is equivalent to $y \Rightarrow \neg x$, 
		which is a conditional contradiction. 
	\item[Incomparable alternatives.] Finally, suppose any $(s', s) \in Rel[G]$, in case the effective aspect of $\mathcal{I}(\{t\}, s')$ 
		has got nothing to do with the effective aspect of $\mathcal{I}(\{t\}, s)$, 
		in the precise formal sense that 
		$\bigwedge \mathcal{I}(\{t\}, s') \not\in {\downarrow \{\bigwedge \mathcal{I}(\{t\}, s)\}} 
		\cup {\uparrow \{\bigwedge \mathcal{I}(\{t\}, s)\}} 
		\cup {\downarrow \{\neg \bigwedge \mathcal{I}(\{t\}, s)\}} 
		\cup {\uparrow \{\neg \bigwedge \mathcal{I}(\{t\}, s)\}}$, then  
		$\bigwedge \mathcal{I}(\{t\}, s')$ is an incomparable alternative to $\bigwedge \mathcal{I}(\{t\}, s)$. 
		If we for example state  
		$s_6$: ``{\it We should buy  
		homeowners insurance.}'' touching upon 
		$y$, that may very well be an attack directed at $s_1$, 
		directed in fact stronger than $s_3$. 
	\end{description} 

\noindent All the above cases put together, 
		we obtain that a statement $s'$ may attack
		 $s$ when $\bigwedge \mathcal{I}(\{t\}, s') 
		 \not= \bigwedge \mathcal{I}(\{t\}, s)$. That is, attack 
		 is more a difference than a logical contradiction. 
\begin{note}\label{note_xxx} \rm  
	 Some of these types of conflicts are informally discussed in the literature of abstract argumentation models 
	 prior to the recent exploration by \citep{Heyninck20} of non-contrary conflicts in structured argumentation. 
	 An example of incomparable alternatives 
	 appears to be given in \citep{Baroni05} capturing the analysis of \citep{Pollock94}.  
	However, the illustration is ultimately informal, unconnected to any semantic structure.   
	 We have positioned it formally. Indeed, we have covered all the possible 
	 categories within a Boolean algebra as above. 
	\hfill$\clubsuit$ 
\end{note} 
\subsection{Understanding supports}  
Let us reason about supports in $\mathcal{D}$ in a similar manner.
\begin{description} 
	\item[Affirmation.] For each aspect, an obvious aspect that is in direct support of it is itself. 
		For $s_1$: ``{\it We should buy fire insurance}'' that 
		touches upon $x$, 
		$s_2$: ``{\it I agree, we should buy fire insurance}'' that touches upon $x$ 
		is clearly for $s_1$. Thus, $s'$ may support $s$ when  
		$\bigwedge \mathcal{I}(\{t\}, s') =  
		\bigwedge \mathcal{I}(\{t\}, s)$.   
	\item[Insurance (Weakening).]  However, if we state 
			$s_3$: ``{\it We should buy fire insurance or homeowners insurance.}'' touching upon $x \vee y$, 
		that may be as much of a support for $s_1$ as a weakened contradiction may be an attack, 
		due to $x \vee y \equiv \neg y \Rightarrow x$ which is a conditional support of $x$. 
We obtain that a statement $s'$ may support $s$ 
		when $\bigwedge \mathcal{I}(\{t\}, s') \in 
		{\uparrow \{\bigwedge \mathcal{I}(\{t\}, s)\}}$.   
	\item[Value-augmentation (Strengthening).] Furthermore, if we state $s_4$: 
		``{\it We should buy fire insurance. We should buy homeowners insurance.}'' touching upon 
$x \wedge y$ and thereby strengthening $x$, that, too, may very well be a
		support for $s_1$ in case you do not mind 
		additionally buying homeowners insurance. 
		We obtain that a statement  
		$s'$ may support $s$ when 
		$\bigwedge 	\mathcal{I}(\{t\}, s') \in 
		{\downarrow\{\bigwedge \mathcal{I}(\{t\}, s)\}}$. 
	\item[Contraries and Weakened contradiction.]  
		 On the other hand, if we state $s_5$: 
		 ``{\it If we buy fire insurance, 
		we should also buy homeowners insurance.}'' 
		({\it i.e.} ``{\it We should not buy fire insurance, 
		or we should buy homeowners insurance.}'') 
		touching upon $\neg x \vee y$,  
                the locution's intention does not support  
		$s_1$ in any way.\footnote{It is possible that 
		$s_6$ is mixed up with ``{\it If we buy homeowners 
		insurance, we should also buy fire insurance.}'' 
		touching upon $\neg y \vee x$ which may 
		be a support. This kind of logical fallacy, in which one takes 
		an expression for another, is not in the scope 
		of the current work, however.}  
		And, of course, a statement 
		that effectively touches upon a contrary 
		of $x$ does not act to support $s_1$. 
	\item[The others.] For any other cases, if we for example state 
		$s_6$: ``{\it We should buy  
		homeowners insurance.}'' touching upon 
		$y$ as an incomparable alternative to $x$, 
		the difference from $s_7$: ``{\it We should also 
		buy homeowners insurance.}'' which touches 
		upon $x \wedge y$ is salient. The former 
		intends to offer an option different from 
		$x$, which is not a support but an attack, 
		while the latter can be  
		value-augmentation (see the value-augmentation case above). 
\end{description} 
From all these cases, we obtain that a statement $s'$ 
may support $s$ when $\bigwedge \mathcal{I}(\{t\}, s') \in 
{\uparrow \{\bigwedge \mathcal{I}(\{t\}, s)\}} \cup 
{\downarrow \{\bigwedge \mathcal{I}(\{t\}, s)\}}$.  

\begin{note}\label{note_xyz} \rm 
	In logic-based structured argumentation formalisms 
	(deductive argumentation \citep{Besnard01} for example), 
	the premise-conclusion relation is given  
	as the proof-theoretical consequence 
	relation which is described as a `support' in \citep{Besnard01}. %which is rigorously the following relation 
%	coming with two parts:  
%	(1) if the premise(s) are true, then the conclusion is 
%	(proof-theoretically) derivable from them 
%	and (2) if some of the premise(s) are false, then the conclusion is not derivable from them. 
	In everyday argumentation, when we support somebody's statement with some statement, 
	we rarely intend that their statement is a proof-theoretical 
	consequence of our statement. As a matter of fact, it is very difficult 
	to casually attain the proof-theoretical 
	consequence (see Remark \ref{note_2}). 
	%satisfying (1) and (2). 
	Rhetorically acceptable nuances of support 
	are wider, and the Value-augmentation is not necessarily 
	the relation of the proof-theoretical consequence  
	since the semantic structure in theme aspect argumentation 
	model is just one single semantic structure. %It is not validity but 
%	satisfiability that it captures.  
	%whereas 
	%the proof-theoretical consequence encompasses all structures. 
	There is some work on structured argumentation \citep{Kristijonas17} importing the 
	nuance of support 
	in abstract argumentation models \citep{Boella10,Nouioua11}. However, the support in abstract argumentation models 
	is of acceptability status. It is not primarily determined 
	by the semantics of statements. 
	\hfill$\clubsuit$ 
\end{note} 

%\cup 
%{\uparrow \{\neg \bigwedge \mathcal{I}(\{t\}, s)\}}$.  
%\begin{note} \rm   
%	The literature on the semantics of `$support$' between 
%	statements (or between structured argumentation arguments
%	in the case of a structured argumentation formalism) 
%	is scarce 
%	even within structured argumentation. There is some formulation \cite{Kristijonas17} that imports 
%	the nuance of support relation within abstract argumentation, in particular 
%	from \cite{Boella10,Nouioua11}. But the abstract argumentation support  
%	concerns acceptability semantics. In other words, the semantics of `$support$' in abstract argumentation   
%	is not determined by the semantics of statements.  
%\end{note} 
\section{Core constraints of theme aspect argumentation model} \label{sec_core_constraints_of_theme_aspect_argumentation_model}
In this section, we probe core constraints on the theme aspect 
argumentation 
model $\taam$. Some are on the typed argumentation graph $\modell$ 
and others are on the theme aspect argumentation model. 
Those on the theme aspect argumentation model are 
meant to be taken collectively; 
however, to give an idea of their individual role 
in formal fallacy identification, we describe them through examples. 
%features that 
%can be only collectively catered for. 
%\ryuta{Write something here. The limitation of individual constraints.} 
%What role each of them plays in formal fallacy identification 
%will be described intuitively using finite Boolean algebras.
We illustrate what theoretical consequences 
ensue when they are combined. 
We conclude this section by showing theoretical
relations between the constraints on the 1st component (the typed argumentation graph $\modell$) 
which is purely graphic/syntactic 
and those on $\taam$ involving the semantics.  
The list of all the constraints in this and the next sections is in {\it \hyperref[app_table_of_constraints]{\labelText{Appendix: table of constraints}{text_table_of_constraints}}}. 
Apart from trivial ones, proofs are listed in {\it \hyperref[app_proofs]{\labelText{Appendix: proofs}{text_proofs}}}. 

As the first constraint of this section, when a statement has a `$support$' or an `$attack$' relation 
to a statement, the two statements should be related by a theme.  
This is achieved by a constraint purely 
on the typed argumentation graph. %$\modell$. 
%This is a constraint 
%on a typed graph $\modell$. 
\begin{definition}[(Graphic) Theme relevance constraint]\label{def_theme_relevance_constraint}\rm 
	Given a theme aspect argumentation model 
	$\langle \modell, \mathcal{D}, \mathcal{I} \rangle$, 
	we say the typed argumentation graph $\modell$ 
	satisfies   
	constraint \pmb{tr} (\pmb{t}heme \pmb{r}elevance)  
	iff for every $(s_1, s_2) \in Rel[G]$, 
	there exists some $t \in \pmb{Themes}$ such that 
	$t \in \Pi((s_1, s_2)) \cap \Pi(s_1) \cap \Pi(s_2)$.  \hfill$\spadesuit$ 
\end{definition}

\begin{example}[The role of \pmb{tr} in formal fallacy identification] \rm \label{ex_role_tr} 
 	%\pmb{tr} guards against remarks that are theme-irrelevant to the statement they respond to.  
	Within the repertoire of informally classified fallacies, 
	there are examples such as 
	{\it ad hominem}, {\it fear appeals}, {\it ad populum} and 
	{\it tu quoque}. All of them are 
	relevance fallacies. 
%	It captures a general principle of theme relevance 
%	and guards against relevance fallacies such as 
%	{\it ad hominem}, {\it fear appeals}, {\it ad populum} and 
%	{\it tu quoque} as are so called  
%	within the informally classified fallacy repertoire.  
%	Core constraints should capture 
%	broad principles around the relevance rather than minute points that differentiate 
%	one type of relevance from another. The \pmb{tr} constraint serves 
%	that purpose. 
	Suppose Angela and Becky 
	are having argumentation on the theme $t$: `{\it healthcare access}'.  
	\begin{itemize} 
		\item Angela $s_1$: ``{\it Everyone should have access to 
	quality healthcare.}". 
		\item Becky $s_2$: ``{\it Isn't it funny how you say that while drinking a soda? You 
	clearly aren't saying that seriously.}'' %care  
	%about health.}"  
	\end{itemize} 
	The rhetorical intention is such that Becky intends to attack Angela's 
	statement. 
	So, the typed argumentation graph 
	for this argumentation is 
	\begin{tikzcd}[row sep=tiny,column sep={1.6cm}]   
		s_2:\{t_2\} 
		\arrow{r}[above]{\{\text{`}attack\text{'},t\}} 
		& s_1:\{t, t_1\} 
%		& s_8:\{t_2,t_8\} \arrow{r}[above]{\{\text{`}attack\text{'},t_2\}} & s_7:\{t_2,t_7\}
%		 \arrow{r}[above]{\{\text{`}support\text{'}, t_2\}} & s_6:\{t_2,t_6\}.  
%		aCostH \arrow[-,ur]\arrow[-,dr] & 	  & \neg aCostH \arrow[-,ul]\arrow[-,dl] & bCostH \arrow[-,ur]\arrow[-,dr] & & \neg bCostH \arrow[-,ul]\arrow[-,dl]\\
%		& \pmb{0} & & & \pmb{0} \\ 
%                & \pmb{1} &   & & \pmb{1} \\ 
%		aP \arrow[-,ur]\arrow[-,dr] & 	  & \neg aP \arrow[-,ul]\arrow[-,dl] & bP \arrow[-,ur]\arrow[-,dr] & & \neg bP 
%		\arrow[-,ul]\arrow[-,dl]\\
%		& \pmb{0} & & & \pmb{0} 
	\end{tikzcd}
	where $t_2$ denotes `Becky' and $t_1$ denotes `Angela'. 
	%It is assumed that $\pmb{Themes} = \{t, t_1, t_2\}$. 
	Here and elsewhere, $s:T$ in the visualisation 
	of a typed argumentation graph indicates
	a node $s$ typed with $T$, and the set of types above or below 
	an edge in the visualisation indicates that the edge is 
	typed with the types. 
	%records an edge from Bob's statement to Alice's statement  
	%with `$attack$' being its type. As for the statements, 
	Angela's statement $s_1$ falls under the theme $t$ 
	`{\it healthcare access}'. On the other hand, 
	while Becky's statement $s_2$ 
	has the appearance of 
	attacking Angela's statement by 
	pointing out her personal choice to drink a soda, 
	it is irrelevant to $t$ `{\it healthcare access}'. %While he appears to attack Alice's statement by pointing out a personal 
%	choice to drink a soda, they do not touch upon the theme 
	%``{\it healthcare access}''. 
%	Alice's statement has the theme as its type, but not Bob's. 
	\pmb{tr} is not satisfied by this kind of typed argumentation graph. %in force, 
	%these kinds can be formally identified fallacious 
	%as they violate the constraint. 
%	A semantic counterpart, the constraint \pmb{i},   
%	will be introduced later. 
%	Let us consider {\it ad hominem}, again between Alice and Bob, 
%	on the theme ``{\it Should we buy fire insurance?}'' 
%	Alice states: ``{\it We should buy fire insurance.}'' and Bob states: 
%	``{\it I don't think so. You are overcautious and see danger in everything.}'' 
%	As we saw in Remark \ref{note_1}, the rhetoric is such that Bob intends to attack Alice's statement. 
%	So, once again, the typed argumentation graph records an edge from Bob's statement to Alice's 
%	statement with `$attack$' being its type. 
	%In Remark \cite{note_1}, 
	 \hfill$\clubsuit$  
	%the rhetoric model of this dialogue records `$attack$' from 
	%Bob's statement to Alice's statement. 
	%\ryuta{There are many types of relevance fallacies bearing various names 
	%{\it ad hominem}, 
	%{\it fear appeals}, {\it ad populum}, {\it tu quoque}, so on and so forth. 
	%In formal fallacy identification, rather than minute differentiations, 
	%broader principles around relevance are captured. Put this somewhere else.} 
\end{example}

Moreover, a pointer statement $s \in PStmt[G]$ 
should actually refer to something. This is again achieved by a constraint 
purely on the typed argumentation graph. 
\begin{definition}[(Graphic) No null pointer constraint]\label{def_no_null_pointer_constraint} \rm 
	Given a theme aspect argumentation model $\langle \modell, \mathcal{D}, \mathcal{I} \rangle$, 
	we say the typed argumentation graph $\modell$ satisfies 
	constraint $\pmb{nnp}$ (\pmb{n}o \pmb{n}ull 
	\pmb{p}ointer) iff the following hold for 
	every $s \in PStmt[G]$. 
	\begin{description} 
		\item[Case $s \equiv t.\mathcal{C}$ for some $t 
			\in \pmb{Themes}$:] {\ }\\
			there is some $s' \in Stmt[G]$ such that 
			$t \in \Pi(s')$. 
		\item[Case $s \equiv t.a$ for some $t 
			\in \pmb{Themes}$ and some  
			$a \in \pmb{A}^{ord}$:]   
			$a \in Stmt[G]$ and 
			$t \in \Pi(a)$. \hfill$\spadesuit$
%			wwonee 
	\end{description}
\end{definition}

\begin{example}[The role of \pmb{nnp} in formal fallacy identification]  \rm \label{ex_role_nnp} 
	There are fallacies 
	involving reference, directly or implicitly, to non-existent or fictional things. %directly 
	%or implicitly. %Within the informal fallacy repertoire,   
	They are known by various names such as 
	{\it appeal to non-existent (false) authorities} and 
	{\it false attribution}  
	 within the repertoire of informally classified 
	fallacies. 
	Let us consider {\it false attribution} here. 
	Ann and Braun are having a discussion on the theme 
	$t$: `{\it investment in XYZ}'.  
	\begin{itemize} 
		\item Ann states $s_1$: ``{\it We should invest 
	in XYZ, shouldn't we?}'' 
		\item Braun fabricates an imaginary evidence and states $s_2$: 
			``{\it Indeed, we should. The Fox News 
	reported this morning that 
	8 out of 10 businesses using XYZ's products saw a 200\% increase in productivity.}''  
	\end{itemize} 
	The rhetorical intention is such that Braun supports Ann's statement. 
	So, the typed argumentation graph  
	for this argumentation is  
	\begin{tikzcd}[row sep=tiny,column sep={1.6cm}]   
		s_2:\{t, t_2\} 
		\arrow{r}[above]{\{\text{`}support\text{'},t\}} 
		& s_1:\{t, t_1\} 
%		& s_8:\{t_2,t_8\} \arrow{r}[above]{\{\text{`}attack\text{'},t_2\}} & s_7:\{t_2,t_7\}
%		 \arrow{r}[above]{\{\text{`}support\text{'}, t_2\}} & s_6:\{t_2,t_6\}.  
%		aCostH \arrow[-,ur]\arrow[-,dr] & 	  & \neg aCostH \arrow[-,ul]\arrow[-,dl] & bCostH \arrow[-,ur]\arrow[-,dr] & & \neg bCostH \arrow[-,ul]\arrow[-,dl]\\
%		& \pmb{0} & & & \pmb{0} \\ 
%                & \pmb{1} &   & & \pmb{1} \\ 
%		aP \arrow[-,ur]\arrow[-,dr] & 	  & \neg aP \arrow[-,ul]\arrow[-,dl] & bP \arrow[-,ur]\arrow[-,dr] & & \neg bP 
%		\arrow[-,ul]\arrow[-,dl]\\
%		& \pmb{0} & & & \pmb{0} 
	\end{tikzcd}
	where $t_2$ denotes `Braun' and $t_1$ denotes `Ann'. 
	 Braun's statement $s_2$ 
	is a pointer statement, say $t_3.a$, where $t_3$ denotes 
	``{\it Fox News this morning}'' 
	and $a$ denotes ``{\it 8 out of 10 businesses 
	using XYZ's products saw 
	a 200\% increase in productivity}''. It points to a 
	non-existent ordinary statement $a$ in the theme 
	$t_3$. 
	%It is assumed that $\pmb{Themes} = \{t, t_1, t_2, t_3\}$. 
	%Both statements and the edge are typed with $t$, so they satisfy 
	%\pmb{tr}. 
%	Braun's statement $s_2$ 
%	is a pointer statement, say $t_3.a$, where $t$ denotes ``{\it Fox News this morning}'' 
%	and $a$ denotes ``{\it 8 out of 10 businesses using XYZ's products saw 
%	a 200\% increase in productivity}''. It points to a non-existent statement $a$ in the theme 
%	$t$. 
	\pmb{nnp} is not satisfied by this kind of typed argumentation graph. 
	\hfill$\clubsuit$ 
	%do not satisfy \pmb{nnp}. %With \pmb{nnp} in force, these kinds can be identified 
%	a formal fallacy %formally 
	%identified fallacious as they violate the constraint. \hfill$\clubsuit$%``{\it Fox News 
	%this morning}''. 
%	The {\it false flags} also refers to non-existent things, 
%	to which extent it also belongs to this category.  
	%{\it argumentum ad lapidem} 
	%rely on reference to non-existent information source.
	%\pmb{nnp} 
\end{example} 

\begin{note} \rm  
	%It will be interesting to see how formal fallacy identification
%consolidates 
%	our understanding of the phenomena of fallacies. %further. 
%There are many informally classified fallacies bearing various names. 
%More than 50 of them are listed in \citep{Walton08} alone.  
%However, they are not necessarily independent  
%of others, and 
%due to the innate difficulties surrounding
%informal properties (see sections \ref{subsec_high_level_research_problem}
%and \ref{subsec_solution_high_level_research_problem}), 
%it may also not be the case that the current 
%repertoire identifies all that fall into it as a fallacy. 
{\it False flag} has become prominent 
	as a tactic to disguise 
	the actual source of responsibility so as to justify 
	one's otherwise unjustifiable behaviour and actions.\footnote{\url{https://en.wikipedia.org/wiki/False_flag}}  
A typical scenario is to accuse first with a concocted `fact' and 
	then to materialise the `fact' next.  
	While it does not seem to be widely discussed as 
	a fallacy in the literature of informally classified fallacies,
	it exhibits the same phenomena 
	as {\it false attribution}. 
%	A full treatment 
%of false flag goes beyond the scope of this paper; however, 
%a formal explanation is given for the following scenario as a violation of \pmb{nnp}, 
%that is, due to \mbox{Theorem \ref{thm_core}}, that of \pmb{Core}.  
Suppose an allegation against X was made by Y that X sabotaged the PC room. 
	Y knows it is not the case, but Y wants X to get arrested for it. Y plans to go and destroy the PC room later.  
	This is {\it false attribution} 
	with an extra condition, the extra condition being 
	that the imaginary evidence may be actually materialised 
	at some point. Indeed, 
	suppose the theme $t$: `{\it Should X be arrested?}', 
	and suppose two other themes $t_1$: `{\it X's deeds}' 
	and $t_2$: `{\it Y's claims}'.  
	Y's allegation $s_1$ is a pointer statement to X's deed, {\it i.e.} $s_1$ is
	$t_1.a$ with $a$: ``{\it X sabotaged the PC room.}''  
	It fails to satisfy \pmb{nnp}.  
	As this case exemplifies, formal constraints can 
	identify the phenomena of fallacies more uniformly. 
		\hfill$\clubsuit$ % and therefore is a 
\end{note} 

Now, as per our analysis in Section \ref{sec_understanding_the_semanics}, `$attack$' should be an attack, and `$support$' should be a support for every concerned set of themes. 
Further, any statement that does not map to any aspect in $D$ (which is the underlying set of a complete Boolean algebra $\mathcal{D}$) with respect to the theme(s) associated  
to an `$attack$' relation or a `$support$' relation into or out of 
the statement shall not attack, be attacked by, support, or be supported by anything. 
\begin{definition}[Attack as attack, support as support constraint]\label{def_attack_as_attack_support_as_support} \rm 
   Given a theme aspect argumentation model $\langle \modell, \mathcal{D}, \mathcal{I} \rangle$, 
	we say it satisfies constraint \pmb{aass} (\pmb{a}ttack as \pmb{a}ttack, 
	\pmb{s}upport as \pmb{s}upport)  
	iff the following hold for every $\emptyset \subset T \subseteq
	\pmb{Themes}$ 
	and every  $(s', s) \in Rel[G]$: 
	$\{`attack$'$\} \cup T \subseteq \Pi((s', s))$ materially implies 
	%and 
	%$\mathcal{I}(T, s) \not= \emptyset$ 
%	and $\mathcal{I}(T, s') \not= \emptyset$ 
%	materially imply  
%	$\emptyset \not= 
	$\emptyset \not= \bigwedge \mathcal{I}(T, s') \not=  
	\bigwedge \mathcal{I}(T, s) \not= \emptyset$;
	%\not= \emptyset$; 
	and $\{`support$'$\} \cup T \subseteq \Pi((s', s))$ %and  
%	$\mathcal{I}(T, s) \not= \emptyset$ 
	materially implies 
	$\bigwedge \mathcal{I}(T, s') \in {\uparrow\! \{\bigwedge \mathcal{I}(T, s)\}} 
	\cup {\downarrow\! \{\bigwedge \mathcal{I}(T, s)\}}$ 
	(where necessarily ${\uparrow\! \{\bigwedge \mathcal{I}(T, s)\}} 
	\cup {\downarrow\! \{\bigwedge \mathcal{I}(T, s)\}} 
	\not= \emptyset$). \hfill$\spadesuit$ 
	%\cup {\uparrow\! \{\neg \bigwedge \mathcal{I}(\{t\}, s)\}}$. 
\end{definition}   

\begin{example}[The role of \pmb{aass} in formal fallacy identification] \rm \label{ex_role_aass} 
	As we saw in Section \ref{sec_understanding_the_semanics}, 
	\pmb{aass} delimits rhetorically reasonable attacks and 
	supports.  %For `$attack$', \pmb{aass} 
%	guards against a statement that restates   
%	the same effective aspect as a way to argue against 
%	a statement. 
	Suppose a discussion 
	about $t$: `{\it healthy lifestyle}' with two statements $s_1$ and $s_2$.  
	\begin{itemize} 
		\item $s_1$: 
			``{\it Eating vegetables and no processed 
			food is key.}"  
		\item $s_2$: ``{\it What? The key is to 
	consume vegetables and quit eating ready meals!}"  
	\end{itemize}  
	The rhetorical intention is such that $s_2$ intends to attack $s_1$.  
	The typed argumentation model of this argumentation 
	is therefore  %$s_2:\{t, t_2\} 
	\begin{tikzcd}[row sep=tiny,column sep={1.6cm}]   
		s_2:\{t\} 
		\arrow{r}[above]{\{\text{`}attack\text{'},t\}} 
		& s_1:\{t\}. 
%		& s_8:\{t_2,t_8\} \arrow{r}[above]{\{\text{`}attack\text{'},t_2\}} & s_7:\{t_2,t_7\}
%		 \arrow{r}[above]{\{\text{`}support\text{'}, t_2\}} & s_6:\{t_2,t_6\}.  
%		aCostH \arrow[-,ur]\arrow[-,dr] & 	  & \neg aCostH \arrow[-,ul]\arrow[-,dl] & bCostH \arrow[-,ur]\arrow[-,dr] & & \neg bCostH \arrow[-,ul]\arrow[-,dl]\\
%		& \pmb{0} & & & \pmb{0} \\ 
%                & \pmb{1} &   & & \pmb{1} \\ 
%		aP \arrow[-,ur]\arrow[-,dr] & 	  & \neg aP \arrow[-,ul]\arrow[-,dl] & bP \arrow[-,ur]\arrow[-,dr] & & \neg bP 
%		\arrow[-,ul]\arrow[-,dl]\\
%		& \pmb{0} & & & \pmb{0} 
	\end{tikzcd}
%	where $t_1$ denotes `Allen' and $t_2$ denotes `Bill'. 
	%	generalisation 
%	of individual core constraints 
%	 
%	\pmb{aass} alone 
	The {\it straw man} fallacy as is so called in the 
	repertoire of informally classified fallacies
	is such that, in attacking (resp. supporting) 
	a statement, 
	the attacker (resp. the supporter) attacks (resp. 
	supports) its oversimplification, exaggeration or misrepresentation 
	to appear defeating (resp. supporting) it.  
	Suppose $\mathcal{D}$ is large enough so any reasonable aspect of $t$ is 
	in $D$ including  
	$x$: `{\it eating vegetables}' and $y$: `{\it eating processed food}'. (Further detail of $\mathcal{D}, \mathcal{I}$ 
	is in \citep{Nakai22}.) 
	In case we treat ready meals and processed food synonymously in the context of 
	$t$, it is reasonable for $\mathcal{I}$ to satisfy 
	$\mathcal{I}(\{t\}, s_1) = \{x \wedge \neg y\}$ and 
	$\mathcal{I}(\{t\}, s_2) = \{x \wedge \neg y\}$.   
	Note that ``{\it What?}" touches upon no aspect of $t$ 
	and thus adds no aspect to $\mathcal{I}(\{t\}, s_2)$.    
	With respect to $\mathcal{D}, \mathcal{I}$,  
	we cay say the following: 
	$s_2$, by 
	appearing as if it is counterarguing  
	$s_1$, gives the illusion that 
	$s_1$ has an issue. 
%	actually restating the same point of view).    
	\pmb{aass} is not satisfied by this kind of argumentation. 
		
%	Suppose $\mathcal{D}$ is large enough so any reasonable aspect  
%	of $t$ is available including the following: 
%	\begin{itemize} 
%		\item 
%	\end{itemize} 
%
%	 
%	Let $\mathcal{I}$ be such that 
%	$\mathcal{I}(\{t,t_1\}, s_1) 
%		As for `$support$',  
	Suppose 
	another theme $t'$: `{\it How should we prepare 
	for emergencies?}' with two statements $s_3$ and $s_4$. 
	\begin{itemize} 
		\item $s_3$: ``{\it We need to keep the preserved food 
			in the storage.}'' 
		\item $s_4$: ``{\it As you say, absolutely, let's 
			clear the storage.}''
	\end{itemize} 
	The rhetorical intention is such that $s_4$ intends to support $s_3$. 
	The typed argumentation graph for this argumentation 
	is  therefore %$s_2:\{t, t_2\} 
	\begin{tikzcd}[row sep=tiny,column sep={1.6cm}]   
		s_4:\{t'\} 
		\arrow{r}[above]{\{\text{`}support\text{'},t'\}} 
		& s_3:\{t'\}. 
%		& s_8:\{t_2,t_8\} \arrow{r}[above]{\{\text{`}attack\text{'},t_2\}} & s_7:\{t_2,t_7\}
%		 \arrow{r}[above]{\{\text{`}support\text{'}, t_2\}} & s_6:\{t_2,t_6\}.  
%		aCostH \arrow[-,ur]\arrow[-,dr] & 	  & \neg aCostH \arrow[-,ul]\arrow[-,dl] & bCostH \arrow[-,ur]\arrow[-,dr] & & \neg bCostH \arrow[-,ul]\arrow[-,dl]\\
%		& \pmb{0} & & & \pmb{0} \\ 
%                & \pmb{1} &   & & \pmb{1} \\ 
%		aP \arrow[-,ur]\arrow[-,dr] & 	  & \neg aP \arrow[-,ul]\arrow[-,dl] & bP \arrow[-,ur]\arrow[-,dr] & & \neg bP 
%		\arrow[-,ul]\arrow[-,dl]\\
%		& \pmb{0} & & & \pmb{0} 
	\end{tikzcd}
	Suppose $\mathcal{D}$ is large enough also for the aspects of $t'$. 
	In particular, $x_1$ `{\it keeping preserved food in the storage}' and 
	$x_2$ `{\it clearing the storage}', %$x_3$ `{\it Allen}' 
%	and $x_4$ `{\it Bill}' 
	are included in $D$.\footnote{$x_1$ itself can be the infimum 
	or the supremum of other aspects of $D$. 
	That is not important here.}  
%	Since clearing the storage is a contrary of keeping something in the storage in the Aristotelian
%	sense of contrariness,  
	Suppose $\mathcal{D}$ reflects the relationship between $x_1$ and $x_2$ that $x_2$ 
	is a contrary of $x_1$ in the Aristotelian sense of contrariness, then $x_2$ is a member 
	of $\downarrow \{\neg x_1\}$ 
	  in $\mathcal{D}$. 
	Suppose $\mathcal{I}$ is such that $\mathcal{I}(\{t'\}, s_3) = \{x_1\}$
	and that $\mathcal{I}(\{t'\}, s_4) = \{x_2\}$. Then, \pmb{aass} is not satisfied.  
	%it is clear that the effective aspect of $\mathcal{I}(\{t'\}, s_4)$ is 
	%comparable with that of $\mathcal{I}(\{t'\}, s_4)$ $s_3$. 
	Hence, with respect to $\mathcal{D}, \mathcal{I}$, we can say that 
	$s_4$ only gives the illusion of supporting 
	$s_3$. This is the {\it straw man} once 
	again. %\pmb{aass} is not satisfied by these kinds of argumentation. 
	%\pmb{aass} guards against the opposite direction of misrepresentation, 
	%which is again a form of straw man fallacy. 
%	Alice states: ``{\it It's important to save money for emergencies,}'' 
%	to which Bob reponds: ``{\it Absolutely, Alice. We should spend everything we earn immediately, 
%	so we are ready for the future, as you rightly judge.}'' 
%	While Bob tries to mask his misrepresentation by slipping ``for the future'' into 
%	his statement to echo with ``for emergencies", on a closer look, Bob is clearly stating 
%	the oppostite of Alice's statement, only giving a pretense of support to his statement 
%	to misrepresent Alice's statement. 
%	With \pmb{aass} in force, these kinds can be formally identified fallacious, 
%	while, as discussed in Section \ref{sec_understanding_the_semanics}, 
%	not overly restricting the semantics of `$attack$' and `$support$'.  
 	\hfill$\clubsuit$ 
\end{example} 
\begin{note}[An important observation] \rm In \mbox{Example \ref{ex_role_aass}}, 
	we interpreted processed food and ready meals synonymously. 
	Other contexts may require differentiation of the two. %however. 
	%What if we are in the middle 
%	of a discussion where the difference in phrasing causes 
	%a difference? 
	%Then there is no longer any theoretical guarantee that $s_1$ and $s_2$ 
	%should be touching upon the same effective aspect in $\mathcal{D}$. 
%	paraphrases must touch upon 
%	exactly the same aspects. 
	Indeed, whether an argumentation is fallacious 
	can only be judged properly with respect to $\mathcal{D}$, $\mathcal{I}$. In the theme aspect argumentation model, 
	the semantic information is explictly accessible 
	from within the model and is thus explainable. 
	%5By contrast, %logic-based argumentation 
	%models (unstructured or structured) only keep 
	%an already interpreted 
	%argumentation that does not refer back to the 
	%rhetoric model of the original argumentation. 
	%Within these logic-based models, 
	%automatic propaganda and fallacy detection 
	%techniques in the natural language processing research 
	%\cite{Vorakitphan21b,Goffredo22} 
	%do not precisely explain the contextual information. 
	%which is accessible from 
	%within the theme aspect argumentation model. 
%	and this is a good incentive to opt for the classification of 
%	fallacies with formal constraints. 
%	
%	\ryuta{Need something more here.} 
	\hfill$\clubsuit$
%	must it be given another name? Axiomatic 
%	reconstruction may promote another way of classifying 
%	fallacies. 
\end{note} 

%\begin{note} \rm 
%	In the above illustration, there is no speaker type (such as Alice, Bob, {\it etc}) 
%	given to $s_1$ and $s_2$. This is intentional. As we stated at the beginning of this section, 
%	\pmb{aass} and other 
%	core constraints are meant to be taken collectively; individually, their coverage is narrower. 
%	Later in Section \ref{sec_application_fallacy_detection_and_prevention}, 
%	we revisit a generalised version of this and other examples. \hfill$\clubsuit$ 
%\end{note} 

\noindent \pmb{aass} forces the typed argumentation graph $\modell$ to not contain `$attack$' from a statement 
to itself.  
\begin{proposition}[Consequence of \pmb{aass}] \label{prop_consequence_aass}
  	Given a theme aspect argumentation model 
	$\langle \modell, \mathcal{D}, \mathcal{I} \rangle$,  
	if there is some $s \in Stmt[G]$ 
%	and some $t \in \pmb{Themes}$ 
	such that $(s, s) \in Rel[G]$ and $`attack$'$ \in \Pi((s, s))$ 
	hold, 
	 then $\langle \modell, \mathcal{D}, \mathcal{I} \rangle$ 
	 does not satisfy \pmb{aass}. 
\end{proposition}  
\vspace{-0.26cm} 
\textbf{Proof.} By the definition of \pmb{aass}. \hfill$\Box$ \\ 
%} 
%\begin{note}[   ] 
%
%\end{note} 

We may assign the following constraint purely on the typed 
argumentation graph $\modell$ to prevent this violation of \pmb{aass} 
from occurring. 
\begin{definition}[(Graphic) No self-attack constraint] \label{def_no_self_attack_constraint} \rm 
	Given a theme aspect argumentation model $\langle \modell, \mathcal{D}, \mathcal{I} \rangle$, we say the typed argumentation graph $\modell$ 
	satisfies constraint \pmb{nsa} (\pmb{n}o \pmb{s}elf-\pmb{a}ttack)  
	iff there is no $s \in Stmt[G]$ such that 
	$(s, s) \in Rel[G]$ and $`attack$'$ \in \Pi((s, s))$ 
	both hold. 
	\hfill$\spadesuit$
\end{definition} 

%\begin{note}[The role of \pmb{nsa} in formal fallacy identification] \rm 
%   
%
%	Suppose 
%\end{note} 

\begin{example}[The role of \pmb{nsa} in formal fallacy identification] \rm \label{ex_role_nsa} 
	Within the repertoire of informally 
	classified fallacies, there are 
	{\it self-refuting fallacies}. 
	Suppose Alice states: ``{\it I criticise every generalization.}'' 
	Since her statement is itself a generalisation, if she indeed intends 
	what she states, she is criticising her own statement. Hence, 
	the typed argumentation graph contains a node for her statement with an edge to itself. 
	\pmb{nsa} is not satisfied by this kind of typed argumentation graph. \hfill$\clubsuit$ 
%	guards against these kinds 
%	The main role of \pmb{nsa} is to guard against a statement that 
%	is rhetorically self-refuting fallacies such as: 
%	``{\it 
%
%
%	obviously  
%	 
%
%
%
%	 
%
%
%
%
%
%	 
%	
%	Self-attacking arguments constitute {\it self-refuting fallacies} within the informally 
%	classified fallacy repertoire such as 
%
%	Self-attacking arguments 
%	are already controversial in both abstract and logic-based argumentation 
%	formalisms, with many examples given in the literature. 
%	As an example, (Alice) ``{\it Nobody can know anything for sure}'' 
\end{example} 

It is natural to enforce containment of aspects of a statement for theme(s) within  
the aspects of the theme(s). %In the remaining of this section, all the illustrations 
%of the role of constraints will be around Alice's insurance. However,
%as we have shown, any other example may be similarly used. 

\begin{definition}[Inclusion constraint] \label{def_inclusion_constraint}  \rm 
	Given a theme aspect argumentation model $\langle \modell, \mathcal{D}, \mathcal{I} \rangle$, we say it satisfies constraint \pmb{i} (\pmb{i}nclusion) iff  
	the following holds for every $s \in Stmt[G]$ and 
	every $T \subseteq \pmb{Themes}$: $\mathcal{I}(T, s) \subseteq \mathcal{I}(T, \omega)$. \hfill$\spadesuit$
\end{definition}        

\begin{example}[The role of \pmb{i} in formal fallacy identification] \rm   \label{ex_role_i} 
	\pmb{i} guards against violation of the semantic boundary of 
	theme(s) such as may be exploited in {\it red herring} 
	in the repertoire of informally classified fallacies. 
	Suppose $t$: `{\it financial considerations when purchasing property insurance for 
	a summer house}' and two statements $s_1$ and $s_2$. 
	\begin{itemize} 
		\item $s_1$: ``{\it   Alice should buy property insurance for her summer house. It does not 
			cost much.}'' 
		\item $s_2$: ``{\it  Well, I disagree. 
	Her summer house is right next to a waterfall, attracting 
	a lot of mosquitoes. It is not worth purchasing it.}'' 
	\end{itemize}   
	Suppose $\mathcal{D}$ is large enough. 
	(The full detail of $\mathcal{D}, \mathcal{I}$ is in 
	\citep{Nakai22}; similarly 
	for all the other examples.) %includes a large number of aspects of various themes. 
	%So, any reasonable aspects for $t$ are available in $\mathcal{D}$, and $\mathcal{D}$ 
	%contains any reasonable aspects for other themes, too.  
	%Suppose Bob states: ``{\it Alice should buy property insurance for her summer house. 
	%It will not cost much.}'', 
	%to which Chris states: ``{\it Well, I diagree. 
	%Her summer house is right next to a waterfall, attracting 
	%a lot of mosquitoes. It is not worth purchasing it.}''   
        Some part 
	of each of the statements touches upon the theme (``{\it It does not cost much}" 
	and ``{\it It is not worth purchasing it}") and thus 
	upon aspects of $\mathcal{I}(\{t\}, \omega)$. 
	% (and thus upon its aspects in $\mathcal{D}$).  
	Both statements are typed with $t$. % since at least 
%	some part 
%	of each of the statements touches upon the theme (and thus its aspects in $\mathcal{D}$).  
	But here, while $s_1$ considers financial considerations, 
	$s_2$  alludes to mosquitoes as a concern against 
	insuring the summer house. 
	Since the proximity to a waterfall and the 
	presence of mosquitoes are not financial 
	considerations, these aspects 
	lie outside $\mathcal{I}(\{t\}, \omega)$ in $\mathcal{D}$. 
	With \pmb{i} in force, they will not be included 
	in $\mathcal{I}(\{t\}, s_2)$. %That is, with \pmb{i} in force,  
	%As far as $t$ is concerned, $s_2$ is interpreted as if it were 
	%``{\it Well, I disagree. It is not worth purchasing it.}''  
	%when \pmb{i} is in force. 
\hfill$\clubsuit$ 
\end{example}

Aspects of a statement should be 
determined only when the theme(s) its aspects 
are derived for are non-empty. 

\begin{definition}[Vacuous interpretation constraint]\label{def_vacuous_interpretation_constraint} \rm 
	Given a theme aspect argumentation model \linebreak 
	$\langle \modell, \mathcal{D}, \mathcal{I} \rangle$, 
	we say it satisfies constraint \pmb{vi} (\pmb{v}acuous \pmb{i}nterpretation) 
	iff  
	the following holds for every $s \in Stmt[G]$: 
	$\mathcal{I}(\emptyset, s) = \emptyset$. 
	\hfill$\spadesuit$
\end{definition} 
\begin{example}[The role of \pmb{vi} in formal fallacy identification] \rm   \label{ex_role_vi} 
%	\pmb{vi} reflects the standpoint that the semantics of a statement must be understood 
%	in the theme(s) it is interpreted.  
	As regards the role of \pmb{vi} in formal fallacy identification,   
	note that statements that do not fall under any theme of ongoing argumentation   
%	should not be considered providing any aspects to the ongoing 
%	argumentation. 
	%provided set of 
	%themes 
	engage in various fallacies such as {\it ad hominem}, 
	{\it red herring}, {\it non-sequitur}, 
	{\it tu quoque}, {\it appeal to emotion} 
	and so on. %and so on.  
	 % under a given theme. 
	%Such kinds of statements engage in 
%	fallacies bearing various terms such as {\it ad hominem},  
	Suppose $t$: `{\it Should Alice buy fire insurance 
	given the cost?}' %,  fire insurance 
	%worth buying?}' 
	and suppose 
	$\pmb{Themes} = \{t\}$. Suppose the following 
	argumentation where each of $s_2$ and $s_3$ %and $s_4$ 
	responds to $s_1$.  
	\begin{itemize} 
		\item $s_1$: ``{\it Alice should not buy fire insurance 
	for her summer house. It costs her a lot.}'' 
				\item $s_2$ (to $s_1$): ``{\it You are an idiot, Mister!}''  ({\it ad hominem})
\item $s_3$ (to $s_1$): ``{\it Right, what Alice should get is a security system which is 
	more important!}''   ({\it non-sequitur} or {\it red herring}) 
%\item $s_4$ (to $s_1$): ``{\it Well, no, think of the peace of mind Alice gets by knowing 
%	she is protected against unforseen disasters! There is nothing better than 
%	feeling secure and worry-free!}''  
	\end{itemize} 
	The rhetorical intention is such that $s_2$ intends to attack $s_1$
	while $s_3$ intends to support it. 
	However, neither of these statements falls under any theme  
	in $\pmb{Themes}$. Such statements are irrelevant to $\pmb{Themes}$ 
	and should not be considered providing any aspects. % to the ongoing 
%	argumentation. 
	With \pmb{vi} in force, it is not permitted to ascribe aspects 
	to any such statements.% in argumentation. 
	 \hfill$\clubsuit$ 
\end{example}

It is desirable 
to define the closure of aspects of themes by complement. However, it is actually also desirable 
to ensure that aspects of themes form a distributive lattice. 

\begin{definition}[Sub-boolean algebra constraint for aspects of themes]\label{def_sub_boolean_algebra_constraint} \rm 
	Given a theme aspect argumentation model $\langle \modell, \mathcal{D}, \mathcal{I} \rangle$, we say it satisfies constraint 
	\pmb{bat} (\pmb{b}oolean \pmb{a}lgebra for \pmb{t}hemes) 
	iff the following holds for every $T \subseteq \pmb{Themes}$: 
	$\mathcal{I}(T, \omega) \not= \emptyset$ 
	materially implies 
	that $(\mathcal{I}(T, \omega), \pmb{1}, \pmb{0}, \neg, 
	\wedge, \vee)$ is a 
	sub-complete Boolean algebra of $\mathcal{D}$.	\hfill$\spadesuit$
\end{definition} 

\begin{example}[The role of \pmb{bat} in formal fallacy identification] \label{ex_role_bat} \rm Without closure by complement, there can be some aspect $x$ which can be argued for in a theme (that is, $x$ is 
	an aspect 
	of the theme) which is, however, not allowed to be argued against in the same 
	theme, or vice versa, hindering free speech. Such purposeful 
	restriction, often imposed by biased moderators, leads 
	to fallacies known  as {\it suppressed evidence}, {\it one-sidedness}, {\it false dichotomy} 
	and {\it cherry-picking} 
	within the repertoire of informally classified fallacies.  
	As for the benefit of a distributive lattice, 
	 if some aspects $x$: ``{\it purchasing fire insurance}'' 
and $y$: ``{\it purchasing homeowners insurance}'' 
	are in $\mathcal{I}(T, \omega)$, 
then $x \vee y$: ``{\it purchasing fire insurance or  
	homeowners insurance}'' should be also in $\mathcal{I}(T, \omega)$, 
as well as $x \wedge y$ which is a more specific aspect than $x$ or $y$ alone that should be 
	discussable if $x$ and $y$ are. If such generalisation or specialisation is disallowed, 
	we will see {\it cherry-picking} or a 
	nuanced version of 
	 {\it false dichotomy} manipulating one into 
	 the belief that only presented options are available to choose from. %It is also 
%	 {\it cherry-picking}.  

	 Suppose $t$: `{\it Should 
	 Alice buy fire insurance for her house?}'  
	 Suppose $\mathcal{D}$ is large enough, 
	 then many reasonable aspects mutually incomparable in $\mathcal{D}$ exist for $t$ 
	 such as 
	 $x_1$: `{\it Alice}', $x_2$: `{\it house}', 
	$x_3$: `{\it purchasing 
	fire insurance}', $x_4$: `{\it costing a lot}' and so on. 
	 But now, the moderator wants  
	 a biased argumentation. In particular, 
	 he/she does not want to hear 
	 any positive opinion about Alice's purchase of fire insurance. 
	 As such,  he/she deliberately 
	 restricts the aspects of $t$, {\it i.e.} 
	 $\mathcal{I}(\{t\}, \omega)$, 
	 to - for example - the closure of $\{x_1, x_2, 
	 x_3, x_4\}$ by $\wedge$, $\vee$ 
	 and $\neg$ without, however, including any 
	 member of $\downarrow \{x_1 \wedge x_3\}$.    
	 Let us see what happens 
	 in the following argumentation with this bias 
	 in place. 
	 %, in his/her attempt 
	 %to manipulate the argumentation for $t$, decides  
	 %to prohibit any affirmative opinion 
	 %by restri
	 %\ryuta{This has to change.}  
%
%	 ka statement from 
%	 touching upon 
%	 enforce $\mathcal{I}(\{t\}, \omega)$ to be 
%	 closure of $\{\neg \textit{purchasing fire insurance}, \neg \textit{purchasing homeowners insurance}, 
%	 Alice, Bob, house, \textit{costing a lot}\}$ by supremum and infimum but not by complement. 
%	  and 
%	 the following argumentation. 
	 \begin{itemize} 
		 \item $s_1$: ``{\it Alice should not buy 
	 fire insurance for her house. It costs her a lot.}'' 
 		\item $s_2$: ``{\it  It is not as though  
	 fire insurance costs \$1,000 a year. 
	 She should definitely buy it for her house.}'' 
	 \end{itemize}
	The typed argumentation graph is 
	\begin{tikzcd}[row sep=tiny,column sep={1.6cm}]   
		s_2:\{t\} 
		\arrow{r}[above]{\{\text{`}attack\text{'},t\}} 
		& s_1:\{t\}. 
%		& s_8:\{t_2,t_8\} \arrow{r}[above]{\{\text{`}attack\text{'},t_2\}} & s_7:\{t_2,t_7\}
%		 \arrow{r}[above]{\{\text{`}support\text{'}, t_2\}} & s_6:\{t_2,t_6\}.  
%		aCostH \arrow[-,ur]\arrow[-,dr] & 	  & \neg aCostH \arrow[-,ul]\arrow[-,dl] & bCostH \arrow[-,ur]\arrow[-,dr] & & \neg bCostH \arrow[-,ul]\arrow[-,dl]\\
%		& \pmb{0} & & & \pmb{0} \\ 
%                & \pmb{1} &   & & \pmb{1} \\ 
%		aP \arrow[-,ur]\arrow[-,dr] & 	  & \neg aP \arrow[-,ul]\arrow[-,dl] & bP \arrow[-,ur]\arrow[-,dr] & & \neg bP 
%		\arrow[-,ul]\arrow[-,dl]\\
%		& \pmb{0} & & & \pmb{0} 
	\end{tikzcd}
	%Suppose $\mathcal{D}$ is large enough, then  
	%with aspects $x_1$ `{\it Alice}', $x_2$ `{\it house}', 
	%$x_3$ `{\it purchasing 
	%fire insurance}' and $x_4$ `{\it costing a lot}', 
	$\mathcal{I}$ can be such that  
	$\mathcal{I}(\{t\}, s_1) = 
	\{(x_1 \wedge x_2) \Rightarrow \neg (x_1 \wedge x_3), 
	(x_1 \wedge x_3) \Rightarrow (x_1 \wedge 
	x_4)\}$ 
	and $\mathcal{I}(\{t\}, s_2) = \{x_3 \Rightarrow \neg x_4, 
	(x_1 \wedge x_2) \Rightarrow (x_1 \wedge x_3)\}$ (\$1,000 a year 
	is interpreted as a lot of cost in $\mathcal{I}(\{t\}, s_2)$ 
	here\footnote{With more concrete an interpretation, say, $\mathcal{I}'$,  
	$\mathcal{I}'(\{t\}, s_2)$ could be 
	$\{x_3 \Rightarrow \neg x', (x_1 \wedge x_2) 
	\Rightarrow (x_1 \wedge x_3)\}$ 
	where $x'$ `{\it costing \$1,000 a year}' satisfies  $x' < x_4$ in $\mathcal{D}$.}).  
	Firstly, both members of $\mathcal{I}(\{t\}, s_1)$ are registered as aspects 
	of $t$. As for $s_2$, it appears reasonable to regard $s_2$ as  counterarguing $s_1$ directly. 
	However, since $\mathcal{I}(\{t\}, \omega)$ does not include $x_1 \wedge x_3$, 
	$s_2$ is regarded as touching upon 
	an aspect outside the theme. 
	\pmb{bat} is not satisfied when this kind of manipulation is applied to the interpretation function. 
	\hfill$\clubsuit$
\end{example}

Next, let us suppose an ordinary statement $a \in OStmt[G]$, and 
let us suppose some theme $t'$ which is not in $\Pi(a)$. Then, 
it is hypothetical to conceive aspects of $a$ for $t'$ 
(``Let us hypothetically assume 
that $a$ were stated for $t'$, then $a$ would represent such and such 
aspects of $t'$"). 
Such counterfactual interpretation can be useful for hypothetical argumentation.
(``Imagine you are to win the lottery...")
However, we are rather to 
prevent counterfactuals in this paper, 
so we enforce that $\mathcal{I}(T \cup \{t'\}, a)$ cannot be strictly 
greater than $\mathcal{I}(T, a)$, {\it i.e.} we define 
an upper bound for aspects of an ordinary statement $a$ based on $\Pi(a)$,
so that no unconnected theme will provide $a$ with further aspects. 
The condition for a pointer statement $t.x \in PStmt[G]$ where $x$ 
is either an ordinary statement or $\mathcal{C}$ is similar,  
except for the following difference. Even if, say, $t \not\in \Pi(t.a)$ (that is, 
$t.a$ is not explicitly expressed for $t$), as long as $t \in \Pi(a)$ holds,  
$t.a$'s aspects are obtainable for $t$. 
\begin{definition}[Proper range constraint]\label{def_proper_range_constraint} \rm 
	Given a theme aspect argumentation model $\langle \modell, \mathcal{D}, \mathcal{I} \rangle$, we say it 
	satisfies constraint \pmb{pr} (\pmb{p}roper \pmb{r}ange) 
	iff the following hold for every $s \in Stmt[G]$ and every $T \subseteq \pmb{Themes}$. 
	\begin{itemize}  
		\item Case $s \in OStmt[G]$:
			$\mathcal{I}(T, s) \subseteq \mathcal{I}(T \cap \Pi(s), s)$.
						\item Case $s \in PStmt[G]$ and $s$ is in the form 
			$t.a$ ($t \in \pmb{Themes}, a \in \pmb{A}^{ord}$): 
					{\small $\mathcal{I}(T, s) \subseteq 
					\mathcal{I}(T \cap (\Pi(s) 
					\cup \{t\}),
					a)$}. 
					\item Case $s \in PStmt[G]$ and $s$ is in 
			the form $t.\mathcal{C}$ ($t \in \pmb{Themes}$):  
					$\mathcal{I}(T, s) \subseteq 
					\mathcal{I}(T \cap (\Pi(s) 
					\cup \{t\}), t.\mathcal{C})$. \hfill$\spadesuit$
%				\item Case $s \in PStmt[G]$ and $s$ is in the form $t_1.pStmt$:  
%					$\mathcal{I}(T, s) = \emptyset$ if $T \cap \Pi(s) = \emptyset$ and 
%					($t_1 \not\in T$ or $\mathcal{I}(T, pStmt) = \emptyset$). 
				
	\end{itemize}
\end{definition}  
\begin{example}[The role of \pmb{pr} in formal fallacy identification]\label{ex_role_pr}   \rm 
	Hypotheticals are exploited in {\it conditional fallacy}, 
	so known 
	within the repertoire of informally classified fallacies, 
	making hypotheticals appear just as substantive as non-hypotheticals. 
	 %The conditional fallacy may be combined with 
%	{\it equivocation fallacy}, so known within the informally classified 
%	fallacy repertoire, manufacturing alternative meanings to given statements. \ryuta{Need 
%	to revise this part.} 
	% exploits hypothetical conditions.  
	Suppose $t$ is `{\it Should Alice buy fire insurance?}' and 
	$t'$ is `{\it What if fire insurance would cost just 
	50 cents per year?}'  
	Consider the following argumentation. 
	\begin{itemize} 
		\item $s_1$: ``{\it Alice should not buy fire insurance. 
	It costs her a lot.}'' 
		\item $s_2$: ``{\it 
	But what if the insurance would cost just 50 cents  
			per year? Then 
			she should be buying it.}''
		\item $s_3$: ``{\it Well, sure, 
			she should 
			be buying it in the hypothetical scenario.}'' 
		\item $s_4$: ``{\it See? Alice should buy fire 
			insurance! }''  
	\end{itemize}  
%	The rhetorical intention is such that   
%	$s_4$ intends to support $s_3$ which intends 
%	to support $s_2$ which intends to attack $s_1$. 
	%As such, 

	Given the rhetorical intention, the typed argumentation graph 
	for this argumentation is as follows. 
	\begin{tikzcd}[row sep=tiny,column sep={1.6cm}]   
		s_4:\{t\} 
		\arrow{r}[above]{\{\text{`}support\text{'},t\}} 
		& s_3:\{t,t'\}  
		\arrow{r}[above]{\{\text{`}support\text{'},t'\}} 
		& s_2:\{t,t'\} 
		\arrow{r}[above]{\{\text{`}attack\text{'},t\}} 
		& s_1:\{t\}. 
%		& s_8:\{t_2,t_8\} \arrow{r}[above]{\{\text{`}attack\text{'},t_2\}} & s_7:\{t_2,t_7\}
%		 \arrow{r}[above]{\{\text{`}support\text{'}, t_2\}} & s_6:\{t_2,t_6\}.  
%		aCostH \arrow[-,ur]\arrow[-,dr] & 	  & \neg aCostH \arrow[-,ul]\arrow[-,dl] & bCostH \arrow[-,ur]\arrow[-,dr] & & \neg bCostH \arrow[-,ul]\arrow[-,dl]\\
%		& \pmb{0} & & & \pmb{0} \\ 
%                & \pmb{1} &   & & \pmb{1} \\ 
%		aP \arrow[-,ur]\arrow[-,dr] & 	  & \neg aP \arrow[-,ul]\arrow[-,dl] & bP \arrow[-,ur]\arrow[-,dr] & & \neg bP 
%		\arrow[-,ul]\arrow[-,dl]\\
%		& \pmb{0} & & & \pmb{0} 
	\end{tikzcd}
		%includes a large number of aspects 
	%for all sorts of themes so that any reasonable aspects for $t$ and $t'$ are 
	%available in the 
	%semantic structure. 
%	such aspects `{\it purchasing fire insurance}', `{\it purchasing property insurance}', 
%	`{\it Alice}', `{\it Bob}', `{\it house}', `{\it costing a lot}' and so on, but also 
%	`{\it house browning}', `{\it window films}', `{\it UV resistent paint}' and so on.  
%	Now, suppose Bob states $s_1$: ``{\it Alice should not buy fire insurance for her summer house. 
%	It costs her a lot.}'', Chris states $s_2$: ``{\it 
%	But what if the insurance would cost just a fraction of the current price? Then 
%	the cost would be negligible, and she should 
%	really be buying it.}'', Bob states $s_3$: ``{\it Well, sure.}'', 
%	and Chris states $s_4$: ``{\it See, Alice should buy it, right?}''  
	Every node is typed with 
	$t$. There are, however, two nodes $s_2$ and $s_3$  
	that contain hypotheticals. Indeed, $s_2$ and $s_3$ are $t'.a$ for an ordinary statement $a$: ``{\it Alice should be buying 
	fire insurance.}'' 
	%Here, $\Pi(a) = \emptyset$ since $a$ has not been stated for $t$. 
	Suppose $\mathcal{D}$ is large enough but finite. 
	Let $x_1$ denote `{\it Alice}', 
	$x_2$ denote `{\it purchasing 
	fire insurance}' and $x_3$ denote `{\it costing a lot}'. 
	Suppose $\mathcal{I}$ 
	is such that $\mathcal{I}(\{t\}, s_1) = \{x_1 \Rightarrow 
	\neg (x_1 \wedge x_2), (x_1 \wedge x_2) 
	\Rightarrow (x_1 \wedge x_3)\}$.  
	Now, $t \not\in \Pi(t'.a)$ as we do not 
	take the hypothetical scenario seriously. However, 
	if we are to take hypoheticals just as seriously 
	and treat them as if contributing to the discussion on $t$,
	then 
	$\mathcal{I}(\{t\}, s_2)$ (and $\mathcal{I}(\{t\}, s_3)$) 
	would  
	include the aspect 
	$(x_2 \Rightarrow \neg x_3) \wedge (x_1 \Rightarrow (x_1 \wedge x_2))$. 
	Here, $\bigwedge \mathcal{I}(\{t\}, s_1) = \neg x_1 \vee \neg x_2 \vee x_3$, and $(x_2 \Rightarrow \neg x_3) \wedge (x_1 \Rightarrow (x_1 \wedge x_2))
	\equiv (\neg x_1 \wedge \neg x_2) \vee 
	(\neg x_1 \wedge \neg x_3) \vee (x_2 \wedge \neg x_3)$.
	By the isomorphism of a finite Boolean algebra to 
	a power set lattice (see section \ref{subsec_lattices_boolean_algebra}),
        $(\neg x_1 \wedge \neg x_2) \vee (\neg x_1 \wedge \neg x_3) 
	\vee (x_2 \wedge \neg x_3) \in {\uparrow \{\neg \bigwedge \mathcal{I}(\{t\},
	s_1)\}}$. Since the aspect from the hypothetical  
	in $s_2$ 
	would be a conditional contradiction of the effective aspect 
	of $s_1$ under $\{t\}$ forming a rhetorical-grade 
	attack on $s_1$ (see Section \ref{sec_understanding_the_semanics}), 
	$s_2$ might sound like genuinely attacking $s_1$. 
	From there, the argumentation could be directed to 
	the conclusion that Alice should buy fire insurance. 
%	which forms a rheotric-grade attack. 
%	which are the negative aspects of `{\it costing a lot}' and 
%	$\neg$ `{\it purchasing fire insurance}' included in $\mathcal{I}(\{t\}, s_1)$, 
%	sounding as if both Bob and Chris agreed Alice should buy fire insurance. 
	With \pmb{pr} in force, $\mathcal{I}(\{t\}, s_2)$ cannot 
	contain more aspects 
	than are included in $\mathcal{I}(\{t\} \cap (\Pi(s_2) \cup 
	\{t'\}), a) (= \mathcal{I}(\emptyset, a))$. %That is, the 
%	hypothetical 
%	adds no additional aspects, as required. 
\hfill$\clubsuit$  
	%With \pmb{vi} also in force, 
	%$\mathcal{I}(\{t\}, a) = \emptyset$, so $\mathcal{I}(\{t\}, s_2)$
%	. That is, $\Pi(a) = \{t\}$. 
%	Here, by `{\it fire insurance}' Bob means 
%	protection from flames. However, let's 
%	assume hypothetically that her summer house is in a very sunny location and hypothetically also that 
%	`{\it fire insurance}' means protection from sunburns. 
%	In the hypothetical situation, $\mathcal{I}(\{t'\},a)$ includes aspects related 
%	to sunburnt houses such as 
%	`{\it house browning}' and so on as well as those aspects related to 
%	sunburn protection such as `{\it window films}', `{\it UV resistent paint}' and so on. 
	%`{\it window films}', `{\it UV resistent paint}' and 
	%other aspects that relate to sunburns or 
\end{example} 

\noindent We obtain the following results. Firstly, when 
\pmb{pr} and \pmb{vi} are both presumed, no statements have 
any aspects for themes disconnected from them. 
\begin{proposition}[Consequence of \pmb{pr} and \pmb{vi}]\label{prop_consequence_pr_vi} 
	Given a theme aspect argumentation model $\langle \modell, \mathcal{D}, \mathcal{I} \rangle$, 
	that it satisfies \pmb{pr} together with \pmb{vi} enforces the following 
	for every $T \subseteq \pmb{Themes}$, every $t \in \pmb{Themes}$, every $a \in OStmt[G]$ and every $t.x \in 
	PStmt[G]$ ($x$ is $\mathcal{C}$ or an ordinary statement):  
	\begin{multicols}{2} 
	\begin{itemize} 
		\item  $\mathcal{I}(T, a) = \emptyset$  
			if $T \cap \Pi(a) = \emptyset$. 
		\item $\mathcal{I}(T, t.x) = \emptyset$  
			if $T \cap (\Pi(t.x) \cup \{t\}) = \emptyset$.  
	\end{itemize} 
	\end{multicols} 
\end{proposition}  
\vspace{-0.26cm} 
\textbf{Proof.} Trivial. \hfill$\Box$ \\ 

\noindent Thus, in the example given in Example \ref{ex_role_pr}, 
$\mathcal{I}(\{t\}, s_2)$ (and 
$\mathcal{I}(\{t\}, s_3)$) become $\emptyset$ 
if both \pmb{pr} and \pmb{vi} are in force. \\ 
%because    
%$\Pi(a) = \emptyset$ (that is, $a$ has not been stated for $t$ in the given argumentation) and so $\mathcal{I}(\{t\}, a) = \emptyset$.   \\

\noindent Presuming \pmb{pr}, \pmb{vi} and \pmb{i}, we obtain the following upper bounds 
for aspects of a statement. 
\begin{proposition}[Consequence of \pmb{pr}, \pmb{vi} and \pmb{i}]\label{prop_consequence_pr_vi_i} 
	Given a theme aspect argumentation model $\langle \modell, \mathcal{D}, \mathcal{I} \rangle$, that it satisfies \pmb{pr} together with \pmb{i} and \pmb{vi} 
	enforces 
	the following for every $T \subseteq \pmb{Themes}$, 
	every $t \in \pmb{Themes}$, every $a \in OStmt[G]$ and every 
	$t.a, t.\mathcal{C} \in PStmt[G]$: 
	\begin{itemize} 
		\item $\emptyset \subseteq \mathcal{I}(T, a) 
			\subseteq \mathcal{I}(T, \omega) 
			\cap \mathcal{I}(T \cap \Pi(a), a)$.  
		\item $\emptyset \subseteq \mathcal{I}(T, t.a) 
			\subseteq \mathcal{I}(T, \omega) 
			\cap \mathcal{I}(T \cap (\Pi(t.a) \cup \{t\}), a)$.  
		\item $\emptyset \subseteq \mathcal{I}(T, t.\mathcal{C}) 
			\subseteq \mathcal{I}(T, \omega) 
			\cap \mathcal{I}(T \cap (\Pi(t.\mathcal{C}) \cup \{t\}), t.\mathcal{C})$.
	\end{itemize}
\end{proposition}  
\textbf{Proof.} Trivial. \hfill$\Box$ \\ 

\noindent In relation to \pmb{aass}, 
the following results hold for 
\pmb{pr} and \pmb{vi}.  
When a typed argumentation graph $\modell$  satisfies both \pmb{tr} and \pmb{nnp}, 
it is impossible to find any $\mathcal{D}$ and $\mathcal{I}$  
satisfying the following condition: 
$\modell$, $\mathcal{D}$ and $\mathcal{I}$  
form a theme aspect argumentation model $\taam$ and 
$\taam$ fails to satisfy \pmb{pr}, \pmb{vi} and \pmb{aass}. 
However, this property ceases to hold in case 
$\modell$ does not satisfy either \pmb{tr} or \pmb{nnp}. 

\begin{theorem}[{\hyperref[proof_consequence_pr_vi_aass]{Consequence of \pmb{pr}, \pmb{vi} and \pmb{aass}}}] \label{thm_consequence_pr_vi_aass} 
	\begin{enumerate}
		\item 
		There exists some $\modell$ such that 
		it does not satisfy \pmb{tr} and 
			that 
			the following holds for every $\mathcal{D}, 
			\mathcal{I}$: that $\modell$, 
			$\mathcal{D}$ and $\mathcal{I}$ form 
			a theme aspect argumentation model 
			$\langle \modell, \mathcal{D}, 
	\mathcal{I}\rangle$ materially implies that $\taam$ does not satisfy \pmb{pr}, \pmb{vi} 
			and \pmb{aass} simultaneously. 
%	iff $\langle \modell, \mathcal{D}, 
%	\mathcal{I}\rangle$ does not satisfy \pmb{aass}. 
\item There exists some $\modell$ such that 
		it does not satisfy \pmb{nnp} and 
			that 
	the following holds 
			for every $\mathcal{D}, 
			\mathcal{I}$: that   
			$\modell$, 
			$\mathcal{D}$ and $\mathcal{I}$ form 
			a theme aspect argumentation model materially implies that 
			$\langle \modell, \mathcal{D}, 
	\mathcal{I}\rangle$ does not satisfy \pmb{pr}, \pmb{vi}, 
			and  \pmb{aass} simultaneously. 
	%iff $\langle \modell, \mathcal{D}, 
	%\mathcal{I}\rangle$ does not satisfy \pmb{aass}.  
\item 	There is no $\modell$ such that 
	it satisfies \pmb{tr} and \pmb{nnp} and that 
			the following holds for every $\mathcal{D}, \mathcal{I}$: 
			that  
			$\modell$, 
			$\mathcal{D}$ and $\mathcal{I}$ form 
			a theme aspect argumentation model 
			$\langle \modell, \mathcal{D}, 
			\mathcal{I}\rangle$ materially implies that $\taam$ does not satisfy 
			\pmb{pr}, \pmb{vi} or \pmb{aass}. 
	\end{enumerate} 
\end{theorem} 
%\textbf{Proof.}  See Appendix:  proofs. \hfill$\Box$ \\ 
\begin{note}[The point of this consequence]\label{note_point_this_consequence} \rm 
Just like the relationship between 
proof-theoretical constraints and constraints on semantics 
in formal logic, we may be interested in learning  
whether some constraints on the typed argumentation 
	graph (corresponding to the proof-theoretical constraints in formal logic) guarantee satisfaction of constraints on 
the theme aspect argumentation model (corresponding to 
the semantic constraints in formal logic). 
The significance of the correspondence is 
in being able to ensure theoretical properties 
in one of them by just ensuring those on the other.  
	There is also an implication to computation which 
	we will visit later in Section 
	\ref{sec_computational_complexities}, specifically in 
	Theorem \ref{thm_model_existence_model_checking}.  
	\hfill$\clubsuit$ 
\end{note} 

Next, we observe two monotonicity constraints. Aspects of a given set of 
theme(s) should be still in a larger set of theme(s), 
which gives rise to the first monotonicity constraint. 
 
\begin{definition}[Monotone aspects of themes constraint] \label{def_monotone_aspects_themes_constraint} \rm 
	Given a theme aspect argumentation model $\langle \modell, \mathcal{D}, \mathcal{I} \rangle$, we say it satisfies 
	constraint \pmb{mat} (\pmb{m}onotone \pmb{a}spects of \pmb{t}hemes)  
	iff the following holds for every $T_1 \subseteq T_2 \subseteq \pmb{Themes}$ 
	and every $x \in \mathcal{I}(T_1, \omega)$: $x \in \mathcal{I}(T_2, \omega)$.  \hfill$\spadesuit$
\end{definition} 

\begin{example}[The role of \pmb{mat} in formal fallacy identification] \label{ex_role_mat}  \rm 
		%{\it Special pleading}, 
	%so known within the informally classified fallacy repertoire, 
	%arises from double standards, applying some principles to some situations while 
	%excluding them in other situations.  
	Suppose two themes $t$: `{\it Should Alice buy insurance for her summer house?}' and 
	$t'$: `{\it Should Alice put physical security measures for her summer house?}' 
%	and $t''$: `{\it Should Alice pay for maintenance and repairs of her summer house?}'.  
	Suppose $\mathcal{D}$ is large enough. %(but finite).  
	Suppose that the following hold. 
	\begin{itemize} 
	      \item `{\it Alice}', `{\it summer house}' and their negative aspects 
			are shared between 
	$\mathcal{I}(\{t\}, \omega)$ and $\mathcal{I}(\{t'\}, \omega)$. 
	%and $\mathcal{I}(\{t''\}, \omega)$. 
			Any supremum and infimum of the shared aspects 
	are also shared between them. 
		\item `{\it purchasing fire insurance}', `{\it purchasing homeowners insurance}', 
			`{\it purchasing insurance}' and their 
			negative aspects are in $\mathcal{I}(\{t\}, \omega)$ but not in 
			$\mathcal{I}(\{t'\}, \omega)$. 
			Any infimum and supremum of $\mathcal{I}(\{t\}, \omega)$-specific 
			aspects 
			except for $\pmb{0}$ and $\pmb{1}$ 
			are also $\mathcal{I}(\{t\}, \omega)$-specific. Note that 
			fire insurance and homeowners insurance are two types of insurance. 
			It is assumed that `{\it purchasing insurance}' is a greater element 
			 of both `{\it purchasing fire insurance}' and 
			`{\it purchasing homeowners insurance}' in $D$. 
			%to $\mathcal{I}(\{t\}, \omega)$. 
		\item `{\it installing security cameras}', `{\it using reinforced windows}', 
			`{\it enforcing physical security measures}' 
			and their negative aspects are 
			in $\mathcal{I}(\{t'\}, \omega)$ but not in $\mathcal{I}(\{t\}, \omega)$. 
			Any infimum and supremum of $\mathcal{I}(\{t'\}, \omega)$-specific 
			aspects 
			except for $\pmb{0}$ and $\pmb{1}$ 
			are also $\mathcal{I}(\{t'\}, \omega)$-specific. 
			Naturally, `{\it enforcing physical security measures}' is a greater element 
			of `{\it installing security cameras}' and `{\it using reinforced windows}' 
			in $D$. 
%		\item `{\it roof maintainance}', `{\it regular inspections for mold}' 
	\end{itemize} 
	A moderator initiates discussion on $\{t, t'\}$ (`{\it Should Alice buy insurance 
	and/or put physical security measures for her summer house?}').  
	While he/she allows participants to discuss physical measures in greater detail, 
	he/she does not want to hear any particular of insurance. 
	Hence, even though aspects such as 
	 `{\it purchasing fire insurance}', `{\it purchasing homeowners insurance}' and so on 
	 can be discussed for $t$, 
	 he/she insists that they not be discussed 
	for $\{t, t'\}$ (that is, that they not be in 
	$\mathcal{I}(\{t, t'\}, \omega)$). This kind of manipulation 
	leads to unsatisfaction of \pmb{mat}. 
	Deliberate restriction   
	on what can be discussed and what cannot be forms 
	{\it cherry-picking} (see also Example \ref{ex_role_bat}). Application of different standards to different discussion themes forms 
	{\it special pleading}. 
	If a moderator needs to reflect his/her preference, 
	a sensible way would have been for him/her to state the condition at the front 
	by 
	having $t_1$: `{\it Should Alice buy insurance for 
	her summer house? (Particular types of insurance need not be discussed)}' instead of $t$. 
	If $t$ is chosen as a theme, discussion on anything touching upon aspects in $\mathcal{I}(\{t\}, \omega)$ 
	should be permitted. \pmb{mat} enforces this requirement. \hfill$\clubsuit$  
\end{example}

 Also, for any statement that is not a pointer 
 to \pmb{s}ummary of a theme its aspects should still be 
 in a larger set of themes. On the other hand, 
 \pmb{s}ummary of a theme depends on how the summarisation 
 is done, which is up to the theme(s) 
 that is interpreting it. It is generally non-monotonic. 
 Consequently, monotonicity does not extend to statements that point to \pmb{s}ummary of a theme. 
\begin{definition}[Monotone aspects constraint for non-\pmb{s}ummary statements]\label{def_monotone_aspects_constraint_non_summary_statements}  \rm 
    	Given a theme aspect argumentation model $\langle \modell, \mathcal{D}, \mathcal{I} \rangle$, we say it satisfies 
	constraint \pmb{manss} (\pmb{m}onotone \pmb{a}spects of \pmb{n}on-\pmb{s}ummary
	 \pmb{s}tatements)  
	iff the following holds for every $T_1 \subseteq T_2 \subseteq \pmb{Themes}$, 
	every $s \in Stmt[G]$ that does not point to \pmb{s}ummary and 
	every $x \in \mathcal{I}(T_1, s)$: 
	$x \in \mathcal{I}(T_2, s)$.  
	\hfill$\spadesuit$
\end{definition} 
Proposition \ref{prop_consequence_pr_vi} saw a property around 
the smallest set of aspects (= $\emptyset$) of a statement.  
The two constraints \pmb{pr} and \pmb{manss} ensure no influence 
of disconnected themes on aspects of ordinary statements. %See Appendix: proofs for the proof. 
\begin{proposition}[{\hyperref[proof_consequence_pr_manss]{Consequence of \pmb{pr} and \pmb{manss}}}]\label{prop_consequence_pr_manss} 
	Given a theme aspect argumentation model $\langle \modell, \mathcal{D}, \mathcal{I} \rangle$, 
	that it satisfies \pmb{pr} and \pmb{manss} enforces the following 
	for every $T \subseteq \pmb{Themes}$ and every $a \in OStmt[G]$: 
%	and every $t_1.a_1 \in PStmt[G]$: 
	${\mathcal{I}(T, a)} = {\mathcal{I}(T \cup T', a)}$ 
	for every $T' \cap \Pi(a) = \emptyset$. 
%	and ${\mathcal{I}(T, t_1.a_1)} = {\mathcal{I}(T \cup T', t_1.a_1)}$ 
%	for: every $T' \cap \Pi(t_1.a_1) = \emptyset$ 
%	if $t_1 \in \Pi(t_1.a_1)$; 
%	and every $(T' \cup \{t_1\}) \cap \Pi(t_1.a_1) = \emptyset$
%	if $t_1 \not\in \Pi(t_1.a_1)$. 
\end{proposition}    

 \noindent We also obtain  that a pointer statement referring to 
an ordinary statement, $t.a_1$, can touch upon an aspect in $T \subseteq \pmb{Themes}$ only when 
$a_1$ touches upon some aspects in $T$.  
%See Appendix: proofs for the proof. 
\begin{lemma}[{\hyperref[proof_consequence_pr_manss_2]{Consequence of \pmb{pr} and \pmb{manss} 2}}] \label{lem_consequence_pr_manss_2}  
   Given a theme aspect argumentation model $\langle \modell, \mathcal{D}, \mathcal{I} \rangle$, 
	that it satisfies \pmb{pr} and \pmb{manss} enforces the following 
	for every $T \subseteq \pmb{Themes}$, every $t \in \pmb{Themes}$, every $a \in OStmt[G]$ and every $t.a \in 
	PStmt[G]$: $\mathcal{I}(T, a) = \emptyset$ 
	materially implies $\mathcal{I}(T, t.a) = \emptyset$. 
\end{lemma} 
%\textbf{Proof.} See Appendix:  proofs. \hfill$\Box$ \\ 

Thus, we obtain the following consequence of \pmb{pr}, \pmb{vi} and \pmb{manss}. %See Appendix: proofs for the proof. 
\begin{proposition}[{\hyperref[proof_consequence_pr_vi_manss]{Consequence of \pmb{pr}, \pmb{vi} and \pmb{manss}}}]\label{prop_consequence_pr_vi_manss} 
     Given a theme aspect argumentation model $\langle \modell, \mathcal{D}, \mathcal{I} \rangle$, 
	that it satisfies \pmb{pr}, \pmb{vi} and \pmb{manss} enforces the following 
	for every $T \subseteq \pmb{Themes}$, every $t \in \pmb{Themes}$, every $a \in OStmt[G]$ and every $t.a \in 
	PStmt[G]$: $\Pi(a) \cap T = \emptyset$ materially implies $\mathcal{I}(T, t.a) = \emptyset$. 
\end{proposition}

\noindent The 3 constraints \pmb{pr}, \pmb{vi} and \pmb{manss} have an implication to \pmb{aass}. %See Appendix: proofs for the proof.
\begin{proposition}[{\hyperref[proof_consequence_pr_vi_manss_aass]{Consequence of \pmb{pr}, \pmb{vi} and \pmb{manss} to \pmb{aass}}}]\label{prop_consequence_pr_vi_manss_aass}
	There exists some $\taam$ such that it satisfies \pmb{pr}, \pmb{vi} and \pmb{manss} only if it does not satisfy \pmb{aass}. 
\end{proposition} 
%\textbf{Proof.} See Appendix:  proofs. \hfill$\Box$\\ 

\noindent The following constraint purely on the typed argumentation graph $\modell$ is a guard against this 
conflict between \pmb{aass} and the three constraints \pmb{pr}, \pmb{vi} and \pmb{manss}.  
It enforces that a referencing to an ordinary statement from a theme can only be made when 
the ordinary statement is in the theme. 
\begin{definition}[(Graphic) Known ordinary statement constraint]\label{def_known_ordinary_statement_constraint} \rm 
	Given a typed argumentation graph $\modell$, we say it satisfies constraint \pmb{kos} (\pmb{k}nown \pmb{o}rdinary \pmb{s}tatement) 
	 iff the following holds for any $t \in \pmb{Themes}$, any $a \in OStmt[G]$ and any $s \in Stmt[G]$: 
	 $s \equiv t.a$ materially implies  
	 $\Pi(t.a) \subseteq \Pi(a)$. 
	 \hfill$\spadesuit$
\end{definition}  

\begin{example}[The role of \pmb{kos} in formal fallacy identification] \label{ex_role_kos} \rm   
	Fallacies such as {\it appeal to authority}, so known 
	within the repertoire of informally classified fallacies, 
	 may resort to technical jargon not shared among discussion participants.  
	With this type of fallacy, a speaker tries to position himself/herself as an expert and 
	tries to attach credibility to his/her statement.  
	Suppose two themes $t$: `{\it Should Alice buy fire insurance for her summer house?}' 
	and $t'$: `{\it advanced architectural and construction principles}'. 
	Suppose any advanced architectural and construction principles 
	are mentioned in $t'$ including the following statement $a$: 
	``{\it A smart home implementing a zoned HVAC system with thermal sensor capabilities 
	  can automatically shut off zones with heat anomalies.}''  
	  Suppose the following argumentation for $t$: 
	  \begin{itemize} 
		  \item $s_1$: ``{\it Alice should not buy fire insurance for her summer house.
			  It does not anyway prevent fire.}"  
		  \item $s_2$: ``{\it Right, no need. 
				According to advanced architectural and construction principles, 
	a smart home implementing a zoned HVAC system with thermal sensor capabilities 
	can automatically shut off zones with heat anomalies.
			  }'' 
	  \end{itemize} 
%	Suppose, in $t$, Bob states $s_1$: ``{\it Alice should not buy 
%	fire insurance for her summer house. It does not anyway prevent fire.}'', 
%	and Chris states $s_2$: ``{\it Right, no need. 
%	According to advanced architectural and construction principles, 
%	a smart home implementing a zoned HVAC system with thermal sensor capabilities 
%	can automatically shut off zones with heat anomalies.}'' 
	The rhetorical intention is such that $s_2$ intends to support 
	$s_1$ by citing 
	$a$ in $t$. Thus, $s_2$ is $t.a$. However, $a$ is unknown within $t$ and, in this example, forms
	{\it appeal to authority} and possibly also {\it red herring}. %in this example.  
	Abrupt reference to an unknown statement does not satisfy \pmb{kos}. 
	 \hfill$\clubsuit$ 
%	Within the informally classified fallacy repertoire, there is a fallacy known as 
%	{\it quoting out of context} which occurs when a statement is extracted from 
%	its original context and used in another context without, however, sharing the original context.  
%	It is used to mislead discussion participants or to cause misunderstanding among them. 
%	Suppose two themes $t$: `{\it an earlier discussion}' 
%	and $t'$: `{\it How should Alice best maintain her summer house?}'  
%	%Suppose the complete Boolean algebra includes a large number of aspects 
%	%kso that any reasonable aspects for $t$ and $t'$ are available in $\mathcal{D}$. 
%	Suppose there was the following in $t$. 
%	Bob stated $s_1$: ``{\it Alice should buy fire insurance for her summer house.}'' 
%	Chris then stated $s_2$: ``{\it Sure. The cost of fire insurance will not be much.}''  
%	%The rhetoric is such that $s_2$ intends to support $s_1$. 
%	%Hence, the typed argumentation graph 
%	Let's say both statements are typed with $t$. 
%	 
%	Now, suppose the following in $t'$. 
%	David states $s_3$: ``{\it It is essential to consider regular inspections and timely repairs 
%	to keep her summer house in good condition.}''. 
%	Erica states $s_4$: ``{\it Well, as they argued in an earlier discussion, Alice should buy fire insurance 
%	for her summer house.}'' Let's say both statements are typed with $t'$. Erica's statement 
%	refers to $s_2$ in $t$, {\it i.e.} $s_4$ is $t.s_2$.  
%
%	Since $s_2$ itself has not been expressed beforehand in $t'$,  
%
%	 
%%	David states ``{\it Come on, that is a totally different context.} 
%	
\end{example} 

%\ryuta{
%\begin{note}[Intuitive interpretation]   \rm 
%	With \pmb{kos}, for every theme $t$, if there is any $t'.a$ in $t$, there must be 
%	also $a$ in $t$. An epistemic interpretation is intuitive. Say $t'$ is a theme: ``{\it should we buy a property insurance?}'' 
%	$t$ is another theme: ``{\it Is there any benefit in a property insurance?}'' 
%\end{note}  
%%}

For aspect consistency and non-triviality, it is natural to filter out any statement that effectively touches upon an inconsistent aspect \pmb{0} 
or a trivial aspect \pmb{1} 
for every theme of its. 
\begin{definition}[Substantial statement constraint]\label{def_substantial_statement_constraint} \rm 
	Given a theme aspect argumentation model\linebreak $\langle \modell, \mathcal{D}, \mathcal{I} \rangle$, 
	we say it satisfies constraint \pmb{ss} (\pmb{s}ubstantial \pmb{s}tatement) iff 
	the following hold for any $s \in Stmt[G]$ 
	and any $T \subseteq \Pi(s)$: $\mathcal{I}(T, s) \not= \emptyset$ 
	materially implies both $\pmb{0} \not= \bigwedge \mathcal{I}(T, s)$ and $\pmb{1} \not= \bigwedge \mathcal{I}(T, s)$.  
	\hfill$\spadesuit$
\end{definition} 

\begin{example}[The role of \pmb{ss} in formal fallacy identification]\rm \label{ex_role_ss} 
     Suppose the theme $t$: `{\it wage}' and the following argumentation. 
	\begin{itemize} 
		\item $s_1$: ``{\it Increasing the minimum wage should be beneficial for low-income workers.}''
		\item $s_2$ (to $s_1$): ``{\it Oh come on... if you increase the amount of money they get, they get more money, 
	that is so obvious!}''
\item $s_3$ (to $s_1$): ``{\it Absolutely. If they earn more money, they will have more money 
	in their pockets.}''
	\end{itemize} 
%     Alice states: ``{\it Increasing the minimum wage should be beneficial for low-income workers,}''
%	Bob states: ``{\it Oh come on... if you increase the amount of money someone gets, they get more money, 
%	that is so obvious!}'' 
%	and Chris states: ``{\it Well, Alice is right. If workers earn more money, they will have more money 
%	in their pockets.}''
%	Everyone knows eating in moderation doesn't work and you need strict diets.}''   
	The rhetorical intention is such that $s_2$ intends to attack $s_1$ 
	and that $s_3$ intends to support $s_1$. The typed 
	argumentation graph is therefore:  
	\begin{tikzcd}[row sep=tiny,column sep={1.6cm}]   
		s_2:\{t\} 
		\arrow{r}[above]{\{\text{`}attack\text{'},t\}} 
		& s_1:\{t\} 		
		& s_3:\{t\} \arrow{l}[above]{\{\text{`}support\text{'},t\}}. 
%		& s_8:\{t_2,t_8\} \arrow{r}[above]{\{\text{`}attack\text{'},t_2\}} & s_7:\{t_2,t_7\}
%		 \arrow{r}[above]{\{\text{`}support\text{'}, t_2\}} & s_6:\{t_2,t_6\}.  
%		aCostH \arrow[-,ur]\arrow[-,dr] & 	  & \neg aCostH \arrow[-,ul]\arrow[-,dl] & bCostH \arrow[-,ur]\arrow[-,dr] & & \neg bCostH \arrow[-,ul]\arrow[-,dl]\\
%		& \pmb{0} & & & \pmb{0} \\ 
%                & \pmb{1} &   & & \pmb{1} \\ 
%		aP \arrow[-,ur]\arrow[-,dr] & 	  & \neg aP \arrow[-,ul]\arrow[-,dl] & bP \arrow[-,ur]\arrow[-,dr] & & \neg bP 
%		\arrow[-,ul]\arrow[-,dl]\\
%		& \pmb{0} & & & \pmb{0} 
	\end{tikzcd}
%	and that Chris intends to support it.  
	Suppose $\mathcal{D}$ is large enough, %(but finite),  
	and suppose $x$ denotes `{\it more money 
	for low-income workers}' in $\mathcal{D}$. 
	$\mathcal{I}$ is such that 
	$\mathcal{I}(\{t\}, s_2) = \{x \Rightarrow x\}$ 
	and $\mathcal{I}(\{t\}, s_3) = \{x \Rightarrow x\}$.  
	With respect to $\mathcal{D}, \mathcal{I}$, 
	$\mathcal{I}(\{t\},s_1)$ and $\mathcal{I}(\{t\}, s_2)$ 
	are both tautological and 
	neither $s_2$ nor $s_3$ consequently provides 
	any meaningful critique 
	or support, engaging in 
	{\it begging the question} or {\it circular reasoning} as is 
	so called within the repertoire of informally 
	classified fallacies. 
	For the same theme $t$ and to the same statement $s_1$, 
	suppose the following two statements respond to it. 
	\begin{itemize} 
		\item $s_4$ (to $s_1$):  ``{\it No way! Paying 
			them more will force companies to hire 
	fewer workers. Look, we are currently paying more, 
			seeing more and more workers 
			getting hired as the consequence! 
			 %Further, they will also 
		%	hire more workers as better pay 
		%	attracts better workers.}''
			}''
\item $s_5$ (to $s_1$): ``{\it 
	That's beneficial all around!
	When you pay workers more,
	they will be happy working fewer hours. 
			Look at the situation now for the evidence: 
			we are paying them more, 
			and they are 
			working a lot more to earn even more!}''  
	\end{itemize} 
	% and Chris states: ``{\it Well, Alice is right. When you pay people more,
	%they can finally afford to work fewer hours. But on the flip side, with better pay, they 
	%will work even more hours to earn even more. That's beneficial all around!}''  
	The typed argumentation graph is: 
	\begin{tikzcd}[row sep=tiny,column sep={1.6cm}]   
		s_4:\{t\} 
		\arrow{r}[above]{\{\text{`}attack\text{'},t\}} 
		& s_1:\{t\} 		
		& s_5:\{t\} \arrow{l}[above]{\{\text{`}support\text{'},t\}} 
%		& s_8:\{t_2,t_8\} \arrow{r}[above]{\{\text{`}attack\text{'},t_2\}} & s_7:\{t_2,t_7\}
%		 \arrow{r}[above]{\{\text{`}support\text{'}, t_2\}} & s_6:\{t_2,t_6\}.  
%		aCostH \arrow[-,ur]\arrow[-,dr] & 	  & \neg aCostH \arrow[-,ul]\arrow[-,dl] & bCostH \arrow[-,ur]\arrow[-,dr] & & \neg bCostH \arrow[-,ul]\arrow[-,dl]\\
%		& \pmb{0} & & & \pmb{0} \\ 
%                & \pmb{1} &   & & \pmb{1} \\ 
%		aP \arrow[-,ur]\arrow[-,dr] & 	  & \neg aP \arrow[-,ul]\arrow[-,dl] & bP \arrow[-,ur]\arrow[-,dr] & & \neg bP 
%		\arrow[-,ul]\arrow[-,dl]\\
%		& \pmb{0} & & & \pmb{0} 
	\end{tikzcd}  
	Let $y_1$ denote `{\it companies hiring more workers}' 
	and $y_2$ denote `{\it workers to work more hours}'. 
%	$y_2$ denote 
	These statements can be interpreted 
	as follows. $\mathcal{I}_1(\{t\}, s_4) 
	=  \{x \Rightarrow \neg y_1, x \wedge y_1\}$ 
	and $\mathcal{I}_1(\{t\}, s_5) = \{x \Rightarrow \neg y_2,
	x \wedge y_2\}$. 
%	A reasonable interpretation $\mathcal{I}_1$ is:  
%	$\mathcal{I}_1(\{t\}, s_4) = \{x \Rightarrow \neg y_1, 
%	x \Rightarrow y_1\}$ and 
%	that $\mathcal{I}_1(\{t\}, s_5) = \{x \Rightarrow \neg y_2, 
%	x \Rightarrow y_2\}$. 
	However, with respect to $\mathcal{D}$ and $\mathcal{I}_1$, 
	both $s_4$ and $s_5$ are inconsistent, 
	since $\bigwedge \mathcal{I}_1(\{t\}, s_4) = 
	\bigwedge \mathcal{I}_1(\{t\}, s_5) = \emptyset$. 
	They are thus engaging 
	in {\it self-contradiction} and possibly also 
	a {\it straw man}.  
	These kinds of argumentation do not satisfy \pmb{ss}. 
%	With \pmb{ss} in force, these kinds can 
%	be formally identified fallacious. 
%	 the minimum wage, 
%	workers have more motivation to spend. Also, p}'' 
	%The typed argumentation graph 
	%is such that it has two nodes typed with the theme and an edge from Bob's statement to Alice's statement 
	%typed with `$attack$' and the theme. 
	%Here, Bob's statement is tautological. 
	 \hfill$\clubsuit$ 
\end{example} 

\begin{note}\rm In deductive argumentation \cite{Besnard01}, 
         deductive argumentation arguments with inconsistent premises or an inconsistent conclusion are prohibited 
	(see section \ref{subsec_logic_based_argumentation}) but not those with tautological 
	premises or a tautological conclusion. 
	As indicated in Example \ref{ex_role_ss}, however, 
	so long as formal fallacy identification is concerned, 
	it is not necessarily desirable to permit them. 
	%for formal fallacy identification. 
%	both contribute to fallacious reasoning and are thus 
%	prevented by \pmb{ss}. 
	%Thus, \pmb{ss} is a core constraint 
%	in the theme aspect argumentation model. %they are 
	\hfill$\clubsuit$ 
\end{note}

Let us call all these constraints 
except for those purely on the typed argumentation graph $\modell$ (which are: \pmb{tr}, \pmb{nnp}, \pmb{nsa} and \pmb{kos}) 
as \pmb{Core} (which is a collection of \pmb{aass}, \pmb{i}, \pmb{vi}, \pmb{bat}, \pmb{pr}, \pmb{mat}, \pmb{manss}, and \pmb{ss}). For \pmb{Core} 
to be meaningful, all its constituents should be simultaneously satisfiable.  
Starting from a simpler one, we have the following.  
%See Appendix: proofs for the proof. 
%the following result.
\begin{theorem}[{\hyperref[proof_existence]{Existence}}] \label{thm_existence}
	There is some $\taam$ satisfying \pmb{Core}. 
\end{theorem} 
%\textbf{Proof.} See Appendix:  proofs. \hfill$\Box$\\ 
%A simpler result would be existence of some $\langle \modell, \mathcal{D}, 
%\mathcal{I} \rangle$ that satisfies \pmb{Core}. \mbox{Theorem \ref{thm_core}} below 

\noindent A more non-trivial result is: for any typed argumentation graph $\modell$ satisfying  \pmb{tr}, \pmb{nnp}, \pmb{nsa},  
and \pmb{kos}, 
there is always some $\mathcal{D}, \mathcal{I}$ such that 
$\taam$ is a theme aspect argumentation model and that 
$\taam$ satisfies \pmb{Core}. %If we should treat $\modell$ as a syntax, 
%this would be the model existence theorem in logic, that every consistent 
%theorem has a model, where a consistent theorem refers to a $\modell$ satisfying 
%\pmb{tr}, \pmb{nnp}, \pmb{nsa} and \pmb{kos} and a model 
%is some $\mathcal{D}, \mathcal{I}$ such that $\taam$ satisfies \pmb{Core}. 
%This is akin to model exitence theorem (every consistent theory has a model), 
%regarding $\modell$ as the syntax and \pmb{tr}, \pmb{nnp}, \pmb{nsa} and \pmb{kos} as its axioms, 
%and $\mathcal{D}$ as the semantics and $\pmb{Core}$ as semantic constraints. 
\mbox{Theorem \ref{thm_core}} below 
also establishes the other direction: for any typed argumentation graph $\modell$, if there is some $\mathcal{D}, \mathcal{I}$  
such that $\taam$ is a theme aspect argumentation model and that $\taam$ satisfies \pmb{Core}, then 
$\modell$ satisfies \pmb{tr}, \pmb{nnp}, \pmb{nsa} and \pmb{kos}, 
generalising the theoretical correspondence 
that we saw earlier in Theorem \ref{thm_consequence_pr_vi_aass}. 
See also Remark \ref{note_point_this_consequence}.

%little preparation is in order. 
%Even  more non-trivial a result is to show a bi-directional correspondence between 
%the constraints purely on $\modell$ and \pmb{Core}, which we show below. This is precisely the syntax-semantic 
%correspondence in formal logic, treating a typed argumentation graph as a syntax, 
%the constraints purely on $\modell$ as axioms on $\modell$, 
%and $\mathcal{D}, \mathcal{I}$ as 
%a semantics, even though we still view $\langle \modell, \mathcal{D}, \mathcal{I} \rangle$ as a whole a model for a good reason 
%that will become evident in subsequent future work. As far as we are aware, this is the first time this kind of 
%a conceptual separation is shown in a formally rigorous way, and 
%a `sound and complete' result is reported, within formal argumentation. Earlier research on identification 
%of graphical constraints such as \citep{Bonzon16,Amgoud17} has been conducted predominently through informal criteria. 

%\noindent Onto the main result: 

%\noindent Theorem \ref{thm_core} generalises the theoretical correspondence 
%that we saw earlier in Theorem \ref{thm_consequence_pr_vi_aass}. 
%See also Remark \ref{note_point_this_consequence} for the significance. 
%to all the constraints purely on the typed argumentation graph 
%and those on the theme aspect argumentation model. 
\begin{theorem}[{\hyperref[proof_core]{\pmb{Core} satisfaction, necessary and sufficient conditions}}] \label{thm_core}
	Given a typed argumentation graph $\modell$, it satisfies \pmb{tr}, \pmb{nsa}, \pmb{nnp} and \pmb{kos} iff 
	there is some pair of  
	$\mathcal{D}$ and $\mathcal{I}$ such that 
	$\modell$, $\mathcal{D}$ and $\mathcal{I}$ form 
	a theme aspect argumentation model and that 
	$\langle \modell, \mathcal{D}, \mathcal{I}\rangle$ 
	satisfies \pmb{Core}. 
\end{theorem} 
%\textbf{Proof.} %\ryuta{The last part of the proof needs to be carefully checked.} 
%See Appendix:  proofs. \hfill$\Box$ 

\noindent The \pmb{Core} constraints we have identified 
may be regarded as forming the semantics of fallacies. 
More broadly, they 
define normal forms of typed argumentation graphs $\modell$ with respect to 
$\mathcal{D}$ and $\mathcal{I}$. 

\begin{definition}[Normal forms and formal fallacies]\label{def_normal_forms_and_fallacies} \rm 
%	Let $\alpha$ denote any subset of $\{\text{\pmb{Core}}, \pmb{E}, \pmb{das}, \pmb{nwci}, \pmb{F}\}$ 
%	that includes \pmb{Core}.  
	Given a theme aspect argumentation model 
	$\langle \modell, \mathcal{D}, \mathcal{I} \rangle$ 
	and a set of constraints $\alpha$, 
	we say the typed argumentation graph $\modell$ is 
	{\it $\alpha$-normal with respect to $\mathcal{D}, 
	\mathcal{I}$} 
	iff $\langle \modell, \mathcal{D}, \mathcal{I} \rangle$ 
	satisfies every member of $\alpha$. 
	We say {\it $\modell$ is a $\alpha$-fallacy with 
	respect to $\mathcal{D}, \mathcal{I}$} iff $\modell$ is not 
	$\alpha$-normal 
	with respect to $\mathcal{D}, \mathcal{I}$. \hfill$\spadesuit$
\end{definition}  
In Examples \ref{ex_role_aass}, \ref{ex_role_i}, 
\ref{ex_role_vi},
\ldots, \ref{ex_role_mat} and \ref{ex_role_ss}, 
we illustrated the individual role of each 
constituent of \pmb{Core} but \pmb{manss} due to its 
similarity to \pmb{mat}. 
The scenarios given in the examples induce 
a $\{$\pmb{Core}$\}$-fallacy with respect to  
the associated complete Boolean algebra and the interpretation 
function. (The full detail of the theme aspect argumentation 
model in these examples is in \citep{Nakai22}.)   
We emphasise that \pmb{Core} constraints 
are meant to be taken collectively. 
To see why, even if, say, \pmb{aass} is enforced for delimiting 
the semantics of `$attack$' and `$support$', it does not prevent 
manipulation on what can be discussed (see Examples \ref{ex_role_bat}
and \ref{ex_role_mat}) for which other 
constraints such as \pmb{bat}, \pmb{mat} and \pmb{manss} 
become important. Similar goes for any other \pmb{Core} constraints. %they are individually not strong enough  

\section{More Constraints} \label{sec_more_constraints}
%We can already prevent a fair number of fallacies with \pmb{Core}, as we will show in Section \ref{sec_application_fallacy_detection_and_prevention} in more detail. However, 
There are still other well-schemed fallacies that slip through the radar of \pmb{Core}. 
To cope with them, we may pose additional constraints on the theme aspect argumentation model. There are other 
miscellaneous constraints that may be also useful in certain circumstances. 
We presume satisfaction of \pmb{Core} throughout this section's discussion and in all examples. As with all the preceding 
examples, we omit the full detail of $\mathcal{D}$ and $\mathcal{I}$, 
which is, however, found in \citep{Nakai22}. 

%The definition
%of normal forms \ryuta{to be written.} 

\subsection{Context immunity constraints} 
Let us talk of the value of a variable held in a certain address in computer memory, 
to draw inspiration for reasonable constraints for theme aspect argumentation model. In a typical programming language, 
once a value is assigned to a variable in the address, 
the variable will retain the very value until some other value is assigned to it. 
When the value of the variable is referred to by a pointer, 
the pointer refers to it exactly in the sense that 
if we are to obtain the value of the variable 
as referred to by the pointer (and if the value of the variable is not modified), we are again sure to obtain the very value, 
no matter from where the reference is being made. 

Analogous properties for the theme aspect argumentation model can be described as context 
immunities of aspects of statements. Since $\mathcal{I}(T, s)$, aspects of $s$ for $T$, 
can be understood equally as aspects of $s$ observed from context $T$, the context immunity  
would be: no matter from which two contexts $T_1$ and $T_2$ they are observed, aspects of $s$ appear the same 
(obviously only in their common region in $\mathcal{D}$).  
\begin{definition}[Exact common region constraints]\label{def_exact_common_region_constraint} \rm   
	Given a theme aspect argumentation model $\langle \modell, \mathcal{D}, \mathcal{I} \rangle$, we say it 
	satisfies constraint \pmb{esr} (\pmb{e}xact \pmb{s}ummary \pmb{r}eference),  
	\pmb{ensr} (\pmb{e}xact \pmb{n}on-\pmb{s}ummary \pmb{r}eference), 
	and respectively \pmb{eos} (\pmb{e}xact \pmb{o}rdinary \pmb{s}tatement)  
	iff (1), (2), and respectively (3) below hold for every $T_1, T_2 \subseteq \pmb{Themes}$. 
	$Common$ denotes $\mathcal{I}(T_1, \omega) \cap \mathcal{I}(T_2, \omega)$. 
	\begin{enumerate}[label={(\arabic*)},leftmargin=0.9cm]  
		\item for every $t \in \pmb{Themes}$ and 
			every $t.\mathcal{C} \in PStmt[G]$, $\mathcal{I}(T_1, t.\mathcal{C}) \cap Common = \mathcal{I}(T_2, t.\mathcal{C}) \cap Common$. 
		\item for every $t \in \pmb{Themes}$, 
			every $a \in \pmb{A}^{ord}$ and every $t.a \in PStmt[G]$, $\mathcal{I}(T_1, t.a) \cap Common = \mathcal{I}(T_2, t.a) \cap Common$. 
		\item for every $a \in OStmt[G]$, $\mathcal{I}(T_1, a) \cap Common = \mathcal{I}(T_2, a) \cap Common$. \hfill$\spadesuit$
	\end{enumerate}
\end{definition}   
\begin{example}[The role of \pmb{esr}, \pmb{ensr} and \pmb{eos}  
	in formal fallacy identification] \label{ex_role_esr_ensr_eos} \rm  
	%A city has held three discussions among its members on Day 1 and Day 2. %and Day 3. 
	Ambiguous reference leads to fallacies such as {\it equivocation} and {\it ad populum} 
	as are so called in the repertoire of informally 
	classified fallacies. 
	Suppose $t_1$: `{\it Should the local park be transformed into a commercial zone?}'  
        was discussed on Day 1 and ended inconclusively as follows.  
	\begin{itemize} 
		\item Allen $s_1$: ``{\it We should transform the park into a commercial zone. It generates revenue for the city 
	and provides job opportunities.}''   
		\item Bianca $s_2$: ``{\it The park 
			contributes to the overall 
			health of the community, so I do not think we should.}'' 
			%Besides, once the park is gone, we cannot get it back.}''   
		\item Allen 
	$s_3$: ``{\it  But think about the revenue. 
			The city will be richer 
			with the commercial zone.}''  
		\item 
	Bianca $s_4$: ``{\it The overall health of the community 
			cannot be sacrificed for 
			the revenue. %is 
			%more important. 
	}"   
\item Allen $s_5$: ``{\it Well, I do not agree, but fine, we agree to disagree.}''   
\end{itemize} 
	%and the following inconclusive argumentation on the theme resulted.  
	%\begin{itemize} 
	%	\item 		\item 		\item 	\item 
%					\item 	\end{itemize} 
	\indent Suppose another theme $t_2$: `{\it  Where should the city's budget be allocated?}' 
	was discussed on Day 2:   
	\begin{itemize} 
		\item Celina $s_6$: ``{\it We should allocate city's budget to the sectors that ensure 
	economic growth.}''  
		\item Dan $s_7$: ``{\it Sure, the previous discussion on Day 1 supports that.}''   
		\item Edwin $s_8$: ``{\it What? The previous discussion on Day 1 was against that!}''  
		\item Frank $s_9$: ``{\it Guys, it was inconclusive!}''
	\end{itemize} 
	Let $t_6$ be `{\it Celina}', $t_7$ be `{\it Dan}', $t_8$ be `{\it Edwin}' and $t_9$ be `{\it Frank}'. %$t_6$ be `Dan' 
	%and $t_7$ be `Edwin'. 
	The rhetorical intention of the discussion on Day 2 is modelled 
	as:  
%	\begin{tikzcd}[row sep=tiny,column sep={1.6cm}]   
%	s_5:\{t_1,t_3\} \arrow{r}[above]{\{\text{`}attack\text{'}, t_1\}} & s_4:\{t_1, t_4\} 
%		 \arrow{r}[above]{\{\text{`}attack\text{'}, t_1\}} & s_3:\{t_1,t_3\}
%		 \arrow{r}[above]{\{\text{`}attack\text{'}, t_1\}} & s_2:\{t_1, t_4\}
	%	 \arrow{r}[above]{\{\text{`}attack\text{'}, t_1\}} & s_1:\{t_1, t_3\}
%		aCostH \arrow[-,ur]\arrow[-,dr] & 	  & \neg aCostH \arrow[-,ul]\arrow[-,dl] & bCostH \arrow[-,ur]\arrow[-,dr] & & \neg bCostH \arrow[-,ul]\arrow[-,dl]\\
%		& \pmb{0} & & & \pmb{0} \\ 
%                & \pmb{1} &   & & \pmb{1} \\ 
%		aP \arrow[-,ur]\arrow[-,dr] & 	  & \neg aP \arrow[-,ul]\arrow[-,dl] & bP \arrow[-,ur]\arrow[-,dr] & & \neg bP 
%		\arrow[-,ul]\arrow[-,dl]\\
%		& \pmb{0} & & & \pmb{0} 
%	\end{tikzcd}
%	{\ }\linebreak on Day 1 and 
	\begin{tikzcd}[row sep=tiny,column sep={1.6cm}]   
		s_9:\{t_2,t_9\} \arrow{r}[above]{\{\text{`}attack\text{'},t_2\}}
		& s_8:\{t_2,t_8\} \arrow{r}[above]{\{\text{`}attack\text{'},t_2\}} & s_7:\{t_2,t_7\}
		 \arrow{r}[above]{\{\text{`}support\text{'}, t_2\}} & s_6:\{t_2,t_6\}.  
%		aCostH \arrow[-,ur]\arrow[-,dr] & 	  & \neg aCostH \arrow[-,ul]\arrow[-,dl] & bCostH \arrow[-,ur]\arrow[-,dr] & & \neg bCostH \arrow[-,ul]\arrow[-,dl]\\
%		& \pmb{0} & & & \pmb{0} \\ 
%                & \pmb{1} &   & & \pmb{1} \\ 
%		aP \arrow[-,ur]\arrow[-,dr] & 	  & \neg aP \arrow[-,ul]\arrow[-,dl] & bP \arrow[-,ur]\arrow[-,dr] & & \neg bP 
%		\arrow[-,ul]\arrow[-,dl]\\
%		& \pmb{0} & & & \pmb{0} 
	\end{tikzcd}
%	on Day 2.  
	$s_7$, $s_8$ and $s_9$ on Day 2 are pointer statements referring to 
	a \pmb{s}ummary of $t_1$, {\it i.e.} $s_9 \equiv s_8 \equiv s_7 \equiv t_1.\mathcal{C}$. 
	But here, Dan, Edwin and Frank all entertain their personal understanding 
	of $t_1$'s \pmb{s}ummary, which can happen with \pmb{Core} because $s_7$ is mapped to $\mathcal{I}(\{t_2, t_7\}, t_1.\mathcal{C})$, 
	$s_8$ is mapped to $\mathcal{I}(\{t_2, t_8\}, t_1.\mathcal{C})$ 
	and $s_9$ is mapped to $\mathcal{I}(\{t_2, t_9\}, t_1.\mathcal{C})$.  
	Although the statement is shared, the contexts interpreting it 
	are not the same. It is possible that 
	$\mathcal{I}(\{t_2, t_7\}, t_1.\mathcal{C}) = \{`more\ revenue' 
	\wedge `commercial\ zone' 
	\wedge \neg `park'\}$, that 
	$\mathcal{I}(\{t_2, t_8\}, t_1.\mathcal{C}) = 
	\{`community's\ overall\ health' \wedge 
	\neg `more\ revenue' \wedge `park'\}$,
	and that 
	$\mathcal{I}(\{t_2, t_9\}, t_1.\mathcal{C})$ is a disjunction of
	some of the aspects taken from Allen's and Bianca's statements.  
	It is easy 
	to select a pair of $\mathcal{D}$ 
	and $\mathcal{I}$ such that 
	the typed argumentation graph is \pmb{Core}-normal 
	with respect to $\mathcal{D}$ and $\mathcal{I}$ 
	while satisfying 
	these conditions. 
%	and satisfying moreover that the typed argumentation graph 
%	is \pmb{Core}-normal with respect to $\mathcal{D}, \mathcal{I}$.
	With \pmb{esr} in force, though, ambiguous reference will not be permitted. So, this typed argumentation graph 
	is an $\{\text{\pmb{esr}}\}$-fallacy 
	with respect to $\mathcal{D}, \mathcal{I}$. 
	To avoid the $\{\text{\pmb{esr}}\}$-fallacy
	with respect to $\mathcal{D}, \mathcal{I}$, 
	Dan, Edwin and Frank must either refer to the same aspect(s) in $\mathcal{I}(\{t_2\}, t_1.\mathcal{C})$   
	when they refer to \pmb{s}ummary of $t_1$, or 
	avoid referring to \pmb{s}ummary of $t_1$ altogether 
	and instead explicitly state what they intend to express. 
	%explicitly.  
%	refer to an ordinary statement in $t_1$. 
	 %This also means that  enforcement of \pmb{esr} 
%	together with \pmb{Core} in the above argumentation causes 
%	violation of \pmb{Core} (specifically \pmb{aass}) 
%	since the aspects of $s_7$, $s_8$ and $s_9$ will match 
%	exactly. 
	
	The ambiguous reference does not have to be to 
	a \pmb{s}ummary of theme(s). It could be 
	to a statement of theme(s), too. \pmb{ensr} and \pmb{eos} are for ensuring 
	the univocal interpretation of every ordinary statement and every 
	reference to it.  \hfill$\clubsuit$ 
%	$s_7$, $s_8$ and $s_9$ must be considered denoting 
%	the same aspects (and thus some of them are lying). 
	%In the above scenario, this constraint imposes a condition 
	%on $t_1$ that it must have a univocal  
%
%
%
%	 
%
%
%	the \pmb{s}ummary of $t_1$ 
	 %= \{`economic growth' \wedge `commercial zone'\}$,
%	There are various methods proposed for obtaining a summary 
%	of argumentation such as by {\it acceptability semantics} in the abstract argumentation 
%	literature, but the detail is unimportant. In this work, 
%	it is sufficient to only assume that there is some external module, 
%	and that the module obtains 
%	\pmb{s}ummary of theme(s) as some aspect(s). 
%	It suffices to only observe the following: in the presence 
%	Since the discussion is inconclusive, one user supporting the {\it economic growth} 
%	and another user supporting the {\it overall health of the community},  

	%There are various methods proposed 

%	We suppose the complete Boolean algebra covers all reasonable aspects of the city's policymaking. 
%
%	$s_5 \rightarrow s_4 \rightarrow s_3 
%	\rightarrow s_2 \rightarrow s_1$ and that on Day 2 is $s_8 \rightarrow s_7\rightarrow s_6$. 
%	 
\end{example} 

\begin{note}[Why is \pmb{esr} not a core constraint?] \rm     
	While it may be ideal that \pmb{s}ummary of theme(s)  
	be always univocal, it is practically unreasonable
	to enforce it, 
	not only because argumentation can be inconclusive but also 
	because a summary of argumentation can come with different granularities. The more
	abstract it is, the more aspects may be omitted or generalised. 
	%\pmb{S}ummary of the 
	%theme(s) a pointer statement within some theme(s) $T$ refers to is interpreted 
	%in $T$. In other words, $T$ decides the type of acceptability semantics 
	%to evaluate it with, which easily leads to different \pmb{s}ummaries. 
	Didactic enforcement of 
	\pmb{esr} fails to recognise the contextual influence over the way argumentation is summarised. 
	\hfill$\clubsuit$
\end{note} 

\begin{note}[Why are \pmb{ensr} and \pmb{eos} not core constraints?] \rm \label{note_ensr_eos}
	Natural language statements are innately vague and additional aspects 
	may be associated to them {\it post hoc}. For instance, 
	$\mathcal{I}(\{t_1\}, a) \cap \mathcal{I}(\{t_1\}, \omega) \subset \mathcal{I}(\{t_1, t_2\}, a) \cap \mathcal{I}(\{t_1\}, \omega)$ is possible. 
	1968's Nixon's statement {\it ``[W]e shall have an honourable  
	end to the war in Vietnam."} is one example, which could have meant a total win. 
	Thus, for the theme of ``{\it American-Vietnamese Conflict around 1968}" ($t_{1968}$ for abbreviation), it is an ordinary statement, say $a_{he}$, which can have 
	such aspects as 
	{\it ``how to win a war"}. On the other hand, 
	{\it ``how to retreat"} is reasonably not $a_{he}$'s aspect
	for $t_{1968}$. 
	However, Nixon's 1973 statement of {\it ``peace with honour''} \citep{Johns10} ($a_{ph}$ for abbreviation) painted {\it post hoc} another meaning to 
	 the honourable end, to pull U.S. troops out of Vietnam. 
	  Hence, for the theme of {\it ``American Politics"} ($t_{AP}$ for abbreviation), 
	  {\it ``how to retreat''} is a reasonable aspect of $a_{he}$ 
	for its sub-theme $t_{1968}$, {\it i.e.} 
	 {\it ``how to retreat''} $\in \mathcal{I}(\{t_{1968}, t_{AP}\}, a_{he}) \cap \mathcal{I}(\{t_{1968}\}, \omega)$. 
	 Making \pmb{ensr} or \pmb{eos} a core constraint is tantamount to assuming the attitude that 
	 every ordinary statement has a fixed particular intrinsic interpretation for every theme(s) regardless of the point 
	 of observation of the statement, which certainly benefits stronger fallacy identification but which can run counter 
	 to the nature of natural language statements.  \hfill$\clubsuit$
\end{note}

\noindent Let \pmb{E} denote \pmb{esr}, \pmb{ensr} and \pmb{eos} 
collectively. \pmb{E} and \pmb{Core} guarantee a stronger 
monotonicity property for statements than \pmb{Core} alone. 
 Moreover, we obtain distributivities. %See Appendix: proofs for the proofs.
\begin{theorem}[{\hyperref[proof_monotone_aspects_of_statements]{Monotone aspects of statements}}] \label{thm_monotone_aspects_of_statements} 
	Given a theme aspect argumentation model $\langle \modell, \mathcal{D}, \mathcal{I} \rangle$,  
	if it satisfies \pmb{Core} and \pmb{esr},
	%and \pmb{ensr}, 
	then the following holds for  every $T_1 \subseteq T_2 \subseteq \pmb{Themes}$, 
	every $s \in Stmt[G]$ and every $x \in \mathcal{I}(T_1, s)$: $x \in \mathcal{I}(T_2, s)$. 
\end{theorem} 
\begin{theorem}[{\hyperref[proof_distribution]{Distribution}}] \label{thm_distribution} 
      Given a theme aspect argumentation model 
	$\langle \modell, \mathcal{D}, \mathcal{I} \rangle$, if it satisfies \pmb{Core} and \pmb{E}, then 
	the following hold for any $T_1, T_2 \subseteq \pmb{Themes}$: 
	$\mathcal{I}(T_1 \cup T_2, s) = \mathcal{I}(T_1, s) \cup \mathcal{I}(T_2, s)$  
	and $\mathcal{I}(T_1 \cap T_2, s) = \mathcal{I}(T_1, s) \cap \mathcal{I}(T_2, s)$. 
\end{theorem} 
%\textbf{Proof.} See Appendix:  proofs. \hfill$\Box$ 

\subsection{Constraints on `attack'/`support'-linked statements}
\pmb{Core} has defined the semantics of `$attack$' and `$support$' via \pmb{aass}. 
As we saw back in Section \ref{sec_understanding_the_semanics}, there are certain cases that can be either a support or an attack 
({\it Insurance (Weakening)} or {\it Weakening}, {\it Value-augmentation (Strengthening)} or {\it Strengthening}). 
We can of course try to eliminate the attack-support ambiguity. 

\begin{definition}[Distinct attack support constraint]\label{def_distinct_attack_support_constraint} \rm 
   Given a theme aspect argumentation model $\langle \modell, \mathcal{D}, \mathcal{I} \rangle$,  
	we say it satisfies constraint \pmb{das} (\pmb{d}istinct
	\pmb{a}ttack \pmb{s}upport) iff 
	all the following hold for every $(s', s) \in Rel[G]$ 
	and every $\emptyset \subset T \subseteq \pmb{Themes}$: 
	\begin{itemize}  
%		\item  $\emptyset \subset \mathcal{I}(\{t\}, s')$.  
%		\item  $\mathcal{I}(\{t\}, s') \cap 
%			\mathcal{I}(\{t\}, s) = \emptyset$. \ryuta{Need to put 
%			an explanation.} 
		\item  $T \cup \{`attack$'$\} \subseteq \Pi((s', s))$ 
			materially implies $\bigwedge 
			\mathcal{I}(T, s')    
			\not\in  {\downarrow \{\bigwedge \mathcal{I}(T, s)\}} \cup 
			{\uparrow \{\bigwedge \mathcal{I}(T, s)\}}$. 
%			\cup ({\uparrow 
%			\{\neg \bigwedge \mathcal{I}(T, s)\}} 
%			\backslash \{\neg \bigwedge \mathcal{I}(T, s)\})$. 
		\item  $T \cup \{`support$'$\} \subseteq 
			\Pi((s', s))$ materially implies 
			$\bigwedge \mathcal{I}(T, s') 
			= \bigwedge \mathcal{I}(T, s)$.  \hfill$\spadesuit$
%			(\neg RlvntAspcts \cup \mathcal{I}(\{t\}, s)) = \emptyset$.  
%		\item  $`support' \in \Pi((s', s))$ materially implies 
%			$\mathcal{I}(\{t\}, s') \subseteq 
%			RlvntAspcts$. 
	\end{itemize} 
\end{definition}    

\begin{example}[The role of \pmb{das} in formal fallacy identification]\label{ex_role_das} \rm 
     We observed the overlap in the nuance of attack and support. 
	The overlapping region (see Section \ref{sec_understanding_the_semanics}) can be exploited 
	in fallacies such as {\it hasty generalisation}, 
	{\it composition fallacy} and {\it straw man} as 
	are so called within the repertoire of informally 
	classified fallacies. 
	Suppose $t$: `{\it Should Alice buy 
	fire insurance?}'  and the following argumentation. 
	\begin{itemize} 
		\item Bob $s_1$: ``{\it Alice should not buy 
			fire insurance. It costs her a lot. 
			Also, it does not benefit her much.}" 
		\item Chris $s_2$: ``{\it Exactly. 
		  	Insurance in general is too expensive, 
			benefiting insurance companies mostly. 
			She should not buy fire insurance.}''
	\end{itemize} 
	The rhetoric is supportive, so the typed argumentation 
	graph is 
	\begin{tikzcd}[row sep=tiny,column sep={1.6cm}]   
		s_2:\{t, t_2\} 
		\arrow{r}[above]{\{\text{`}support\text{'},t\}}
		& s_1:\{t,t_1\} 
		%\arrow{r}[above]{\{\text{`}attack\text{'},t_2\}} & s_7:\{t_2,t_7\}
		% \arrow{r}[above]{\{\text{`}support\text{'}, t_2\}} & s_6:\{t_2,t_6\}.  
%		aCostH \arrow[-,ur]\arrow[-,dr] & 	  & \neg aCostH \arrow[-,ul]\arrow[-,dl] & bCostH \arrow[-,ur]\arrow[-,dr] & & \neg bCostH \arrow[-,ul]\arrow[-,dl]\\
%		& \pmb{0} & & & \pmb{0} \\ 
%                & \pmb{1} &   & & \pmb{1} \\ 
%		aP \arrow[-,ur]\arrow[-,dr] & 	  & \neg aP \arrow[-,ul]\arrow[-,dl] & bP \arrow[-,ur]\arrow[-,dr] & & \neg bP 
%		\arrow[-,ul]\arrow[-,dl]\\
%		& \pmb{0} & & & \pmb{0} 
	\end{tikzcd}
	where $t_1$ denotes `{\it Bob}' and $t_2$ 
	denotes `{\it Chris}'. 
	Here, 
	%Let us suppose a complete Boolean algebra. Lemma 
	%\ref{lem_disjoint_combination} makes the construction 
	%of $\mathcal{D}$ very simple and compositional.  
        Chris' statement misrepresents ({\it straw man}) 
	and/or hastily generalises ({\it hasty generalisation})  
	Bob's. 
	Suppose $\mathcal{D}$ is large enough. 
	Suppose the following aspects $x_1$: `{\it Alice}', 
	$x_2$: `{\it purchasing insurance}', 
	$x'_2$: `{\it purchasing fire insurance}', 
	$x_3$: `{\it costing a lot}', 
	$x_4$: `{\it benefiting a lot}', 
	and $x_5$: `{\it insurance companies}' 
	are in $D$. 
	It is assumed that $x'_2 < x_2$. 
	Suppose our interpretation $\mathcal{I}$ 
	is such that $\mathcal{I}(\{t\}, s_1) = 
	\{\neg (x_1 \wedge x'_2), (x_1 \wedge x'_2) 
	\Rightarrow (x_1 \wedge x_3), (x_1 \wedge x'_2) 
	\Rightarrow \neg (x_1 \wedge x_4)\}$  
	and that $\mathcal{I}(\{t\}, s_2) = 
	\{x_2 \Rightarrow (x_3 \wedge x_4 \wedge x_5),
	\neg (x_1 \wedge x'_2)\}$.   
	Then, $\bigwedge \mathcal{I}(\{t\}, s_1) =  
	\neg (x_1 \wedge x_2')$  
	and $\bigwedge \mathcal{I}(\{t\}, s_2) = 
	\neg (x_1 \wedge x_2') \wedge 
	(x_2 \Rightarrow (x_3 \wedge x_4 \wedge x_5))$. 
	Hence, $\bigwedge \mathcal{I}(\{t\}, s_2) < \bigwedge \mathcal{I}(\{t\}, s_1)$. 
	% \ryuta{Compute them.}
	This kind of argumentation 
	is a $\{\text{\pmb{das}}\}$-fallacy. 
	%does not satisfy \pmb{das}. %With \pmb{das} in force, the theme aspect argumentation model 
%	bars this kind of argumentation. %are barred.  
	With \pmb{das} in force, if Chris' intention 
	is to agree with $s_1$ and to add the extra 
	information 
	that Alice should not buy it for the reason 
	in $s_2$, then Chris needs to just 
	agree with $s_1$ without attaching anything else; 
	the extra part needs to be 
	expressed as a fresh statement 
	to be separately discussed. 
	%expresses the other part as another fresh statement. 
	
	As for the restriction on the semantics of `$support$' 
	by \pmb{das}, 
	note that only the effective aspect 
	of two statements needs to be the same. 
	In addition to the simple agreement ``{\it I agree.}",  
	it is possible to express support by 
	stating that some aspect $x$ that is omitted 
	from consideration is not relevant to 
	the effective aspect of $s$. For example, 
	another statement $s_3$ may be 
	responding to $s_1$. $s_3$: ``{\it I am with you, 
	she should not buy it. 
	If she buys  
	homeowners insurance, fire insurance is too much of an expense 
	 and not much of a benefit. 
	 But even if she does not buy it, %buy homeowners insurance, 
	 fire insurance alone is already quite costly 
	 and simply does not offer much 
	 benefit to her.}'' 
	 Let $x_6$ denote `{\it purchasing homeowners insurance}'. 
	 By assuming {\it too much of an expense} and 
	 {\it quite costly} both mean a lot of cost,  
	 $\mathcal{I}(\{t\}, s_3) = 
	 \{\neg (x_1 \wedge x_2'), (x_1 \wedge x_6) 
	 \Rightarrow (((x_1 \wedge x'_2) 
	\Rightarrow (x_1 \wedge x_3)) \wedge  
	((x_1 \wedge x'_2) 
	\Rightarrow \neg (x_1 \wedge x_4))),   
	\neg (x_1 \wedge x_6) 
	 \Rightarrow (((x_1 \wedge x'_2) 
	\Rightarrow (x_1 \wedge x_3)) \wedge  
	 ((x_1 \wedge x'_2) 
	\Rightarrow \neg (x_1 \wedge x_4)))
	 \}$. Clearly, $\bigwedge \mathcal{I}(\{t\}, s_3) 
	 = \bigwedge \mathcal{I}(\{t\}, s_1)$. 
%	 adds the new information that purchasing homeowners insurance 
%	 or otherwise is irrelevant in the eyes of David. 
	 \hfill$\clubsuit$ 
%	\ryuta{Need to write this part.}   
%	Suppose $t$ `{\it Should Alice buy fire insurance for 
%	her summer house?}' and suppose that the complete Boolean algebra 
%	$\mathcal{D}$ covers a large number of aspects 
%	of the theme. 
%	\begin{itemize} 
%		\item Bob $s_1$: ``{\it Alice should not buy 
%			fire insurance. It costs her a lot.}''  
%		\item Chris $s_2$: ``{\it Exactly. Insurance 
%			is a waste of money for us. 
%			 
%			 
%			 }''
%	\end{itemize} 
%
%	 
%	allacies 
%     This can be exploited in fallacious argumentation 
%     Suppose a theme $t$: `{\it Should Alice buy fire insurance for her summer house?}'  
\end{example} 
Weakened contradictions \citep{Heyninck20} and incomparable alternatives are still permitted in \pmb{das}. 
To bring the semantics of `$attack$' closer to logic-based argumentation 
models', we may impose: 
\begin{definition}[No weakened contradictions / incomparables constraint] \label{def_no_weakened_contradiction_incomparables_constraint} \rm 
	Given a theme aspect argumentation model $\taam$, we say it satisfies constraint \pmb{nwci} (\pmb{n}o \pmb{w}eakened \pmb{c}ontradictions / 
	\pmb{i}ncomparables) iff the following hold for every $(s', s) \in Rel[G]$ and every $\emptyset \subset 
	T \subseteq \pmb{Themes}$: 
	if $\{`attack$'$\} \cup T \subseteq \Pi((s', s))$, then: 
	\begin{itemize} 
		\item $\bigwedge \mathcal{I}(T, s') \not\in ({\uparrow \{\neg \bigwedge \mathcal{I}(T, s)\})} 
			\backslash \{\neg \bigwedge \mathcal{I}(T, s)\}$. 
		\item 
	$\bigwedge \mathcal{I}(T, s') \in {\downarrow \{\bigwedge \mathcal{I}(T, s)\}} \cup {\uparrow \{\bigwedge 
	\mathcal{I}(T, s)\}} \cup {\downarrow \{\neg \bigwedge \mathcal{I}(T, s)\}} \cup
	{\uparrow \{\neg \bigwedge \mathcal{I}(T, s)\}}$.  \hfill$\spadesuit$
	\end{itemize} 
\end{definition} 

\begin{note}[Why is \pmb{das} or \pmb{nwci} not a core constraint?] \rm 
	At least \pmb{das} is a semi-core constraint; only, 
	it is a stronger requirement than the core constraints that 
	can also block rhetorically reasonable statements' interactions. 
	\hfill$\clubsuit$ 

\end{note} 
%\begin{note} \rm 
%A little more detail on \pmb{das}, the restriction  
%	on the semantics of `$support$' does not otherwise bar 
%	introduction of new supporting aspects 
%	to an existing statement from getting introduced.  
%	Suppose  
%	the effective  
%
%
%	\hfill$\clubsuit$ 
%\end{note}
%\begin{note} 
%   Amgoud's compiled arguments, ad hocness. 
%\end{note} 
%\noindent \pmb{das}, \pmb{esr} and \pmb{ensr} ensure that no relation may be typed both $`attack'$ and $`support'$ 
%for any theme $t \in \pmb{Themes}$ typing the relation.   
\noindent \pmb{das} ensures that no relation may be typed with both `$attack$' and `$support$' 
for any theme $t \in \pmb{Themes}$ typing the relation. Furthermore, if the relation 
$s_1 \rightarrow s_2$ is typed `$attack$', no statement $s$ can have a support relation 
to both $s_1$ and $s_2$ for a shared theme or can be supported by both 
$s_1$ and $s_2$ for a shared theme. %See \hyperref[app_proofs]{Appendix: proofs} for the proof. 
%\begin{proposition}[Consequence of \pmb{das}, \pmb{esr} and \pmb{ensr}]\label{prop_consequence_das} 
	\begin{proposition}[{\hyperref[proof_consequence_das]{Consequence of \pmb{das}}}]\label{prop_consequence_das} 
	Given a theme aspect argumentation model $\langle \modell, \mathcal{D}, \mathcal{I} \rangle$,
	that it satisfies \pmb{Core} and \pmb{das} % \pmb{esr} and \pmb{ensr}, then  
	materially implies the following for every $(s_1, s_2) \in Rel[G]$: if $`attack$'$ \in \Pi((s_1, s_2))$, 
		then: 
	\begin{itemize} 
		\item $`support$'$\ \not\in \Pi((s_1, s_2))$. 
		\item for every $s \in Stmt[G]$ and every $i, j \in \{1,2\}$ with $i \not= j$, $(s, s_i) \in Rel[G]$ 
			and $`support$'$ \in \Pi((s, s_i))$ materially imply either    
			$(s, s_j) \not\in Rel[G]$, or $`support$'$ \not\in \Pi((s, s_j))$, 
			or there is no $t \in \pmb{Themes}$ such that $t \in \Pi((s, s_i)) \cap \Pi((s, s_j))$. 
		\item for every $s \in Stmt[G]$ and every $i, j \in \{1,2\}$ with $i \not= j$, $(s_i, s) \in Rel[G]$ 
			and $`support$'$ \in \Pi((s_i, s))$ materially imply either   
			$(s_j, s) \not\in Rel[G]$, or $`support$'$ \not\in \Pi((s_j, s))$, 
			or there is no $t \in \pmb{Themes}$ such that $t \in \Pi((s_i, s)) \cap \Pi((s_j, s))$. 
\end{itemize}
\end{proposition} 

\noindent We may impose the following constraint purely on the typed argumentation graph $\modell$ to prevent  
the conflict between \pmb{Core} and \pmb{das} from arising. 

\begin{definition}[(Graphic) No simultaneous neighbouring supports constraint]\label{def_no_simultaneous_neighbouring_supports_constraint} \rm 
	Given a typed argumentation graph $\modell$, we say $\modell$ satisfies constraint \pmb{nss} (\pmb{n}o \pmb{s}imultaneous 
	neighbouring 
	\pmb{s}upport) iff the following hold for every $(s_1, s_2) \in Rel[G]$: if $`attack$'$ \in \Pi((s_1, s_2))$, then: 
	\begin{itemize} 
		\item $`support$'$\ \not\in \Pi((s_1, s_2))$. 
		\item for every $s \in Stmt[G]$ and every $i, j \in \{1,2\}$ with $i \not= j$, $(s, s_i) \in Rel[G]$ 
			and $`support$'$ \in \Pi((s, s_i))$ materially imply either    
			$(s, s_j) \not\in Rel[G]$, or $`support$'$ \not\in \Pi((s, s_j))$, 
			or there is no $t \in \pmb{Themes}$ such that $t \in \Pi((s, s_i)) \cap \Pi((s, s_j))$. 
		\item for every $s \in Stmt[G]$ and every $i, j \in \{1,2\}$ with $i \not= j$, $(s_i, s) \in Rel[G]$ 
			and $`support$'$ \in \Pi((s_i, s))$ materially imply either   
			$(s_j, s) \not\in Rel[G]$, or $`support$'$ \not\in \Pi((s_j, s))$, 
			or there is no $t \in \pmb{Themes}$ such that $t \in \Pi((s_i, s)) \cap \Pi((s_j, s))$. \hfill$\spadesuit$
	\end{itemize}  
\end{definition} 
%\begin{note}[The role of \pmb{nss} in formal fallacy identification] \rm 
%    While a statement $s_1$ can both attack and support  
%	another statement $s_2$ without becoming fallacious, 
%	for example by pointing out pros and cons of 
%	multiple aspects in $s_1$, 
%	that $s_2$ may also appear 
%	to attack and support $s_1$ via deceptive rhetoric. 
%	Suppose $t$ `{\it wage}'. 
%	\begin{itemize} 
%		\item Bob $s_1$: ``{\it Increasing the minimum wage 
%			should be beneficial for low-income workers.}'' 
%		\item Chris $s_2$: 
%			``{\it That's right. That helps them 
%			have more spending power. But also, 
%			increasing the minimum wage 
%			}'' 
%	\end{itemize} 
%\end{note} 
We may enforce constraints to preclude 
an argumentation with repeated expressions 
of the same statements, 
both width-wise and depth-wise. 
Let us define the following notations: {\it sub-theme typed argumentation graph} (Definition \ref{def_sub_theme}) 
for obtaining a sub-graph of a typed argumentation graph for some theme, and 
{\it width-statements-sets} and {\it depth-statements-sets} (Definition \ref{def_theme_related_statements_sets}) 
which formally define which set of statements is considered width-wise or depth-wise connected to a statement 
with respect to 
a given theme. 

\begin{definition}[Sub-theme $\modell$] \label{def_sub_theme} \rm 
%	Given $\modell$ and $T \subseteq \pmb{Themes}$, 
%	we define $(G_T, \Pi_T)$ to be such that 
%	\begin{itemize}  
%				\item $Stmt[G_T]$ includes every $s \in Stmt[G]$ such that \mbox{$\Pi(s) \cap T \not= \emptyset$}, but nothing else. 
%		\item $Rel[G_T]$ includes every $(s', s) \in Rel[G]$ such that $\Pi((s', s)) \cap T \not= \emptyset$, but nothing else. 
%		\item $\Pi_T$ is such that, for every $x \in Stmt[G_T] \cup Rel[G_T]$, 
%			$\Pi_T(x) = \Pi(x) \cap (T \cup \{`attack', `support'\})$, 
%			and for every $x \not\in Stmt[G_T] \cup Rel[G_T]$, 
%			$\Pi_T(x) = \emptyset$. \hfill$\spadesuit$
%	\end{itemize} 
	Given a typed argumentation graph 
	$\modell$ and some $t \in \pmb{Themes}$, 
		we define $(G_t, \Pi_t)$,  
		{\it a sub-theme $\modell$}, 
		to be such that 
	\begin{itemize}  
		\item $Stmt[G_t]$ includes every $s \in Stmt[G]$ 
			such that $t \in \Pi(s)$, but nothing else. 
		\item $Rel[G_t]$ includes every $(s', s) \in Rel[G]$ 
			such that $t \in \Pi((s', s))$, but nothing else. 
		\item $\Pi_t$ is such that, for every 
			$x \in Stmt[G_t] \cup Rel[G_t]$, 
			$\Pi_t(x) = \Pi(x) \cap (\{t\} 
			\cup \{`attack$'$, `support$'$\})$, 
			and for every $x \not\in Stmt[G_t] \cup Rel[G_t]$, 
			$\Pi_t(x) = \emptyset$. \hfill$\spadesuit$
	\end{itemize} 
\end{definition} 
\begin{proposition}[Well-definedness] \label{prop_well_definedness} 
%	Given $\modell$ and $T \subseteq \pmb{Themes}$, 
%	$(G_T, \Pi_T)$ is well-defined iff $\range(\Pi) \cap T \not= \emptyset$. 
	Given a typed argumentation graph $\modell$ and some $t \in \pmb{Themes}$, 
	$(G_t, \Pi_t)$ is well-defined iff $t \in \range(\Pi)$. 
\end{proposition}  
\vspace{-0.2cm} 
\textbf{Proof.} Obvious. \hfill$\Box$ \\ 

\begin{definition}[Theme related statements-sets]\label{def_theme_related_statements_sets} \rm 
	Given a typed argumentation graph $\modell$, some $s \in Stmt[G]$, $t \in \pmb{Themes}$ and $trel \in \{$`$attack$'$,
	$`$support$'$\}$, 
	we say that $S \subseteq Stmt[G]$ 
	is {\it a width-statements-set of $s$ with respect to $t$ for $trel$} 
	iff $S$ is a maximal set that satisfies all the following conditions:  
	\begin{itemize} 
		\item $S \subseteq Stmt[G_{t}]$. 
		\item $s \in S$. 
		\item $s' \in S$ and $s \not= s'$ materially imply $(s', s) \in Rel[G_{t}]$ and $trel \in \Pi((s',s))$. %In another word,  
%			$s'$ is $s$'s predecessor. 
%		\item There exists $trel \in \{`attack$'$, `support$'$\}$ such that $trel \in \Pi((s', s))$ for every $s' \in S$ 
%			with $(s', s) \in Rel[G_{t}]$. 
%		\item $t \in \Pi((s', s)) \cap \Pi(s') \cap \Pi(s)$ for 
	%		every $s' \in S$ with $(s', s) \in Rel[G]$. 
	\end{itemize}
	and is {\it a depth-statements-set of $s$ with respect to $t$ for $trel$} iff $S$ is a maximal 
	set that satisfies all the following conditions: 
	\begin{itemize}  
		\item $S \subseteq Stmt[G_{t}]$. 
		\item  
%			$S \subseteq \{s' \in Stmt[G_{\{t\}}] \mid (s', s) \in Rel[G_{\{t\}}]^*\}$ where $Rel[G_{\{t\}}]^*$ is 
%			the reflexive-transitive closure of $Rel[G_{\{t\}}]$. 
			Let $n$ denote $|S|$ and let $s_n$ denote $s$, then  
			there is some $m$ at least as large as $n$ and 
			a finite sequence $s_1, \ldots, s_{m-1}, s_n$ of members of $S$ 
			such that the following hold: $S$ is the set consisting of 
			all the distinct members of $\{s_1, \ldots, s_{m-1}, s_n\}$; 
			for every $1 \leq i \leq m - 2$, 
			$(s_i, s_{i+1}) \in Rel[G_{t}]$ and $trel \in \Pi((s_i, s_{i+1}))$; 
			and $(s_{m-1}, s_n) \in Rel[G_{t}]$ and $trel \in \Pi((s_{m-1}, s_n))$. 
%		 \item There exists $trel \in \{`attack$'$, `support$'$\}$ such that $trel \in \Pi((s', s))$ for every $s' \in S$ 
%			with $(s', s) \in Rel[G_{t}]$. 
	\end{itemize}
	We may use $WidS[s, t, `attack$'$], WidS[s, t, `support$'$],$ $DepS[s, t, `attack$'$],$ and 
	$DepS[s, t, `support$'$]$  
	to denote the set of all width-statements-sets of $s$ with respect to $t$ for `$attack$'s, 
        the set of all width-statements-sets of $s$ with respect to $t$ for `$support$'s, 
	the set of all depth-statements-sets of $s$ with respect to $t$ for `$attack$'s, 
	and the set of all depth-statements-sets of $s$ with respect to $t$ for `$support$'s. \hfill$\spadesuit$
\end{definition}   
\pagebreak
\begin{note}[Width-/depth-statements-sets]\label{ex_width_depth_statements_sets} \rm
	The following example provides  
	an illustration. %By $s_6:\{t_1, t_2\}$ we mean that $s_6$ is typed with $t_1$ and $t_2$. 
%	By $\{`support$'$, t_1\}$ right next to the edge from $s_3$ to $s_6$ we mean 
%	that the edge is typed with `$support$' and $t_1$. Similarly for all the other nodes and edges. 
	\\%\ryuta{I need to remove attack+support off 
%	the example.} \\
	\begin{center} 
	\includegraphics[scale=0.5]{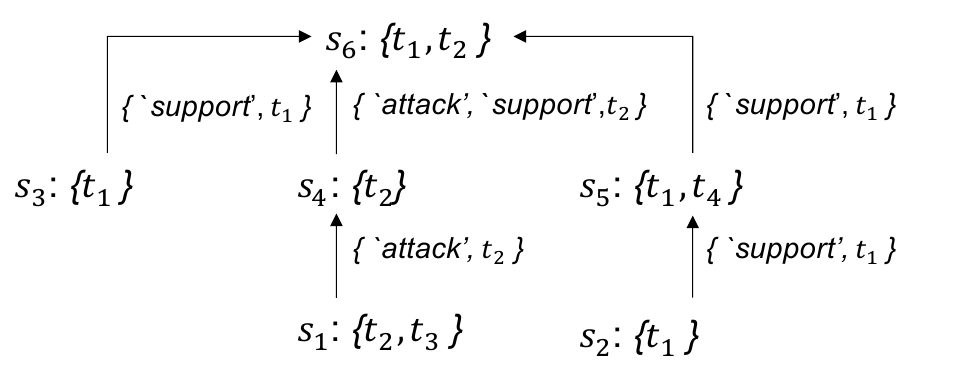} 
	\end{center} 
	\noindent		{\small {$WidS[s_6, t_2, `attack$'$] = \{\{s_4, s_6\}\}$}, 
		 {$WidS[s_6, t_1, `support$'$] = \{\{s_3, s_5, s_6\}\}$},  
		 {$DepS[s_6, t_2, `attack$'$] = \{\{s_1, s_4, s_6\}\}$},  \linebreak
		  {$DepS[s_6, t_1, `support$'$] = \{\{s_2, s_5, s_6\}, \{s_3, s_6\}\}$}}.   
		 
		 By the definition of a depth-statements-set and a width-statements-set, 
		 these outputs do not change if we put another edge $(s_6, s_6)$ 
		 typed with $\{t_1, t_2, `attack$'$, `support$'$\}$. 
		  \hfill$\clubsuit$\\

%   \indent For example, $widS[s_6, t_1, `support']$ looks at all the statements
%	attacking $s_6$ for $t_1$, $depS[s_6, t_2, `attack']$ 
%	looks at all the statements in a chain 
%	 of attacks for $t_2$ below $s_6$ in the above diagram.  
\end{note}  
\begin{proposition}[On the cardinality] \label{prop_uniqueness} 
	Given a theme aspect argumentation model $\langle \modell, \mathcal{D}, \mathcal{I} \rangle$, the following 
	hold for every 
	$s \in Stmt[G]$ and every $t \in \Pi(s)$:  
	\begin{multicols}{2} 
	\begin{itemize} 
		\item $0 \leq |WidS[s, t, `attack$'$]| \leq 1$. 
		\item $0 \leq |WidS[s, t, `support$'$]| \leq 1$.
		\item $0 \leq |DepS[s, t, `attack$'$]|$. 
		\item $0 \leq |DepS[s, t, `support$'$]|$. 
	\end{itemize} 
	\end{multicols} 
\end{proposition}  
\vspace{-0.2cm} 
\textbf{Proof.} Straightforwardly generalisable from {Remark \ref{ex_width_depth_statements_sets}}. \hfill$\Box$ \\

\noindent These notations capture both width-wise and depth-wise connected statements for each theme. 
\begin{note}[On the depth-statements-sets]    \rm 
   The second condition of a depth-statements-set allows the same member of $S$ 
   to appear at least once but possibly more than once in the sequence. The following 
   example should make it clear why it is not formulated with a permutation of $S$ instead. Denote the graph below by $G$, 
	and assume that $\range(\Pi) = \{t, `attack$'$\}$ where $\range(...)$ is the range of $...$ and that 
	$(G, \Pi)$ is well-formed. 
	\begin{center} 
    \begin{tikzcd}[column sep=tiny,row sep=tiny]   
	        &               & s_4 \arrow[dl] \\    
	    s_6 & s_5 \arrow[l] \arrow[dr] &  & s_3 \arrow[ul] & s_1 \arrow[l]    \\  
		&  & s_2 \arrow[ur] 
%		aCostH \arrow[-,ur]\arrow[-,dr] & 	  & \neg aCostH \arrow[-,ul]\arrow[-,dl] & bCostH \arrow[-,ur]\arrow[-,dr] & & \neg bCostH \arrow[-,ul]\arrow[-,dl]\\
%		& \pmb{0} & & & \pmb{0} \\ 
%                & \pmb{1} &   & & \pmb{1} \\ 
%		aP \arrow[-,ur]\arrow[-,dr] & 	  & \neg aP \arrow[-,ul]\arrow[-,dl] & bP \arrow[-,ur]\arrow[-,dr] & & \neg bP 
%		\arrow[-,ul]\arrow[-,dl]\\
%		& \pmb{0} & & & \pmb{0} 
	\end{tikzcd}
	\end{center}  
	By tracing nodes from $s_6$ through the edges in the reversed direction, we can visit all the members of $Stmt[G]$. 
	Therefore, we would like to obtain $\{s_1, \ldots, s_6\}$ for $DepS[s_6, t, `attack$'$]$ 
	as those that are depth-wise connected to $s_6$. 
	The path, however, contains a loop in the middle, and no  permutation 
	of $Stmt[G]$ can be a simple path of all the 6 nodes. \hfill$\clubsuit$
\end{note} 
%To say that 
Next, we define two formal conditions: {\it a minimal representation} of aspects of a statement (\mbox{Definition 
\ref{def_minimal_representation}}); and {\it redundancy} (\mbox{Definition \ref{def_redundancy}}) 
for forming a precise formal sense of `repeatedly expressing' the same statements in theme aspect argumentation 
model. 
 
\begin{definition}[Minimal representation of aspects of a statement]\label{def_minimal_representation} \rm  
	Given a theme aspect argumentation model $\langle \modell, \mathcal{D}, \mathcal{I} \rangle$,  
	we define a binary order $\preceq$ over 
	$\mathfrak{p}(D)$ to be such that,  
	for any $D_1, D_2 \in \mathfrak{p}(D)$, 
	$D_1 \preceq D_2$ holds iff both 
	$D_1 \subseteq D_2$ and $\bigwedge D_1 = \bigwedge D_2$ 
	hold. 
	Given a theme aspect argumentation model $\langle \modell, \mathcal{D}, \mathcal{I} \rangle$, 
	$s \in Stmt[G]$, $T \subseteq \pmb{Themes}$ and 
	$D' \subseteq D$, 
	we say that $D'$ is {\it a minimal representation of 
	$\mathcal{I}(T, s)$} iff 
	$D'$ is a minimal member of 
	$\{D_1 \subseteq D \mid D_1 \preceq \mathcal{I}(T, s)\}$. \hfill$\spadesuit$
\end{definition}  

\begin{note} \rm 
	In deductive argumentation \citep{Besnard01},  
	the premise-conclusion relation is such that a strict subset of the premise does not proof-theoretically 
	derive the conclusion (see \ref{subsec_logic_based_argumentation}). In Remark \ref{note_effective}, 
	we mentioned a naive (!) viewpoint which regards  
	$(\mathcal{I}(T, s), \bigwedge \mathcal{I}(T, s))$ as 
	the premise-conclusion relation for the interpretation of $s$ under $T$ in our model. 
	Assume with no loss of generality that $\{D_1, \ldots, D_n\}$ 
	is the set of all non-empty minimal representations of $\mathcal{I}(T, s)$, 
	then $(D_i, \bigwedge \mathcal{I}(T, s))$ for each $1 \leq i \leq n$ 
	satisfies the minimality condition of the premise in deductive argumentation 
	(see section \ref{subsec_logic_based_argumentation}).   
	A recent work \citep{Amgoud21} studies the minimality 
	also for the syntax of propositional formulas 
	through CNF conversion and 
	 minimisation of the number of propositional variables.\footnote{A note on the syntactic 
	minimality:  
	in Boolean algebra, 
		there is no difference among $x$, 
		$(x \vee y) \wedge (y \Rightarrow x)$, 
		$x \wedge x$, $x \wedge (y \vee \neg y)$, $\neg (\neg x)$. They are equal so long as they denote some member of the underlying 
		set.   
		In proof theory \cite{Kleene52,351148}, however, 
	        formulas are more syntactic 
		and the algebraic equality is often not enforced. 
		Hence, $x$ and $x \wedge (y \vee \neg y)$ may not be equal 
		and $x$ may be actually smaller than $x \wedge (y \vee \neg y)$.}
	%The sugggested CNF conversion into a minimal syntax, however, is 
	%not (known to be) a polynomial-time transformation as it does not 
	%introduce new propositional variables. The known polynomial-time 
	%CNF conversion can only ensure equisatisfiability between 
	%the original and the converted propositional formulas by 
	%permitting the use of fresh variables. 
	%discusses 
	%syntax-oriented minimality. However, this still admits } 
		%Note, however, that this encoding is naive 
	%as we pointed out.
	\hfill$\clubsuit$ 
\end{note} 

\begin{definition}[Redundancy]\rm \label{def_redundancy} 
	Given a theme aspect argumentation model $\langle \modell, \mathcal{D}, \mathcal{I} \rangle$, 
	$S \subseteq Stmt[G]$ and $t \in \pmb{Themes}$, we say that  
	{\it $S$ contains redundancy with respect to $t$}  iff 
	there are some $s_1, s_2 \in S$ with $s_1 \not= s_2$ and 
	some $D_1 \subseteq D$ such that 
	$D_1$ is a non-empty minimal representation of $\mathcal{I}(\{t\}, s_1)$ and of $\mathcal{I}(\{t\}, s_2)$. \hfill$\spadesuit$
\end{definition}  
\begin{note}[Illustration of redundancy] \rm 
Suppose that $\mathcal{D}$ is the Boolean algebra 
in \mbox{Figure \ref{fig_boolean_algebra}}. 
For $\{s_2, s_5, s_6\} 
\in DepS[s_6, t_1, `support$'$]$, suppose $\mathcal{I}(\{t_1\}, s_6) = \{\neg x\}$
with $\bigwedge \mathcal{I}(\{t_1\}, s_6) = \neg x$. 
	Suppose also $\mathcal{I}(\{t_1\}, s_5) = \{{x \Rightarrow y}, {\neg x \vee \neg y}\}$ with $\bigwedge \mathcal{I}(\{t_1\}, s_5) = \neg x$. Then in both \pmb{Core} and (\pmb{Core} + \pmb{das}), 
$s_5$ supports $s_6$. Now, while $s_5$ and $s_6$ effectively touch upon the same aspect $\neg x$,  
still, $\{s_5, s_6\}$ does not contain redundancy with respect 
	to $t_1$ (only 
$\{x \Rightarrow y, \neg x \vee \neg y\}$ is a minimal representation of $\mathcal{I}(\{t_1\}, s_5)$, 
and  only $\{\neg x\}$ is that of $\mathcal{I}(\{t_1\}, s_6)$). However, 
it is possible that $\mathcal{I}(\{t_1\}, s_2) = \{x \Rightarrow y, \neg x\}$ which contains 
a minimal representation $\{\neg x\}$ of $\mathcal{I}(\{t_1\}, s_6)$, leading to redundancy 
	of $\{s_2, s_5, s_6\}$ with respect to $t_1$. \hfill$\clubsuit$
\end{note} 
\begin{note}\label{note_concept_redundancy} \rm The concept of redundancy in some sense is discussed both in 
	abstract argumentation models and logic-based argumentation models (structured or unstructured). 
	In abstract argumentation models, 
	dialogue games (see section \ref{subsec_misleading_deeds} for references)  
	prohibit the same statement from appearing twice. 
	In deductive argumentation \citep{Besnard01}, 
	since it concerns proof-theoretical syntactic equality (see \ref{subsec_logic_based_argumentation}),   
	given some propositional letters $p_1$ and $p_2$, 
	$p_1 \wedge p_2$ and $p_2 \wedge p_1$, 
	or even $p_1$ and $p_1 \wedge p_1$, 
	are regarded as unequal. Hence, there is an infinite number of  
	expressions (as strings) that are mutually derivable from one other. 
	The redundancy is curtailed by choosing one representative (canonical) expression 
	for each equivalent set of syntactic expressions.
	There are works on identifying similarities between 
	logic-based argumentation arguments \citep{Amgoud18c}.   
	Compared to these works comparing two objects, the redundancy above requires  
	comparisons between two sets of minimal representations associated to 
	two objects $\mathcal{I}(\{t\}, s_1)$ and $\mathcal{I}(\{t\}, s_2)$.  \hfill$\clubsuit$ 
%	More recently,  
%	 in \cite{Amgoud18c,Amgoud22}. 
\end{note}

%\begin{proposition}[Consequence of \pmb{ar} 1]\label{prop_consequence_ar_1} 
%	Given $\langle \modell, \mathcal{D}, \mathcal{I} \rangle$,
%	if it satisfies \pmb{Core} and \pmb{ar}, then  
%	there is no $s \in Stmt[G]$ and $t \in \pmb{Themes}$ 
%	and $\mu \in \{`attack', `support'\}$ such that  
%	$(s, s) \in Rel[G]$ and that $\{t, \mu\} \subseteq \Pi((s, s))$.  
%\end{proposition} 
%\textbf{Proof.} 
% Suppose otherwise, then there is some $s \in Stmt[G]$ 
% and $t \in \pmb{Themes}$ such that $(s, s) \in Rel[G]$ and that 
% $\{t, `support'\} \subseteq \Pi((s, s))$. 
% By the first condition of \pmb{ar}, $\emptyset \subset \mathcal{I}(\{t\}, s)$. 
% By the third condition of \pmb{ar}, $\mathcal{I}(\{t\}, s) \cap 
% (\neg RlvntAspects \cup \mathcal{I}(\{t\}, s)) = \emptyset$. 
% However, $\emptyset \subset \mathcal{I}(\{t\}, s) \subseteq 
% \mathcal{I}(\{t\}, s) \cap  (\neg RlvntAspects \cup \mathcal{I}(\{t\}, s))$, 
% contradiction. 
%\hfill$\Box$ \\ 

\noindent The following constraints ensure no redundancy in a depth-statements-set and in a 
width-statements-set. 
\begin{definition}[Fresh aspect DepS constraint]\label{def_fad} \rm 
	Given a theme aspect argumentation model $\langle \modell, \mathcal{D}, \mathcal{I} \rangle$, we say it satisfies 
	constraint \pmb{faD} (\pmb{f}resh \pmb{a}spect \pmb{D}epS) iff it satisfies the following 
	condition 
	for any $s \in Stmt[G]$, any $t \in \Pi(s)$ and  
	any depth-statements-set $S$ of $s$ with respect to $t$ (for `{\it attack}' or `{\it support}'):   
	$S$ does not contain redundancy with respect to $t$.  \hfill$\spadesuit$
%	$\{  \} \cap \{   \} \not= \emptyset$ 
%	materially implies $s_1 = s_2$. 
\end{definition}  

\begin{example}[The role of \pmb{faD} in formal fallacy identification]\label{ex_role_fad} \rm  
	Let us look at a simple case. 
	Suppose $t$ `{\it Should Alice 
	consider security measures for her house?}' % to combat climate change?}' 
	and the following argumentation. 
	\begin{itemize} 
		\item Bob $s_1$: ``{\it Alice should  
			install security cameras for her house.}'' 
%			for safety.}'' 
%			We need to transition to renewable energy sources to combat 
%			cllimate change.}'' 
%			fire insurance. It costs her a lot.}''   
		\item Chris $s_2$: ``{\it But 
			it is costly to install and keep them.}'' 
%			requires an initial investment.}'' 
		\item  David $s_3$: ``{\it 
			What is costly is to keep protection 
			dogs. Keeping the security cameras 
			is not costly.}'' 
		\item Chris $s_4$: ``{\it Were you listening?
			The cost of installing and 
			maintaining them is too costly.}'' 
			%But if Alice's house catches fire, 
			%the financial loss will be huge.  
			%Alice should consider long-term 
			%protection a fire insurance can offer. 
%			%}''  
%		\item Bob $s_3$: ``{\it That's a fair point.} 
	\end{itemize}  
	Given the rhetoric, the typed argumentation graph 
	for this argumentation is:\\ 
	\begin{tikzcd}[row sep=tiny,column sep={1.6cm}]   
		s_4:\{t,t_2\} 
		\arrow{r}[above]{\{\text{`}attack\text{'},t\}}
		& s_3:\{t,t_3\} 
		\arrow{r}[above]{\{\text{`}attack\text{'},t\}}
		& s_2:\{t,t_2\}
		\arrow{r}[above]{\{\text{`}attack\text{'},t\}}
		& s_1:\{t,t_1\}
		%\arrow{r}[above]{\{\text{`}attack\text{'},t_2\}} & s_7:\{t_2,t_7\}
		% \arrow{r}[above]{\{\text{`}support\text{'}, t_2\}} & s_6:\{t_2,t_6\}.  
%		aCostH \arrow[-,ur]\arrow[-,dr] & 	  & \neg aCostH \arrow[-,ul]\arrow[-,dl] & bCostH \arrow[-,ur]\arrow[-,dr] & & \neg bCostH \arrow[-,ul]\arrow[-,dl]\\
%		& \pmb{0} & & & \pmb{0} \\ 
%                & \pmb{1} &   & & \pmb{1} \\ 
%		aP \arrow[-,ur]\arrow[-,dr] & 	  & \neg aP \arrow[-,ul]\arrow[-,dl] & bP \arrow[-,ur]\arrow[-,dr] & & \neg bP 
%		\arrow[-,ul]\arrow[-,dl]\\
%		& \pmb{0} & & & \pmb{0} 
	\end{tikzcd}
	{\ }\\
	\noindent where $t_3$ denotes `David', $t_2$ `Chris' and $t_1$ `Bob'.  Suppose $\mathcal{D}$ is large enough, 
	and suppose $x_1$: `{\it Alice}',
	$x_2$: `{\it house}', 
	$x_3$: `{\it installing security 
	cameras}', $x_4$: `{\it keeping security 
	cameras}', $x_5$: `{\it keeping protection dogs}',  
	and $x_6$: `{\it costing a lot}' are in $D$. 
	It is assumed that none of these aspects are comparable 
	in $\mathcal{D}$.  
	We suppose the following interpretation $\mathcal{I}$. 
	$\mathcal{I}(\{t\}, s_1) = \{(x_1 \wedge x_2)  
	\Rightarrow (x_1 \wedge x_3)\}$, 
	$\mathcal{I}(\{t\}, s_2) = \{(x_3 \wedge x_4) 
	\Rightarrow x_6\}$,  
	$\mathcal{I}(\{t\}, s_3) = \{x_5 \Rightarrow x_6, 
	x_4 \Rightarrow \neg x_6\}$,  
	$\mathcal{I}(\{t\}, s_4) = \{(x_3 \wedge x_4) 
	\Rightarrow x_6\}$.  
	Chris, while not repeating the exact 
	same statement, touches upon the same effective aspect. 
	This is {\it Ignoratio Elenchi} as is so called 
	within the repertoire of informally classified fallacies  
	since he ignores and avoids engaging with the aspects raised 
	by David. 
	 
	Let $\taam$ denote the theme aspect argumentation model 
	consisting of the typed argumentation graph given above 
	and a pair of $\mathcal{D}, \mathcal{I}$ 
	satisfying the above requirements and 
	$\modell$'s \pmb{Core}-normality with respect to 
	$\mathcal{D}, \mathcal{I}$. 
%	possible 
%	that $\taam$ be \pmb{Core}-normal. 
%	That is, it is possible that the above argumentation  
%	is not identified fallacious with \pmb{Core} alone. 
	$\{s_1, \ldots, s_4\}$ 
	contains redundancy with respect to $t$, 
	so it is a $\{\text{\pmb{faD}}\}$-fallacy 
	with respect to $\mathcal{D}, \mathcal{I}$. 
	\hfill$\clubsuit$ 
\end{example} 

\begin{note}[Why is \pmb{faD} not a core constraint?]   \rm 
       It is not uncommon to continue argumentation 
       by reminding the audience of previously stated aspects before providing new aspects. 
	In real-life argumentation with a large depth-statements-set, occasional reminders help the audience understand 
	the earlier statements better.  
       Nonetheless, \pmb{faD} presupposes perfect recall on argumentation participants.  \hfill$\clubsuit$ 
\end{note}

\noindent Constraints that are similar to \pmb{faD} but for a width-statements-set can be defined, preventing 
a deliberate strengthening or weakening 
with the same aspect(s) of a statement.

\begin{definition}[Fresh aspect WidS constraint] \label{def_faw} \rm  
Given a theme aspect argumentation model $\langle \modell, \mathcal{D}, \mathcal{I} \rangle$, we say it 
	satisfies constraint \pmb{faW} (\pmb{f}resh \pmb{a}spect \pmb{w}idS)  
	iff it satisfies the following condition for any $s \in Stmt[G]$, 
	any $t \in \Pi(s)$ and any width-statements-set $S$ of $s$ 
	with respect to $t$ (for `$attack$' or `$support$'): $S$ does not 
	contain redundancy with respect to $t$. \hfill$\spadesuit$
%	and every $t \in \Pi(s)$: for every $s_1$ and $s_2$ 
%			in a width-statements-set of $s$ with respect to 
%			$t$ (for $`attack'$ or for $`support'$), 
%			$\mathcal{I}(\{t\}, s_1) \cap \mathcal{I}(\{t\}, s_2) 
%			\not= \emptyset$ materially implies 
%			$s_1 = s_2$. 

%	\begin{itemize} 
%		\item  If $WidS[s, t, `attack'] \not= \emptyset$, 
%			then 
%			for every $s_1$ and $s_2$ 
%			in the width-statements-set of $s$ with respect to 
%			$t$ (for $`attack'$ or for $`support'$), 
%			$\mathcal{I}(\{t\}, s_1) \cap \mathcal{I}(\{t\}, s_2) 
%			\not= \emptyset$ materially implies 
%			$s_1 = s_2$. 
		%\item If $widS[s, t, `support'] \not= \emptyset$, 
		%	then for every $s_1, s_2 \in widS[s, t, `support'] \cup \{s\}$, 
		%	 $\mathcal{I}(\{t\}, s_1) 
		%	 \cap \mathcal{I}(\{t\}, s_2) \not= \emptyset$ 
		%	 materially implies $s_1 = s_2$.  
%	\end{itemize}
\end{definition}   
\begin{note}[Why is \pmb{faW} not a core constraint?] \rm 
	Pedantic enforcement of \pmb{faW} would prohibit 
	a simple agreement ``I (do not) like [this aspect] of the idea.''    
	to be expressed twice to a given statement, which 
	is unreasonable in practice, since 
	some statement may refer to the fact that 
	everyone (, a majority, and so on) has agreed to a certain statement (as a summary of the theme 
	given by the statement). 
	There, counting is necessary which \pmb{faW} 
	does not permit.\hfill$\clubsuit$  
\end{note} 

Let \pmb{F} denote the constraints 
\pmb{faW} and \pmb{faD} collectively. 
Then, we have \pmb{tr}, \pmb{nnp}, \pmb{nsa}, \pmb{kos}, \pmb{nss}
as the constraints purely on the typed argumentation graph, 
and \pmb{Core}, \pmb{E}, \pmb{das}, \pmb{nwci} and \pmb{F} as the constraints 
on the theme aspect argumentation model. The constraints 
are pairwise independent. 
\begin{proposition}[{\hyperref[proof_deciding_subtest_and_subtestall]{Pairwise independence of constraints}}] 
	\label{prop_pairwise_independence} 
	Let \pmb{x} and \pmb{y} be some constraints 
	either on the typed argumentation graph or on the theme aspect argumentation model, 
	then 
	$\pmb{x} \not= \pmb{y}$ holds iff 
	there is some theme aspect argumentation model $\taam$ such that 
	the satisfaction statuses of $\pmb{x}$ and $\pmb{y}$ 
	do not coincide.  
	%satisfaction of $\pmb{x}$ by 
	%$\modell$ or $\taam$ 
%
%
%	 satisfies 
%	\pmb{x} 
%	 
%	there is no $\taam$ such that 
%	both $\pmb{x}$ and $\pmb{y}$ are simultaneously satisfied by 
%	$\modell$ or $\taam$ 
%	either: satisfies both $\pmb{x}$ and $\pmb{y}$; or unsatisfies both of them. 
%
%
%	holds iff 
%	The following holds for any constraint $\pmb{x}$ either on the typed graph or on the theme aspect argumentation model 
%	and any constraint $\pmb{y}$ either on the typed graph or on the theme aspect argumentation model: 
%	there is no $\taam$ such that 
%	both $\pmb{x}$ and $\pmb{y}$ are satisfied by 
%	$\modell$ or $\taam$ 
%	either: satisfies both $\pmb{x}$ and $\pmb{y}$; or unsatisfies both of them. 
	\end{proposition} 
Existence holds, too. 
\begin{theorem}[Existence] There is some theme aspect argumentation 
	model $\taam$ satisfying \pmb{Core}, \pmb{E}, \pmb{das}, \pmb{nwci} 
	and \pmb{F}. 
\end{theorem}  
\vspace{-0.26cm} 
\textbf{Proof.} By the same setting in the \hyperref[proof_existence]{proof} of \mbox{Theorem \ref{thm_existence}}. \hfill$\Box$ \\\\
For satisfaction, the following direction is guaranteed. %The other direction is
%unclear to us at this moment. 
%\noindent We claim the following. The point of it is not so much to show that such existence 
%is guaranteed when all these optional constraints are adopted in addition to \pmb{Core} 
%as to show the guarantee for any combination that extends \pmb{Core} (as 
%follows immediately from the claim). 
\begin{proposition}[Satisfaction of \pmb{Core} + E +
	\pmb{das} + \pmb{nwci} + \pmb{F} (Necessary condition)] \label{thm_model_existence_for_every_axiom}     
	Given a typed argumentation graph $\modell$, it satisfies 
	\pmb{tr}, \pmb{nnp}, \pmb{nsa}, \pmb{kos}, and \pmb{nss} %\pmb{nss} and \pmb{ntst} 
	if there is some pair of $\mathcal{D}$ and $\mathcal{I}$ 
	such that $\langle \modell, \mathcal{D}, \mathcal{I} \rangle$ satisfies 
	\pmb{Core}, \pmb{E},  \pmb{das}, \pmb{nwci}, and \pmb{F}. 
\end{proposition}  
\vspace{-0.26cm} 
\textbf{Proof.} The contrapositive follows from  
\mbox{Proposition \ref{prop_consequence_aass}}, 
the first two sub-results of \mbox{Theorem \ref{thm_consequence_pr_vi_aass}}, \mbox{Proposition 
\ref{prop_consequence_pr_vi_manss_aass}}, and \mbox{Proposition \ref{prop_consequence_das}}. \hfill$\Box$ \\
%\textbf{Proof.} \pmb{E} is satisfied by the $\mathcal{D}$ and $\mathcal{I}$ constructed for \mbox{Theorem \ref{thm_core}}. 
%\hfill$\Box$ \\ 

%\noindent Intuition speaks that the following claim also holds. 
%\begin{claim}[Uniqueness]    
%	\ryuta{need to change this.} 
%	Let $\alpha$ denote any subset of $\{\pmb{Core}, \pmb{E}, \pmb{IC}, \pmb{ar}, \pmb{FA}\}$ that includes \pmb{Core}. 
%        Every two distinct combinations result in two distinct sets of models such that neither is subsumed in the other.
%\end{claim} 
%\ryuta{This. needs to be defined for all the other preceding parts.
%\noindent The constraints we have identified in this and the previous sections 
%may be regarded as the semantics of fallacy detection. More broadly, they 
%define normal forms of typed argumentation graphs $\modell$ with respect to 
%$\mathcal{D}$ and $\mathcal{I}$.  

%\begin{definition}[Normal forms and fallacies]
%	Let $\alpha$ denote any subset of $\{\text{\pmb{Core}}, \pmb{E}, \pmb{das}, \pmb{nwci}, \pmb{F}\}$ 
%	that includes \pmb{Core}.  
%	Given a theme aspect argumentation model $\langle \modell, \mathcal{D}, \mathcal{I} \rangle$, 
%	we say the typed argumentation graph $\modell$ is: 
%	{\it $\alpha$-normal with respect to $\mathcal{D}, \mathcal{I}$} iff $\langle \modell, \mathcal{D}, \mathcal{I} \rangle$ 
%	satisfies every member of $\alpha$. 
%	We say {\it $\modell$ is an $\alpha$-fallacy with respect to $\mathcal{D}, \mathcal{I}$} iff $\modell$ is not $\alpha$-normal 
%	with respect to $\mathcal{D}, \mathcal{I}$. \hfill$\spadesuit$
%\end{definition}  
%}
\noindent It becomes harder and harder to satisfy 
a set of constraints as it becomes larger. This intuition 
is proper: 
\begin{theorem}[Monotonicity]  
       Let $\alpha$ and $\beta$ denote any subset of 
	$\{\pmb{Core}, \pmb{E}, \pmb{das}, \pmb{nwci}, \pmb{F}\}$  
	that includes \pmb{Core}. 
	Given a theme aspect argumentation model $\langle \modell, \mathcal{D}, \mathcal{I} \rangle$,  
	$\alpha \subseteq \beta$ and $\modell$ being $\beta$-normal with respect to 
	$\mathcal{D}, \mathcal{I}$ materially 
	imply $\modell$ being $\alpha$-normal with respect to $\mathcal{D}, \mathcal{I}$. 
\end{theorem}   
\vspace{-0.26cm} 
\textbf{Proof.} None of the constraints modifies other constraints. 
\hfill$\Box$ \\
\begin{note} \rm It is obvious that this result holds if we drop  \pmb{Core},  save we do not care much about pre-\pmb{Core} normal forms.  \hfill$\clubsuit$
\end{note} 
\subsection{Logico-rhetorical conclusion for logical fallacy identification} \label{sec_logico_rhetorical_acceptance} 
There are fallacies that are driven more by one's logical errors. %An example is {\it argumentum ad ignorantiam}, 
%where one concludes from an unproved statement that its opposite holds. 
%This includes the claim 
%that God exists because it is not proven that God does not exist. 
With the theme aspect argumentation model, identification of this sort of logical fallacy 
can be done smoothly, and generally, since it models the rhetoric of an argumentation in typed argumentation graph  
$\modell$ 
and the semantics of it in $\mathcal{D}, \mathcal{I}$. What can be inferred as a conclusion 
of $\modell$ is in fact a {\it rhetorical conclusion} ({\it acceptability semantics} of abstract argumentation models \citep{Dung95}), 
that is to say, a conclusion on the surface. 
This rhetorical conclusion taken as a whole should be free of logical inconsistency. But the logical consistency 
check is readily done, since we have access to the semantics of the conclusion. 
A rhetorical conclusion with its logical consistency also checked is a {\it logico-rhetorical} conclusion (Definition 
\ref{def_logico_rhetorical_conclusion}). 

Let us go into detail. There are a number of types of rhetorical conclusions 
(acceptability semantics) in the literature 
{\it e.g.} \citep{Baroni07} for typical ones for `$attack$'s,  and 
\citep{Cayrol13,ArisakaSantini19,Arisaka2020} for `$attack$'s and `$support$'s. For our purpose, 
it matters little what specificities each of them assumes. We only 
need a broad perspective - their purpose - 
that they choose subsets of them from among all the statements in argumentation graph, say 
$\{s_1, s_2, s_3\}$, $\{s_3, s_4, s_5\}$ and $\{s_1, s_6\}$, 
which are disjunctively understood to be rhetorically acceptable 
(that is, the first set, or the second set, or the third set of statements is rhetorically acceptable), 
except in our case any conclusion is tied to a chosen set of themes.
%\begin{note}\rm Newer schools 
%have named various other entities acceptability semantics. For example, in a recent work \cite{Amgoud22}, 
%each member of the acceptability semantics in the above traditional sense is 
%described as (labelled) acceptability semantics, and some others, {\it e.g.} \cite{ADSV22}, term a set 
%of sub-graphs of a given argumentation graph 
%the acceptability semantics. We stick to the traditional one in this paper. \hfill$\clubsuit$ 
%\end{note} 

Formally, the following is the definition of rhetorical conclusion in theme aspect argumentation model. 
We generalise the sub-theme typed argumentation graph notation $(G_t, \Pi_t)$ 
introduced earlier in Definition \ref{def_sub_theme}. Given a typed argumentation graph $\modell$ and $T \subseteq \pmb{Themes}$,  
by $G_T$ we denote $(\bigcup_{t \in T} Stmt[G_t], \bigcup_{t \in T} Rel[G_t])$ 
and by $\Pi_T$ we denote the following typing function: for every $x \in Stmt[G_T] \cup Rel[G_T]$, 
			$\Pi_T(x) = \Pi(x) \cap (T \cup \{`attack$'$, `support$'$\})$, and for 
			every $x \not\in Stmt[G_T] \cup Rel[G_T]$, $\Pi_T(x) = \emptyset$. 
\begin{definition}[Rhetorical conclusion]\label{def_rhetoric_conclusion} \rm 
	Given a typed argumentation graph $\modell$ and some $T \subseteq \pmb{Themes}$, suppose an acceptability semantics $\mathfrak{s}$ 
	which takes $\modell$ and $T$ as its inputs and which outputs a subset of 
	$\mathfrak{p}(\mathfrak{p}(Stmt[G]))$. 
	We say $\mathfrak{s}(\modell, T)$ is {\it the rhetorical conclusion 
	of $\modell$ with respect to  $T$ and $\mathfrak{s}$}. 
	We define that the rhetorical conclusion of $\modell$ 
	with respect to $T$ and $\mathfrak{s}$ is $\emptyset$ if 
	$(G_T, \Pi_T)$ is not well-defined.  \hfill$\spadesuit$
\end{definition} 
Now, let $\bigvee D_1$ for any $D_1 \subseteq D$ be a supremum of $D_1$ if $D_1 \not= \emptyset$; 
and $\emptyset$ if $D_1 = \emptyset$, then: 
\begin{definition}[Logico-rhetorical conclusion]\label{def_logico_rhetorical_conclusion} \rm 
%	Given $\modell$ and $T \subseteq \pmb{Themes}$, %suppose an acceptability semantics $\mathfrak{s}$ 
%	which takes $\modell$ and some $T \subseteq \pmb{Themes}$ as its inputs and outputs a subset of 
%	$\mathfrak{p}(\mathfrak{p}(Stmt[G]))$. 
%	we say $\mathfrak{s}(\modell, T)$ is the rhetorical conclusion 
%	of $\modell$ with respect to  $T$ and $\mathfrak{s}$.  
	Given a theme aspect argumentation model $\langle \modell, \mathcal{D}, \mathcal{I} \rangle$, 
	some $T \subseteq \pmb{Themes}$ 
	and $\mathfrak{s}$, we say 
	$\bigvee_{S \in \mathfrak{s}(\modell, T)}\bigwedge_{s \in S}\mathcal{I}(T, s)$ 
	%\{x \in D \mid x = \bigwedge \mathcal{I}(T, s)\}$ 
	is {\it the logico-rhetorical conclusion of $\taam$ with respect to $T$ and 
	$\mathfrak{s}$}. We define that 
	the logico-rhetorical conclusion of $\taam$ with respect to $T \subseteq \pmb{Themes}$ and $\mathfrak{s}$ 
	is $\emptyset$ if there is no $\emptyset  \subset S \subseteq Stmt[G_T]$ such that 
	$S \in \mathfrak{s}(\modell, T)$. \hfill$\spadesuit$ 
\end{definition}

We obtain the following nice property that, unless 
a logico-rhetorical conclusion does not exist ({\it i.e.} is $\emptyset$),  
it is always found on $\mathcal{D}$. 
\begin{theorem}[Existence on $\mathcal{D}$] 
	Given a theme aspect argumentation model $\langle \modell, \mathcal{D}, \mathcal{I} \rangle$, 
	some $T \subseteq \pmb{Themes}$
	and $\mathfrak{s}$, the logico-rhetorical conclusion
	of $\taam$ with respect to $T$ and $\mathfrak{s}$ 
	is a member of $D$ iff there is some $\emptyset \subset S \subseteq Stmt[G_T]$ such that 
	$S \in \mathfrak{s}(\modell, T)$. 
	%$\mathfrak{s}(\modell) \not= \emptyset$ 
	%materially implies that the logico-rhetorical 
	%conclusion of $\taam$ with respect to $\mathfrak{s}$ 
	% is some $x \in D$. 
\end{theorem} 
\vspace{-0.26cm} 
\textbf{Proof.} Obvious. \hfill$\Box$
\begin{note}[One warning.] \label{note_warning} \rm   
     This property holds for $\taam$ because the semantic structure of $\taam$ 
	is a Boolean algebra with  well-defined algebraic operations $\wedge$ and $\vee$. 
	It is not generally true for other structures with 
	no conditions on the existence of supremum and infimum of its elements. \hfill$\clubsuit$
\end{note} 
We make use of the logico-rhetorical conclusion for checking the presence of a fallacy in a rhetorical conclusion. 
\begin{definition}[Fallacy of a rhetorical conclusion] \label{def_fallacy_rhetoric_conclusion} \rm 
	Given a theme aspect argumentation model $\taam$ and some $\mathfrak{s}$, we say 
	$(\modell, \mathfrak{s})$ is {\it a logical fallacy with respect to  
	$\mathcal{D}, \mathcal{I}$} iff  
	there is some $T \subseteq \pmb{Themes}$  
	such that the logico-rhetorical conclusion of $\taam$ with respect to $T$ and $\mathfrak{s}$ 
	is $\pmb{0}$. \hfill$\spadesuit$
\end{definition} 
As repeatedly observed in the literature of abstract argumentation models, 
a rhetorical conclusion 
by an acceptability semantics, when it is used to draw a conclusion about the content of an argumentation, 
can result in a logically problematic conclusion. %Though, one 
%may simplistically declare  that it is a fallacy to claim that an acceptability semantics 
%in abstract argumentation whose purpose is really to draw a rhetoric conclusion is equally 
%suitable for drawing a logical conclusion, the pessimism is only half true and ultimately vague.  
\mbox{Definition \ref{def_fallacy_rhetoric_conclusion}}  
provides a formal characterisation of precisely when it is logically fallacious and when it is not.  
All in all, fallacy check on a typed argumentation graph $\modell$ 
%\langle \modell, \mathcal{D}, 
%\mathcal{I} \rangle$ 
with respect to $\mathcal{D}, \mathcal{I}$ can proceed 
in two steps, by first checking $\alpha$-normality for some $\alpha$ 
(if it is not $\alpha$-normal, $\modell$ is an $\alpha$-fallacy with respect to 
$\mathcal{D}, \mathcal{I}$), and by 
then checking whether $(\modell, \mathfrak{s})$ with a chosen $\mathfrak{s}$ 
is a logical fallacy with respect to $\mathcal{D}, \mathcal{I}$. 
However, if we detect a fallacy in the first step, there is no need 
of the second step, achieving incremental fallacy checks.

\section{Computational Complexity} \label{sec_computational_complexities}
%We have experimentally obtained the likelihood that satisfiability of all the constraints on the typed graph 
%can be efficiently computed. In this section, we formally show 
%that those satisfiability problems are indeed in P. 
%On the other hand, checking whether a given $\taam$ is $\{$\pmb{Core}, $\Gamma\}$-normal with respect to 
%$\mathcal{I}, \mathcal{D}$ for $\Gamma \subseteq 
%\{$\pmb{E},\pmb{das},\pmb{nwci},\pmb{F}$\}$ has turned out to be not as efficient. 
%We formally establish that the satisfiability of \pmb{Core} is $\mathsf{\Pi}_2$-complete 
%by first proving that 
%$\{$\pmb{Core}, \pmb{E}, \pmb{das}, \pmb{nwci}$\}$ is. 
%Upper complexity bounds 
%of \pmb{faW} and \pmb{faD} 
%will be also given. %More detailed complexity results 
%such as tighter complexity bounds 
%of individual constraints and, practically importantly, 
%identification of efficiently computable sub-decision problems  
%are left to future work. 
%We formulated the formal constraints in Sections \ref{sec_core_constraints_of_theme_aspect_argumentation_model} and \ref{sec_more_constraints}, 
%illustrating their role through examples.   
But how expensive is it to check the $\alpha$-normality? 
To answer this question, let us identify the computational complexities of deciding the satisfiability 
of the constraints introduced in Sections \ref{sec_core_constraints_of_theme_aspect_argumentation_model} and \ref{sec_more_constraints} 
%and 
%demonstrated formal fallacy identification through them. 
%We have not yet analysed the computational 
%complexities of deciding the satisfiability 
%of the formal constraints. %  \ryuta{Must be something, and the sentence must be complete.} 
%on this section. 
%verifying them. 
%In this section, therefore, we identify    
%the complexity for the constraints %introduced in Sections 
%\ref{sec_core_constraints_of_theme_aspect_argumentation_model} and \ref{sec_more_constraints} under 
under a set of restrictions on the theme aspect argumentation model (to be mentioned later).   
We assume readers are familiar with the terminologies of 
complexity theory. 
For the oracle complexity classes, we follow the 
notational convention in \citep{Wegener05}.  
In particular, 
%We assume certain familiarity with the key concepts 
%of the computational complexity \citep{Wegener05} 
%on the part of readers. In particular, 
%we assume the knowledge of 
%the basic complexity classes P, NP and \mbox{co-NP}, 
%that of the complexity of well-known  
%decision and other algorithmic problems such as SAT (is in NP), TAUT (is in co-SAT = co-NP), 
%the shortest path problem for a graph (in NP-hard) or for a direct acyclic graph (in P), 
%or the awareness of the difference between polynomial-time many-one reducibility and 
%polynomial-time Turing reducibility. All of them are in the textbook \citep{SipserWegener05}. 
%Furthermore, 
suppose X is a decision problem, then 
P(X) is the complexity class comprising 
all the decision problems that can be decided 
by a polynomial-time algorithm with access to an oracle 
for X. Suppose C is a complexity class comprising decision problems, 
then P(C) = $\bigcup_{X \in C}$ P(X).  
Similarly, for a decision problem X, 
NP(X) is the complexity class comprising 
all the decision problems that can be decided  
in polynomial time 
by a non-deterministic algorithm with access to an oracle for X. 
If C is a complexity class, 
then NP(C) = $\bigcup_{X \in C}$ NP(X). 
%Similarly for NP(X) and NP(C) for a given decision problem X or a given complexity class 
%C with a non-deterministic polynomial-time algorithm. co-NP(X) and co-NP(C)  
%are 
 P($\underbrace{\text{NP(NP($\cdots$)}}_{\text{NP } n \text{ times.}}$)) is often denoted by $\mathsf{\Delta}_{n+1}$, 
NP($\underbrace{\text{NP(NP($\cdots$)}}_{\text{NP } n \text{ times.}}$)) by $\mathsf{\Sigma}_{n+1}$ and 
co-NP($\underbrace{\text{NP(NP($\cdots$)}}_{\text{NP } n \text{ times.}}$)) by $\mathsf{\Pi}_{n+1}$.  
Also, $\mathsf{\Delta}_0,  \mathsf{\Sigma}_0,$ and $\mathsf{\Pi}_0$ are defined to be P. 
Given X and Y each of which is either an algorithmic problem or 
a complexity class, we will write X = Y to mean that any algorithmic problem in either of them is 
polynomial-time many-one reducible to 
an algorithmic problem in the other. If X is some complexity class, 
X = Y means Y is X-complete. 
%However, we will use these symbols sparingly and only when 
%there is a space issue.
%For the polynomial hierarchy, 
%suppose C is a computational complexity class, 
%then P(C), NP(C) and co-NP(C) are also computational complexity 
%classes  

%is, what the computational complexity classes P, NP and co-NP 
%are, 

%Let us recall some key notions and results 
%from the computational complexity 
%literature \citep{Wegener05}. 
%A decision problem L with its input $x$ of length $n$ is: 
%\begin{itemize} 
%	\item in the computational complexity 
%class P just when: it can be solved by an algorithm; 
%and there is a polynomial $f$ 
%such that the number of steps 
%the algorithm needs to solve it is bounded by   
%		$f(n)$ for some polynomial $f$. 
%	\item in the computational complexity class NP 
%	just when: there is a decision problem L' in P 
%		with its input $x' \equiv (x, 
%		\{0, 1\}^{|g(n)|})$ such that 
%	    
%\end{itemize} 
%
%%
\begin{lemma}[from 10.4.2 in \citep{Wegener05}] \label{lem_subsumption_complexity_classes} 
	The following are known to hold for any $0 \leq n$. 
	\begin{itemize} 
		\item $\mathsf{\Delta}_n$ = co-$\mathsf{\Delta}_n$ = P$(\mathsf{\Delta}_n)$ 
			$\subseteq$ $\mathsf{\Sigma}_n \cap \mathsf{\Pi}_n \subseteq 
			\mathsf{\Sigma}_n \cup \mathsf{\Pi}_n \subseteq 
			\Delta_{n+1}$ = P$(\mathsf{\Pi}_n)$. 
%			P $\subseteq$ (NP $\cap$ co-NP) 
%			$\subseteq$ (NP $\cup$ co-NP) 
%			$\subseteq$ P(NP) $\subseteq$ \\
%			(NP(NP) $\cap$ co-NP(NP)) $\subseteq$ 
%			P(NP(NP)) $\subseteq$\\ NP(NP(NP) $\cap$ 
%			co-NP(NP)) $\subseteq \cdots$.  
%			\item P(NP) = P(co-NP) = P(P(NP)).  
%				\item NP(NP) = NP(co-NP) = NP(P(NP)).  
		\item $\mathsf{\Sigma}_{n+1}$ = NP$(\mathsf{\Pi}_n)$ = NP$(\mathsf{\Delta}_{n+1})$. 
		\item $\mathsf{\Pi}_{n+1}$ = co-NP$(\mathsf{\Pi}_n)$ = co-NP$(\mathsf{\Delta}_{n+1})$. 
%		\item co-NP(NP) = co-NP(co-NP) = co-NP(P(NP)).      
						%\item P(NP) = P(co-NP) = P(P(NP)). 
	%	\item P(P(C)) = P(C) and NP(P(C)) = NP(C) 
	%		for every 
	%		complexity class C.   
%		\item P $\subseteq$ NP and P $\subseteq$ co-NP. 
	\end{itemize} 
\end{lemma} 

\noindent Apart from the background knowledge, 
we enforce the following conditions 
on $\taam$, in particular on $\mathcal{D}$ and $\mathcal{I}$, 
so the length of inputs to them will be 
bounded by a polynomial of 
$(|Stmt[G]| + |Rel[G]| + 
|\pmb{Themes}| + \log(|\pmb{Atom}(\mathcal{D})|) + \kappa)$. 
The $\kappa$ acts as padding. It is some natural number that is set to 0 normally.
A non-zero value is used only for 
establishing hardness results. %polynomial-time many-one reducibility  
%of other decision problems to our satisfiability problem (in this paper, only to that of \pmb{aass}). 
%If unbounded, an input can require an infinite length in the first place. For example, 
%$\mathcal{D}$ can be an infinite lattice. 
%$\mathcal{I}$ can be also chosen in the way that its outputs on given inputs are an arbitrary 
%infinite set of elements of $D$.
%As some conditions that need to be verified for some constraint \pmb{$x$} on $\taam$ 
%require set comparisons for subsumption judgement and others, it holds that not all constraints on 
%any arbitrary $\taam$ are decidable.
%Even if it is finite, it can still become exponentially long with respect to $n$. 
%We thus assume the following conditions on $\taam$ throughout this section. 
%\begin{definition}[Representation of $\mathcal{D}$] \rm 
% $\mathcal{D}$ satisfies $|\pmb{Atom}(\mathcal{D})| = 2^m$ 
%		for some finite $m$. 
%		$\mathcal{D}$ is represented by a list of 
%		$\log(|\pmb{Atom}(\mathcal{D})|)$ (= $m$) 
%		words. Assume it is $[x_1, \ldots, x_m]$, 
%		then for every member of $\pmb{Atom}(\mathcal{D})$, 
%		it is algebraically equal to some Boolean algebraic 
%		expression $z_1 \wedge \cdots \wedge z_m$ 
%		where, for each $1 \leq i \leq m$, 
%		$z_i$ is either $x_i$ or else $\neg x_i$. 
%\end{definition} 

\begin{itemize} 
	\item $\mathcal{D}$ satisfies $|\pmb{Atom}(\mathcal{D})| = 2^m$ 
		for some finite $m$. 
		$\mathcal{D}$ is represented by a list of 
		$\log(|\pmb{Atom}(\mathcal{D})|)$ (= $m$) 
		words. Assume it is $[w_1, \ldots, w_m]$, 
		then for every member of $\pmb{Atom}(\mathcal{D})$, 
		it is algebraically equal to some Boolean  
		expression $z_1 \wedge \cdots \wedge z_m$ 
		where, for each $1 \leq i \leq m$, 
		$z_i$ is either $w_i$ or else $\neg w_i$.\footnote{A note on the `algebraic equality': in Boolean algebra, 
		there is no difference among $x$, 
		$(x \vee y) \wedge (y \Rightarrow x)$, 
		$x \wedge x$, $\neg (\neg x)$. They are equal.  
		However, the computational representation  
		is a string and does not perforce enforce the algebraic 
		equality.   
		We are making it explicit that  
		the sense of the equality here is of the algebraic equality. 
		}  
		 
	\item Let $\sigma$ be a polynomial-time computable function  
		which takes inputs (1) a list of words $[w_1, \ldots, w_m]$ 
		and (2) a $m$-bit pattern (in the $m$-bit sequence $\{0,1\}^m$) 
		for outputting a list of words $[w_{i_1}, \ldots, w_{i_{end}}]$ 
		satisfying the following conditions:%\pagebreak
		\begin{itemize} 
			\item $\{w_{i_1}, \ldots, w_{i_{end}}\} \subseteq \{w_1, \ldots, w_m\}$. 
			\item for any $1 \leq j \leq m$ and any $w_j \in \{w_{i_1}, \ldots, w_{i_{end}}\}$,  
				the $j$-th bit in the bit pattern is 1. 
			\item for any $1 \leq j \leq m$ and any $w_j \not\in \{w_{i_1}, \ldots, w_{i_{end}}\}$,  
				the $j$-th bit in the bit pattern is 0. 
		\end{itemize}
		%$(\bigcup_{n \in \mathbb{N}}\{0, 1\}^n) \times 
		%\mathcal{D} \rightarrow \mathcal{D}$ 
		%Given $\mathcal{D}$ represented by $[x_1, \ldots, x_m]$, 
		%we denote by a $m$-bit sequence $\{0, 1\}^m$ 
		%a Boolean algebra $\mathcal{D}'$ represented by 
		%$\delta(\{0, 1\}^m, [x_1, \ldots, x_m])$. 
	 $\mathcal{I}$ is a polynomial-time 
		computable function. 
			For every $T \subseteq \pmb{Themes}$ and 
				every $s \in Stmt[G]$, 
				%\cup \{\omega\}$, 
				$\mathcal{I}(T, s)$ is  
				a string representing a set 
				of propositional Boolean formulas
				%either a string representing a set 
				%of Boolean algebraic 
				%expressions,
				and $\mathcal{I}(T, \omega)$ is 
				either such a string or 
				a $\log (|\pmb{Atom}(\mathcal{D})|)$-bit 
				pattern (in 
				the 
				$\log (|\pmb{Atom}(\mathcal{D})|)$-bit sequence $\{0,1\}^{\log (|\pmb{Atom}(\mathcal{D})|)}$).  
				If $\mathcal{I}(T, \omega)$ is a bit pattern, 
				let $b$ denote 
				the bit pattern. Then, 
				$b$ is considered to denote a Boolean algebra represented by a list of words  
				$\sigma([w_1, \ldots, w_{\log(|\pmb{Atom}(\mathcal{D})|)}], b)$. 
\end{itemize} 

\begin{example}[Representation of $\mathcal{D}$ and $\mathcal{I}$]  \rm 
	Let us suppose the following Boolean algebra $\mathcal{D}$.\\
%        \begin{figure}[!t]  
	\begin{center} 
%	\begin{wrapfigure}{l}[3pt]{3cm}
	\resizebox{6cm}{!}{ 
	\begin{tikzpicture} %lxy    x means the level (down the greater), y means the width (right the greater) 
		\node (top) at (0, 0) {$\pmb{1}$}; 
		%%%%%%%l1 starts 
		\node [below left of=top] (l12) {$y \Rightarrow x$}; 
		\node [below right of=top] (l13) {$x \Rightarrow y$}; 
		\node [left of=l12,xshift=-0.5cm] (l11) {$x \vee y$}; 
		\node [right of=l13,xshift=0.5cm] (l14) {$\neg x \vee \neg y$};  
		\draw [out=-120,in=90] (top) to (l12);
		\draw [out=-60,in=90] (top) to (l13);
		\draw [out=-160,in=40] (top) to (l11);
		\draw [out=-20,in=130] (top) to (l14);
		%%%%%%%%l2 starts
		\node [below of=l11] (l22) {$y$}; 
		\node [left of=l22,xshift=-0.1cm] (l21) {$x$};
		\node [below of=l12] (l23) {$x \Leftrightarrow y$};  
		\node [below of=l13] (l24) {$x \Leftrightarrow \neg y$}; 
		\node [below of=l14] (l25) {$\neg y$}; 
		\node [right of=l25,xshift=0.1cm] (l26) {$\neg x$};   
		\draw (l11) to (l22);
		\draw [out=-160,in=30] (l13) to (l22);
		\draw (l12) to (l23);
		\draw [out=-110,in=40] (l13) to (l23); 
		\draw (l14) to (l25);
		\draw [out=-160,in=60] (l11) to (l21);
		\draw [out=-160,in=30] (l12) to (l21);
		\draw [out=-30,in=150] (l12) to (l25);
		\draw [out=-10,in=100] (l14) to (l26);
		\draw [out=-20,in=150] (l13) to (l26);
		\draw [out=-30,in=150] (l11) to (l24);
		\draw [out=-140,in=50] (l14) to (l24);
		%%%%%%%l3 starts  
		\node [below of=l22] (l31) {$x \wedge y$}; 
		\node [below of=l23] (l32) {$x \wedge \neg y$};
		\node [below of=l24] (l33) {$\neg x \wedge y$};  
		\node [below of=l25] (l34) {$\neg x \wedge \neg y$};   
		\draw (l25) to (l34);
		\draw [out=20,in=-150] (l32) to (l25);
		\draw (l24) to (l33);
		\draw [out=70,in=-120] (l32) to (l24); 
		\draw (l31) to (l22);
		\draw [out=20,in=-120] (l34) to (l26);
		\draw [out=20,in=-150] (l33) to (l26);
		\draw [out=150,in=-30] (l33) to (l22);
		\draw [out=170,in=-80] (l31) to (l21);
		\draw [out=160,in=-30] (l32) to (l21);
		\draw [out=150,in=-30] (l34) to (l23);
		\draw [out=20,in=-130] (l31) to (l23);
		%%%%%%%l4 starts 
		\node [below left of=l33] (l41) {$\pmb{0}$};  
		\draw [out=120,in=-90] (l41) to (l32);
		\draw [out=60,in=-90] (l41) to (l33);
		\draw [out=160,in=-40] (l41) to (l31);
		\draw [out=20,in=-130] (l41) to (l34);
%		\node [left of=l12,xshift=-0.5cm] (l11) {$x \vee y$}; 
%		\node [right of=l13,xshift=0.5cm] (l14) {$\neg x \vee \neg y$}; 
	\end{tikzpicture}   
	}
\end{center} 
	%\end{wrapfigure} 
	
	$|\pmb{Atom}(\mathcal{D})| = 2^2$. We represent $\mathcal{D}$ as $[x, y]$. 
	Suppose some $T \subseteq \pmb{Themes}$, some $G$ and some $s \in Stmt[G]$, 
	and suppose $\mathcal{I}(T, \omega)$ is a sub-Boolean algebra $\mathcal{D}'$ 
	with $\pmb{Atom}(\mathcal{D'}) = \{x, \neg x\}$. Then, $\mathcal{D'}$ 
	is represented as $\sigma([x,y], 10) = [x]$, so  
	the representation of $\mathcal{I}(T, \omega)$ can be the 2-bit pattern 10.  
	It can be represented as a string 
	$\{\pmb{0}, \pmb{1}, x, \neg x\}$, too, alternatively, so long as the length of the string is polynomially bounded.
	 \hfill$\clubsuit$ 
%	\caption{A complete Boolean algebra. $y \Rightarrow x$ abbreviates $\neg y \vee x$ and $x \Leftrightarrow y$ abbreviates 
%	$(x \Rightarrow y) \wedge (y \Rightarrow x)$. Similarly for the remaining.} 
%	\label{fig_boolean_algebra} 
%\end{figure} 

\end{example} 

\noindent We assume the above restrictions on $\mathcal{D}$ and $\mathcal{I}$. 
%Agreeing on 
The following decision problems will be used regularly 
in the subsequent proofs. %See Appendix: proofs for the proofs. %\begin{lemma}[Decision problems in P] \label{lem_p_problems} 
%	Given $\taam$, some $T \subseteq \pmb{Themes}$ 
%	and some $s \in Stmt[G]$, verification 
%	of each of the following is P. 
%	\begin{itemize} 
%		\item $\mathcal{I}(T, s) \not= \emptyset$ 
%			and $\mathcal{I}(T, \omega) \not= \emptyset$. 
%		\item $\bigwedge \mathcal{I}(T, s) \not= \emptyset$. 
%		%	and $\mathcal{I}(T, \omega) \not= \emptyset$. 
%	\end{itemize}
%\end{lemma} 
%\textbf{Proof.} The first one is obvious. 
%The second one is also obvious due to $\mathcal{D}$ being 
%a complete lattice: unless $\mathcal{I}(T, s) = \emptyset$, 
%$\bigwedge \mathcal{I}(T, s)$ cannot be $\emptyset$.  \hfill$\Box$ 

\begin{lemma}[{\hyperref[proof_np_problems]{Decision problems in NP}}] \label{lem_np_problems}
	Given a theme aspect argumentation model $\taam$, 
	let each of $T_1, \ldots, T_{n}, T'_1, \ldots, T'_{n}$ 
	be a subset of $\pmb{Themes}$  
	and let each of $s_1, \ldots, s_{m}, s'_1, \ldots, s'_m$ 
	be a member of $Stmt[G]$. 
	Let  $n$ and $m$ be bounded by a polynomial 
	of the length of the representation of $\taam$,  
%	and $m \leq q(|Stmt[G]|)$ 
%	for some polynomial functions $p$ and $q$,   
	then each of the following decision 
	problems is NP-complete. 
	%verification of each of the following is NP-complete. 
	\begin{itemize} 
		\item the decision problem 
			which takes an input $\taam$ 
			for verifying 
	$\bigwedge \mathcal{I}(T_i, s_j) 
	\not= \bigwedge \mathcal{I}(T'_i, s'_j)$ 
	for every $1 \leq i \leq n$ and every $1 \leq j \leq m$. 
%	is verifiable in NP.   
\item the decision problem which takes an input $\taam$ 
	for verifying 
		$\pmb{0} \not= \bigwedge \mathcal{I}(T_i, s_j) \not= \pmb{1}$ 
	for every $1 \leq i \leq n$ and every $1 \leq j \leq m$. 
	\end{itemize} 
%	the verification 
%	of $\bigwedge_{1 \leq i \leq n, 1 \leq j \leq m} 
%	(\bigwedge \mathcal{I}(T_i, s_j) 
%	\not= \bigwedge \mathcal{I}(T'_i, s'_j)$ is in NP. 
\end{lemma} 
%\textbf{Proof.} See Appendix:  proofs. \hfill$\Box$ \\ 

\begin{lemma}[{\hyperref[proof_co_np_problems]{Decision problems in co-NP}}] \label{lem_co_np_problems}
	Given a theme aspect argumentation model $\taam$, let each of $T_1, \ldots, T_n, T'_1, \ldots, T'_n$ 
	be a subset of $\pmb{Themes}$, 
	and let each of $s_1, \ldots, s_m. s'_1, \ldots, s'_m$ 
	be a member of $Stmt[G]$. Let $n$ and $m$ 
	be bounded by a polynomial of the length of 
	the representation of $\taam$, then 
	each of the following decision problems is \mbox{co-NP}-complete. 
%	\leq p (|\pmb{Themes}|)$ 
%	and $m \leq q(|Stmt[G]|)$ for 
%	some polynomial functions $p$ and $q$, then 
%	some $T, T' \subseteq \pmb{Themes}$ 
%	and some $s, s' \in Stmt[G]$, the verification 
%	of 
%	the verification of each of the following is \mbox{co-NP}-complete. 
	\begin{itemize} 
		\item the decision problem 
			which takes an input $\taam$ 
			for verifying 
			$\bigwedge \mathcal{I}(T'_i, s'_j) \in 
			{\uparrow\{\bigwedge \mathcal{I}(T_i, s_j)\}}$ 
			   for every $1 \leq i \leq n$ 
			   and every $1 \leq j \leq m$. 
		\item the decision problem which takes an input 
			$\taam$ for verifying $\bigwedge \mathcal{I}(T'_i, s'_j) \in 
			{\downarrow\{\bigwedge \mathcal{I}(T_i, s_j)\}}$ 
			for every $1 \leq i \leq n$ and every 
			$1 \leq j \leq m$. 
		\item the decision problem which takes an input 
			$\taam$ for verifying $\mathcal{I}(T_i, s_j) \subseteq \mathcal{I}(T'_i, s'_j)$ for every $1 \leq i \leq n$ and every $1 \leq j \leq m$. 
		\item the decision problem which takes an input 
			$\taam$ for verifying $\mathcal{I}(T_i, s_j) \subseteq \mathcal{I}(T'_i, \omega)$ for every $1 \leq i \leq n$ and every $1 \leq j \leq m$. 
		\item the decision problem which takes an input 
			$\taam$ for verifying $\mathcal{I}(T_i, \omega) \subseteq \mathcal{I}(T'_i, \omega)$ for every $1 \leq i \leq n$ and every $1 \leq j \leq m$. 
%		\item $(\mathcal{I}(T, \omega), \pmb{1}, \pmb{0}, 
%			\neg, \wedge, \vee)$ is 
%			a sub-complete Boolean lattice of $\mathcal{D}$. 
	\end{itemize}
\end{lemma} 
%\textbf{Proof.} See Appendix:  proofs. \hfill$\Box$\\ 

Now, recall there are 5 constraints on typed argumentation 
graphs: \pmb{tr}, \pmb{nnp}, 
\pmb{nsa}, \pmb{kos}, and \pmb{nss}
and 15 constraints on theme aspect argumentation models: 
\pmb{aass}, \pmb{i}, \pmb{vi}, \pmb{bat}, \pmb{pr}, \pmb{mat}, 
\pmb{manss}, \pmb{ss}, \pmb{esr}, \pmb{ensr}, \pmb{eos}, \pmb{das}, 
\pmb{nwci}, \pmb{faD} and \pmb{faW}, as listed in \hyperref[app_table_of_constraints]{Appendix: table of constraints}.
\noindent The central decision problems are as follows. 
\begin{definition}[Decision problems]\label{def_decision_problems} \rm 
	Let $\pmb{x}$ be a member 
	of $\{$\pmb{tr}, \pmb{nnp}, 
	\pmb{nsa}, \pmb{kos}, \pmb{nss}$\}$, 
	we denote by $\Sat{$x$}$ the decision problem 
	which takes any finite-length input $\taam$  
	and which decides whether $\modell$ satisfies $\pmb{x}$. 
	Let $\pmb{x}$ be a member of 
	$\{$\pmb{aass}, \pmb{i}, \pmb{vi}, \pmb{bat}, \pmb{pr}, \pmb{mat}, 
\pmb{manss}, \pmb{ss}, \pmb{esr}, \pmb{ensr}, \pmb{eos}, \pmb{das}, 
	\pmb{nwci}, \pmb{faD}, \pmb{faW}$\}$,  
	let $n$ denote the input length, 
%	and let $p(n)$ be a polynomial of $n$, 
	we denote by $\Sat{$x$}$ and $\SatOne{$x$}$ 
	the following decision problems. 
	%, \SatOne{$x$}, \SatP{$x$}$ and 
	%$\SatOneP{$x$}$ the following decision problems. 
	\begin{description}
		\item[$\Sat{$x$}$] the decision problem 
	which takes any finite-length input $\taam$ 
			and which decides whether $\taam$ satisfies $\pmb{x}$. 
\item[$\SatOne{$x$}$] the decision problem  
	which takes any finite-length input $\taam$  with $2^{|\pmb{Themes}|} \leq p(n)$ 
			for a polynomial $p(n)$ 
			and which decides whether $\taam$ satisfies $\pmb{x}$. 
%		\item[$\SatP{$x$}$]  the decision problem 
%	which takes any finite-length input $\taam$ 
%			with $\taam$ being the power-set abstraction of itself  
%			and $\pmb{Themes}$ and which decides whether $\taam$ satisfies $\pmb{x}$. 
%\item[$\SatOneP{$x$}$] 
%	the decision problem  
%	which takes any finite-length input $\taam$ 
%			with $\taam$ being the power-set abstraction of itself 
%	and $\pmb{Themes}$ with $|\pmb{Themes}| = 1$ 
%%			and which decides whether $\taam$ satisfies $\pmb{x}$. \hfill$\spadesuit$
	\end{description} 
	We define $\Sat{ \pmb{Core}}$ to be $\bigcup_{\pmb{x} \in \pmb{Core}}\Sat{$x$}$, 
	$\Sat{ \pmb{All}$^{- \text{\pmb{F}}}$}$ to be $(\Sat{ \pmb{Core}}) \cup 
	(\bigcup_{\pmb{x} \in \pmb{E}}\Sat{$x$}) \cup \Sat{das} \cup \Sat{nwci}$, and $\Sat{ \pmb{All}}$ to be $(\Sat{ \pmb{All}$^{- \text{\pmb{F}}}$})
	\cup (\bigcup_{\pmb{x} \in \pmb{F}}\Sat{$x$})$.% \cup \Sat{das} \cup \Sat{nwci}$, 
	\hfill$\spadesuit$  
\end{definition}   
We establish the complexity of $\Sat{$x$}$ for each constraint $\pmb{x}$ on the typed argumentation graph 
first. 
%it turns out that $\Sat{$x$}$ is in P. 
We will then establish that $\Sat{ \pmb{Core}}$ and 
$\Sat{ \pmb{All}$^{- \text{\pmb{F}}}$}$
are both \mbox{co-NP(N}P)-complete ({\it i.e.} $\mathsf{\Pi}_2$-complete) but that 
$\Sat{ \pmb{All}}$ is likely not contained in 
\mbox{co-NP(N}P). To establish the upper bound of co-NP(NP) 
for \pmb{Core}, we identify the complexity of $\SatOne{$x$}$ for each 
constraint $\pmb{x}$ on theme aspect argumentation models. 
\hypersetup{
%  colorlinks   = true, %Colours links instead of ugly boxes
%  urlcolor     = blue, %Colour for external hyperlinks
%  linkcolor    = blue, %Colour of internal links
%  citecolor   = red, %Colour of citations 
  linkbordercolor = {white}
%  hidelinks = true
}
\makeatletter
\DeclareRobustCommand*{\nameref}{%
\color{blue}%
        \@ifstar\T@nameref\T@nameref
        }%
\makeatother
\makeatletter
\DeclareRobustCommand*{\ref}{%
\color{blue}%
        \@ifstar\T@ref\T@ref
        }%
\makeatother

\begin{table*}[!t]
\ra{1.3}
	\begin{tabular}{@{}rcccc|ccc@{}}		
		{\small Thm.} &  {\small \nameref{thm_satisfaction_tr}} & 
		{\small \nameref{thm_satisfaction_nnp}} & 
		{\small \nameref{thm_satisfaction_nsa}} & 
		\multicolumn{1}{c}{{\small \nameref{thm_satisfaction_kos}}} & 
		{\small \nameref{thm_satisfaction_nss}} \\
		\toprule
	$\modell$ & \pmb{tr} & \pmb{nnp} & \pmb{nsa} & \pmb{kos} & 
	\pmb{nss}  \\ \midrule   
	$\quad$ Sat: & P & P & P & P & P \\
	\bottomrule\\ 
\end{tabular}    
	{\ }\\ %\vspace{.3cm} 
%	{\ }\\ %$\quad$   
\begin{tabular}{@{}rc|c|ccccccc@{}}	
	{\small Thm.} & \multicolumn{1}{c}{{\small \ref{thm_sat_core_complete}}} & 
	\multicolumn{1}{c}{{\small   
	\ref{thm_sat_core_complete}}} & 
	\multicolumn{1}{c}{{\small \ref{thm_sat_all}}}\\ 
	\toprule 
	  	$\taam$ & \pmb{Core} & \pmb{All}$^{- \text{\pmb{$F$}}}$ & 
	\pmb{All}\\\midrule %& \pmb{faD} & 
	%\pmb{faW}  \\ \midrule   
	$\quad$ Sat: & $\mathsf{\Pi}_2$-complete & 
	$\mathsf{\Pi}_2$-complete & 
	$\mathsf{\Pi}_3$ and $\mathsf{\Pi}_2$-hard  \\
	%& 
	%$\mathsf{\Pi}_3$ & $\mathsf{\Pi}_3$ \\
	\bottomrule\\
\end{tabular}    
	{\ } \\
\begin{tabular}{@{}rcccccccc|ccccccc@{}}
	\multicolumn{1}{r}{\small Thm.} & {\small \nameref{thm_sat_aass}} & {\small \nameref{thm_sat_i}} 
	& {\small \nameref{thm_sat_vi}}  & {\small \nameref{thm_sat_bat}}   
	& {\small \nameref{thm_sat_pr}}  & {\small \nameref{thm_sat_mat}} 
	& {\small \nameref{thm_sat_manss}} & \multicolumn{1}{c}{{\small \nameref{thm_sat_ss}}} 
	& {\small \nameref{thm_sat_esr}} & {\small \nameref{thm_sat_esr}} 
	& {\small \nameref{thm_sat_esr}} & {\small \nameref{thm_sat_das}} 
	& {\small \nameref{thm_sat_nwci}} & {\small \ref{thm_sat_faw}} 
	& {\small \ref{thm_sat_fad}} \\\toprule  
		$\taam$ & \pmb{aass} & \pmb{i} & \pmb{vi} & \pmb{bat} & 
	\pmb{pr} & \pmb{mat} & \pmb{manss} & \pmb{ss} 
	& \pmb{esr} & \pmb{ensr} & \pmb{eos} & \pmb{das} 
	& \pmb{nwci} & \pmb{faW} & \pmb{faD} \\ \midrule   
%	$\quad$ Sat: & $\mathsf{\Pi}_2$ & $\mathsf{\Pi}_1$ & P & $\mathsf{\Pi}_2$ & $\mathsf{\Pi}_2$ & $\mathsf{\Pi}_2$ 
%	& $\mathsf{\Pi}_2$ & $\mathsf{\Pi}_2$ & $\mathsf{\Pi}_2$ & 
%	$\mathsf{\Pi}_2$ & $\mathsf{\Pi}_2$ & $\mathsf{\Pi}_2$ & $\mathsf{\Pi}_2$ & $\mathsf{\Pi}_3$-hard & $\mathsf{\Pi}_2$-hard\\
	$\quad$ p-Sat: &  $\mathsf{\Delta}_2$  & $\mathsf{\Pi}_1$ & P &
	$\mathsf{\Delta}_2$ & $\mathsf{\Pi}_1$ & $\mathsf{\Pi}_1$ & $\mathsf{\Pi}_1$ & NP & $\mathsf{\Delta}_2$
	& $\mathsf{\Delta}_2$ & $\mathsf{\Delta}_2$ & $\mathsf{\Delta}_2$ & $\mathsf{\Delta}_2$ & $\mathsf{\Pi}_3$ & 
	$\mathsf{\Pi}_3$ %\\
%	\midrule
%	$\mathfrak{p}$-Sat: & $\mathsf{\Pi}_1$ & $\mathsf{\Pi}_1$ & P & $\mathsf{\Pi}_1$ & $\mathsf{\Pi}_1$ & $\mathsf{\Pi}_1$ 
%	& $\mathsf{\Pi}_1$ & $\mathsf{\Pi}_1$ & $\mathsf{\Pi}_1$ & 
%	$\mathsf{\Pi}_1$ & $\mathsf{\Pi}_1$ & $\mathsf{\Pi}_1$ & $\mathsf{\Pi}_1$ & $\mathsf{\Pi}_2$-hard & $\mathsf{\Pi}_1$-hard\\
%	1-$\mathfrak{p}$-Sat: &  P  & P & P &
%	P & P & P & P & P & P  
%	& P & P & P & P & $\mathsf{\Pi}_2$-hard & $\mathsf{\Pi}_1$-hard \\
\end{tabular}

	\caption{The complexity classes Sat and p-Sat decision problems 
	belong to. 	    %$\Sat{$x$}$ and $\SatOne{$x$}$ for some constraint $\pmb{x}$.    
	%    \pmb{All}$^{- \text{\pmb{F}}}$ is the set of all constraints on theme aspect argumentation 
%	5    models but \pmb{F}. 
	%    $\Sat{ Core} \equiv \bigcup_{\pmb{x} \in \pmb{Core}} \Sat{$x$}$  
	    %$\SatP{$x$}$ and 
	    %$\SatOneP{$x$}$ for some constraint $\pmb{x}$.  
	    $\mathsf{\Delta}_2$ is P(NP). 
	    $\mathsf{\Pi}_1$ is co-NP. $\mathsf{\Pi}_2$ is co-NP(NP). 
	    $\mathsf{\Pi}_3$ is co-NP(NP(NP)).  
	    The theorem numbers are indicated. 
%	    $\mathsf{\Pi}_4$ is co-NP(NP(NP(NP))).   
	    %Completeness holds for all but P, $\mathsf{\Delta}_2$, $\mathsf{\Pi}_2$-hard, 
	    %and $\mathsf{\Pi}_3$-hard decision problems.
	    }
	    %Completeness holds for all 
%	but for those in P, $\mathsf{\Delta}_2$ or $\mathsf{\Pi}_k$-hard.} 
		%Families of informal fallacies and thematic and referencing issues they present. 
%	A cross indicates that some specific examples of the fallacy 
%	group can be identified as a fallacy with the constraint. 
%	For example, 
%	this table indicates that some straw man fallacy can be identified as a fallacy with \pmb{das} 
%	as it does not satisfy it. An indication as to where 
%	to find an example is given for those that have a specific example 
%	in this paper.} 
	\label{tbl_computational_complexity} 
\end{table*}
\hypersetup{
  colorlinks   = true, %Colours links instead of ugly boxes
%  urlcolor     = blue, %Colour for external hyperlinks
%  linkcolor    = blue, %Colour of internal links
%  citecolor   = red, %Colour of citations 
  linkbordercolor = {red}
}
\makeatletter
\DeclareRobustCommand*{\nameref}{%
\color{black}%
        \@ifstar\T@nameref\T@nameref
        }%
\makeatother
\makeatletter
\DeclareRobustCommand*{\ref}{%
\color{black}%
        \@ifstar\T@ref\T@ref
        }%
\makeatother

Table \ref{tbl_computational_complexity} shows to which complexity 
class these decision problems belong. 
%the computational 
%complexities of the decision problems. 
%We formally establish the result in the remaining of this section, 
Starting with the constraints on the typed argumentation graph: 
\begin{multicols}{2} 
\begin{description}  
	\item[({\hyperref[proof_satisfaction_tr]{Theorem 
		\labelText{9}{thm_satisfaction_tr}}})] 
		{\it Sat:\pmb{tr} is in P.} 
	\item[({\hyperref[proof_satisfaction_nnp]{Theorem \labelText{10}{thm_satisfaction_nnp}}})]
		{\it Sat:\pmb{nnp} is in P.} 
	\item[({\hyperref[proof_satisfaction_nsa]{Theorem \labelText{11}{thm_satisfaction_nsa}}})] 
	 {\it Sat:\pmb{nsa} is in P.} 
 \item[({\hyperref[proof_satisfaction_kos]{Theorem 
	 \labelText{12}{thm_satisfaction_kos}}})] 
	 {\it Sat:\pmb{kos} is in P.} 
 \item[({\hyperref[proof_satisfaction_nss]{Theorem 
	 \labelText{13}{thm_satisfaction_nss}}})] 
 {\it Sat:\pmb{nss} is in P.} \item
\end{description} 
\end{multicols} 
For $\Sat{ \pmb{Core}}$, $\Sat{ \pmb{All}$^{- \text{\pmb{F}}}$}$ 
and $\Sat{ \pmb{All}}$, 
we first identify the complexity of p-Sat decision problems  
for each of the individual constraints on 
theme aspect argumentation models. 
%See Appendix: proofs for non-vacuous proofs.  
\setcounter{theorem}{13} 

\begin{multicols}{2} 
	\begin{description}
		\item[({\hyperref[proof_sat_aass]{Theorem \labelText{14}{thm_sat_aass}}})] $\SatOne{aass}$ is in \mbox{P(NP)}. 
		\item[(Theorem \labelText{15}{thm_sat_i}$^*$)] $\SatOne{i}$ is in \mbox{co-NP}. 
		\item[(Theorem \labelText{16}{thm_sat_vi}$^*$)] $\SatOne{vi}$ is in P.  
		\item[({\hyperref[proof_sat_bat]{Theorem \labelText{17}{thm_sat_bat}}})] $\SatOne{bat}$ is in \mbox{P(NP)}. 
		\item[({\hyperref[proof_sat_pr]{Theorem \labelText{18}{thm_sat_pr}}})] $\SatOne{pr}$ is in co-NP. 
		\item[({\hyperref[proof_sat_mat]{Theorem \labelText{19}{thm_sat_mat}}})] $\SatOne{mat}$ is in \mbox{co-NP}. 
		\item[(Theorem \labelText{20}{thm_sat_manss}$^*$)] $\SatOne{manss}$ is in co-NP.  
		\item[({\hyperref[proof_sat_ss]{Theorem \labelText{21}{thm_sat_ss}}})] $\SatOne{ss}$ is in NP. 
		\item[({\hyperref[proof_sat_esr]{Theorem \labelText{22}{thm_sat_esr}}})] $\SatOne{esr}$, 
			$\SatOne{ensr}$ and $\SatOne{eos}$ 
			are in P(NP). %Let \pmb{$x$} be any of \pmb{esr}, \pmb{ensr} and \pmb{eos}. 
%	$\SatOne{$x$}$ is in P(NP). 
		\item[({\hyperref[proof_sat_das]{Theorem \labelText{23}{thm_sat_das}}})] $\SatOne{das}$ is in P(NP). 
		\item[(Theorem \labelText{24}{thm_sat_nwci}$^*$)] $\SatOne{nwci}$ is in P(NP). 
	\end{description} 
\end{multicols} 
Theorem \nameref{thm_sat_i} is proved straightforwardly by Lemma \ref{lem_co_np_problems}. 
Theorem \nameref{thm_sat_vi} is obvious. Theorem \nameref{thm_sat_manss} is proved similarly to 
Theorem \nameref{thm_sat_mat}. Theorem \nameref{thm_sat_nwci} is proved similarly to Theorem 
\nameref{thm_sat_das}. 
These straightforward theorems are marked with an asterisk $^*$ in 
the above results list. 
See \hyperref[app_proofs]{Appendix: proofs} for the proof of all the others. %\begin{theorem}[p-Sat:\pmb{aass}] \label{thm_sat_aass} 
\setcounter{theorem}{24} 

\indent From the complexity of these p-Sat decision problems, we shall obtain the required complexity 
of $\Sat{ \pmb{Core}}$ together with that of $\Sat{ \pmb{All}$^{- \text{\pmb{F}}}$}$. 
Let us first note the following. In the worst case, every one of the constraints in 
(\pmb{All}$^{- \text{\pmb{F}}} \backslash \{$\pmb{vi}$\}$)
on theme aspect argumentation models $\taam$ %except for \pmb{faD} and \pmb{faW} 
requires checking a certain set of conditions for all or almost all the subsets of $\pmb{Themes}$  
- or in the case of \pmb{mat}, \pmb{manss} and \pmb{E} all the pairs of the subsets of $\pmb{Themes}$. 
However, nothing prevents us from partially verifying them just for one 
member - 
or in the case of \pmb{mat}, \pmb{manss} and \pmb{E} just one pair of members 
- of 
$\mathfrak{p}(\pmb{Themes})$. %$T_1, \ldots, T_n$ for a polynomially bounded $n$. %where $n$ is bounded by some 
%polynomial $p(|\pmb{Themes}|)$. 
Given a constraint \pmb{$x$} on theme aspect argumentation models $\taam$, 
let us denote the algorithmic problem to partially verify 
\pmb{$x$} for one member - or in the case 
of \pmb{mat}, \pmb{manss} and \pmb{E} one pair of members 
- of $p(|\pmb{Themes}|)$ 
%$\mathfrak{p}(\pmb{Themes})$ 
by \mbox{1-P}artial:\pmb{$x$}. %and its sub-problem when 
%the first input $\taam$ to it is restricted to the power-set abstraction of itself 
%by \mbox{$\mathfrak{p}$-S}at:$x_{\leq p}$, 
We have the following result. 

\begin{lemma}[1-Partial:\pmb{$x$}] \label{lem_partial} 
	Let \pmb{$x$} be any constraint 
	in (\pmb{All}$^{- \text{\pmb{F}}} \backslash \{$\pmb{vi}$\}$).  
        1-Partial:\pmb{$x$} is in the complexity class of 
	\mbox{p-S}at:\pmb{$x$}.  
%	Suppose \mbox{S}at:$x_{\leq p}$ and \mbox{$\mathfrak{p}$-S}at:$x_{\leq p}$  
%	are the decision problems 
%	for verifying  
	%\mbox{1-S}at:$x$ and respectively \mbox{$\mathfrak{p}$-S}at:$x$ for 
	%every $T_1, \ldots, T_n \in \pmb{Themes}$  
	%where $n$ is bounded by a polynomial of the input. Then, 
%	Sat:$x_{\leq p}$ is in \mbox{P(1-Sat:$x$)}. 
%	\mbox{$\mathfrak{p}$-S}at:$x_{\leq p}$ is in \mbox{P(1-$\mathfrak{p}$-Sat:$x$)}. 
\end{lemma}  
\vspace{-0.2cm} 
\textbf{Proof.} 1-Partial:\pmb{$x$} is a sub-algorithmic problem of 
\mbox{p-S}at:\pmb{$x$}. Vacuous. \hfill$\Box$ \\
%\textbf{Proof.} Since the number of subsets of $\pmb{Themes}$ to partially 
%verify $x$ for is bounded by a polynomial $p(|\pmb{Themes}|)$, Sat:$x_{\leq p}$ 
%is straightforwardly polynomial-time Turing reducible to 
%\mbox{1-Sat:$x$} and \mbox{$\mathfrak{p}$-S}at:$x_{\leq p}$ 
%to 1-$\mathfrak{p}$-Sat:$x$.  \hfill$\Box$ 

%\begin{corollary}
%	Let $x$ be any constraint on $\taam$ but \pmb{faD} and \pmb{faW}, then 
%	\mbox{$\mathfrak{p}$-S}at:$x_{\leq p}$ is in P.  
%\end{corollary} 
%\textbf{Proof.} 1-$\mathfrak{p}$-Sat:$x$ is in P and 
%P(P) = P by Lemma \ref{lem_subsumption_complexity_classes}. \hfill$\Box$ \\

\noindent Lemma \ref{lem_partial} leads to the following inclusion 
for $\Sat{  \pmb{All}$^{- \text{\pmb{F}}}$}$. 
%See Appendix: proofs 
%for the proof. 
The inclusion for $\Sat{ \pmb{Core}}$ is then immediate 
since $\Sat{ \pmb{Core}}$ is a sub-algorithmic problem of 
$\Sat{  \pmb{All}$^{- \text{\pmb{F}}}$}$. 
\begin{theorem}[{\hyperref[proof_inclusion_sat_all_f]{Inclusion of $\Sat{  \pmb{All}$^{- \text{\pmb{F}}}$}$}}]
	\label{thm_sat_core}
	$\Sat{ \pmb{All}$^{- \text{\pmb{F}}}$}$ is in \mbox{co-NP(N}P). 
\end{theorem} 

\begin{corollary}[Inclusion of $\Sat{ \pmb{Core}}$] \label{cor_sat_core_complete} 
	\hspace{0.15cm} $\Sat{ \pmb{Core}}$ is in co-NP(NP). 
\end{corollary} 

For the hardness, it suffices to show 
\mbox{co-NP(N}P)-hardness of  
$\Sat{$x$}$ for one 
$\pmb{x}$ in \pmb{Core}. 
$\Sat{aass}$ is one of them. %See Appendix: proofs for the proof. 
\begin{lemma}[{\hyperref[proof_hardness_aass]{Hardness of $\Sat{aass}$}}] \label{lem_hardness_aass} 
	$\Sat{aass}$ is \mbox{co-NP(N}P)-hard. 
\end{lemma} 
%\textbf{Proof.} See Appendix: proofs. \hfill$\Box$ \\ 

\begin{theorem}[{\hyperref[proof_hard_core_all]{Hardness of $\Sat{ \pmb{Core}}$, $\Sat{  \pmb{All}$^{- \text{\pmb{F}}}$}$ and $\Sat{  \pmb{All}}$}}] \label{thm_hard_core_all} 
    $\Sat{ \pmb{Core}}$ is \mbox{co-NP(N}P)-hard. 
	$\Sat{  \pmb{All}$^{- \text{\pmb{F}}}$}$ is 
	\mbox{co-NP(N}P)-hard. %Further, let \pmb{All} 
	%be \pmb{All}$^{- \text{\pmb{F}}} \cup \{\text{\pmb{F}}\}$, then 
	$\Sat{  \pmb{All}}$ is 
	\mbox{co-NP(N}P)-hard. 
\end{theorem} 
\vspace{-0.26cm} 
\textbf{Proof.}  $\Sat{aass}$ is a sub-algorithmic problem 
of $\Sat{ \pmb{Core}}$, $\Sat{  \pmb{All}$^{- \text{\pmb{F}}}$}$ 
and $\Sat{  \pmb{All}}$. 
Lemma \ref{lem_hardness_aass}. \hfill$\Box$\\

These results identify the complexity class of 
$\Sat{ \pmb{Core}}$ and $\Sat{  \pmb{All}$^{- \text{\pmb{F}}}$}$. 
\begin{theorem}[$\Sat{ \pmb{Core}}$ and $\Sat{  \pmb{All}$^{- \text{\pmb{F}}}$}$] \label{thm_sat_core_complete}	$\Sat{ \pmb{Core}}$ is \mbox{co-NP(N}P)-complete. 
	$\Sat{  \pmb{All}$^{- \text{\pmb{F}}}$}$ is 
	\mbox{co-NP(N}P)-complete. 
\end{theorem} 
\vspace{-0.26cm} 
\textbf{Proof.} By Theorems \ref{thm_sat_core} 
and \ref{thm_hard_core_all} and Corollary \ref{cor_sat_core_complete}.  \hfill$\Box$ \\

By comparing $\Sat{Core}$ and $\Sat{tr} \cup \Sat{nnp} \cup \Sat{nsa} \cup \Sat{kos}$, 
we can observe that, unless P = NP, it is computationally cheaper to tell 
if a typed argumentation graph $\modell$ has some $\mathcal{D}, \mathcal{I}$ such that $\taam$  
satisfies \pmb{Core} than if, 
given some $\mathcal{D}_1$ and $\mathcal{I}_1$, 
$\langle \modell, \mathcal{D}_1, \mathcal{I}_1 \rangle$ satisfies \pmb{Core}. %See Appendix: proofs for the proof. 
\begin{theorem}[{\hyperref[proof_thm_model_existence_model_checking]{\pmb{Core}-model existence and \pmb{Core}-model checking}}]\label{thm_model_existence_model_checking} 
	Given a typed argumentation graph $\modell$, the existence of some $\mathcal{D}, \mathcal{I}$ 
	such that $\langle \modell, \mathcal{D}, \mathcal{I} \rangle$ satisfies  
	\pmb{Core} is polynomial-time verifiable. However, given   
	some $\mathcal{D}_1$ and $\mathcal{I}_1$ 
	such that $\modell$, $\mathcal{D}_1$ and $\mathcal{I}_1$  
	from a theme aspect argumentation model, 
	whether $\langle \modell, \mathcal{D}_1, \mathcal{I}_1 \rangle$ satisfies 
	\pmb{Core} is not polynomial-time verifiable unless P = NP. 
\end{theorem}
\subsection{Preparation for the upper bound of Sat:\pmb{All}} 
Now, for the upper bound of $\Sat{ \pmb{All}}$, 
we need to know the upper bounds of $\Sat{faD}$ (= $\SatOne{faD}$ trivially) and $\Sat{faW}$ 
(= $\SatOne{faW}$ trivially), 
which is  more laborious a process than with the other 
constraints. We start by obtaining preliminary complexity results, 
in particular the complexity of  
obtaining width-statements-sets 
(Lemma \ref{lem_computing_width_statements_sets}),   
that of deciding whether a set of statements  
is a depth-statements-set (Lemmas 
\ref{lem_polynomialtime_transformation_g_dagg} and 
\ref{lem_polynomial_time_transformation_dag_labelled_dag} 
and Proposition \ref{prop_decide_membership_deps}) 
and that of decision problems for judging the absence 
of redundancy (Proposition \ref{lem_deciding_subtest_and_subtestall}). 
%Theorems \ref{thm_equivalence} and 
%Trivial proofs are embedded in the text. 
%See Appendix: proofs for the others. 

\begin{lemma}[{\hyperref[proof_computing_width_statements_sets]{Computing width-statements-sets}}]\label{lem_computing_width_statements_sets}
	Let WSS be an algorithmic problem with inputs 
	(1)  some $\modell$, (2) some $s \in Stmt[G]$, 
	(3) some $trel 
	\in \{`attack$'$, `support$'$\}$ and (4) some $t \in \pmb{Themes}$ 
	for outputting  $WidS[s, t, trel]$. 
	WSS is in P. 
\end{lemma} 
%\textbf{Proof.} See Appendix:  proofs. \hfill$\Box$\\ 

For the depth-statements-sets, a naive approach
and representation of the output 
would make this enumeration alone NP-hard.
\begin{example}\label{ex_graph}  \rm 
	Suppose that $G$ is the graph shown below, that $\bigcap_{s \in Stmt[G]} \Pi(s) = 
	\bigcup_{s \in Stmt[G]} \Pi(s) = \{t\}$, that 
	$\bigcap_{(s', s) \in Rel[G]} \Pi((s', s)) = \bigcup_{(s', s) \in Rel[G]} \Pi((s', s)) = \{t, `attack$'$\}$,
	and that $\Pi(x) = \emptyset$ for any $x \not\in Stmt[G] \cup Rel[G]$. Then 
	$(G, \Pi)$ is well-formed. %\ryuta{The earlier example contains an error then.}  
    \begin{center} 
    \begin{tikzcd}[column sep=tiny,row sep=tiny]   
	    & s_7 \arrow[dl]  \arrow[dd] & s_5 \arrow[l] \arrow[ddl]   & s_3 \arrow[l] \arrow[ddl]  \\ 
	    s_8 & &  & & s_1 \arrow[dl] \arrow[ul]    \\  
		& s_6 \arrow[uu] \arrow[ul] & s_4 \arrow[l] \arrow[uul] & s_2 \arrow[l] \arrow[uul]
%		aCostH \arrow[-,ur]\arrow[-,dr] & 	  & \neg aCostH \arrow[-,ul]\arrow[-,dl] & bCostH \arrow[-,ur]\arrow[-,dr] & & \neg bCostH \arrow[-,ul]\arrow[-,dl]\\
%		& \pmb{0} & & & \pmb{0} \\ 
%                & \pmb{1} &   & & \pmb{1} \\ 
%		aP \arrow[-,ur]\arrow[-,dr] & 	  & \neg aP \arrow[-,ul]\arrow[-,dl] & bP \arrow[-,ur]\arrow[-,dr] & & \neg bP 
%		\arrow[-,ul]\arrow[-,dl]\\
%		& \pmb{0} & & & \pmb{0} 
	\end{tikzcd}
	\end{center}   
	All the depth-statements sets with respect to $s_8$ and $t$ may be obtained
	by tracing all potential reverse walks ({\it i.e.} walks in the opposite direction of edges). Starting at $s_8$, 
	we have two alternative walks, to $s_7$ or to $s_6$. 
	At $s_7$ as well as at $s_6$, we find that walk(s) to the other (and back) 
	are possible. So we find that every member of $DepS[s_8, t, `attack$'$]$ 
	includes $s_8, s_7$ and $s_6$ at minimum. From $s_6$ or from $s_7$, 
	we have other two walks, to $s_5$ or to $s_4$. From $s_5$ or from $s_4$, 
	we again have two walks, to $s_3$ or to $s_2$. Then, from $s_3$ or from $s_2$, 
	there is a walk to $s_1$. At this point, we learn that 
	a member of $DepS[s_8, t, `attack$'$]$ contains $s_8, s_7, s_6$ 
	and in addition to them either $s_4$ or $s_5$, either $s_2$ or $s_3$, 
	and lastly $s_1$. This intuitive traversal, given an arbitrary 
	$\modell$ and some starting node $s \in Stmt[G]$, is at least as algorithmically complex as 
	the longest path problem which is already NP-hard. \hfill$\clubsuit$ 
\end{example} 
By utilising two observations,
however, it becomes possible to obtain a graph-based representation of depth-statements-sets
in polynomial time.   

\begin{observation} \rm 
	While a sequence $s_1.\ldots.s_n$ of statements  
	forms a walk, or in our case a reverse walk in the opposite direction,  
	a depth-statements-set does not perforce  
	demand the knowledge of the exact reverse walk as long as 
	we can obtain all the statements in the walk.   
\end{observation} 
The consequence of this observation is a reduction of a general graph walk into 
a DAG (Direct Acyclic Graph) path traversal. Up till now, 
when we mentioned a finite graph, it was a finite graph of statements. 
For the convenience of abstract graph manipulation, however, we now extend $\mathcal{G}$ to $\mathcal{G}^+$ 
such that $\mathcal{G}^+ \backslash \mathcal{G}$ contains 
any other finite graphs $(V, E)$ with $V \subseteq \pmb{V}$ 
and $E \subseteq V \times V$ unrelated to argumentation. $\pmb{V}$ is an infinite set of nodes 
unrelated to argumentation. 
\begin{definition}[Sub-theme $\modell$ for $trel \in \{`attack$'$, `support$'$\}$]\label{def_sub_theme_trel} \rm 
	Given a typed argumentation graph $\modell$, $t \in \pmb{Themes}$ and $trel \in \{`attack$'$, `support$'$\}$,
	we define ${(G_{t, trel}, \Pi_{t, trel})}$ to be $(G', \Pi')$ satisfying 
	all the following conditions.  
	\begin{itemize}   
			\begin{multicols}{2}
		\item $Stmt[G'] = Stmt[G_{t}]$. 
		\item $Rel[G'] = \{(s', s) \in Rel[G_{t}] \mid trel \in \Pi_{t}((s', s))
			 \}$. 
			\end{multicols}  
			\vspace{-0.4cm} 
			\begin{multicols}{2} 
		 \item for any $x \in Stmt[G'] \cup Rel[G']$, $\Pi'(x) = \Pi_{t}(x) \cap \{t, trel\}$. 
			 %for any $s \in \pmb{A}$, $s \in  Stmt[G']$ materially implies $\Pi(s) = \Pi'(s)$, 
			 %and $s \not\in Stmt[G']$ implies $\Pi(s) = \emptyset$. 
		 \item for any $x \not\in Stmt[G'] \cup Rel[G']$, $\Pi'(x) = \emptyset$. 
			 \hfill$\spadesuit$
			\end{multicols} 
	\end{itemize} 
\end{definition} 
\begin{definition}[DAG transformation] \label{def_dag_transformation} \rm       
	Given $G \in \mathcal{G}$, let $condense_{G}: Stmt[G] \rightarrow \pmb{V}$ be such that 
	it satisfies the following condition: 
		for every $s_1, s_2 \in Stmt[G]$, $condense_G(s_1) = condense_G(s_2)$ holds iff there 
			is a strongly connected component in $G$ that contains both $s_1$ and $s_2$.     
	
	Let $EXPAND$ be the set of all functions $expand: \pmb{V} \rightarrow \mathfrak{p}(\pmb{A})$, 
	we define $dag: \mathcal{G} \rightarrow ((\mathcal{G}^+ \backslash \mathcal{G}) \times EXPAND)$ to be such that 
	$dag(G)$ is $((V, E), expand)$ satisfying all the following conditions.
	\begin{itemize}  
		\item $V = \bigcup_{s \in Stmt[G]} condense_G(s)$. 
		\item for any $v', v \in V$, $(v', v) \in E$ holds iff there exist $s', s \in Stmt[G]$ 
			such that $v' = condense_G(s') \not= condense_G(s) = v$ and that 
			$(s', s) \in Rel[G]$. 
		\item for any $v \in V$, $expand(v) = \{s \in Stmt[G] \mid condense_G(s) = v\}$. \hfill$\spadesuit$
	\end{itemize} 
\end{definition}  

\begin{example} \rm 
	The left graph below was introduced in Example \ref{ex_graph}. 
	\begin{center} 
		\begin{tikzcd}[column sep=tiny,row sep=tiny]   
	    & s_7 \arrow[dl]  \arrow[dd] & s_5 \arrow[l] \arrow[ddl]   & s_3 \arrow[l] \arrow[ddl]  \\ 
	    s_8 & &  & & s_1 \arrow[dl] \arrow[ul]    \\  
		& s_6 \arrow[uu] \arrow[ul] & s_4 \arrow[l] \arrow[uul] & s_2 \arrow[l] \arrow[uul]
%		aCostH \arrow[-,ur]\arrow[-,dr] & 	  & \neg aCostH \arrow[-,ul]\arrow[-,dl] & bCostH \arrow[-,ur]\arrow[-,dr] & & \neg bCostH \arrow[-,ul]\arrow[-,dl]\\
%		& \pmb{0} & & & \pmb{0} \\ 
%                & \pmb{1} &   & & \pmb{1} \\ 
%		aP \arrow[-,ur]\arrow[-,dr] & 	  & \neg aP \arrow[-,ul]\arrow[-,dl] & bP \arrow[-,ur]\arrow[-,dr] & & \neg bP 
%		\arrow[-,ul]\arrow[-,dl]\\
%		& \pmb{0} & & & \pmb{0} 
	\end{tikzcd}  
		\hspace{3cm} 
		\begin{tikzcd}[column sep=tiny,row sep=tiny]   
	    &    & v_5 \arrow[dl]   & v_3 \arrow[l] \arrow[ddl]  \\ 
	    v_8 & v_6 \arrow[l] &  & & v_1 \arrow[dl] \arrow[ul]    \\  
		& & v_4 \arrow[ul]  & v_2 \arrow[l] \arrow[uul]
%		aCostH \arrow[-,ur]\arrow[-,dr] & 	  & \neg aCostH \arrow[-,ul]\arrow[-,dl] & bCostH \arrow[-,ur]\arrow[-,dr] & & \neg bCostH \arrow[-,ul]\arrow[-,dl]\\
%		& \pmb{0} & & & \pmb{0} \\ 
%                & \pmb{1} &   & & \pmb{1} \\ 
%		aP \arrow[-,ur]\arrow[-,dr] & 	  & \neg aP \arrow[-,ul]\arrow[-,dl] & bP \arrow[-,ur]\arrow[-,dr] & & \neg bP 
%		\arrow[-,ul]\arrow[-,dl]\\
%		& \pmb{0} & & & \pmb{0} 
	\end{tikzcd}
	\end{center}   
%    \end{center} 
	Denote it by $G$, 
	$dag(G)$ is then $((V, E), expand)$ where $(V, E)$ is the right graph above.
%	\begin{center} 
    	$expand$ is such that $expand(v_i) = \{s_i\}$ for $i \in \{1, \ldots, 5, 8\}$ 
	and that $expand(v_6) = \{s_6, s_7\}$. \hfill$\clubsuit$ \\
\end{example} 
This transformation causes no loss of generality as is formally stated below. 
By a terminal node of a DAG, we shall mean a node with no incoming edges. This is 
the opposite of the usual definition, but in our reverse walk/path, 
a node with no incoming edges indeed becomes a terminal. 
\begin{proposition}[Equivalence] \label{prop_equivalence} 
	Given a typed argumentation graph $\modell$, some $s \in Stmt[G]$, $t \in \pmb{Themes}$ and $trel \in \{`attack$'$, `support$'$\}$, 
	let $((V,E), expand)$ be $dag(G_{t, trel})$, then all the following hold. 
	\begin{itemize} 
		\item for every $S \in DepS[s, t, trel]$, 
			there is some terminal node $v' \in V$ 
			and some sequence $v_1, \ldots, v_n$ of members of $V$ 
			such that all the following conditions hold. 
			\begin{itemize} 
				\item $v_n, \ldots, v_1$ is a reverse path of $(V, E)$ from $v_n$ to $v_1$. 
				\item $v_1 = v'$ and 
					$v_n = condense_{G_{t, trel}}(s)$. 
				\item  $(\bigcup_{v \in \{v_1, \ldots, v_n\}} expand(v)) = S$.   
			\end{itemize} 
		\item for every terminal node $v' \in V$ and every sequence $v', \ldots, condense_{G_{t,trel}}(s)$ of members of $V$, 
			if $condense_{G_{t,trel}}(s), \ldots, v'$ is a reverse path from $condense_{G_{t,trel}}(s)$ to $v'$ in $(V, E)$, then \linebreak 
			$\bigcup_{v \in \{v', \ldots, condense_{G_{t,trel}}(s)\}} expand(v)$ is a member of $DepS[s, t, trel]$. 
	\end{itemize} 
\end{proposition} 
\vspace{-0.26cm} 
\textbf{Proof.} By the correspondence of $G$ and $dag(G)$. \hfill$\Box$ \\
\begin{lemma}[Polynomial-time transformation of $G$ into $dag(G)$] \label{lem_polynomialtime_transformation_g_dagg}
	Given a finite argumentation graph $G \in \mathcal{G}$, $dag(G)$ is polynomial-time computable. 
\end{lemma}
\vspace{-0.26cm} 
\textbf{Proof.} The condensation of a finite graph is a well-known polynomial-time algorithmic problem. \hfill$\Box$ \\

%\noindent Whether we traverse a DAG or a general graph 
%(note, in the reverse direction) is computationally significant. Indeed, in the case of a DAG, 
%the longest-path problem is the shortest-path problem (which is in P) 
%with a negative weight on each edge. 
%The same 
%approach is not applicable to a general graph since a loop would cause the traversal cost to continually 
%drop. 

Unfortunately, if we attempt to enumerate all the members of $DepS[s, t, trel]$ ($trel \in \{`attack$'$, `support$'$\}$) 
and not just one of them, the number of $|DepS[s, t, trel]|$ is still exponential in the worst case.  
\begin{example}\label{ex_dag} \rm 
	The graph shown below is a DAG of 11 nodes.  
	\begin{center} 
	\begin{tikzcd}[column sep=tiny,row sep=tiny]   
		&  v_{10} \arrow[dl] & v_8 \arrow[l]\arrow[ddl]   & v_6 \arrow[l] \arrow[ddl]  & v_4 \arrow[l] \arrow[ddl] & v_2 \arrow[l] \arrow[ddl] \\ 
	v_{11} &  &  & &     \\  
		& v_{9} \arrow[ul] & v_7 \arrow[uul] \arrow[l] & v_5 \arrow[l] \arrow[uul] & v_3 \arrow[l] \arrow[uul] & v_1 \arrow[l] \arrow[uul]
%		aCostH \arrow[-,ur]\arrow[-,dr] & 	  & \neg aCostH \arrow[-,ul]\arrow[-,dl] & bCostH \arrow[-,ur]\arrow[-,dr] & & \neg bCostH \arrow[-,ul]\arrow[-,dl]\\
%		& \pmb{0} & & & \pmb{0} \\ 
%                & \pmb{1} &   & & \pmb{1} \\ 
%		aP \arrow[-,ur]\arrow[-,dr] & 	  & \neg aP \arrow[-,ul]\arrow[-,dl] & bP \arrow[-,ur]\arrow[-,dr] & & \neg bP 
%		\arrow[-,ul]\arrow[-,dl]\\
%		& \pmb{0} & & & \pmb{0} 
	\end{tikzcd}
	\end{center} 
	Any reverse path from $v_{11}$ to a terminal node contains 5 edges. 
	Thus, there are $\sqrt{2}^{10}$ distinct reverse paths from $v_{11}$ to a terminal node. \hfill$\clubsuit$ 
\end{example} 

\noindent Here, however, comes our second observation. 
\begin{observation} \rm 
	For (p-)Sat:\pmb{faD}, our primary concern is to judge 
	the absence of redundancy. While we certainly need some representation of 
	$DepS[s, t, trel]$, it does not have to be in the form of sets. 
\end{observation} 
The consequence of this observation is representing  $DepS[s, t, trel]$ as a labelled DAG together 
with an $expand$ function. We define the concept of subsumption of  
a path into another path (Definition \ref{def_path_subsumption}) 
and then define the labelled DAG (Definition \ref{def_labelled_dag}) 
based on it. 

\begin{definition}[Subsumption of paths]\label{def_path_subsumption} \rm 
	Given a DAG $(V, E) \in (\mathcal{G}^+ \backslash \mathcal{G})$ 
	and two reverse paths $rpath_1$ and $rpath_2$ in $(V, E)$, 
	we say that $rpath_2$ subsumes  $rpath_1$ iff $rpath_2$ traverses 
	in the reverse direction 
	all the nodes in $rpath_1$. \hfill$\spadesuit$ 
\end{definition} 

\begin{definition}[Labelled DAG] \label{def_labelled_dag} \rm  
	Given a DAG $(V, E) \in (\mathcal{G}^+ \backslash \mathcal{G})$,   
	let $RPath((V, E), v)$ be the set of 
	all the reverse paths from $v$ to some terminal node in $(V, E)$.  
	Let $RMaxiPath((V, E, v)$ be a subset 
	of $RPath((V, E), v)$ comprising all the maximal 
	members of $RPath((V, E), v)$ with respect to the path subsumption. 
%	denote  
%	$\{(v, \ldots, v') \mid v' \text{ is a terminal node and } (v, \ldots, v') \text{ is 
%	a longest reverse}\linebreak \text{path from } v \text{ to } v' \text{ in } (V, E)\}$. 
	Let $label: V \times V \rightarrow \mathfrak{p}(V)$  be such that 
	$label(v, v') = \{v'' \in V \mid \exists rpath \in RMaxiPath((V, E), v).\linebreak
	{v' \in rpath} \text{ and } {v'' \in rpath} \text{ and } (v', v'') \in E\}$. 
%	\text{there is a member of } 
%	LongestRPath((V, E), v) \text{ such that } v'' \text{ is }\linebreak v'\text{'s 
%	immediate precedent node in the reverse path.}\}$, then 
	We call $((V, E), label)$ a labelled DAG.  \hfill$\spadesuit$ 
\end{definition} 
In $((V, E), label)$, the first parameter $v$ to the function $label$ determines 
the first node of one of the maximal reverse paths (to a terminal node) that it 
be $v$ and the second parameter $v'$ is the node that 
is supposed to be in the maximal reverse path(s) from $v$. 
Intuitively, the output $label(v, v')$ assigns 
a set of nodes that can appear in some of the maximal reverse path(s) from 
$v$ (to a terminal node) that occur immediately 
before $v'$. %Any reverse path that is not maximal is removed. 
\begin{example}  \rm 
	Denote the DAG introduced in Example \ref{ex_dag} by $(V, E)$. Suppose $label$ is such that 
%	\begin{center} 
%	\begin{tikzcd}[column sep=tiny,row sep=tiny]   
%		&  v_{10} \arrow[dl] & v_8 \arrow[l]\arrow[ddl]   & v_6 \arrow[l] \arrow[ddl]  & v_4 \arrow[l] \arrow[ddl] & v_2 \arrow[l] \arrow[ddl] \\ 
%	v_{11} &  &  & &     \\  
%		& v_{9} \arrow[ul] & v_7 \arrow[uul] \arrow[l] & v_5 \arrow[l] \arrow[uul] & v_3 \arrow[l] \arrow[uul] & v_1 \arrow[l] \arrow[uul]
%%		aCostH \arrow[-,ur]\arrow[-,dr] & 	  & \neg aCostH \arrow[-,ul]\arrow[-,dl] & bCostH \arrow[-,ur]\arrow[-,dr] & & \neg bCostH \arrow[-,ul]\arrow[-,dl]\\
%%		& \pmb{0} & & & \pmb{0} \\ 
%%                & \pmb{1} &   & & \pmb{1} \\ 
%%		aP \arrow[-,ur]\arrow[-,dr] & 	  & \neg aP \arrow[-,ul]\arrow[-,dl] & bP \arrow[-,ur]\arrow[-,dr] & & \neg bP 
%%		\arrow[-,ul]\arrow[-,dl]\\
%%		& \pmb{0} & & & \pmb{0} 
%	\end{tikzcd}
%	\end{center} 
%	Denote this DAG by $(V, E)$. Suppose $label$ is such that 
	\begin{itemize}  
		\item $label(v_{11}, v_{11}) = \emptyset$. 
		\item $label(v_{2i}, v_{j}) =  label(v_{2i-1}, v_j) = \emptyset$ 
			for every $1 \leq i \leq 5$ and every $2i-1 \leq j$.  
%		\item $label(v_{2i-1}, v_j) = \emptyset$ for evey $1 \leq i \leq 5$ and every $2i -1 \leq j$.    
		\item $label(v_{2i+1}, v_{2i}) = label(v_{2i+1}, v_{2i-1}) = \{v_{2i+1}\}$ for every $1 \leq i \leq 5$. 
		\item $label(v_{2i+2}, v_{2i}) = label(v_{2i+2}, v_{2i-1}) = \{v_{2i+2}\}$ for every $1 \leq i \leq 4$.  
		\item $label(v_{2i+1}, v_{2j}) = label(v_{2i+1}, v_{2j-1}) = \{v_{2j+1}, v_{2j+2}\}$ for every $1 \leq i \leq 5$ 
			and every $1 \leq j \leq i-1$.  
		\item $label(v_{2i+2}, v_{2j}) = label(v_{2i+2}, v_{2j-1}) = \{v_{2j+2}, v_{2j+1}\}$ for every $1 \leq i \leq 4$
			and every $1 \leq j \leq i-1$. 
%		\item $label(v_{11}, v_{10}) = label(v_{11}, v_{9}) = \{v_{11}\}$. 
%		\item $label(v_{11}, v_{2i}) = label(v_{11}, v_{2i-1}) = \{v_{2i+1}, v_{2i+2}\}$ for every $1 \leq i \leq 4$.
%		\item $label(v_{10}, v_{11}) = label(v_{10}, v_{10}) = label(v_{10}, v_9) = \emptyset$. 
%		\item $label(v_{10}, v_{8}) = label(v_{10}, v_{7}) = \{v_{10}\}$. 
%		\item $label(v_{10}, v_{6}) = label(v_{10}, v_5) = \{v_{8}, v_{7}\}$.  
	\end{itemize}
	Then $((V, E), label)$ is the labelled DAG of $(V, E)$.  \hfill$\clubsuit$ \\
\end{example}  
%\pagebreak

\begin{example}\label{ex_maximal_reverse_path} \rm 
	A maximal reverse path is obviously not always a longest reverse path. 
	Also, it is not always the case that $label(v, v')$  
	is the set of all the previous nodes of $v'$ in some reverse path 
	from $v$ to $v'$. To illustrate these two points, 
	suppose the DAG of 6 nodes shown below. 
	We only detail $label(v_6, v_i)$ for illustration. 
	\begin{center} 
	\begin{tikzcd}[column sep=small,row sep=small]   
		v_{6} & v_{4} \arrow[l]  & v_{2} \arrow[l] \\ 
		v_{5} \arrow[u] & v_{3} \arrow[ur] \arrow[l] & v_{1} \arrow[l] \arrow[ul] 
%		&  v_{10} \arrow[dl] & v_8 \arrow[l]\arrow[ddl]   & v_6 \arrow[l] \arrow[ddl]  & v_4 \arrow[l] \arrow[ddl] & v_2 \arrow[l] \arrow[ddl] \\ 
%	v_{11} &  &  & &     \\  
%		& v_{9} \arrow[ul] & v_7 \arrow[uul] \arrow[l] & v_5 \arrow[l] \arrow[uul] & v_3 \arrow[l] \arrow[uul] & v_1 \arrow[l] \arrow[uul]
%		aCostH \arrow[-,ur]\arrow[-,dr] & 	  & \neg aCostH \arrow[-,ul]\arrow[-,dl] & bCostH \arrow[-,ur]\arrow[-,dr] & & \neg bCostH \arrow[-,ul]\arrow[-,dl]\\
%		& \pmb{0} & & & \pmb{0} \\ 
%                & \pmb{1} &   & & \pmb{1} \\ 
%		aP \arrow[-,ur]\arrow[-,dr] & 	  & \neg aP \arrow[-,ul]\arrow[-,dl] & bP \arrow[-,ur]\arrow[-,dr] & & \neg bP 
%		\arrow[-,ul]\arrow[-,dl]\\
%		& \pmb{0} & & & \pmb{0} 
	\end{tikzcd}
%		{\ }\quad\quad\quad\quad\quad
%	\begin{tikzcd}[column sep=small,row sep=small]   
%		v_{6} & v_{4} \arrow[l]  & v_{2} \arrow[l] \\ 
%		v_{5} \arrow[u] & v_{3} \arrow[ur] \arrow[l] & v_{1} \arrow[l] 
%		&  v_{10} \arrow[dl] & v_8 \arrow[l]\arrow[ddl]   & v_6 \arrow[l] \arrow[ddl]  & v_4 \arrow[l] \arrow[ddl] & v_2 \arrow[l] \arrow[ddl] \\ 
%	v_{11} &  &  & &     \\  
%		& v_{9} \arrow[ul] & v_7 \arrow[uul] \arrow[l] & v_5 \arrow[l] \arrow[uul] & v_3 \arrow[l] \arrow[uul] & v_1 \arrow[l] \arrow[uul]
%		aCostH \arrow[-,ur]\arrow[-,dr] & 	  & \neg aCostH \arrow[-,ul]\arrow[-,dl] & bCostH \arrow[-,ur]\arrow[-,dr] & & \neg bCostH \arrow[-,ul]\arrow[-,dl]\\
%		& \pmb{0} & & & \pmb{0} \\ 
%                & \pmb{1} &   & & \pmb{1} \\ 
%		aP \arrow[-,ur]\arrow[-,dr] & 	  & \neg aP \arrow[-,ul]\arrow[-,dl] & bP \arrow[-,ur]\arrow[-,dr] & & \neg bP 
%		\arrow[-,ul]\arrow[-,dl]\\
%		& \pmb{0} & & & \pmb{0} 
%	\end{tikzcd}
	\end{center}   
	There are two maximal reverse paths from $v_6$: 
	$(v_6, v_4, v_2, v_3, v_1)$ traversing 5 nodes 
	and $(v_6, v_5, v_3, v_1)$ traversing 4 nodes. 
	 This verifies the first point. Now,  
	 $label(v_6, v_6) = \emptyset$, $label(v_6, v_5) = label(v_6, v_4) = \{v_6\}$, 
	$label(v_6, v_3) = \{v_2, v_5\}$, $label(v_6, v_2) = \{v_4\}$. 
	In other words, $label(v_6, v_i)$ with $i \not= 1$   
	is the set of all the previous nodes of $v_i$ in the reverse path 
	from $v_6$ to $v_i$. However,  
	$label(v_6, v_1)$ is not $\{v_3, v_4\}$ even though 
	both $v_3$ and $v_4$ are nodes that precede $v_1$ 
	in some reverse path from $v_6$ to $v_1$. This is 
	because the reverse path $(v_6, v_4, v_1)$ 
	is subsumed in the reverse path $(v_6, v_4, v_2, v_3, v_1)$. 
	Hence, $label(v_6, v_1) = \{v_3\}$, verifying the second point. 
%	It is worth noting that if we remove the edge $(v_1, v_4)$  
%	from this DAG into the DAG shown on the right hand side above, 
%	we still obtain the same output for $label(v_6, v_i)$ 
%	for each $1 \leq i \leq 6$. 
 	\hfill$\clubsuit$ 
\end{example}

\begin{theorem}[Equivalence] \label{thm_equivalence} \rm    
	Given a typed argumentation graph $\modell$, some $s \in Stmt[G]$, $t \in \pmb{Themes}$ 
	and $trel \in \{`attack$'$, `support$'$\}$, 
	let $((V, E), expand, label)$ be such that 
	$((V, E), expand)$ is $dag(G_{t, trel})$ 
	and that $((V, E), label)$ is the labelled DAG of $(V, E)$, 
	then all the following hold. 
	\begin{itemize} 
		\item for every $S \in DepS[s, t, trel]$, there is some terminal node $v' \in V$ 
			and some sequence $v_n, \ldots, v_1$ of members of $V$ 
			such that all the following conditions hold. 
			\begin{itemize} 
				\item $v_1 = v'$ and $v_n = condense_{G_{t,trel}}(s)$. 
				\item $v_n, \ldots, v_1$ is such that, for every $1 \leq i \leq n-1$, 
					$v_{i+1} \in label(v_n, v_i)$ holds.   
				\item  $(\bigcup_{v \in \{v_n, \ldots, v_1\}} expand(v)) = S$.   
			\end{itemize} 
		\item for every terminal node $v' \in V$ and every sequence $v_n, \ldots, v_1$ 
			of members of $V$, 
			if $v_1 = v'$, $v_n = condense_{G_{t,trel}}(s)$ and $v_{i+1} \in label(v_n, v_i)$ 
			for every $1 \leq i \leq n-1$, then 
			$\bigcup_{v \in \{v_n, \ldots, v_1\}} expand(v)$ is a member of $DepS[s, t, trel]$. 
	\end{itemize} 
\end{theorem}  
\vspace{-0.26cm} 
\textbf{Proof.} Follows from Proposition \ref{prop_equivalence} and Definition \ref{def_labelled_dag}. \hfill$\Box$ \\

\noindent There is an efficient algorithm for deriving the labelled DAG of a DAG. Moreover, 
given a labelled DAG as the rerpesentation of a depth-statements-set, 
we can efficiently tell whether a set of statements belongs to the depth-statements-set.

\begin{lemma}[{\hyperref[proof_polynomial_time_transformation_dag_labelled_dag]{Polynomial-time transformation of a DAG into a labelled DAG}}] \label{lem_polynomial_time_transformation_dag_labelled_dag}
   Given a DAG $(V, E) \in (\mathcal{G}^+ \backslash \mathcal{G})$, 
	the labelled DAG of $(V, E)$ is polynomial-time computable. 
%	let $(V, E')$ be a DAG satisfying $(V, E') \leq_{\mathsf{DAG}} (V, E)$, 
%	then $(V, E')$ is polynomial-time computable.  
\end{lemma} 

\begin{proposition}[{\hyperref[proof_decide_membership_deps]{Deciding the membership of a depth-statements-set}}] \label{prop_decide_membership_deps} 
	Given a typed argumentation graph $\modell$, some $s \in Stmt[G]$, $t \in \pmb{Themes}$, 
	$trel \in \{`attack$'$, `support$'$\}$  and $S \subseteq Stmt[G]$, 
	deciding whether $S$ is a member of $DepS[s,t,trel]$ is in P. 
\end{proposition}

\noindent Deciding the absence of redundancy in a set of statements 
with respect to a theme involves an optimisation problem around minimality. 
\begin{definition}[NOREDUNDANT] \rm 
	We define \mbox{NOREDUNDANT} to be a decision problem with inputs (1) some $\taam$, 
	 (2) some $S_1 \subseteq Stmt[G]$ 
%	$\{S_1, \ldots , S_n\} \subseteq \mathfrak{p}(Stmt[G])$ for some $n$ bounded by
%	a polynomial of the length of the representations of $\taam$ and $\pmb{Themes}$ %$p(|Stmt[G]|)$ 
	and (3) some $t \in \pmb{Themes}$ for verifying that 
	$S_1$ contains no redundancy with respect to $t$. \hfill$\spadesuit$ 
%	\in \{S_1, \ldots, S_n\}$ 
%	contains redundancy with respect to $t$.   \hfill$\spadesuit$ 
\end{definition} 
To obtain an upper bound of (p-)$\Sat{faW}$ and (p-)$\Sat{faD}$, %\pmb{faW} and \pmb{faD}, 
we make use of the following decision problems to over-approximate 
the complexity of \mbox{NOREDUNDANT}. %problem. 
%the complexity 
%overapproximate 
%
%To obtain an upper bound of NOREDUNDANT, we make use of the following 
%decision problems. 
%However, since we are obtaining an upper complexity bound,  
%it sufficies to make use of the following decision problem. %But since we are deriving an upper bound, it suffices 
%to make use of the following decision problem.  
\begin{definition}[SUBTEST] \rm 
We define \mbox{SUBTEST} to be a decision problem which takes 
	inputs (1) some $\taam$, 
	(2) some $S_1 \subseteq Stmt[G]$,  
	%$\{S_1, \ldots, S_n\}$ for some $n$ bounded by  
	%a polynomial of the length of the representations of $\taam$ and $\pmb{Themes}$, 
	(3) some $t \in \pmb{Themes}$, (4) some $s_1 \in S_1$, (5) some $s_2 \in S_1$ 
	and (6) some $D_1 \subseteq \mathcal{I}(\{t\}, s_1)$ for verifying  
	that at least one of the following conditions holds: 
 that $s_1 = s_2$, that $D_1 = \emptyset$, 
	that the effective aspects of $D_1$, $\mathcal{I}(\{t\}, s_1)$ and 
$\mathcal{I}(\{t\}, s_2)$ do not coincide, or that 
$D_1 \not\subseteq \mathcal{I}(\{t\}, s_2)$.  \hfill$\spadesuit$ 
\end{definition} 
\begin{definition}[SUBTESTSUCSOME] \rm 
	We define \mbox{SUBTESTSUCSOME} to be a decision problem 
	with the same inputs as \mbox{SUBTEST}
	for verifying that $s_1 \not= s_2$ and that there is 
	some non-empty $D_2 \subseteq D_1$ 
	such 
	that (A) the effective aspects of $D_2$, $\mathcal{I}(\{t\}, s_1)$ and 
	$\mathcal{I}(\{t\}, s_2)$ coincide, and that (B) 
$D_2 \subseteq \mathcal{I}(\{t\}, s_2)$. \hfill$\spadesuit$ 
\end{definition} 
%(B) 
%$D_1$ is a minimal representation of $\mathcal{I}(\{t\}, s_1)$.   
Of course, it is only the complexity of \mbox{NOREDUNDANT} that is being over-approximated. As for the accept/reject 
judgement, \mbox{NOREDUNDANT} and the pair of \mbox{SUBTEST} and \mbox{SUBTESTSUCSOME} make the following
equivalent judgement. 
\begin{proposition} 
	Given a theme aspect argumentation model $\taam$, some $S_1 \subseteq Stmt[G]$ 
	and some $t \in \pmb{Themes}$, NOREDUNDANT accepts  
	$(\taam, S_1, t)$ 
	iff 
	the following holds for every $s_1, s_2 \in S_1$ and every $D_1 \subseteq \mathcal{I}(\{t\}, s_1)$: 
	either SUBTEST or SUBTESTSUCSOME accepts 
	$(\taam, S_1, t, s_1, s_2, D_1)$. 
\end{proposition}  
\vspace{-0.26cm} 
\textbf{Proof.} Obvious by the definition of $\preceq$ and that of minimal representation 
(Definition \ref{def_minimal_representation}). \hfill$\Box$  \\

%\noindent For any input $(\taam, \pmb{Themes}, S_1, t, s_1, s_2, D_1)$, 
%if either \mbox{SUBTEST} or accepts it, either $s_1$ $D_1 \not\preceq \mathcal{I}(\{t\}, s_1)$ or 
%$D_1 \not\preceq \mathcal{I}(\{t\}, s_2)$ holds, 
%and if \mbox{SUBTESTSUCSOME} accepts it, $D_1$ is not a non-empty minimal representation 
%of $\mathcal{I}(\{t\}, s_1)$ and $\mathcal{I}(\{t\}, s_2)$. 

\begin{proposition}[{\hyperref[proof_deciding_subtest_and_subtestall]{Deciding SUBTEST and \mbox{SUBTESTSUCSOME}}}] 
	\label{lem_deciding_subtest_and_subtestall} {\ }\\ 
\mbox{SUBTEST} is \mbox{NP}.  
	\mbox{SUBTESTSUCSOME} is in \mbox{NP(N}P). 
%by Lemma \ref{lem_co_np_problems} since it is a tautology check on 
%$(\bigwedge D_1 = \bigwedge \mathcal{I}(\{t\}, s_1)) \wedge   
%(\bigwedge \mathcal{I}(\{t\}, s_1) = \bigwedge \mathcal{I}(\{t\}, s_2))
%\wedge (D_1 \subseteq \mathcal{I}(\{t\}, s_2))$.  
\end{proposition}  
\subsection{Upper bounds of (p-)Sat:\pmb{faW}, (p-)Sat:\pmb{faD} 
and Sat:\pmb{All}}  
We establish the following inclusions 
for (p-)Sat:\pmb{faW} and (p-)Sat:\pmb{faD} and obtain  
Sat:\pmb{All}. 
\begin{theorem}[(p-)Sat:\pmb{faW}] \label{thm_sat_faw}  
	$\Sat{faW} = \SatOne{faW}$. 
	$\Sat{faW}$ is in co-NP(NP(NP)).  
%	$\SatP{faW}$ and $\SatOneP{faW}$ are in co-NP-hard. 
\end{theorem}  
\vspace{-0.2cm} 
\textbf{Proof.}  
$\Sat{faW}$ considers singleton sets of $\pmb{Themes}$. Thus, it is obvious that $\Sat{faW} = \SatOne{faW}$.  

There is a polynomially-bounded bit sequence $\{0,1\}^{\lceil \log |Stmt[G]| \rceil + \lceil \log |\pmb{Themes}| \rceil + 1 + 2\cdot\lceil \log |Stmt[G]| \rceil + m}$ 
where $m$ is no smaller than $\max_{t \in \pmb{Themes}, s \in Stmt[G]}|\mathcal{I}(\{t\}, s)|$.  
%(By the restriction on $\mathcal{I}$, $m$ is indeed bounded by a polynomial of the length of the representation 
%of $\taam$. 0 is assumed for the padding $\kappa$.)  
This bit sequence uniquely identifies a member $s$ of $Stmt[G]$ by 
the first $\lceil \log |Stmt[G]| \rceil$ bits, 
a member $t$ of $\pmb{Themes}$ by the next $\lceil \log |\pmb{Themes}| \rceil$ bits 
and a member $trel$ of $\{`attack$'$, `support$'$\}$ by the next 1 bit.   
By Proposition \ref{prop_uniqueness}, 
%for any $s \in Stmt[G]$, $t \in \pmb{Themes}$ and $trel \in \{`attack', `support'\}$,
either $WidS[s,t,trel] = \emptyset$ 
or else there is only 1 member 
in $WidS[s, t, trel]$. By Lemma \ref{lem_computing_width_statements_sets}, $WidS[s,t,trel]$ is polynomial-time 
computable.  %Since
%there are $|Stmt[G]|$ members in $Stmt[G]$ and $|\pmb{Themes}|$ members in $\pmb{Themes}$,   
If $WidS[s,t,trel] = \emptyset$, we have nothing to do. Otherwise, denote the member of 
$WidS[s,t,trel]$ by $S$. $S$ contains no redundancy 
with respect to $t$ just when the following holds for every $s_1, s_2 \in S$ and every $D_1 \subseteq 
\mathcal{I}(\{t\}, s_1)$: \mbox{SUBTEST} or \mbox{SUBTESTSUCSOME} accepts 
$(\taam,  S, t, s_1, s_2, D_1)$.   
Now, let us denote the $(\lceil \log |Stmt[G]| \rceil + \lceil \log |\pmb{Themes}| \rceil + 1)$ -bit pattern that identifies the above-described $s$, $t$ and $trel$ by $b$, and let $\Gamma$ denote all the bit patterns of the bit sequence that start with $b$. 
Then, $\Gamma$ contains all the bit patterns that identify 
every such $s_1$ and $s_2$ uniquely in the $2 \cdot \lceil \log |Stmt[G]| \rceil$ bits 
from the $(\lceil \log |Stmt[G]| \rceil + \lceil \log |\pmb{Themes}| \rceil + 2)$-th bit 
to the $(\lceil \log |Stmt[G]| \rceil + \lceil \log |\pmb{Themes}| \rceil + 1 + 2 \cdot \lceil \log |Stmt[G]| \rceil)$-th bit 
and every such subset $D_1$ of $\mathcal{I}(\{t\}, s_1)$ in the next 
$|\mathcal{I}(\{t\}, s_1)|$ bits of the remaining bits. Of course, those bits may identify some $s_1 \not\in S$, 
$s_2 \not\in S$ or $D_1 \not\subseteq \mathcal{I}(\{t\}, s_1)$. We simply skip those bit patterns. We also 
simply ignore any bit in the remaining $(m - |\mathcal{I}(\{t\}, s_1)|)$ bits. 
So, we only need to check the following 
for each bit pattern $b_{full} \in \Gamma$ in the $\{0,1\}^{\lceil \log |Stmt[G]| \rceil + \lceil \log |\pmb{Themes}| \rceil + 1 + 2\cdot\lceil \log |Stmt[G]| \rceil + m}$ 
bit sequence: $b_{full}$ is either an irrelevant bit pattern as described above or else 
it is such that $s,t$ and $trel$ identified 
by $b_{full}$ identifies $S \in WidS[s,t,trel]$,  
that $s_1$ and $s_2$ identified by $b_{full}$ 
are members of $S$, that $D_1 \subseteq \mathcal{I}(\{t\}, s_1)$ is identified by $b_{full}$, 
and that 
SUBTEST or SUBTESTSUCSOME accepts 
$(\taam, S, t, s_1, s_2, D_1)$.  For any other bit patterns that do not start with $b$ in the first $\lceil \log |Stmt[G]| \rceil + \lceil \log |\pmb{Themes}| \rceil + 1$ bits, too, 
we only need to check all the bit patterns in the remaining bits in the bit sequence and make 
the same judgement. Hence, there is actually no need 
of fixing the $|b|$ bits: it suffices to make such a judgement for each bit pattern in the bit sequence. 
Since $|\mathfrak{p}(\mathcal{I}(\{t\}, s_1))|$ 
is not guaranteed to be polynomially-bounded, $\Sat{faW}$ is in co-NP(\mbox{SUBTEST} $\cup$ \mbox{SUBTESTSUCSOME}) 
which is in co-NP(NP(NP)) by Lemma \ref{lem_subsumption_complexity_classes}. \hfill$\Box$ \\
%See Appendix:  proofs. \hfill$\Box$\\ 

\noindent Interestingly, (p-)$\Sat{faD}$ is also contained in the same complexity class. 
\begin{theorem}[(p-)Sat:\pmb{faD}]  \label{thm_sat_fad}
	$\Sat{faD} =  \SatOne{faD}$. 
	$\Sat{faD}$ is in \mbox{co-NP(NP(N}P)). 
%	$\SatP{faD}$ and $\SatOneP{faD}$ 
%	are in \mbox{co-NP(N}P)-hard. 
%      Given $\taam$, verification of $\taam$'s satisfiability of \pmb{nwci} is P.  
\end{theorem} 
\vspace{-0.2cm} 
\textbf{Proof.}   
$\Sat{faD}$ considers singleton sets of $\pmb{Themes}$. Thus, it is obvious that $\Sat{faD} = \SatOne{faD}$.   

The key difference from $\Sat{faW}$ is in that $DepS[s,t,trel]$ ($s \in Stmt[G]$, $t \in \pmb{Themes}$,  $trel \in 
\{`attack$'$, \linebreak `support$'$\}$) can contain an exponential number of members; see Example \ref{ex_dag}. 
By the labelled DAG representation, however, it suffices to consider the following 
polynomially-bounded bit sequence $\{0,1\}^{n}$ 
where $n \equiv \lceil \log |Stmt[G]| \rceil + \lceil \log | \pmb{Themes} | \rceil + 1 + |Stmt[G]| + 2\cdot \lceil \log | Stmt[G]|\rceil + m$. Again, $m$ is no smaller than $\max_{t \in \pmb{Themes}, s \in Stmt[G]}|\mathcal{I}(\{t\}, s)|$.   
Compared to the bit sequence used for $\Sat{faW}$,  
this bit sequence contains extra $|Stmt[G]|$ bits. The bit sequence uniquely identifies 
a member $s$ of $Stmt[G]$ by 
the first $\lceil \log |Stmt[G]| \rceil$ bits, 
a member $t$ of $\pmb{Themes}$ by the next $\lceil \log |\pmb{Themes}| \rceil$ bits 
and a member $trel$ of $\{`attack$'$, `support$'$\}$ by the next 1 bit, just like the bit sequence used for $\Sat{faW}$ does. 
The next $|Stmt[G]|$ bits uniquely identify a subset $S$ of $Stmt[G]$. 

By Lemmas \ref{lem_polynomialtime_transformation_g_dagg} and \ref{lem_polynomial_time_transformation_dag_labelled_dag}, 
construction of the labelled DAGs for $t$ and $trel$  
is polynomial-time doable. By Proposition \ref{prop_decide_membership_deps}, checking $S \in DepS[s,t,trel]$ is polynomial-time doable. 
$S$ contains no redundancy with respect to $t$ just when the following holds for every 
$s_1, s_2 \in S$ and every $D_1 \subseteq \mathcal{I}(\{t\}, s_1)$: 
\mbox{SUBTEST} or \mbox{SUBTESTSUCSOME} accepts $(\taam, S, t, s_1, s_2, D_1)$.  
The given bit pattern uniquely identifies $s_1, s_2 \in S$ in the $2 \cdot \lceil \log |Stmt[G]| \rceil$ bits 
from the $(\lceil \log |Stmt[G]| \rceil + \lceil \log |\pmb{Themes}| \rceil + 1 + |Stmt[G]| + 1)$-th bit 
to the $(\lceil \log |Stmt[G]| \rceil + \lceil \log |\pmb{Themes}| \rceil + 1 + |Stmt[G]| + 2 \cdot \lceil \log |Stmt[G]| \rceil)$-th bit 
and every such subset $D_1$ of $\mathcal{I}(\{t\}, s_1)$ in the next 
$|\mathcal{I}(\{t\}, s_1)|$ bits of the remaining bits. 
Thus, by the same reasoning as in Theorem \ref{thm_sat_faw}, $\Sat{faD}$ is in \mbox{co-NP(S}UBTEST $\cup$ 
SUBTESTSUCSOME) which is in \mbox{co-NP(NP(N}P)) by Lemma \ref{lem_subsumption_complexity_classes}. 
In the worst case, we need to check almost all the members of $\mathfrak{p}(\mathcal{I}(\{t\}, s_1))$ and, in addition, 
of $\mathfrak{p}(Stmt[G])$. Neither $|\mathfrak{p}(\mathcal{I}(\{t\}, s_1))|$ nor 
$|\mathfrak{p}(Stmt[G])|$
is guaranteed to be polynomially bounded. \hfill$\Box$ 
%See Appendix:  proofs. \hfill$\Box$\\ 

\begin{theorem}[$\Sat{ \pmb{All}}$] \label{thm_sat_all} 
	$\Sat{ \pmb{All}}$ is co-NP(NP)-hard and 
	is in co-NP(NP(NP)). 
\end{theorem}  
\vspace{-0.2cm} 
\textbf{Proof.} By Theorems \ref{thm_hard_core_all}, \ref{thm_sat_core_complete}, \ref{thm_sat_faw} and \ref{thm_sat_fad}. \hfill$\Box$\\

{\it \hyperref[app_implementation_detail]{Appendix: implementation, gathered data 
and run-time evaluation}} 
shows a decent alignment of the actual run times to the theoretical expectations. 

We believe it unlikely  
that the upper bound of $\Sat{ \pmb{All}}$ can be 
tightened further (unless P = NP). %Theoretically, too, 
The next lower co-NP(X) below \mbox{co-NP(NP(N}P)) is \mbox{co-NP(N}P), which seems to imply  
that the upper bound of $\bigcup_{\pmb{x} \in \pmb{F}}\Sat{$x$}$ becomes 
co-NP(NP) if the tightening is possible at all. Since 
SUBTEST (which is NP by Proposition \ref{lem_deciding_subtest_and_subtestall}) 
is called in $\bigcup_{\pmb{x} \in \pmb{F}}\Sat{$x$}$,
it cannot become as low as \mbox{co-N}P unless NP = \mbox{co-N}P. 
But then, %\mbox{co-NP(S}UBTEST) is already in \mbox{co-NP(N}P), so 
achieving \mbox{co-NP(N}P) for $\bigcup_{\pmb{x} \in \pmb{F}}\Sat{$x$}$ seems to require 
the confirmation 
that making the minimality judgement around redundancy is algorithmically no harder than SUBTEST, 
which is unlikely to be so obtained 
since $\mathfrak{p}(\mathcal{I}(\{t\},s_1))$ ($t \in \pmb{Themes}, s_1 \in Stmt[G]$) is not guaranteed to be polynomially bounded.  
%However, as NOREDUNDANT is an optimisation problem which we over-approximated 
%with SUBTEST and SUBTESTSUCSOME, we do not have 
%a formal proof to conclude in the either. %The uncertainty 
%extends to the lower-bound, too, whether 
%it is also $\mathsf{\Gamma}_2$-hard. 
%at this stage. 
%we are not clear 
%at this stage if 
Second, while we focused on \pmb{Core} + $\alpha$,  not every one of $\Sat{$x$}$, $\pmb{x} \in \text{\pmb{Core}}$, 
is \mbox{co-NP(N}P)-hard. For instance, if we take just $\Sat{vi}$, it is 
trivially in P, and similarly, if we just take $\Sat{i}$, since $\SatOne{i}$ is in co-NP, it is not difficult to foresee that 
$\Sat{i}$ is also in \mbox{co-NP}. The identification of 
tighter bounds for individual constraints may be of (at least) some technical interest. 
Apart from these theoretical matters, there is a significant practical interest 
in finding suitable restrictions on $\taam$ 
so the formal constraints can be verified in polynomial time. These we leave as open problems to be addressed in future work. 
%further explorations to future work. 
%\hide{ 
%\section{Implementation and Evaluation} 
%\subsection{Implementation and run-time evaluation} \label{subsec_implementation}  
%\section{Comparisons of Methods to Reduce the Computational Complexity} 

\section{Conclusions}  
Seeing the need to explain how and why argumentation is judged fallacious,  
we proposed hybrid argumentation models of abstract and (unstructured) logic-based argumentation models 
%and logic-based 
%argumentation 
for modelling different aspects of argumentation. Rhetorical intentions,  
best modelled in abstract argumentation models, are modelled in an abstract argumentation model, 
while a semantic structure is provided for the semantics of statements 
as the basis of the semantics of 
`$attack$'s and `$support$'s. The syntax-semantics insulation 
accommodates context-dependent one-to-many mappings 
of statements to elements of the semantic structure.  
\begin{itemize} 
	\item In our instantiation with the theme aspect argumentation model, 
we examined the semantics of `$attack$'s and `$support$'s with a concrete 
Boolean algebra, identifying types of 
rhetorical-grade attacks and supports 
which have been seldom explored in logic-based argumentation models, structured or unstructured. Refer to Remarks \ref{note_xxx} and \ref{note_xyz}. 

%One of them was recently identified in \citep{Heyninck20}. 
	\item We then went through core constraints, 
and more constraints later, establishing  
the correspondence between \pmb{Core} constraints and  
the graphic constraints \pmb{tr}, \pmb{nnp}, \pmb{nsa} and 
		\pmb{kos} along the way. Refer to \mbox{Theorem \ref{thm_core}}. 
		Examples 
		\ref{ex_role_tr} through \ref{ex_role_fad} facilitate 
	 	intuitive understanding of the constraints. 
%		The full detail of the complete Boolean algebra(s)
%		and interpretation function(s) in these examples 
%		is in \citep{Nakai22}. 
		We characterised normal forms 
		for identifying fallacies 
		with formal 
		constraints. 
	%	formally classifying and identifying fallacies.  
		%The formal fallacy identification through 
		%the constraints is robust for all but 
		%\pmb{F} constraints in the following sense. 
		%When 
%Computational complexities  
%of deciding satisfiability of the constraints were reported, and the implemented 
%system was used to corroborate 
We also formed the notion of logico-rhetorical conclusion for testing 
the rhetorical conclusion ({\it i.e.} abstract argumentation models' acceptability semantics) against 
the semantic structure of a given hybrid argumentation model to see if 
		any logical fallacy lurks in the rhetorical conclusion. 
In future work, this concept may be explored 
		for empirical evaluation of acceptability semantics 
		(refer to \cite{Dung95}).  
\item We investigated the computational complexities of deciding 
	the satisfiability of the formal constraints. % also 
  	%obtaining the actual run-times through evaluation 
%	with an implemented system. 
		%establishing several key results.  
		It was identified that 
$\Sat{ \pmb{Core}} = 
\Sat{ \pmb{All}$^{- \text{\pmb{F}}}$} (= $
\mbox{co-NP(N}P)-complete$)$ (refer to \mbox{Theorem \ref{thm_sat_core_complete}}): given a theme aspect argumentation model $\taam$,   
checking the satisfiability of 
all the constraints but \pmb{F} is as theoretically complex 
as checking that of \pmb{Core}. 
%As such, when one is in indecision which 
%extra constraints in addition to \pmb{Core} should or should not be 
%		verified, 
%		he/she may as well just check all but \pmb{F} constraints. 
%		That obviates any recomputation at a later time. %for omitted 
		%constraints. One observation 
%		Satisfiability of these constraints need not be 
%%		recomputed 
%		even if some new nodes are added to $\modell$, 
%		unlike 
		 On the other hand, $\Sat{ \pmb{All}}$ 
		 is likely 
		 more complex. Refer to 
		Theorem \ref{thm_sat_all}.    
		For \pmb{Core} constraints, 
		we showed in 
		Theorem \ref{thm_model_existence_model_checking} 
		that, 
		given a typed argumentation graph $\modell$, 
		whether there exists some $\mathcal{D},\mathcal{I}$ 
		such that $\taam$ satisfies them 
		can be verified in polynomial time. 
		As theorems \nameref{thm_sat_aass} through 
		\nameref{thm_sat_nwci} indicate, 
		the complexity 
		gets lowered for p-Sat decision problems 
		for all the constraints on the theme aspect 
		argumentation model except for \pmb{F} constraints.
		%polynomial-time verifiable. 
%When satisfiability of the theme aspect argumentation model $\taam$ is judged,
\end{itemize} 
%As with any formal characterisation of an informal concept, 
%formal fallacy identification 
%This marks the end. 
We showed a successful initial formal fallacy identification 
as a viable alternative to traditional 
fallacy identification by informal criteria, 
with transparency and explainability. 

%including 
% and carried out evaluations of the degree of association of (un)satisfaction 
%of constraints and of the run time to obtain empirical results. We then explored computational 
%complexities of the constraints to confirm the alignment of the practical and theoretical 
%expectation of time complexity.

%\subsection{Future research directions}  

\subsection{Future research directions} 
\subsubsection{Other instantiations of the hybrid argumentation model} 
The theme aspect argumentation model is only one instantiation 
of the hybrid argumentation model. Any abstract argumentation model can be used 
for the first component, and some other semantic structure 
may be used for the second component. To be fair, 
the complete Boolean algebra is already general insofar as 
a finite domain is sufficient. %easily accommodating 
%temporal progression, causal relation and so on. 
To wit, suppose a finite Boolean algebra 
			    $\mathcal{D}$ without any 
			    temporal information contained within. 
			    Suppose a Boolean algebra 
			    $\mathcal{D}_1$ 
			    with $\pmb{Atom}(\mathcal{D}_1) 
			    = \{t_0, \neg t_0\}$ denoting 
			    one time step $t_0$. 
			    $\mathcal{D}$ and $\mathcal{D}_1$ 
			    can be combined into  
			    a larger complete Boolean algebra 
			    $\mathcal{D}'$ where  
			    $\mathcal{D}$'s components  
			    are associated with the temporal 
			    information in $\mathcal{D}_1$. 
			    Cf. Inner-Lemma 
			    \ref{innerlem_disjoint_combination}. 
			    $\mathcal{D}'$ can be further extended with 
			    other time steps so long as the number 
			    is finite. Similarly, causal relation 
			    is expressible so long as 
			    the number of states to transition into is finite. 
Notwithstanding, for some situations,  the cylindric algebra or a 
modal algebra may be a better fit, and, for intuitionism, Heyting algebra 
is required. 

\subsubsection{Investigation into scalability, precision and automatability of 
formal fallacy identification} 
Just as with verification of a computer program, it is impossible to verify 
everything. Naturally, some information must be 
abstracted or else ignored; refer to Example 
\ref{ex_illustration_semantic_granularity}
and other examples. A pertinent question to ask is 
if we can guarantee sound abstraction \cite{Cousot77}. 
Exploring techniques that achieve it  
should be instrumental for the scalable and meaningful 
application of the theme aspect argumentation model in reasoning 
about fallacies. 
Finding suitable restrictions on 
$\taam$ is an alternative way to reduce the time complexity. 
For bolstering the automation of formal fallacy identification,   
it is important not only to extract the rhetoric model (for which
argumentation mining techniques can be used), 
but also to find an appropriate way to automate generation of  
$\mathcal{I}$, given $\modell$ and $\mathcal{D}$. 
Relevant formal encoding 
methodologies from computational linguistics  
might offer insights here. 
As for the balance between scalability and precision, 
theoretical investigations should continue for enhancing 
explainability. Concurrently, 
there is a need to further 
neuro-symbolic research. %should be furthered at the same time. 
Integrating 
a streamlined version of the theme aspect argumentation model 
into large language models appears to be a promising research direction.

\section*{Acknowledgements}    
This work was supported by JSPS KAKENHI Grant Number 21K12028. 
The first author was also supported by JST Sakigake Tokutei Kadai Chosa (Contract Number 21-211035057). 
%The first author thanks Yusuke Kawamoto for valuable comments on the presentation 
%of the 1st and the 2nd drafts.  
%In conducting this paper's research, 
%the first author was supported in part by JST Sakigake Tokutei Kadai Chosa (Contract Number 21-211035057) 
%and in part by JSPS KAKENHI 21K12028, 
%the second author was supported by JSPS KAKENHI 21K12028. 
%and the third author was supported by JSPS KAKENHI 21K12028. 
%and the fourth author was supported by KAKENHI ( ). 
%This work was 
%The first author was supported by JST SAkigake 
%This work was supported by JST Sakigake Tokutei Kadai Chosa (Contract Number 21-211035057). 

\pagebreak 
\section*{{\hyperref[text_table_of_notations]{Appendix: table of notations}} (stating notations, descriptions 
and their 1st appearance)} \label{app_table_of_notations}
%\begin{table}[htbp]  
	\begin{center}% used the environment to augment the vertical space
		{\scriptsize 
		\begin{longtable}{r c p{10cm} p{2.4cm}}
			% between the caption and the table
		%\begin{tabular}{X[r] X[c] X[p{10cm}] X[l] }
\toprule  
			Notation & & Description & 1st appears in \\ 
			\midrule 
			\multicolumn{3}{c}{\underline{Class-like Sets}} \\ 
%\midrule
			$\pmb{A}$ & $\triangleq$ & infinite set of 
			statements & Sec. \ref{sec_theme_aspect_argumentation_model} \\
%			$A$ & $\triangleq$ & some finite subset of 
%			statements (subsumed in $\pmb{A}$) & 
%			Sec. \ref{sec_technical_preliminaries}\\
			$\pmb{A}^{ord}$ & $\triangleq$ & non-empty set of 
			ordinary statements (subsumed in $\pmb{A}$) & Sec. \ref{sec_theme_aspect_argumentation_model}, Def. \ref{def_pointer_and_ordinary_statements} \\
			$\pmb{A}^{pnt}$ & $\triangleq$ & non-empty set of 
			pointer statements (subsumed in $\pmb{A}$) & Sec. \ref{sec_theme_aspect_argumentation_model}, Def. 
			\ref{def_pointer_and_ordinary_statements} \\  
									$EXPAND$ & $\triangleq$ & set of all functions $expand: \pmb{V} \rightarrow \mathfrak{p}(\pmb{A})$ & 
			Sec. \ref{sec_computational_complexities}, Def. \ref{def_dag_transformation}\\
			$\mathcal{G}$ & $\triangleq$ & non-empty set of all finite 
			argumentation graphs (subsumed in $\mathcal{G}^+$)
			& Sec. \ref{sec_theme_aspect_argumentation_model} \\  
			$\mathcal{G}^+$ & $\triangleq$ & non-empty set of all finite graphs & Sec. \ref{sec_computational_complexities}\\
					%	$R$ & $\triangleq$ & set of statement-to-statement 
%			relations & Sec. \ref{sec_technical_preliminaries}\\ 
			%$T$ & $\triangleq$ & subset of $\pmb{Themes}$ & Sec. \ref{sec_understanding_the_semanics}\\
			$\pmb{Themes}$ & $\triangleq$ & countable set of theme 
			constants (subsumed in $\pmb{Types}$) & Sec. \ref{sec_theme_aspect_argumentation_model} \\ 
			$\pmb{Types}$ & $\triangleq$ & countable set of types & Sec. \ref{sec_theme_aspect_argumentation_model}\\
			$\pmb{V}$ & $\triangleq$ & infinite set of graph nodes unrelated to argumentation & 
			Sec. \ref{sec_computational_complexities}\\
									\midrule 
									\multicolumn{3}{c}{\underline{Variables}} \\  
      $a$ & & denotes an ordinary statement & Sec. \ref{sec_theme_aspect_argumentation_model}, Def. \ref{def_pointer_and_ordinary_statements}\\ 
 $A$ & & denotes a finite set of statements (subsumed in $\pmb{A}$) & Sec. \ref{sec_theme_aspect_argumentation_model}\\
 $b$ & & denotes a bit pattern & Sec. \ref{sec_computational_complexities}\\
  C & & denotes a complexity class & Sec. \ref{sec_computational_complexities}\\  
  $D$ &  & denotes a partially ordered set or the underlying partially ordered set of a complete Boolean algebra $\mathcal{D}$ 
			& Sec. \ref{sec_theme_aspect_argumentation_model}\\
			$\mathcal{D}$ & & denotes a complete Boolean algebra & Sec. \ref{sec_theme_aspect_argumentation_model}, 
			Def. \ref{def_complete_boolean_algebra_aspects}\\
			$E$ & & denotes a set of edges of a finite directed graph unrelated to argumentation & Sec. \ref{sec_computational_complexities}\\
			$G$ & & denotes a finite argumentation graph  
			(a member of $\mathcal{G}$) 
			& Sec. \ref{sec_theme_aspect_argumentation_model} \\
			$i$ (, $j$, $k$, $l$, $m$, $n$) & & denotes some entity, most notably a natural number & Sec. \ref{sec_theme_aspect_argumentation_model}, Ex. \ref{ex_illustration_semantic_granularity}\\
%			\ref{sec_core_constraints_of_theme_aspect_argumentation_model}, Lem. \ref{lem_disjoint_combination}\\
  $L$ &  & denotes the underlying partially ordered 
			set of a lattice $\mathcal{L}$ & Sec. \ref{sec_theme_aspect_argumentation_model}\\
		$\mathcal{L}$ & & denotes a lattice & Sec. \ref{sec_theme_aspect_argumentation_model}\\ 
%			 $N$ & & denotes a sample size (a natural number) & Sec. \ref{sec_implementation}\\
			  $R$ & & denotes a set of statement-to-statement 
			relations & Sec. \ref{sec_theme_aspect_argumentation_model}\\ 
			$s$ & & denotes a statement & Sec. \ref{sec_theme_aspect_argumentation_model}\\  
%			${s::t}$ & & is a string indicating $t \in \pmb{Themes}$ and $s \in Stmt[G]$ & Appendix: proofs, Proof of Thm. 10 \\
 $S$ & & denotes a set of statements in typed argumentation graph & Sec. \ref{sec_more_constraints}, Def. \ref{def_theme_related_statements_sets}\\
 $t$ & & denotes a theme constant (often simply written a theme) & Sec. \ref{sec_theme_aspect_argumentation_model} \\
 $T$ & & denotes a subset of $\pmb{Themes}$ & Sec. \ref{sec_theme_aspect_argumentation_model}\\
  $t.a$ & & denotes a pointer statement referring to 
 an ordinary statement $a$ in theme $t$ & Sec. \ref{sec_theme_aspect_argumentation_model}\\
 $t.\mathcal{C}$ & & denotes a pointer statement referring to 
 a theme $t$ & Sec. \ref{sec_theme_aspect_argumentation_model}\\  
 $trel$ & & denotes either `$attack$' or `$support$' & Sec. \ref{sec_more_constraints}, Def. \ref{def_theme_related_statements_sets}\\  
 $type$ & & denotes a member of $\pmb{Types}$ & Sec. \ref{sec_theme_aspect_argumentation_model}\\ 
%  $u$ & & denotes either $\mathsf{F}$ or $\mathsf{T}$ & Sec. \ref{sec_implementation}\\
  $v$ & & denotes a node of a finite directed graph unrelated to argumentation & Sec. \ref{sec_computational_complexities}, Def. \ref{def_dag_transformation}\\
 $V$ & & denotes a set of nodes of a finite directed graph unrelated to argumentation & Sec. \ref{sec_computational_complexities}\\
% $\mathcal{V}$ & & denotes a Cram{\'e}r's $V$ & Sec. \ref{sec_implementation}\\
 $w$ & & denotes a word & Sec. \ref{sec_computational_complexities}\\ 
 $x$ (, $y$, $z$) &  & denotes a member of some set or 
 a special symbol & Sec. \ref{sec_theme_aspect_argumentation_model}\\ 
 $\pmb{x}$ (, $\pmb{y}$) & & denotes a constraint either on the typed argumentation graph or on 
 the theme aspect argumentation model & Sec. \ref{sec_computational_complexities}, Def. \ref{def_decision_problems}\\    
 X (, Y) & & denotes some entity, most notably a decision problem or a complexity class & Sec. \ref{sec_introduction}\\
%  \pmb{X} (, \pmb{Y}) & & denotes a set of constraints either 
%  on the typed argumentation graph or on the theme aspect argumentation 
%  model & Sec. \ref{sec_core_constraints_of_theme_aspect_argumentation_model}, Def. \ref{def_normal_forms_and_fallacies}\\
   $\alpha$ (, $\beta$) & & denotes a set of constraints & Sec. \ref{sec_more_constraints}, Def. \ref{def_normal_forms_and_fallacies}\\  
 $\kappa$ & & is padding for matching the input length of a decision problem to the length 
 of the input to 3-SAT & Sec. \ref{sec_computational_complexities}\\ 
 $(G, \Pi)$ &  & denotes a typed 
			finite argumentation graph & Sec. \ref{sec_theme_aspect_argumentation_model}\\ 
			$\modell$ & & denotes a well-formed 
			typed finite argumentation graph & Sec. \ref{sec_theme_aspect_argumentation_model}\\
			$(D, \vee, \wedge)$ &  & 
			denotes a lattice & Sec. \ref{sec_theme_aspect_argumentation_model}\\
			$(D, \pmb{1}, \pmb{0}, \neg, \vee, \wedge)$ & 
			& denotes a Boolean algebra & Sec. \ref{sec_theme_aspect_argumentation_model}\\
			$\taam$ & & denotes a theme aspect argumentation 
			model & Sec. \ref{sec_theme_aspect_argumentation_model}, Def. \ref{def_theme_aspect_argumentation_model}\\
			$(\modell, \mathfrak{s})$ & & denotes a logical fallacy with respect to some $\mathcal{D}, 
			\mathcal{I}$ & Sec. \ref{sec_more_constraints}, Def. \ref{def_fallacy_rhetoric_conclusion}\\
  \midrule 
\multicolumn{3}{c}{\underline{Constants}}\\   
						`$attack$' & $\triangleq$ & member of 
			$\pmb{Types}$ & Sec. \ref{sec_theme_aspect_argumentation_model}\\
			`$support$' & $\triangleq$ & member of 
			$\pmb{Types}$ & Sec. \ref{sec_theme_aspect_argumentation_model}\\
			$\omega$ & $\triangleq$ & special symbol 
			as the 2nd parameter of interpretation function $\mathcal{I}$ & 
			Sec. \ref{sec_theme_aspect_argumentation_model}\\
			$\mathcal{C}$ & $\triangleq$ & special symbol 
			used in the syntax of 
			a pointer statement $t.\mathcal{C}$ 
			referring to \pmb{s}ummary of a theme $t$ 
			& Sec. \ref{sec_theme_aspect_argumentation_model}, Def. \ref{def_pointer_and_ordinary_statements}\\  
%			$\mathsf{F}$ (, $\mathsf{T}$) & 
%			$\triangleq$ & constant signifying 
%			a satisfiability status & Sec. \ref{sec_implementation}\\
			$\pmb{1}$ & $\triangleq$ & top element of a lattice (in lattice) or the truth (in propositional formula) & Sec. \ref{sec_theme_aspect_argumentation_model}\\
			$\pmb{0}$ & $\triangleq$ & bottom element of a lattice (in lattice) or the falsehood (in propositional formula) & Sec. \ref{sec_theme_aspect_argumentation_model}\\
\midrule 
			\multicolumn{3}{c}{\underline{Functions}} \\ 
			$\pmb{Atom}$ & $\triangleq$ & function that takes a lattice as its input 
			for outputting its atoms & Sec. \ref{sec_theme_aspect_argumentation_model}\\
			$condense_G$ & $\triangleq$ & function that takes a statement (a member of $Stmt[G]$) for outputting 
			a node (a member of $\pmb{V}$), used for condensation of a finite argumentation graph $G$ into DAG & Sec. \ref{sec_computational_complexities}, Def. \ref{def_dag_transformation}\\
			$dag$ & $\triangleq$ & function that takes a finite argumentation graph $G$ for outputting 
			some $((V, E), expand)$ where $(V, E)$ is the condensation of $G$ and where $expand$ 
			is a function associating the nodes in $G$ with those in $(V, E)$ & Sec. \ref{sec_computational_complexities},
			Def. \ref{def_dag_transformation}\\
			$DepS$ & $\triangleq$ & function that takes inputs (1) some statement $s$, (2) 
			some theme constant $t$ and (3) $trel$ which is either `$attack$' or `$support$' 
			for outputting a set of depth-statements-sets of $s$ with respect to $t$ for $trel$ & Sec. \ref{sec_more_constraints}, Def. \ref{def_theme_related_statements_sets}\\ 
			$expand$ & $\triangleq$ & function that takes a node of DAG (a member of $\pmb{V}$) 
			for outputting a set of statements (a member of  $\mathfrak{p}(\pmb{A})$), used for obtaining 
			the set of statements in $G$ condensed into the node of DAG via $condense_G$ &  
			Sec. \ref{sec_computational_complexities}, Def. \ref{def_dag_transformation}\\ 
%			$f$ & $\triangleq$ & function that takes a natural number $n$ for outputting  
%			$\lfloor n^2 \cdot r_1 \cdot r_2^{0.3} \rfloor$ where each of $r_1$ and $r_2$  
%			is a randomly selected number between 0 and 1 inclusive & Sec. \ref{sec_implementation} \\
			$\mathcal{I}$ & $\triangleq$ & interpretation 
			function which takes inputs (1) a subset of $\pmb{Themes}$ 
			and (2) either a statement or 
			a special symbol $\omega$ 
			for outputting a subset of a partially 
			ordered set & Sec. \ref{sec_theme_aspect_argumentation_model} \\ 
%			$\pmb{JI}$ & $\triangleq$ & function that takes a lattice as its input 
%			for outputting its join-irreducibles & Sec. \ref{sec_core_constraints_of_theme_aspect_argumentation_model}, 
%			Def. \ref{def_join_irreducibles_and_atoms}\\
			$label$ & $\triangleq$ & function that takes two nodes $v$ and $v'$ (members of a subset $V$ of $\pmb{V}$ 
			in some DAG $(V, E)$) 
			for outputting a subset of $V$, used for assigning $v'$ with 
			a set of nodes that can appear in some of the reverse path(s) 
			from $v$ that occur immediately before $v'$ & Sec. \ref{sec_computational_complexities}, Def. \ref{def_labelled_dag}\\
			$\mathfrak{p}$ & $\triangleq$ & function that takes a set for outputting its power set & Sec. \ref{sec_theme_aspect_argumentation_model}\\  
			$\range$ & $\triangleq$ & function that takes a function for outputting 
			its range. & Sec. \ref{sec_more_constraints} \\  
			$\mathfrak{s}$ & $\triangleq$ & function that takes a typed argumentation graph
			$\modell$ and a subset $T$ of $\pmb{Themes}$ for outputting a subset of $\mathfrak{p}(\mathfrak{p}(Stmt[G]))$ 
			($\mathfrak{s}(\modell, T)$ is the rhetorical conclusion of $\modell$ with respect to $T$ and 
			$\mathfrak{s}$) 
			& Sec. \ref{sec_more_constraints}, Def. \ref{def_rhetoric_conclusion} \\ 
			$WidS$ & $\triangleq$ & function that takes inputs (1) some statement $s$, (2) 
			some theme constant $t$ and (3) $trel$ which is either `$attack$' or `$support$' 
			for outputting a set of width-statements-sets of $s$ with respect to $t$ for $trel$ & Sec. \ref{sec_more_constraints}, Def. \ref{def_theme_related_statements_sets}\\ 
			$\eta_D$ & $\triangleq$ & function that takes an element $x$ 
			of the underlying set $D$ of a finite Boolean algebra $\mathcal{D}$ for outputting a set of 
			atoms of $\mathcal{D}$ that are smaller than or equal to $x$ & Sec. \ref{sec_theme_aspect_argumentation_model} \\
			$\sigma$ & $\triangleq$ &  polynomial-time computable function which takes inputs (1) 
			a list of words $[w_1, \ldots, w_m]$ and (2) a $m$-bit sequence $\{0,1\}^m$ 
			for outputting a list of words containing only those words $w_x$ in $\{w_1, \ldots, w_m\}$ 
			for which the $x$-th bit in the bit sequence is 1 & Sec. \ref{sec_computational_complexities}\\
			$\Pi$ & $\triangleq$ & typing function taking 
			a statement or a statement-to-statement relation 
			for outputting a subset of $\pmb{Types}$, used for assigning 
			types to the input  
			& Sec. \ref{sec_theme_aspect_argumentation_model}\\ 
			%			\midrule 
%			\multicolumn{3}{c}{\underline{Relations}} \\ 
%			$R$ & $\triangleq$ & statement-to-statement 
%			relation & Sec. \ref{sec_technical_preliminaries}\\ 
			\midrule 
			\multicolumn{3}{c}{\underline{Operators (selected ones only)}} \\ 
			${\uparrow D}$ & $\triangleq$ 
			& the smallest up set of a partially ordered set $D$ & Sec. \ref{sec_theme_aspect_argumentation_model}\\
			${\downarrow D}$ & $\triangleq$ & 
			the smallest down set of a partially ordered set $D$ & Sec. \ref{sec_theme_aspect_argumentation_model}\\  
%			$\bigwedge D$ & $\triangleq$ & infimum of a partially ordered set $D$ if $D \not= \emptyset$; $\emptyset$, otherwise & 
%			Sec. \ref{sec_understanding_the_semanics}, Def. \ref{def_effective_aspect_of_statement}\\
%			$\bigvee D$ & $\triangleq$ & supremum of a partially ordered set $D$ if $D \not= \emptyset$; $\emptyset$, otherwise & 
%			Sec. \ref{sec_application_fallacy_detection_and_prevention}\\
			$D_1 \preceq D_2$ & $\triangleq$ & $D_1 \subseteq D_2$ and $\bigwedge D_1 = \bigwedge D_2$ 
			for partially ordered sets $D_1$ and $D_2$ & Sec. \ref{sec_more_constraints}, Def. \ref{def_minimal_representation}\\
%			$\bigwedge_i F_i$ & $\triangleq$ &  ($F_i$ for each $i$ is assumed a proposition in classical logic) 
% true 
%			iff $F_i$ is true for each $i$ 
%			& Sec. \ref{sec_understanding_the_semanics}, Rem. \ref{note_effective}\\
%			$\bigvee_i F_i$ & $\triangleq$ &  ($F_i$ for each $i$ is assumed a proposition in classical logic) 
% true 
%			iff $F_i$ is true for at least one $i$ 			& Appendix: proofs, Proof of Lem. \ref{lem_np_problems}\\
			$Stmt[G]$ & $\triangleq$ & set of statements  
			in a finite argumentation graph $G$ & Sec. \ref{sec_theme_aspect_argumentation_model}\\ 
			$OStmt[G]$ & $\triangleq$ & set of ordinary 
			statements in a finite argumentation graph $G$ (subsumed in $Stmt[G]$) 
			& Sec. \ref{sec_theme_aspect_argumentation_model}, Def. \ref{def_pointer_and_ordinary_statements}\\
			$PStmt[G]$ & $\triangleq$ & set of pointer 
			statements in a finite argumentation graph $G$ (subsumed in $Stmt[G]$)
			& Sec. \ref{sec_theme_aspect_argumentation_model}, Def. \ref{def_pointer_and_ordinary_statements}\\
			$Rel[G]$ & $\triangleq$ & set of statement-to-statement relations (= directed edges) 
			in a finite argumentation graph $G$ & Sec. \ref{sec_theme_aspect_argumentation_model}\\ 
			$G_t$ & $\triangleq$ & sub-theme finite argumentation graph (a sub-graph of $G$):  
			$Stmt[G_t]$ includes every $s \in Stmt[G]$ such that $t \in \Pi(s)$ but nothing else, 
			and $Rel[G_t]$ includes every $(s', s) \in Rel[G]$ such that $t \in \Pi((s', s))$ but nothing else 
			& Sec. \ref{sec_more_constraints}, Def. \ref{def_sub_theme} \\ 
			$G_T$ & $\triangleq$ & $(\bigcup_{t \in T}Stmt[G_t], \bigcup_{t \in T}Rel[G_t])$ & Sec. 
			\ref{sec_more_constraints}\\
			$G_{t, trel}$ & $\triangleq$ & sub-theme 
			finite argumentation graph for $trel$ which is either `$attack$' 
			or `$support$' (a sub-graph 
			of $G_t$): $Stmt[G_{t, trel}] = Stmt[G_t]$ and 
			$Rel[G_{t, trel}] = \{(s', s) \in Rel[G_t] \mid trel \in \Pi_t((s', s))\}$
			& Sec. \ref{sec_computational_complexities}, Def. \ref{def_sub_theme_trel}\\ 
%			$\neg x$ & $\triangleq$ & (for lattices) a complement of $x$ in a lattice containing $x$ & Sec. \ref{sec_technical_preliminaries}\\
%			$\neg x$ & $\triangleq$ & (for classical logic) true iff the proposition $x$ is false & 
%			Appendix: proofs, Proof of Lem. \ref{lem_np_problems}\\
%			$x \wedge y$  & $\triangleq$ & (for partially ordered sets) an infimum of $\{x, y\}$ where $x$ and $y$ 
%			are some element of a partially ordered set & Sec. \ref{sec_technical_preliminaries}\\
%			$x \wedge y$ & $\triangleq$ & (for classical logic) true iff two propositions $x$ and $y$ are both 
%			true & Appendix: proofs, Proof of Lem. \ref{lem_np_problems} \\
%			$x \vee y$ & $\triangleq$ & (for partially ordered sets) a supremum of $\{x, y\}$ where $x$ and $y$ 
%			are some element of a partially ordered set & Sec. \ref{sec_technical_preliminaries}\\
%			$x \vee y$ & $\triangleq$ & (for classical logic) true iff at least one of the propositions $x$ and $y$ is 
%			true & Appendix: proofs, Proof of Lem. \ref{lem_np_problems} \\
			$\Pi_t$ & $\triangleq$ & sub-theme typing function 
			with $t \in \pmb{Themes}$ 
			satisfying: for every $x \in Stmt[G_t] \cup 
			Rel[G_t]$, $\Pi_t(x) = \Pi(x) \cap (\{t\} \cup 
			\{`attack$'$, `support$'$\})$ and for 
			every $x \not\in Stmt[G] \cup Rel[G_t]$, 
			$\Pi_t(x) = \emptyset$ 
			& Sec. \ref{sec_more_constraints}, Def. \ref{def_sub_theme}\\  
			$\Pi_T$ & $\triangleq$ & is such that, for every $x \in Stmt[G_T] \cup Rel[G_T]$, 
			$\Pi_T(x) = \Pi(x) \cap (T \cup \{`attack$'$, `support$'$\})$, and for 
			every $x \not\in Stmt[G_T] cup Rel[G_T]$, $\Pi_T(x) = \emptyset$ & Sec. \ref{sec_more_constraints}\\
			$\Pi_{t, trel}$ & $\triangleq$ & sub-theme typing function for $trel$ which is either `$attack$' 
			or `$support$' satisfying the following: for every 
			$x \in Stmt[G_{t, trel}] \cup Rel[G_{t, trel}]$,  
			$\Pi_{t, trel}(x) = \Pi_t(x) \cap \{t, trel\}$ 
			and for every $x \not\in Stmt[G_{t,trel}] \cup 
			Rel[G_{t,trel}]$, $\Pi_{t,trel}(x) = \emptyset$ 
			& Sec. \ref{sec_computational_complexities}, Def. \ref{def_sub_theme_trel}\\
			%\multicolumn{3}{c}{\underline{Structures and Tuples}} \\ 
						%$(A, R)$ & $\triangleq$ & finite argumentation 
		%	graph (a member of $\mathcal{G}$) & Sec. \ref{sec_technical_preliminaries}\\
			%$i$ & $\triangleq$ & index value for store locations\\
%${T}_{c}$ & $\triangleq$ & A very long description of this specific variable and is needed in the research and looks good when wrapped and aligned to the left.\\
%$TC$ & $\triangleq$ & Total overall cost(\$)\\  
%\multicolumn{3}{c}{}\\
%\midrule 
\bottomrule 
	\label{tbl_symbols} 
\end{longtable} 
}
\end{center}
%\end{longtabu}
%	\end{table} 
\pagebreak 
\section*{{\hyperref[text_table_of_constraints]{Appendix: table of constraints}}} \label{app_table_of_constraints}

\begin{center}% used the environment to augment the vertical space
		{\scriptsize 
		\begin{longtable}{r c p{10cm} p{2.4cm}}
			% between the caption and the table
		%\begin{tabular}{X[r] X[c] X[p{10cm}] X[l] }
\toprule  
			Constraint & & Description & 1st appears in \\ 
			\midrule 
						\multicolumn{3}{c}{\underline{Constraints on the theme aspect argumentation model $\taam$}}\\   
			\pmb{aass} & $\triangleq$ & \pmb{a}ttack as \pmb{a}ttack, \pmb{s}upport as \pmb{s}upport constraint 
			on the theme aspect argumentation model $\taam$, ensuring the following 
			for every $\emptyset \subset T \subseteq 
				\pmb{Themes}$ 
	and every  $(s', s) \in Rel[G]$: 
	$\{`attack$'$\} \cup T \subseteq \Pi((s', s))$ materially implies 
	%and 
	%$\mathcal{I}(T, s) \not= \emptyset$ 
%	and $\mathcal{I}(T, s') \not= \emptyset$ 
%	materially imply  
%	$\emptyset \not= 
	$\emptyset \not= \bigwedge \mathcal{I}(T, s') \not=  
	\bigwedge \mathcal{I}(T, s) \not= \emptyset$;
	%\not= \emptyset$; 
	and $\{`support$'$\} \cup T \subseteq \Pi((s', s))$ %and  
%	$\mathcal{I}(T, s) \not= \emptyset$ 
	materially implies 
	$\bigwedge \mathcal{I}(T, s') \in {\uparrow\! \{\bigwedge \mathcal{I}(T, s)\}} 
	\cup {\downarrow\! \{\bigwedge \mathcal{I}(T, s)\}}$ 
	(where necessarily ${\uparrow\! \{\bigwedge \mathcal{I}(T, s)\}} 
	\cup {\downarrow\! \{\bigwedge \mathcal{I}(T, s)\}} 
	\not= \emptyset$)
			 & Sec. \ref{sec_core_constraints_of_theme_aspect_argumentation_model}, Def. \ref{def_attack_as_attack_support_as_support}\\ 
			\pmb{i} & $\triangleq$ & \pmb{i}nclusion constraint on the theme aspect argumentation model 
			$\taam$, ensuring the following for every $s \in Stmt[G]$ and 
	every $T \subseteq \pmb{Themes}$: $\mathcal{I}(T, s) \subseteq \mathcal{I}(T, \omega)$ & Sec. \ref{sec_core_constraints_of_theme_aspect_argumentation_model}, Def. \ref{def_inclusion_constraint}\\ 
			\pmb{vi} & $\triangleq$ & \pmb{v}acuous \pmb{i}nterpretation constraint  
			on the theme aspect argumentation model $\taam$, ensuring 
	the following for every $s \in Stmt[G]$: $\mathcal{I}(\emptyset, s) = \emptyset$ & Sec. \ref{sec_core_constraints_of_theme_aspect_argumentation_model}, Def. \ref{def_vacuous_interpretation_constraint}\\
			\pmb{bat} & $\triangleq$ & 
			\pmb{b}oolean \pmb{a}lgebra for \pmb{t}hemes constraint on the theme aspect argumentation model 
			$\taam$, 
			ensuring 
	the following  for every $T \subseteq \pmb{Themes}$: 
	$\mathcal{I}(T, \omega) \not= \emptyset$ 
	materially implies 
	that $(\mathcal{I}(T, \omega), \pmb{1}, \pmb{0}, \neg, 
	\wedge, \vee)$ is a 
	sub-complete Boolean algebra of $\mathcal{D}$ & Sec. \ref{sec_core_constraints_of_theme_aspect_argumentation_model}, 
	Def. \ref{def_sub_boolean_algebra_constraint}\\
			\pmb{pr} & $\triangleq$ & 
			\pmb{p}roper \pmb{r}ange constraint on the theme aspect argumentation model $\taam$, 
			ensuring 
	the following for every $s \in Stmt[G]$ and every $T \subseteq \pmb{Themes}$. 
		 Case $s \in OStmt[G]$:
			$\mathcal{I}(T, s) \subseteq \mathcal{I}(T \cap \Pi(s), s)$.
						Case $s \in PStmt[G]$ and $s$ is in the form 
			$t.a$ ($t \in \pmb{Themes}, a \in \pmb{A}^{ord}$): 
					{\small $\mathcal{I}(T, s) \subseteq 
					\mathcal{I}(T \cap (\Pi(s) 
					\cup \{t\}),
					a)$}. 
					Case $s \in PStmt[G]$ and $s$ is in 
			the form $t.\mathcal{C}$ ($t \in \pmb{Themes}$):  
					$\mathcal{I}(T, s) \subseteq 
					\mathcal{I}(T \cap (\Pi(s) 
					\cup \{t\}), t.\mathcal{C})$. 
%				\item Case $s \in PStmt[G]$ and $s$ is in the form $t_1.pStmt$:  
%					$\mathcal{I}(T, s) = \emptyset$ if $T \cap \Pi(s) = \emptyset$ and 
%					($t_1 \not\in T$ or $\mathcal{I}(T, pStmt) = \emptyset$).  
					& Sec. \ref{sec_core_constraints_of_theme_aspect_argumentation_model}, Def. \ref{def_proper_range_constraint}
			\\
			\pmb{mat} & $\triangleq$ & \pmb{m}onotone \pmb{a}spects of \pmb{t}hemes constraint on the 
			theme aspect argumentation model $\taam$, ensuring 
	the following for every $T_1 \subseteq T_2 \subseteq \pmb{Themes}$ 
	and every $x \in \mathcal{I}(T_1, \omega)$: $x \in \mathcal{I}(T_2, \omega)$ & Sec. \ref{sec_core_constraints_of_theme_aspect_argumentation_model}, Def. \ref{def_monotone_aspects_themes_constraint}\\
			\pmb{manss} & $\triangleq$ & \pmb{m}onotone \pmb{a}spects of \pmb{n}on-\pmb{s}ummary
	 \pmb{s}tatements constraint on the theme aspect argumentation model $\taam$, ensuring 
	the following for every $T_1 \subseteq T_2 \subseteq \pmb{Themes}$, 
	every $s \in Stmt[G]$ that does not point to \pmb{s}ummary and 
	every $x \in \mathcal{I}(T_1, s)$: $x \in \mathcal{I}(T_2, s)$ & Sec. \ref{sec_core_constraints_of_theme_aspect_argumentation_model}, Def. \ref{def_monotone_aspects_constraint_non_summary_statements}\\ 
			\pmb{ss} & $\triangleq$ & \pmb{s}ubstantial \pmb{s}tatement constraint on the theme aspect argumentation 
			model $\taam$, ensuring 
	the following for any $s \in Stmt[G]$ 
	and any $T \subseteq \Pi(s)$: $\mathcal{I}(T, s) \not= \emptyset$ 
	materially implies both $\pmb{0} \not= \bigwedge \mathcal{I}(T, s)$ and $\pmb{1} \not= \bigwedge \mathcal{I}(T, s)$ & 
	Sec. \ref{sec_core_constraints_of_theme_aspect_argumentation_model}, Def. \ref{def_substantial_statement_constraint}\\
			\pmb{esr} & $\triangleq$ & \pmb{e}xact \pmb{s}ummary \pmb{r}eference constraint on 
			the theme asepct argumentation model $\taam$, ensuring the following  
			for every $T_1, T_2 \subseteq \pmb{Themes}$, every $t \in \pmb{Themes}$ and 
			every $t.\mathcal{C} \in PStmt[G]$: $\mathcal{I}(T_1, t.\mathcal{C}) \cap (\mathcal{I}(T_1, \omega) \cap \mathcal{I}(T_2, \omega)) = \mathcal{I}(T_2, t.\mathcal{C}) \cap (\mathcal{I}(T_1, \omega) \cap \mathcal{I}(T_2, \omega))$ & Sec. \ref{sec_more_constraints}, Def. \ref{def_exact_common_region_constraint}\\
			\pmb{ensr} & $\triangleq$ & 
	\pmb{e}xact \pmb{n}on-\pmb{s}ummary \pmb{r}eference constraint on the theme aspect argumentation model $\taam$, 
	ensuring the following for every $T_1, T_2 \subseteq \pmb{Themes}$, every $t \in \pmb{Themes}$, 
			every $a \in \pmb{A}^{ord}$ and every $t.a \in PStmt[G]$: $\mathcal{I}(T_1, t.a) \cap (\mathcal{I}(T_1, \omega) \cap \mathcal{I}(T_2, \omega)) = \mathcal{I}(T_2, t.a) \cap (\mathcal{I}(T_1, \omega) \cap \mathcal{I}(T_2, \omega))$ & Sec. \ref{sec_more_constraints}, Def. \ref{def_exact_common_region_constraint}\\
			\pmb{eos} & $\triangleq$ & 
	\pmb{e}xact \pmb{o}rdinary \pmb{s}tatement constraint on the theme aspect argumentation model $\taam$, ensuring the following 
	for every $T_1, T_2 \subseteq \pmb{Themes}$ and 
		every $a \in OStmt[G]$: $\mathcal{I}(T_1, a) \cap (\mathcal{I}(T_1, \omega) \cap \mathcal{I}(T_2, \omega)) = \mathcal{I}(T_2, a) \cap (\mathcal{I}(T_1, \omega) \cap \mathcal{I}(T_2, \omega))$ & Sec. \ref{sec_more_constraints}, Def. \ref{def_exact_common_region_constraint}\\
			\pmb{das} & $\triangleq$ & 
			\pmb{d}istinct
	\pmb{a}ttack \pmb{s}upport constraint on the theme aspect argumentation model $\taam$, 
	ensuring both of the following for every $(s', s) \in Rel[G]$ 
	and every $\emptyset \subset T \subseteq \pmb{Themes}$: 
		(1) $T \cup \{`attack$'$\} \subseteq \Pi((s', s))$ 
			materially implies $\bigwedge 
			\mathcal{I}(T, s')    
			\not\in  {\downarrow \{\bigwedge \mathcal{I}(T, s)\}} \cup 
			{\uparrow \{\bigwedge \mathcal{I}(T, s)\}}$, and 
%			\cup ({\uparrow 
%			\{\neg \bigwedge \mathcal{I}(T, s)\}} 
%			\backslash \{\neg \bigwedge \mathcal{I}(T, s)\})$. 
		(2) $T \cup \{`support$'$\} \subseteq 
			\Pi((s', s))$ materially implies 
			$\bigwedge \mathcal{I}(T, s') 
			= \bigwedge \mathcal{I}(T, s)$ & Sec. \ref{sec_more_constraints}, Def. \ref{def_distinct_attack_support_constraint}\\
			\pmb{nwci} & $\triangleq$ & \pmb{n}o \pmb{w}eakened \pmb{c}ontradictions / 
	\pmb{i}ncomparables constraint on the theme aspect argumentation model $\taam$, ensuring 
	the following for every $(s', s) \in Rel[G]$ and every $\emptyset \subset 
	T \subseteq \pmb{Themes}$: if %\linebreak\linebreak
	$\{`attack$'$\} \cup T \subseteq \Pi((s', s))$, then: 
		(1) $\bigwedge \mathcal{I}(T, s') \not\in ({\uparrow \{\neg \bigwedge \mathcal{I}(T, s)\})} 
			\backslash \{\neg \bigwedge \mathcal{I}(T, s)\}$, and (2) 
	$\bigwedge \mathcal{I}(T, s') \in {\downarrow \{\bigwedge \mathcal{I}(T, s)\}} \cup {\uparrow \{\bigwedge 
	\mathcal{I}(T, s)\}} \cup {\downarrow \{\neg \bigwedge \mathcal{I}(T, s)\}} \cup
	{\uparrow \{\neg \bigwedge \mathcal{I}(T, s)\}}$ & Sec. \ref{sec_more_constraints}, Def. \ref{def_no_weakened_contradiction_incomparables_constraint}\\ 
	\pmb{faD} & $\triangleq$ &  \pmb{f}resh \pmb{a}spect \pmb{D}epS constraint on the theme aspect argumentation 
	model $\taam$, ensuring the following 
	for any $s \in Stmt[G]$, any $t \in \Pi(s)$ and  
	any depth-statements-set $S$ of $s$ with respect to $t$ (for `{\it attack}' or `{\it support}'):   
	$S$ does not contain redundancy with respect to $t$ & Sec. \ref{sec_more_constraints}, Def. \ref{def_fad}\\
	\pmb{faW} & $\triangleq$ & \pmb{f}resh \pmb{a}spect \pmb{w}idS constraint on the theme aspect argumentation 
	model $\taam$, ensuring 
	the following for any $s \in Stmt[G]$, 
	any $t \in \Pi(s)$ and any width-statements-set $S$ of $s$ 
	with respect to $t$ (for `$attack$' or `$support$'): $S$ does not 
	contain redundancy with respect to $t$ & Sec. \ref{sec_more_constraints}, Def. \ref{def_faw}\\
	\midrule 
			\pmb{Core} & $\triangleq$ & \pmb{aass}, \pmb{i}, \pmb{vi}, \pmb{bat}, \pmb{pr}, \pmb{mat}, \pmb{manss}, and \pmb{ss} & Sec. \ref{sec_core_constraints_of_theme_aspect_argumentation_model}\\ 
			\pmb{E} & $\triangleq$ & \pmb{esr}, \pmb{ensr} and \pmb{eos} & Sec. \ref{sec_more_constraints}\\
			\pmb{F} & $\triangleq$ & \pmb{faW} and \pmb{faD} & Sec. \ref{sec_more_constraints}\\ 
			\pmb{All}$^{- \text{\pmb{F}}}$ & $\triangleq$ & 
			\pmb{Core}, \pmb{E}, \pmb{das} and \pmb{nwci} & Sec. \ref{sec_computational_complexities}\\
			\pmb{All} & $\triangleq$ & \pmb{All}$^{- \text{\pmb{F}}}$ and \pmb{F} & Sec. \ref{sec_computational_complexities}\\\pagebreak
			\midrule 
	\multicolumn{3}{c}{\underline{Constraints on the typed argumentation graph $\modell$}}\\   
\pmb{tr} & $\triangleq$ & \pmb{t}heme \pmb{r}elevance constraint on the typed argumentation graph 
			$\modell$, ensuring theme relevance 
			of direct-edge-connected statements in $Stmt[G]$ 
			(for every $(s_1, s_2) \in Rel[G]$, 
	there exists some $t \in \pmb{Themes}$ such that 
	$t \in \Pi((s_1, s_2)) \cap \Pi(s_1) \cap \Pi(s_2)$)  
			& Sec. \ref{sec_core_constraints_of_theme_aspect_argumentation_model}, Def. \ref{def_theme_relevance_constraint}\\
			\pmb{nnp} & $\triangleq$ & \pmb{n}o \pmb{n}ull \pmb{p}ointer constraint on the typed argumentation 
			graph $\modell$, ensuring that every pointer statement $s \in PStmt[G]$ refers to something 
			in $\modell$ 
			(for 
	every $s \in PStmt[G]$, 
		if $s \equiv t.\mathcal{C}$ for some $t 
			\in \pmb{Themes}$, then 
			there is some $s' \in Stmt[G]$ such that 
			$t \in \Pi(s')$, and 
		if $s \equiv t.a$ for some $t 
			\in \pmb{Themes}$ and some  
			$a \in \pmb{A}^{ord}$, then 
			$a \in Stmt[G]$ and 
			$t \in \Pi(a)$) 
			& Sec. \ref{sec_core_constraints_of_theme_aspect_argumentation_model}, Def. \ref{def_no_null_pointer_constraint}\\ 
			\pmb{nsa} & $\triangleq$ & \pmb{n}o \pmb{s}elf-\pmb{a}ttack constraint on the typed argumentation graph,  
			ensuring no `$attack$'-typed relation from a statement to itself in $G$ (there is no $s \in Stmt[G]$ such that 
	$(s, s) \in Rel[G]$ and $`attack$'$ \in \Pi((s, s))$ 
	both hold) 
			& Sec. \ref{sec_core_constraints_of_theme_aspect_argumentation_model}, Def. \ref{def_no_self_attack_constraint}\\
			\pmb{kos} & $\triangleq$ & \pmb{k}nown \pmb{o}rdinary \pmb{s}tatement constraint on the typed 
			argumentation graph $\modell$, ensuring that when 
			a reference to an ordinary statement is made from within a set of themes, 
			then the ordinary statement is within the set of themes (for any $t \in \pmb{Themes}$, any $a \in OStmt[G]$ and any $s \in Stmt[G]$,  
	 $s \equiv t.a$ materially implies  
	 $\Pi(t.a) \subseteq \Pi(a)$) & Sec. \ref{sec_core_constraints_of_theme_aspect_argumentation_model}, Def. \ref{def_known_ordinary_statement_constraint}\\  
			\pmb{nss} & $\triangleq$ & \pmb{n}o \pmb{s}imultaneous neighbouring \pmb{s}upports constraint on 
			the typed argumentation graph $\modell$, ensuring (1) that no statement-to-statement relation may be typed with both 
			`$attack$' and `$support$' for any theme typing the relation and (2) that  
			for any $(s_1, s_2) \in Rel[G]$ if $(s_1, s_2)$ is typed with `$attack$', 
			then no statement $s \in Stmt[G]$ `$support$'s 
			both $s_1$ and $s_2$ for a shared theme or can be `$support$'ed by both $s_1$ and $s_2$ 
			for a shared theme (for every $(s_1, s_2) \in Rel[G]$ such that $`attack$'$ \in \Pi((s_1, s_2))$, it holds that (1) 
		 $`support$'$ \not\in \Pi((s_1, s_2))$, 
		 (2) 
		for every $s \in Stmt[G]$ and every $i, j \in \{1,2\}$ with $i \not= j$, $(s, s_i) \in Rel[G]$ 
			and $`support$'$ \in \Pi((s, s_i))$ materially imply either    
			$(s, s_j) \not\in Rel[G]$, or $`support$'$ \not\in \Pi((s, s_j))$, 
			or there is no $t \in \pmb{Themes}$ such that $t \in \Pi((s, s_i)) \cap \Pi((s, s_j))$, and (3) 
		for every $s \in Stmt[G]$ and every $i, j \in \{1,2\}$ with $i \not= j$, $(s_i, s) \in Rel[G]$ 
			and $`support$'$ \in \Pi((s_i, s))$ materially imply either   
			$(s_j, s) \not\in Rel[G]$, or $`support$'$ \not\in \Pi((s_j, s))$, 
			or there is no $t \in \pmb{Themes}$ such that $t \in \Pi((s_i, s)) \cap \Pi((s_j, s))$) 
 & Sec. \ref{sec_more_constraints}, Def. \ref{def_no_simultaneous_neighbouring_supports_constraint}\\
\bottomrule 
	\label{tbl_symbols} 
\end{longtable} 
}
\end{center}

\pagebreak  
\section*{{\hyperref[text_proofs]{Appendix: proofs}}}  \label{app_proofs}
\noindent \textbf{Proof of Theorem \ref{thm_consequence_pr_vi_aass}.} 
\phantomsection \label{proof_consequence_pr_vi_aass}
For 1., suppose $\modell$ is such that: $Stmt[G] = \{a_1, a_2\}$; 
$Rel[G] = \{(a_1, a_2)\}$; $\Pi(a_1) = \Pi(a_2) = \{t_1\}$; 
and that $\Pi((a_1, a_2)) = \{t_2, `attack$'$\}$. It does not satisfy 
\pmb{tr}. 
For every $\mathcal{D}, \mathcal{I}$, if 
$\langle \modell, \mathcal{D}, \mathcal{I} \rangle$ 
satisfies \pmb{pr} and \pmb{vi}, then 
by Proposition \ref{prop_consequence_pr_vi},  
$\mathcal{I}(\{t_2\}, a_1)$ is $\emptyset$. 
If $\langle \modell, \mathcal{D}, \mathcal{I} \rangle$ should 
additionally satisfy \pmb{aass}, then by the definition of \pmb{aass}, 
$\mathcal{I}(\{t_2\}, a_1)$ would not be $\emptyset$, a contradiction. 

For 2., suppose $\modell$ is such that: 
$Stmt[G] = \{t_2.a_1, a_2\}$; 
$Rel[G] = \{(t_2.a_1, a_2)\}$; $\Pi(t_2.a_1) = \Pi(a_2) =  
\{t_1\}$; and $\Pi((t_2.a_1, a_2)) = \{t_1, `attack$'$\}$. 
For every $\mathcal{D}, \mathcal{I}$, if 
$\langle \modell, \mathcal{D}, \mathcal{I} \rangle$ 
satisfies \pmb{pr} and \pmb{vi}, then:  
by the second condition of \pmb{pr}, $\mathcal{I}(\{t_1\}, t_2.a_1) \subseteq \mathcal{I}(\{t_1\}, a_1)$; 
by the first condition of \pmb{pr}, $\mathcal{I}(\{t_1\}, a_1) \subseteq \mathcal{I}(\emptyset, a_1)$; 
together, $\mathcal{I}(\{t_1\}, t_2.a_1) \subseteq \mathcal{I}(\emptyset, a_1)$. 
By \pmb{vi}, $\mathcal{I}(\emptyset, a_1) = \emptyset$, thus $\mathcal{I}(\{t_1\}, t_2.a_1) = \emptyset$, too.  
If $\langle \modell, \mathcal{D}, \mathcal{I} \rangle$ should additionally satisfy \pmb{aass}, 
then by the definition of \pmb{aass}, $\mathcal{I}(\{t_1\}, t_2.a_1)$ would not be $\emptyset$, a contradiction. 

An immediate corollary of  \mbox{Theorem \ref{thm_core}} is 3.. As such, we defer the proof 
till \mbox{Theorem \ref{thm_core}}.  \hfill$\Box$ \\ 

\noindent \textbf{Proof of Proposition \ref{prop_consequence_pr_manss}.} 
\phantomsection \label{proof_consequence_pr_manss}
When $T' \cap \Pi(a) = \emptyset$, $\mathcal{I}(T \cup T', a) 
\subseteq \mathcal{I}((T \cup T') \cap \Pi(a), a) 
\subseteq \mathcal{I}((T \cap \Pi(a)) \cup (T'\cap \Pi(a)), a) 
\subseteq \mathcal{I}(T \cap \Pi(a), a)$ via \pmb{pr}. 
Both $\mathcal{I}(T, a) \subseteq \mathcal{I}(T \cup T', a)$ and 
$\mathcal{I}(T \cap \Pi(a), a) \subseteq \mathcal{I}(T, a)$ by \pmb{manss}. 
Hence, $\mathcal{I}(T, a) \subseteq \mathcal{I}(T \cup T', a) \subseteq \mathcal{I}(T, a)$, as required.      
%When $T' \cap \Pi(t_1.a_1) = \emptyset$ and $t_1 \in \Pi(t_1.a_1)$, we have \linebreak $\mathcal{I}(T \cup T', t_1.a_1) 
%\subseteq \mathcal{I}((T \cup T') \cap (\Pi(t_1.a_1) \cup \{t_1\}), a_1) \subseteq \linebreak
%\mathcal{I}((T \cup T') \cap \Pi(t_1.a_1), a_1) \subseteq \mathcal{I}(T \cap \Pi(t_1.a_1), a_1)$ via \pmb{pr}. 
%With \pmb{manss}, $\mathcal{I}(T, t_1.a_1) \subseteq \mathcal{I}(T \cup T', t_1.a_1) \subseteq \mathcal{I}(T, a_1)$. 
 \hfill$\Box$\\

\noindent \textbf{Proof of Lemma \ref{lem_consequence_pr_manss_2}.} 
\phantomsection \label{proof_consequence_pr_manss_2}
By cases. 
\begin{itemize} 
	\item Case $t \in \Pi(t.a)$: $\mathcal{I}(T, t.a) \subseteq \mathcal{I}(T \cap (\Pi(t.a) \cup \{t\}), a) 
		\subseteq \mathcal{I}(T \cap \Pi(t.a), a)$ via \pmb{pr}. Since $\mathcal{I}(T, a) = \emptyset$ by assumption, 
		and since $\taam$ satisfies \pmb{manss}, 
		 it must hold that $\mathcal{I}(T', a) = \emptyset$ for any $T' \subseteq T$. 
		 Thus, necessarily ${\mathcal{I}(T \cap \Pi(t.a), a)} = \emptyset$. Then necessarily 
		 $\mathcal{I}(T, t.a) \subseteq \emptyset$.  
	 \item Case $t \not\in \Pi(t.a)$ and $t \not\in T$:  \\
		 $\mathcal{I}(T, t.a) \subseteq 
		 \mathcal{I}(T \cap (\Pi(t.a) \cup \{t\}), a) \subseteq  
		 \mathcal{I}(T \cap \Pi(t.a), a)$ via \pmb{pr}. Same with the first case. 
	 \item Case $t \not\in \Pi(t.a)$ and $t \in T$: $\mathcal{I}(T, t.a) \subseteq 
		 \mathcal{I}(T \cap (\Pi(t.a) \cup \{t\}), a) \subseteq 
		 \mathcal{I}(\{t\} \cup (T \cap \Pi(t.a)), a)$ via \pmb{pr}. 
		 Since $\taam$ satisfies \pmb{manss},     
		 $\mathcal{I}(\{t\} \cup (T \cap \Pi(t.a)), a) \subseteq 
		 \mathcal{I}(\{t\} \cup T, a) \subseteq \mathcal{I}(T, a)$. However, $\mathcal{I}(T, a) = \emptyset$ by asumption. 
		 Thus, necessarily $\mathcal{I}(T, t.a) \subseteq \emptyset$. \hfill$\Box$ \\
\end{itemize} 

\noindent \textbf{Proof of Proposition \ref{prop_consequence_pr_vi_manss}.}
\phantomsection \label{proof_consequence_pr_vi_manss}
By \pmb{pr}, $\mathcal{I}(T, a) \subseteq \mathcal{I}(T \cap \Pi(a), a)$. 
By assumption, $\mathcal{I}(T \cap \Pi(a), a) \subseteq \mathcal{I}(\emptyset, a)$. 
By \pmb{vi}, $\mathcal{I}(\emptyset, a) \subseteq \emptyset$. Thus, $\mathcal{I}(T, a) \subseteq \emptyset$. 
Lemma \ref{lem_consequence_pr_manss_2}. \hfill$\Box$ \\

\noindent \textbf{Proof of Proposition \ref{prop_consequence_pr_vi_manss_aass}.}
\phantomsection \label{proof_consequence_pr_vi_manss_aass}
Suppose  $\modell$ is such that $Stmt[G] = \{t_1.a_1, a_2\}$, $Rel[G] = \{(t_1.a_1, a_2)\}$, 
$\Pi(t_1.a_1) = \Pi(a_2) = \{t_2\}$, and $\Pi((t_1.a_1, a_2)) = \{`attack$'$, t_2\}$. 
Since $\Pi(a_1) \cap \{t_2\} = \emptyset$, by Proposition \ref{prop_consequence_pr_vi_manss}, 
for any $\mathcal{D}, \mathcal{I}$, if $\langle \modell, \mathcal{D}, \mathcal{I} \rangle$ satisfies 
\pmb{pr}, \pmb{vi} and \pmb{manss}, then $\mathcal{I}(\{t_2\}, t_1.a_1) = \emptyset$. \pmb{aass} expects 
$\mathcal{I}(\{t_2\}, t_1.a_1)$ to be a non-empty set, a contradiction. \hfill$\Box$ \\

\noindent \textbf{Proof of Theorem \ref{thm_existence}.} 
\phantomsection \label{proof_existence}
Assume a typed argumentation graph with one ordinary statement  
and no relations, {\it e.g.} $\modell$  with 
$Stmt[G] = \{a_1\}$ and $\Pi(a_1) = \{t_1\}$. 
Let $\mathcal{D}$ be $(\{x, \neg x, \pmb{1}, \pmb{0}\}, \pmb{1}, \pmb{0}, \neg, \wedge, \vee)$ 
and let $\mathcal{I}$ be such that it satisfies the following for every $T \subseteq \pmb{Themes}$. 
\begin{itemize} 
	%\item $\mathcal{I}(\emptyset, a_1) = \emptyset$. 
	\item $\mathcal{I}(\{t_1\}, a_1) = \{x\}$. 
	\item $\mathcal{I}(T, a_1) = \mathcal{I}(\{t_1\}, a_1)$ if $t_1 \in T$.   
	\item $\mathcal{I}(T, a_1) = \emptyset$ if $t_1 \not\in T$. 
%	\item $\mathcal{I}(\emptyset, \omega) = \emptyset$. 
	\item $\mathcal{I}(\{t_1\}, \omega) = \{x, \neg x, \pmb{1}, \pmb{0}\}$.   
	\item $\mathcal{I}(T, \omega) = \mathcal{I}(\{t_1\}, \omega)$ if $t_1 \in T$. 
	\item $\mathcal{I}(T, \omega) = \emptyset$ if $t_1 \not\in T$. 
\end{itemize}
Then, $\taam$ vacuously satisfies \pmb{aass}, as there is no `$attack$'/`$support$' relation. By the definition 
of $\mathcal{I}$, it satisfies \pmb{i}, \pmb{vi}, \pmb{bat}, \pmb{pr}, \pmb{mat}, \pmb{manss}, and \pmb{ss}. \hfill$\Box$ \\

\noindent\textbf{Proof of Theorem \ref{thm_core}.}
\phantomsection \label{proof_core} 
A few peripheral observations are in order. 
\begin{innerdefinition}[Join irreducibles \citep{Davey02}] \rm 
	Let $\mathcal{L}$ be a lattice. 
	For any $x \in L$ ($L$ is the underlying set of $\mathcal{L}$), $x$ is a {\it join-irreducible} 
	iff the following hold: (1) $x \not= \pmb{0}$ in case $\pmb{0} \in L$; and (2) for any $y, z \in L$, 
	 $x = y \vee z$ materially implies either $x = y$ or $x = z$. 
	Given a lattice $\mathcal{L}$, 
	  $\pmb{JI}(\mathcal{L})$ denotes all the 
	 join-irreducibles of $\mathcal{L}$.  \hfill$\spadesuit$ 
\end{innerdefinition}
	 % and $x_3$ are the atoms of the lattice.  
%They are also all the join-irreducibles. 
As far as a finite Boolean algebra is concerned,  
the set of all atoms (see section \ref{subsec_technical_preliminaries}) 
is that of all join-irreducibles. 
\begin{innerlemma}[Correspondence between join-irreducibles and atoms; from 5.3 in \citep{Davey02}]\label{innerlem_correspondence_join_irreducible_and_atoms}{\ }  
	Let $\mathcal{D}$ be a finite Boolean algebra. Then  
	$\pmb{JI}(\mathcal{D}) = \pmb{Atom}(\mathcal{D})$.  
\end{innerlemma} 
%Also, as per Section, any member of (the underlying set of) a finite Boolean algebra is characterised 
%by a distinct set of atoms. %This is a well-known correspondence between a Boolean algebra  
%and a power-set algebra. 
%\begin{lemma}[Isomorphism and representation; 5.5 in \citep{Davey02}]\label{lem_isomorphism}   
%	Given a finite Boolean algebra $\mathcal{D}$, $\eta_D: x \mapsto \{y \in \pmb{Atom}(\mathcal{D}) \mid y \leq x\}$ is an isomorphism 
%	of $D$ onto $\mathfrak{p}(\pmb{Atom}(\mathcal{D}))$. 
%	%Given some set $X$, a Boolean algebra $(\mathfrak{p}(X), X, \emptyset, - , \cap, \cup)$ 
%%	%where $-$ is a unary operation satisfying $- x = X \backslash x$ for every $x \in \mathfrak{p}(X)$ 
%	%is called a powerset algebra. 
%	%Let $\pmb{Atom}(\mathfrak{p}({X}))$ be 
%	%$\{x \in \mathfrak{p}(X) \mid \emptyset \subset x \text{ and }   
%	%  \forall y \in \mathfrak{p}(X).\emptyset \subset 
%	%  y \subseteq x 
%	%\}$ 
%\end{lemma}  
We obtain the following result  
(Inner-Lemma \ref{innerlem_disjoint_combination})    
through %Inner-Lemma 
Inner-Lemma \ref{innerlem_correspondence_join_irreducible_and_atoms} 
and the isomorphism of a finite Boolean algebra 
to a power set lattice (see section \ref{subsec_technical_preliminaries}). % following concerning composition of almost 
It concerns composition of almost disjoint two finite 
Boolean algebras $\mathcal{D}_1$ 
and $\mathcal{D}_2$ apart from 
\pmb{1} and \pmb{0} into 
a larger finite Boolean algebra $\mathcal{D}_{\star}$. 
%The proof is in 
%{\it \hyperref[app_proofs]{\labelText{Appendix: proofs}{text_proofs}}}. 
%satisfying certain properties. Crucially, 

\begin{innerlemma}[{\hyperref[proof_disjoint_combination]{Disjoint combination}}]\label{innerlem_disjoint_combination} 
	Suppose two finite Boolean algebras $\mathcal{D}_1 \equiv (D_1, \pmb{1}, \pmb{0}, \neg, \wedge, \vee)$ and 
	$\mathcal{D}_2 \equiv (D_2, \pmb{1}, \pmb{0}, \neg, \wedge, \vee)$, 
	and one more finite Boolean algebra $\mathcal{D}_{\star} \equiv (D_{\star}, \pmb{1}, \pmb{0}, \neg, \wedge, \vee)$ 
	such that they satisfy 
	all the following conditions. 
	\begin{itemize} 
		\item $\mathcal{D}_1$ and $\mathcal{D}_2$ are sub-Boolean algebras of $\mathcal{D}_{\star}$. 
			(A sub-Boolean algebra 
	of a Boolean algebra $(D, \pmb{1}, \pmb{0}, \neg, \wedge, \vee)$ 
	is a Boolean algebra $(D', \pmb{1}, \pmb{0}, \neg, \wedge, \vee)$ 
			with $D' \subseteq D$.) 
		\item $|\pmb{Atom}(\mathcal{D}_1)| = 2^i$ and $|\pmb{Atom}(\mathcal{D}_2)| = 2^j$ for some integers $i$ and $j$ strictly greater than 0. 
		\item $|\pmb{Atom}(\mathcal{D}_1)| \times |\pmb{Atom}(\mathcal{D}_2)| = |\pmb{Atom}(\mathcal{D}_{\star})|$. 
		\item for every $x \in D_1$ and every $y \in D_2$, 
			if $x$ and $y$ are comparable in $\mathcal{D}_{\star}$,  
			then either of them is $\pmb{1}$ or $\pmb{0}$. 
	\end{itemize}
	Such $\mathcal{D}_1, \mathcal{D}_2$ and $\mathcal{D}_{\star}$ 
	exist. For every $x \in D_1 \backslash \{\pmb{1}, \pmb{0}\}$ and every $y_1, y_2 \in D_2 \backslash \{\pmb{1},\pmb{0}\}$, 
	$y_1 < y_2$ materially implies ${\pmb{0} < x \wedge y_1 < x \wedge y_2}$  and 
	${x \vee y_1 < x \vee y_2 < \pmb{1}}$ in $D_{\star}$. 
\end{innerlemma} 
\noindent \textbf{Inner-proof.} 
%\phantomsection \label{proof_disjoint_combination}
The existence follows straightforwardly from 
the isomorphism. To discharge the proof obligation of 
${\pmb{0} < x \wedge y_1 < x \wedge y_2}$,  
note by \mbox{Inner-Lemma 
\ref{innerlem_correspondence_join_irreducible_and_atoms}} that   
there exists some $y_3 \in D_2$ such that $y_1$ and $y_3$ are incomparable and that 
$y_2 = y_1 \vee y_3$. Hence, we need to show ${x \wedge y_1 < x \wedge (y_1 \vee y_3) 
= (x \wedge y_1) \vee (x \wedge y_3)}$. But then, it suffices to show that 
$x \wedge y_1$ and $x \wedge y_3$ are incomparable. 

By the isomorphism, there exist 
some $Y_1, Y_3 \subseteq \pmb{Atom}(\mathcal{D}_2)$ 
such that $\eta_{D_2}(y_1) = Y_1$ and that $\eta_{D_2}(y_3) = Y_3$, and there exists 
some $X \subseteq \pmb{Atom}(\mathcal{D}_1)$ such that $\eta_{D_1}(x) = X$.    
%By the given assumptions, we have: 
%\begin{itemize} 
%	\item $\eta_{D_3}(x) = \{z \in \pmb{Atom}(D_3) \mid \exists x \in X\ \exists z_2 \in \pmb{Atom}(D_2).x \wedge z_2\}$.  
%	\item $\eta_{D_3}(y_1) = \{z \in \pmb{Atom}(D_3) \mid \exists y_1 \in Y_1\ \exists z_1 \in \pmb{Atom}(D_1).y_1 \wedge z_2\}$.  
%	\item $\eta_{D_3}(y_3) = \{z \in \pmb{Atom}(D_3) \mid \exists y_3 \in Y_3\ \exists z_1 \in \pmb{Atom}(D_1).y_3 \wedge z_2\}$.  
%\end{itemize} 
%To explain this a little more in detail, 
With no loss of generality, we assume the following. 
\begin{itemize} 
	\item $\{e_1, \ldots, e_{i}, \neg e_1, \ldots, \neg e_i\} \subseteq D_1$. 
	\item $\{g_1, \ldots, g_{j}, \neg g_1, \ldots, \neg g_j\} \subseteq D_2$. 
	\item Each atom in $\mathcal{D}_1$ is in the form: $c_1 \wedge \cdots \wedge c_i$, where $c_k$ ($1 \leq k \leq i$) is either $e_k$ or $\neg e_k$. 
	\item Each atom in $\mathcal{D}_2$ is in the form: $d_1 \wedge \cdots \wedge d_j$, where $d_l$ ($1 \leq l \leq j$) is either $g_l$ or $\neg g_l$. 
	\item Each atom in $\mathcal{D}_3$ is in the form: $c_1 \wedge \cdots \wedge c_i \wedge d_1 \wedge \cdots d_j$, 
		where $c_k$ ($1 \leq k \leq i$) is either $e_k$ or $\neg e_k$, 
		and $d_l$ ($1 \leq l \leq j$) is either $g_l$ or $\neg g_l$.
		
\end{itemize}
Then, it is straightforward to see that, for any $x \in \pmb{Atom}(\mathcal{D}_1)$ and any $y' \in \pmb{Atom}(\mathcal{D}_2)$, there exists 
some $y'' \in \pmb{Atom}(\mathcal{D}_2)$ such that $(x \wedge y') \vee (x \wedge y'') = x$ in $\mathcal{D}_3$. It is also straightforward 
to see that, for any $y \in \pmb{Atom}(\mathcal{D}_2)$ and any $x' \in \pmb{Atom}(\mathcal{D}_1)$, there exists some 
$x'' \in \pmb{Atom}(\mathcal{D}_1)$ such that $(x' \wedge y) \vee (x'' \wedge y) = y$ in $\mathcal{D}_3$. We indeed have: 
\begin{itemize} 
	\item $\eta_{D_3}(x) = \{z \in \pmb{Atom}(\mathcal{D}_3) \mid \exists x \in X\ \exists z_2 \in \pmb{Atom}(\mathcal{D}_2).z = x \wedge z_2\}$.  
	\item $\eta_{D_3}(y_1) = \{z \in \pmb{Atom}(\mathcal{D}_3) \mid \exists y_1 \in Y_1\ \exists z_1 \in \pmb{Atom}(\mathcal{D}_1).z = y_1 \wedge z_2\}$.  
	\item $\eta_{D_3}(y_3) = \{z \in \pmb{Atom}(\mathcal{D}_3) \mid \exists y_3 \in Y_3\ \exists z_1 \in \pmb{Atom}(\mathcal{D}_1).z = y_3 \wedge z_2\}$.  
\end{itemize} 
Now, $\eta_{D_3}(x \wedge y_1) = \eta_{D_3}(x) \cap \eta_{D_3}(y_1) = 
\{z \in \pmb{Atom}(\mathcal{D}_3) \mid \exists x \in X\ \exists y_1 \in Y_1.z = x \wedge y_1\} \not= \emptyset$, 
and similarly, $\eta_{D_3}(x \wedge y_3) = \{z \in \pmb{Atom}(\mathcal{D}_3) \mid \exists x \in X\ \exists y_3 \in Y_3.z = x \wedge y_3\} \not= \emptyset$. 
Hence $\pmb{0} < x \wedge y_1$. 
Since $y_1 \not= y_3$, both $\eta_{D_3}(y_1) \not\subseteq \eta_{D_3}(y_3)$ and $\eta_{D_3}(y_3) \not\subseteq \eta_{D_3}(y_1)$ hold by 
the isomorphism. 
 Hence, both $\eta_{D_3}(x \wedge y_1) \not\subseteq \eta_{D_3}(x \wedge y_3)$ and $\eta_{D_3}(x \wedge y_3) \not\subseteq 
\eta_{D_3}(x \wedge y_1)$ hold. By the isomorphism again, 
$x \wedge y_1$ and $x \wedge y_3$ are incomparable. This concludes 
the proof of $\pmb{0} < x \wedge y_1 < x \wedge y_2$. 
Dually for  $x \vee y_1 < x \vee y_2 < \pmb{1}$. \hfill {\it Q.E.D.} \\

Inner-Lemma \ref{innerlem_disjoint_combination} guarantees the following property: suppose 
$x$ is an element of $\mathcal{D}_1$ different from  
\pmb{1} and \pmb{0} and suppose 
$y$ is an element of $\mathcal{D}_2$ different from 
\pmb{1} and \pmb{0}, then 
if another element $y'$ of $\mathcal{D}_2$ different from 
\pmb{1} and \pmb{0} is comparable to 
$y$ but is not $y$ itself, 
then it is impossible that either $x \wedge y = x \wedge y'$ 
or $x \vee y = x \vee y'$ holds in $\mathcal{D}_{\star}$. 
This concludes our preparation.  \\

Now, for the `if' direction of the main proof obligation, its contrapositive is:
$\modell$ not satisfying \pmb{tr}, \pmb{nsa}, \pmb{nnp} or \pmb{kos} materially implies 
$\taam$'s not satisfying \pmb{Core}. But this follows from \mbox{Proposition \ref{prop_consequence_aass}}, 
the first two sub-results of \mbox{Theorem \ref{thm_consequence_pr_vi_aass}}, and \mbox{Proposition 
\ref{prop_consequence_pr_vi_manss_aass}}.  

For the `only if' direction, by the definition of $\modell$ and $\pmb{Themes}$, $|Stmt[G]|$ and $|\pmb{Themes}|$ are a natural number. 
Thus, with no loss of generality we assume: 
$\pmb{Themes} = \{t_1, \ldots, t_m\}$;  
and $Stmt[G] \subseteq \{a_1, \ldots, a_n, t_1.a_1, \ldots,$ 
$t_1.a_n, 
\ldots,$ 
$t_m.a_1, \ldots,$ $t_m.a_n, t_1.\mathcal{C}, \ldots,$ $t_m.\mathcal{C}\}$, 
for some $m, n \in \mathbb{N}$ and some $a_1, \ldots, a_n \in \pmb{A}^{ord}$.     
It holds that $|Stmt[G]| \leq n + nm + m$. We show one case where 
$Stmt[G] = \{a_1, \ldots, a_n, t_1.a_1, \ldots, t_1.a_n,\linebreak \ldots, t_m.a_1, \ldots, t_m.a_n, t_1.\mathcal{C}, \ldots, t_m.\mathcal{C}\}$; 
the other cases are all similar.  

Assume $\mathcal{D}$ is such that $|\pmb{Atom}(\mathcal{D})| = 2^{m(n + nm + m)}$. Assume 
$\mathcal{D}_1, \ldots, \mathcal{D}_m$ are such that they satisfy the following. 
\begin{itemize} 
	\item for every $1 \leq j \leq m$, $\mathcal{D}_j$ is a sub-Boolean algebra 
		of $\mathcal{D}$ with $|\pmb{Atom}(\mathcal{D}_j)| = 2^{n+nm+m}$. 
	\item for every $1 \leq i \not= j \leq m$, 
		$\mathcal{D}_i$ 
		and $\mathcal{D}_j$ share only $\pmb{1}$ and $\pmb{0}$ 
		as their common elements in $\mathcal{D}$. 
	\item for every permutation $\mathcal{D}'_1, \ldots, \mathcal{D}'_m$ 
		of $\mathcal{D}_1, \ldots, \mathcal{D}_m$ 
		and for every $1 \leq i \leq m -1$, 
		let $\mathcal{D}_{\star'}$ and $\mathcal{D}_{\star''}$ 
		be: $\mathcal{D}_{\star'}$ is a minimal sub-Boolean algebra 
		of $\mathcal{D}$ which contains every $\mathcal{D}'_1, \ldots, 
		\mathcal{D}'_{i}$; $\mathcal{D}_{\star''}$ is a minimal 
		sub-Boolean algebra of $\mathcal{D}$ which contains 
		every $\mathcal{D}'_{i+1}, \ldots, \mathcal{D}'_m$; 
		and $\mathcal{D}_{\star'}$ and $\mathcal{D}_{\star''}$ 
		share only $\pmb{1}$ and $\pmb{0}$ as their common elements. 
\end{itemize}
Such $\mathcal{D}_1, \ldots, \mathcal{D}_m$ exist by Inner-Lemma 
\ref{innerlem_disjoint_combination}. 
Further, assume the following for each $1 \leq i \leq m$. 
\begin{itemize} 
	\item $\mathcal{D}_{i,1}$, \ldots, $\mathcal{D}_{i, n+nm+m}$ are sub-Boolean algebras of 
		$\mathcal{D}_i$ such that $|\pmb{Atom}(\mathcal{D}_{i,j})| = 2^j$ for 
		every $1\leq j \leq n+nm+m$. 
	\item for every $1 \leq j < k \leq n+nm+m$,  $\mathcal{D}_{i,j}$ 
		is a strict sub-Boolean algebra of $\mathcal{D}_{i, k}$.  
\end{itemize}

Now, for every $T \subseteq \pmb{Themes}$, we define $\mathcal{I}$ 
to satisfy the following condition. Let $M$ denote $T \cap \{t_1, \ldots, t_m\}$, then 
$\mathcal{I}(T, \omega)$ is $\emptyset$ if $M = \emptyset$; otherwise,  
for every $t_{l} \in M$, $\mathcal{I}(T, \omega)$ is the underlying set of a minimal
	sub-Boolean algebra 
		of $\mathcal{D}$ containing every $\mathcal{D}_{l}$.  
Then, $\langle \modell, \mathcal{D}, \mathcal{I}\rangle$ satisfies \pmb{bat} and \pmb{mat}. 

Next, for every $1 \leq j \leq m$ and every $1 \leq i \leq n+nm+m$, we define $x_{j,i} \in D_j$ as follows. 
\begin{itemize} 
	\item $x_{j, i} \in D_{j,i}$ and, if $i \not= n+nm+m$, then $x_{j, i} \not\in D_{j, (i+1)}$.  
	\item  $\pmb{0} < x_{j, i} < x_{j, k} < \pmb{1}$ for every $i < k \leq n+nm+m$.  
\end{itemize}
With them, we define $\mathcal{I}$ to further satisfy the following conditions for any $1 \leq j \leq m$. 
		\begin{enumerate} 
			\item for every $1 \leq i \leq n$, $\mathcal{I}(\{t_j\}, a_i)$ is: $\emptyset$ 
				if $t_j \not\in \Pi(a_i)$; $\{x_{j,i}, \ldots, x_{j,(n+nm+m)}\}$, otherwise. 
			\item for every $1 \leq l \leq m$ and every $1 \leq i \leq n$,  
				$\mathcal{I}(\{t_j\}, t_l.a_i)$ is:   
				$\emptyset$ 
				if $\mathcal{I}(\{t_j\}, a_i) = \emptyset$ or $t_j \not\in \{t_l\} \cup \Pi(t_l.a_i)$; 
				$\{x_{j,(n+(l-1)n+i)}, \ldots, x_{j, (n+nm+m)}\}$, otherwise. 
			\item for every $1 \leq l \leq m$, 
				$\mathcal{I}(\{t_j\}, t_l.\mathcal{C})$ is:  
				$\emptyset$ if $t_j \not\in \{t_l\} \cup \Pi(t_l.\mathcal{C})$; 
				$\{x_{j, (n + nm + l)}\}$, otherwise. 
%				 $t_j \not\in \{t_j\} \cup \Pi(t_l.a_i)$   
%				for every $2 \leq i \leq n$,  
%				let $x_{j,i}$ be 
%				such that  
%				$|\mathcal{I}(\{t_j\}, a_i) = $ 
%				$\eta_{D_{j, i}}(x_{j,i})
%				= \{z \in \pmb{Atom}(D_{j,i}) \mid 
%				\exists x \in \eta_{D_{j,(i-1)}}( \ \exists y \in \pmb{Atom}\}$. 
%				$|\mathcal{I}(\{t_j\}, a_i)| = 1$ 
%				and $\eta_{D_{j,i}}(\mathcal{I}(\{t_j\}, a_i)) 
%				\in \pmb{Atom}(w)$.  
		\end{enumerate} 
		By \pmb{kos} and \pmb{nnp}, 2. incurs no loss of generality.  By \pmb{nnp}, 3.  
		incurs no loss of generality.  

		Now, we define $\mathcal{I}$ to satisfy the following condition for every $T \subseteq \pmb{Themes}$ 
		and every $s \in Stmt[G]$. 
		\begin{itemize} 
			\item $\mathcal{I}(T, s) = \bigcup_{1 \leq j \leq m}\{\mathcal{I}(\{t_j\}, s) \subseteq D_{j} \mid 
				t_j \in T\}$. 
		\end{itemize} 
		Then, clearly $\taam$ satisfies \pmb{i}, \pmb{vi}, \pmb{pr}, and \pmb{manss}. Further, 
		by the definition 
		of $\mathcal{D}_j$, $1 \leq j \leq m$, and by \mbox{Inner-Lemma \ref{innerlem_disjoint_combination}},  
		it holds that $\pmb{0} < \bigwedge \mathcal{I}(T, s) < \pmb{1}$ 
		for any $s \in Stmt[G]$ and any $T \subseteq \pmb{Themes}$ such that  
		$\bigwedge \mathcal{I}(T, s) \not= \emptyset$. 
		Thus, $\taam$ satisfies \pmb{ss}. 
		Finally, by the definition of $\mathcal{D}_j$, $1 \leq j \leq m$, and 
		by \pmb{tr}, \pmb{nnp} and \pmb{kos}, 
		for any $(s_1, s_2) \in Rel[G]$ and for any $\emptyset \subset T \subseteq \Pi((s_1, s_2))$, it holds that    
		neither $\mathcal{I}(T, s_1)$ nor $\mathcal{I}(T, s_2)$ is an empty set; and 
		furthermore, for any $t \in \Pi((s_1, s_2))$,  
		$\bigwedge \mathcal{I}(\{t\}, s_1) < \bigwedge \mathcal{I}(\{t\}, s_2)$ 
		or $\bigwedge \mathcal{I}(\{t\}, s_2) < \bigwedge \mathcal{I}(\{t\}, s_1)$ holds,  
		where the inequality is due to \pmb{nsa} forcing $s_1 \not= s_2$. 
		Then by \mbox{Inner-Lemma \ref{innerlem_disjoint_combination}}, $\bigwedge \mathcal{I}(T, s_1) < \bigwedge \mathcal{I}(T, s_2)$  
		or $\bigwedge \mathcal{I}(T, s_2) < \bigwedge \mathcal{I}(T, s_1)$  (for $\emptyset \subset T \subseteq \Pi((s_1, s_2))$).  
		It follows that $\taam$ satisfies \pmb{aass}, and this concludes the proof. \hfill$\Box$ \\

\noindent \textbf{Proof of Theorem \ref{thm_monotone_aspects_of_statements}.}
\phantomsection \label{proof_monotone_aspects_of_statements}
Case $s \in OStmt[G]$: vacuous from the definition of \pmb{Core}.   
Case $t.a \in PStmt[G]$: vacuous 
from the definition of \pmb{Core}. 
Case  $t.\mathcal{C} \in PStmt[G]$: 
\pmb{esr} forces monotonicity of $\mathcal{I}(T, t.\mathcal{C})$ 
for any $T \subseteq \pmb{Themes}$ 
over the increase of $T$ by set-inclusion. \hfill$\Box$\\

\noindent \textbf{Proof of Theorem \ref{thm_distribution}.}
\phantomsection \label{proof_distribution}
To prove $\mathcal{I}(T_1 \cup T_2, s) \subseteq  
\mathcal{I}(T_1, s) \cup \mathcal{I}(T_2, s)$, 
suppose there is some $x \in \mathcal{I}(T_1 \cup T_2, s)$ 
such that $x \not\in \mathcal{I}(T_1, s) \cup \mathcal{I}(T_2, s)$. 
Then $x \not\in \mathcal{I}(T_1, s)$ and $x \not\in \mathcal{I}(T_2, s)$. 
By \pmb{Core}, \pmb{esr}, \pmb{ensr}, and \pmb{eos}, 
$x \not\in \mathcal{I}(T_1, s)$ materially implies 
$x \not\in \mathcal{I}(T_1 \cup T_2, s)$, a contradiction. 
To prove $\mathcal{I}(T_1, s) \cup \mathcal{I}(T_2, s) \subseteq 
\mathcal{I}(T_1 \cup T_2, s)$, suppose there is some  
$x \in \mathcal{I}(T_1, s) \cup \mathcal{I}(T_2, s)$ such that 
$x \not\in \mathcal{I}(T_1 \cup T_2, s)$. 
Then at least either $x \in \mathcal{I}(T_1, s)$ or 
$x \in \mathcal{I}(T_2, s)$ holds. Then, 
by \pmb{Core}, \pmb{esr},  \pmb{ensr},  \pmb{eos}, 
it holds that $x \in \mathcal{I}(T_1 \cup T_2, s)$, a contradiction. 

To prove $\mathcal{I}(T_1 \cap T_2, s) \subseteq \mathcal{I}(T_1, s) \cap 
\mathcal{I}(T_2, s)$, suppose there is some 
$x \in \mathcal{I}(T_1 \cap T_2, s)$ such that 
$x \not\in \mathcal{I}(T_1, s) \cap \mathcal{I}(T_2, s)$. 
We consider every possibility. If $x \in \mathcal{I}(T_1, s)$, 
then $x \not\in \mathcal{I}(T_2, s)$, from which 
follows $x \not\in \mathcal{I}(T_1 \cap T_2, s)$ 
(Cf. \mbox{Theorem \ref{thm_monotone_aspects_of_statements}}), 
a contradiction. If $x \not\in \mathcal{I}(T_1, s)$, 
then similarly it follows $x \not\in \mathcal{I}(T_1 \cap T_2, s)$, 
a contradiction. To prove $\mathcal{I}(T_1, s) \cap \mathcal{I}(T_2, s) 
\subseteq \mathcal{I}(T_1 \cap T_2, s)$, 
suppose there is some $x \in \mathcal{I}(T_1, s) \cap \mathcal{I}(T_2, s)$ 
such that $x \not\in \mathcal{I}(T_1 \cap T_2, s)$. 
From $x \in \mathcal{I}(T_1, s) \cap \mathcal{I}(T_2, s)$ 
follow both $x \in \mathcal{I}(T_1, s)$ and $x \in \mathcal{I}(T_2, s)$. 
From $x \not\in \mathcal{I}(T_1 \cap T_2, s)$ and 
$x \in \mathcal{I}(T_1, s)$ follows 
$x \not\in \mathcal{I}(T_2, s)$, a contradiction.   \hfill$\Box$ \\

\noindent \textbf{Proof of Proposition \ref{prop_consequence_das}.} 
\phantomsection \label{proof_consequence_das}
For the first one, suppose otherwise, then by \pmb{Core} and the second 
condition of \pmb{das}, 
the following holds for some $t \in \Pi((s_1, s_2))$\footnote{This holds for every such $t$, but what this proof 
needs is the existence of one such $t$.}: 
$\emptyset \not= \bigwedge \mathcal{I}(\{t\}, s_1) 
= \bigwedge \mathcal{I}(\{t\}, s_2)$. Therefore,  
$\bigwedge \mathcal{I}(\{t\}, s_1) \in {\downarrow \{\bigwedge \mathcal{I}(\{t\}, s_2)\}} \cup 
			{\uparrow \{\bigwedge \mathcal{I}(\{t\}, s_2)\}}$, 
contradicting the first condition 
	of \pmb{das}. For the second one, suppose otherwise, then  
	there is some $t \in \Pi((s, s_i)) \cap \Pi((s, s_j))$ 
	such that 
	$\emptyset \not= \bigwedge \mathcal{I}(\{t\}, s) = \bigwedge \mathcal{I}(\{t\}, s_i) = \bigwedge \mathcal{I}(\{t\}, s_j)$, 
	contradicting the first condition of \pmb{das}. Similarly for the third one. \hfill$\Box$ \\

\noindent \textbf{Proof of Proposition \ref{prop_pairwise_independence}.}
\phantomsection \label{proof_pairwise_independence_constraints}  
%\pmb{If}: We take the contrapositive. When $\pmb{x} = \pmb{y}$, 
%the satisfaction statuses of $\pmb{x}$ and $\pmb{y}$ 
%coincide. That is, there is no $\taam$ such that 
%the satisfaction statuses of $\pmb{x}$ and $\pmb{y}$ do not coincide. 
%\pmb{Only if}: 
The \pmb{if} condition is obvious. Checking the \pmb{only if} 
direction by hand is tedius and error-prone. However, if the constraints 
are indeed pairwise independent, then for each distinct pair 
$(\pmb{x}, \pmb{y})$, it suffices to 
find one specific theme aspect argumentation model 
satisfying $\pmb{x}$ but not $\pmb{y}$ and another specific 
theme aspect argumentation model satisfying 
$\pmb{y}$ but not $\pmb{x}$. This is what we need to show minimally. 
However, we can of course show more than the minimal evidence 
with the collected data of 118,073 randomly generated 
theme aspect argumentation models (see {\it \hyperref[app_implementation_detail]{Appendix: implementation, gathered data and run-time evaluation}}). Specifically, let us see whether (un)satisfactions of the constraints 
are associated, and if so, how strongly, with \mbox{Cram{\'e}r's $V$} 
\citep{Cramer99} used for measuring the degree of association. 
Let $\pmb{x}$ and $\pmb{y}$ be a constraint, let $N$ be the sample size of 118,073, 
and let $(\pmb{x}, \pmb{y})_{\mathsf{F}.}, (\pmb{x}, \pmb{y})_{\mathsf{T}.}, (\pmb{x}, \pmb{y})_{.\mathsf{F}}, (\pmb{x},\pmb{y})_{.\mathsf{T}}$ be as defined below.   
\begin{multicols}{2} 
\begin{itemize}  
	\item $(\pmb{x}, \pmb{y})_{\mathsf{F}.}$ is the number of times $\pmb{x}$ was not satisfied.  
	\item $(\pmb{x}, \pmb{y})_{\mathsf{T}.}$ is the number of times $\pmb{x}$ was satisfied. 
	\item $(\pmb{x}, \pmb{y})_{.\mathsf{F}}$ is the number of times $\pmb{y}$ was not satisfied.  
	\item $(\pmb{x}, \pmb{y})_{.\mathsf{T}}$ is the number of times $\pmb{y}$ was satisfied. 
\end{itemize}
\end{multicols} 
For any $u_1,u_2 \in \{\mathsf{F}, \mathsf{T}\}$, let $(\pmb{x}, \pmb{y})_{u_1.u_2}$ be the number of times 
$\pmb{x}$ was: not satisfied if $u_1 = \mathsf{F}$; satisfied if $u_1 = \mathsf{T}$ and at the same time 
$\pmb{y}$ was: not satisfied if $u_2 = \mathsf{F}$; satisfied if $u_2 = T$. Then, Cram{\'e}r's $V$, which we denote by 
$\mathcal{V}$, for each pair 
of $\pmb{x}$ and $\pmb{y}$ is: \\

$\mathcal{V} = \sqrt{\frac{\chi^2}{N}}$ where $\chi^2$ is the chi-squared statistic $\Sigma_{u_1,u_2}\frac{((\pmb{x}, \pmb{y})_{u_1,u_2} - \frac{(\pmb{x},\pmb{y})_{u_1.}
(\pmb{x},\pmb{y})_{.u_2}}{N})^2}{\frac{(\pmb{x},\pmb{y})_{u_1.}
(\pmb{x},\pmb{y})_{.u_2}}{N}}$. \\

\begin{figure}[!h]  
	\centering 
   \includegraphics[scale=0.43]{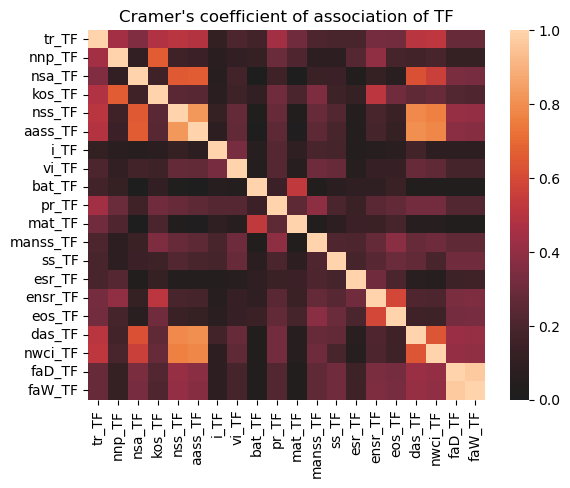}   
	\caption{The heat map of Cram{\'e}r's $V$ of each pair of constraints.}  
	\label{fig_cramersV} 
\end{figure}

\noindent The closer $\mathcal{V}$ is to 1, the more strongly associated  
the (un)satisfaction statuses of $\pmb{x}$ and $\pmb{y}$ 
are. In particular,  
when $\mathcal{V}$ is 1, (un)satisfaction of $\pmb{x}$ completely determines  
the (un)satisfaction status of $\pmb{y}$. \footnote{As the possibilities are $2 \times 2$, 
$\mathcal{V}$ coincides the absolute value of Phi coefficient. For the $2 \times 2$ case, 
$\mathcal{V}$ above 0.3 tends to indicate 
a medium association and $\mathcal{V}$ above 0.5 tends to indicate a large association.} 
Figure \ref{fig_cramersV} shows the heat map of \mbox{Cram{\'e}r's $V$} for every pair of  constraints. 

%It follows straightforwardly that no constraint is equivalent to any others 
%but itself. %See Appendix: proofs for the proof. \\ 

\noindent From Figure \ref{fig_cramersV}, for each 
pair of constraints $\pmb{x}$ and $\pmb{y}$, $\mathcal{V} = 1$ iff 
$\pmb{x} = \pmb{y}$. %\pmb{If}: We take the contrapositive. When $\pmb{x} = \pmb{y}$, 
%the satisfaction statuses of $\pmb{x}$ and $\pmb{y}$ 
%coincide. That is, there is no $\taam$ such that 
%the satisfaction statuses of $\pmb{x}$ and $\pmb{y}$ do not coincide. 
\hfill$\Box$  \\

\noindent \textbf{Proof of Lemma \ref{lem_np_problems}.}
\phantomsection \label{proof_np_problems}
For the first obligation, with no loss of generality, assume $\mathcal{I}(T_i,s_j) = \{x^{(i,j)}_1, 
\ldots, x^{(i,j)}_{k_{(i,j)}}\}$ 
and $\mathcal{I}(T'_i, s'_j) = \{y^{(i,j)}_1, \ldots, 
y^{(i,j)}_{l_{(i,j)}}\}$ 
for every $1 \leq i \leq n$ and every $1 \leq j \leq m$. 
By the conditions on the representation of $\mathcal{D}$ and $\mathcal{I}$, for every $1 \leq i \leq n$ and every 
$1 \leq j \leq m$, $|\mathcal{I}(T_i, s_j)|$ and $|\mathcal{I}(T'_i, s'_j)|$ are both polynomially bounded. 
Now, $\bigwedge_{1 \leq i \leq n}\bigwedge_{1 \leq j \leq m}
(((x^{(i,j)}_1 \wedge \cdots \wedge x^{(i,j)}_{k_{(i,j)}}) 
\wedge \neg (y^{(i,j)}_1 \wedge \cdots \wedge y^{(i,j)}_{l_{(i,j)}})) 
\vee 
((y^{(i,j)}_1 \wedge \cdots \wedge y_{l_{(i,j)}}) 
\wedge \neg (x^{(i,j)}_1 \wedge \cdots \wedge 
x^{(i,j)}_{k_{(i,j)}})))$ is satisfiable iff 
$\bigwedge \mathcal{I}(T_i, s_j) \not= \bigwedge \mathcal{I}(T'_i, s'_j)$ 
for every $1 \leq i \leq n$ and every $1 \leq j \leq m$. 
The satisfiability of this expression is polynomial-time many-one reducible to SAT, and SAT = NP ({\it i.e.} SAT is NP-complete). Hence, 
it is in NP. 

For the hardness, we need to show that \mbox{3-S}AT is polynomial-time 
many-one reducible to this decision problem. We set $\kappa$ 
just as large as $(|Stmt[G]| + |Rel[G]| + |\pmb{Themes}| + \log (|\pmb{Atom}(\mathcal{D})|) + \kappa)$ 
is no smaller than the length of the input to 3-SAT. 
Then, %$\bigwedge \mathcal{I}(T_i, s_j)$ and 
%$\bigwedge \mathcal{I}(T'_i, s'_j)$ can be any 
%element in $D$ and $\mathcal{D}$ is a complete Boolean algebra. 
%As such, 
given any 3-CNF expression $z$, it is straightforward to choose $\mathcal{D}$ in the way that 
$\bigwedge \mathcal{I}(T_1, s_1) = z$. 
Choose $i = 1$ and $j = 1$, and set  
$\mathcal{I}(T'_1, s'_1)$ to $\{\pmb{0}\}$, 
then $z$ is satisfiable iff $\bigwedge \mathcal{I}(T_1, s_1) \not= \bigwedge \mathcal{I}(T'_1, s'_1)$. 
The second obligation is similar. 
\hfill$\Box$ \\

\noindent \textbf{Proof of Lemma \ref{lem_co_np_problems}.}  
\phantomsection \label{proof_co_np_problems}
We require a renaming for parts of this proof. The reason will be 
given in the relevant parts. 
\begin{innerdefinition}[Renaming]   \rm   
	Given a theme aspect argumentation model $\taam$, 
%	let $k$ be a natural number, 
%	let $n$ be the length of 
%	the representation of $\taam$, 
	let $W_{\mathcal{D}}$ be the set of words representing  
	$\mathcal{D}$ and let $W$ be a set of words satisfying 
	$p(n) \cdot |W_{\mathcal{D}}| \leq |W| \leq q(n)$ 
	where $p(n)$ and $q(n)$ 
	are both a polynomial of the length $n$ of the representation 
	of $\taam$. For every $k \leq p(n)$, we define  
	a function 
	$\rho_k: W_{\mathcal{D}} \rightarrow W$ satisfying the following 
	conditions. 
%	\begin{multicols}{2} 
	\begin{itemize} 
		\item $(\rho_k^{-1} \circ \rho_k)(w) = w$ 
			for every $w \in W_{\mathcal{D}}$. 
		\item for every $k_1, k_2 \leq p(n)$, 
			if $k_1 \not= k_2$, 
			$\{w \in W \mid \exists w' \in W_{\mathcal{D}}.
			\rho_{k_1}(w') = w\} \cap  
			\{w \in W \mid \exists w' \in W_{\mathcal{D}}.
			\rho_{k_2}(w') = w\} = \emptyset$. 
	\end{itemize}  
	Let $x$ be a string representing an element of $\mathcal{D}$, 
	then, by $x^{\rho_k}$, we denote a string 
	that obtains by simultaneously replacing each word $w$ occurring in $x$ 
	with $\rho_k(w)$. 
%	\end{multicols} 
	\hfill$\spadesuit$ 
\end{innerdefinition}  
Given a theme aspect argumentation model $\taam$ 
and a polynomially long string $x$ representing an element of $\mathcal{D}$, 
it is obvious that $x^{\rho_k}$ ($k \leq p(n)$) 
is polynomial-time obtainable. 

Now for the proof, with no loss of generality, assume $\mathcal{I}(T_i,s_j) = \{x^{(i,j)}_1, 
\ldots, x^{(i,j)}_{k_{(i,j)}}\}$ 
and $\mathcal{I}(T'_i, s'_j) = \{y^{(i,j)}_1, \ldots, 
y^{(i,j)}_{l_{(i,j)}}\}$ 
for every $1 \leq i \leq n$ and every $1 \leq j \leq m$, 
then: 
%Assume $\mathcal{I}(T,s) = \{x_1, \ldots, x_n\}$ 
%and $\mathcal{I}(T', s') = \{y_1, \ldots, y_m\}$ with no loss of 
%generality, then:  
\begin{itemize} 
	\item $\bigwedge_{1 \leq i\leq n}\bigwedge_{1\leq j \leq m}(\neg (x_1^{(i,j)} \wedge \cdots \wedge x_{k_{(i,j)}}^{(i,j)}) \vee (y^{(i,j)}_1 \wedge \cdots \wedge 
		y_{l_{(i,j)}}^{(i,j)}))$ is a tautology iff 
		$\bigwedge \mathcal{I}(T'_i, s'_j) \in 
{\uparrow\{\bigwedge \mathcal{I}(T_i, s_j)\}}$ holds for every 
		$1\leq i \leq n$ and every $1 \leq j \leq m$. 
	\item $\bigwedge_{1\leq i\leq n}\bigwedge_{1\leq j\leq m}(\neg (y_1^{(i,j)} \wedge \cdots 
		\wedge y_{l_{(i,j)}}^{(i,j)}) \vee (x_1^{(i,j)} \wedge 
		\cdots \wedge x_{k_{(i,j)}}^{(i,j)}))$ is a tautology iff 
		$\bigwedge \mathcal{I}(T'_i, s'_j) \in 
			{\downarrow\{\bigwedge \mathcal{I}(T_i, s_j)\}}$ 
			holds for every $1\leq i \leq n$ and every 
			$1 \leq j \leq m$.   
		\item Let $x = y$ denote 
			$(\neg x \vee y) \wedge (\neg y \vee x)$ and 
			let $\rho^{(i,j)}_z$ denote $\rho_{z + \Sigma_i \Sigma_j k_{(i,j)} \cdot l_{(i,j)}}$, 
			then 
		    {\small $\bigwedge_{1\leq i \leq n}\bigwedge_{1\leq 
		    j\leq m}(((x^{(i,j)}_1 = y^{(i,j)}_1)^{\rho^{(i,j)}_1} 
		    \vee \cdots \vee (x^{(i,j)}_1 = y^{(i,j)}_{l_{(i,j)}})^{\rho^{(i,j)}_{l_{(i,j)}}}) 
		    \wedge \cdots \wedge 
		((x_{k_{(i,j)}}^{(i,j)} = y^{(i,j)}_1)^{\rho^{(i,j)}_{(k_{(i,j)}-1) \cdot l_{(i,j)} + 1}} \vee 
		\cdots \vee (x_{k_{(i,j)}}^{(i,j)} =y_{l_{(i,j)}}^{(i,j)})^{\rho^{(i,j)}_{(k_{(i,j)}\cdot l_{(i,j)}}}))$} is a tautology 
			iff $\mathcal{I}(T_i, s_j) \subseteq 
			\mathcal{I}(T'_i, s'_j)$ for every 
			$1 \leq i \leq n$ and every $1 \leq j \leq m$. As each equality judgement  
			needs to be made independently, we need to ensure independence among equality 
			expressions. We make use of the renaming. 
		\item  Let $[w_1, \ldots, w_{\log(|\pmb{Atom}(\mathcal{D})|)}]$ 
			denote the representation of $\mathcal{D}$. 
			In case $\mathcal{I}(T_i', \omega)$ 
			is a bit pattern $b$, let 
			$\overline{b}$ denote 
			a bit pattern that reverses 1 and 0 
			in $b$. Let $W$ denote 
			$\{w \in \sigma([w_1, \ldots, w_{\log(|\pmb{Atom}(\mathcal{D})|)}], \overline{b}) \mid 
			\exists x \in \mathcal{I}(T_i, s_j). 
			  w \text{ occurs in } x\}$ and with no loss of generality, 
			  assume $W = \{w_1, \ldots, w_{end}\}$ with $end \leq \log(|\pmb{Atom}(\mathcal{D})|)$. 
			Further, for any $x \in \mathcal{I}(T_i, s_j)$, 
			let $x_{w \leftarrow \pmb{0}}$ denote the Boolean formula 
			that is almost exactly $x$ except that $w$ is replaced by \pmb{0} 
			and similarly let $x_{w \leftarrow \pmb{1}}$ denote the Boolean formula 
			that is almost exactly $x$ except that $w$ is replaced by \pmb{1}.  
			Then:
			$\bigwedge_{1 \leq i \leq n}\bigwedge_{1 \leq j \leq m}\bigwedge_{1 \leq z \leq end} 
			(x^{(i,j)}_{w_z \leftarrow \pmb{0}} = x^{(i,j)}_{w_z \leftarrow \pmb{1}})^{\rho_{i+j+z}}$ 
			is a tautology iff 
			$\mathcal{I}(T_i, s_j) \subseteq \mathcal{I}(T'_i, \omega)$. 
			The case where $\mathcal{I}(T_i', \omega)$ is not a bit pattern 
			is similar to the previous case. 
		\item Similarly for $\mathcal{I}(T_i, \omega) \subseteq 
			\mathcal{I}(T'_i, \omega)$. 
\end{itemize}  
By the conditions on the representation of $\mathcal{D}$ and $\mathcal{I}$, 
the length of any output of $\mathcal{I}$ is polynomially bounded. 
Hence, they are in co-SAT = co-NP. 

For the hardness, it suffices to 
show that co-SAT = TAUT is polynomial-time many-one reducible to 
each of the above decision problems. We set $\kappa$ 
just as large as 
$(|Stmt[G]| + |Rel[G]| + |\pmb{Themes}| + \log (|\pmb{Atom}(\mathcal{D})|) + \kappa)$ 
is no smaller than the length of the input to TAUT. 
For the 1st decision problem,  
choose $i = j = 1$ and set $\mathcal{I}(T_1, s_1)$ to $\{\pmb{1}\}$, then 
for any input $e$ to TAUT, we simply choose some $\mathcal{D}$ such that $e = \bigwedge \mathcal{I}(T'_1, s'_1)$. 
It is straightforward to see that $e$ is a tautology iff $\bigwedge \mathcal{I}(T'_1, s'_j) \in {\uparrow \{\bigwedge 
\mathcal{I}(T_1, s_1)\}}$ holds. Similarly for the 2nd decision problem. 
For the 3rd decision problem, choose $i = j = 1$ and set $\mathcal{I}(T'_1, s'_1)$ to $\{\pmb{1}\}$. 
Then, for any expression $e$, we simply choose some $\mathcal{D}$ such that $\mathcal{I}(T_1, s_1) = \{e\}$. 
It is straightforward to see that $e$ is a tautology iff $\mathcal{I}(T_1, s_1) \subseteq \mathcal{I}(T'_1, s'_1)$ holds. 
Similarly for the remaining two decision problems. \hfill$\Box$ \\ 

\noindent \textbf{Proof of Theorem \nameref{thm_satisfaction_tr}.} 
\phantomsection 
\label{proof_satisfaction_tr}
Given a theme aspect argumentation model $\taam$, for every $(s_1, s_2) \in Rel[G]$, we only need to examine fewer than $|\Pi((s_1, s_2)) \cup \Pi(s_1) \cup \Pi(s_2)|$ 
themes. Straightforward. \hfill$\Box$  \\

\noindent \textbf{Proof of Theorem \nameref{thm_satisfaction_nnp}.} \phantomsection \label{proof_satisfaction_nnp}
Given a theme aspect argumentation model $\taam$, for any $t \in \pmb{Themes}$ and any $s \in Stmt[G]$, let us use the following string  
${t::s}$ to represent $t \in \Pi(s)$. 
Then there are at most $|Stmt[G]| \cdot |\Pi(s)|$ 
polynomial-time enumerable such strings 
${t::s}$ for $t \in \pmb{Themes}$ and $s \in Stmt[G]$. 
For every $s \in PStmt[G]$, 
if $s \equiv t.\mathcal{C}$ for some $t \in \pmb{Themes}$, then we only need to check the existence of a string ${t::s'}$ 
for some $s' \in Stmt[G]$, and if 
$s \equiv t.a$ for some $t \in \pmb{Themes}$ and some $a \in OStmt[G]$, 
then we only need to check the existence of a string ${t::a}$ 
for some $a \in OStmt[G]$.  \hfill$\Box$ \\

\noindent \textbf{Proof of Theorem \nameref{thm_satisfaction_nsa}.}  
\phantomsection \label{proof_satisfaction_nsa}
 Given a theme aspect argumentation model $\taam$, it takes only to check  if `$attack$' is 
 in $\Pi((s', s))$ for every $(s', s) \in Rel[G]$. \hfill$\Box$ \\

\noindent \textbf{Proof of Theorem \nameref{thm_satisfaction_kos}.} 
\phantomsection \label{proof_satisfaction_kos} 
We use the ${t::s}$ strings we introduced in the proof of \mbox{Theorem \nameref{thm_satisfaction_nnp}}. Given a theme aspect argumentation model $\taam$, 
 for every $s \in PStmt[G]$, if $s \equiv t.a$ for some $t \in \pmb{Themes}$ and $a \in OStmt[G]$, 
 then both $U_1 \equiv \{t' \in \pmb{Themes} \mid {t'::s}\}$ and $U_2 \equiv \{t' \in \pmb{Themes} \mid {t'::a}\}$ are polynomial-time enumerable, and 
 the number of comparisons for checking the set-subsumption is capped by 
 $|U_1| \cdot |U_2|$. \hfill$\Box$ \\

\noindent \textbf{Proof of Theorem \nameref{thm_satisfaction_nss}.}  
\phantomsection \label{proof_satisfaction_nss} 
Given a theme aspect argumentation model $\taam$, 
for each $s \in Stmt[G]$, let 
$URight(s)$ denote $\{(s, s_1) \in Rel[G] \mid s_1 \in Stmt[G]\}$ 
and let $ULeft(s)$ denote $\{(s_1, s) \in Rel[G] \mid s_1 \in Stmt[G]\}$. Then,  
for any $s \in Stmt[G]$, 
it holds that $|URight(s) \cup ULeft(s)| \leq 2 \cdot |Stmt[G]|^2$, 
and that $|\Pi(URight(s)) \cup \Pi(ULeft(s))| \leq 2 + |\pmb{Themes}|$. 
Thus, for each $s \in Stmt[G]$, the number of comparisons required for checking the 3 conditions of \pmb{nss} is below 
$(2+|\pmb{Themes}|)^2 \cdot |Stmt[G]|^4$. \hfill$\Box$  \\

\noindent \textbf{Proof of Theorem \nameref{thm_sat_aass}.} 
\phantomsection \label{proof_sat_aass}
Suppose a decision problem SUBAASS with inputs 
(1) $\taam$, (2) one subset $T$ of $\pmb{Themes}$ 
and (3) one $(s', s) \in Rel[G]$ 
for verifying that
$\{`attack$'$\} \cup T \subseteq \Pi((s', s))$ materially 
implies $\emptyset \not= \bigwedge \mathcal{I}(T, s) 
\not= \bigwedge \mathcal{I}(T, s) \not= \emptyset$ and  
$\{`support$'$\} \cup T \subseteq \Pi((s', s))$ materially implies $\bigwedge \mathcal{I}(T, s') \in 
{\uparrow \{\bigwedge \mathcal{I}(T, s) \}} 
		\cup {\downarrow \{\bigwedge \mathcal{I}(T, s)\}}$.   
SUBAASS is in the complexity class 
that includes both NP and co-NP, {\it i.e.} in P(NP) since   
\begin{itemize} 
	\item the verification of  $\{`attack$'$\} \cup T \subseteq \Pi((s', s))$ 
		is trivially in P.
	\item that of $\emptyset \not= \bigwedge \mathcal{I}(T, s') \not= \bigwedge \mathcal{I}(T, s) \not= \emptyset$ is in NP by 
		Lemma \ref{lem_np_problems} and 
		by the straightforward observation that the verification of 
		$\bigwedge \mathcal{I}(T, s) \not= \emptyset \not= \bigwedge \mathcal{I}(T, s')$ is 
		in P. 
	\item that of $\{`support$'$\} \cup T \subseteq \Pi((s', s))$ is trivially in P. 
	\item that of $\bigwedge \mathcal{I}(T, s') \in {\uparrow \{\bigwedge \mathcal{I}(T, s) \}} 
		\cup {\downarrow \{\bigwedge \mathcal{I}(T, s)\}}$ is in 
		\mbox{co-NP} by Lemma \ref{lem_co_np_problems}. 
\end{itemize}   
Suppose a more general decision problem \mbox{SUBAASSPAIRS} with inputs
(1) $\taam$ and (2) one subset of $\pmb{Themes}$ 
for verifying that
$\{`attack$'$\} \cup T \subseteq \Pi((s', s))$ materially implies $\emptyset \not= \bigwedge \mathcal{I}(T, s') 
\not= \bigwedge \mathcal{I}(T, s) \not= \emptyset$; and that 
$\{`support$'$\} \cup T \subseteq \Pi((s', s))$ materially implies $\bigwedge \mathcal{I}(T, s') \in 
{\uparrow \{\bigwedge \mathcal{I}(T, s) \}} 
		\cup {\downarrow \{\bigwedge \mathcal{I}(T, s)\}}$ for every $(s', s) \in Rel[G]$. 
	There are only $|Stmt[G]|^2$ pairs of members of 
$Stmt[G]$. Thus, \mbox{SUBAASSPAIRS} is polynomial-time Turing 
reducible to \mbox{SUBAASS}. %by Lemma \ref{lem_subsumption_complexity_classes}.  
As $\SatOne{aass}$ is polynomial-time Turing reducible to \mbox{SUBAASSPAIRS}, 
it holds that $\SatOne{aass}$ is in P(P(P(NP))) = P(NP) by 
%1-Sat:\pmb{aass} is in \mbox{SUBAASSPAIRS} and 
Lemma \ref{lem_subsumption_complexity_classes}. 
%$\SatOne{aass}$ is then in P(NP).  
\hfill$\Box$ \\

\noindent \textbf{Proof of Theorem \nameref{thm_sat_bat}.} 
\phantomsection \label{proof_sat_bat}
Suppose a decision problem SUBBAT with inputs (1) $\taam$ and (2) 
some subset $T$ of $\pmb{Themes}$ for verifying:  $\mathcal{I}(T, \omega) \not= \emptyset$ 
materially implies that $(\mathcal{I}(T, \omega), \pmb{1}, \pmb{0}, \neg, \wedge, \vee)$ 
is a sub-complete Boolean algebra of $\mathcal{D}$. 
SUBBAT is in P(NP) since 
\begin{itemize}  
	\item the verification of $\mathcal{I}(T, \omega) \not= \emptyset$ is clearly in 
		P. 
	\item that of $\pmb{0}, \pmb{1} \in \mathcal{I}(T, \omega)$ is 
              that of $\{\pmb{0}, \pmb{1}\} \subseteq \mathcal{I}(T, \omega)$ which is in 
		co-NP by Lemma \ref{lem_co_np_problems}.
	\item provided $\pmb{0}, \pmb{1} \in \mathcal{I}(T, \omega)$, 
		the verification of $(\mathcal{I}(T, \omega), \wedge, \vee)$ 
		being a complete lattice is vacuous. 
	\item provided $\pmb{0}, \pmb{1} \in \mathcal{I}(T, \omega)$, 
		the verification of 
		$(\mathcal{I}(T, \omega), \wedge, \vee)$ being a 
		distributive lattice is in co-NP as 
		it is a tautology 
		check on $\bigwedge_{(x, y, z) \in \mathcal{I}(T, \omega)^{\times 3}} (x \wedge (y \vee z) = (x \wedge y) \vee (x \wedge z))$ which 
				is polynomially bounded in length. 
			\item If $\mathcal{I}(T, \omega)$ is 
				represented by a bit 
				pattern, then $\mathcal{I}(T, \omega)$ 
				is assumed a Boolean algebra, 
				so it is vacuous that 
				$\mathcal{I}(T, \omega)$ 
				includes the complement of 
				each one of the elements  
				in the Boolean algebra. Therefore, 
				assume   
				$\mathcal{I}(T, \omega) = 
				\{x_1, \ldots, x_n\}$ with no loss of 
				generality, 
				then 
		the verification that every $x \in \mathcal{I}(T, \omega)$ has a complement $y \in \mathcal{I}(T, \omega)$ 
				is in NP as 
				it is a satisfiability 
				check on 
				$((x_1 \vee y_1 = \pmb{1}) \wedge 
				(x_1 \wedge y_1 = \pmb{0}))
				\wedge \cdots \wedge 
				((x_n \vee y_n = \pmb{1}) \wedge 
				(x_n \wedge y_n = \pmb{0}))$ 
				for $y_1, \ldots, y_n \in \mathcal{I}(T, \omega)$ 
				which is polynomially bounded in length. 
\end{itemize}   
$\SatOne{bat}$ is polynomial-time Turing reducible to SUBBAT which is vacuously polynomial-time many-one reducible to P(NP).   
Thus, $\SatOne{bat}$ is in P(P(NP)) = P(NP) by Lemma \ref{lem_subsumption_complexity_classes}. 
\hfill$\Box$ \\

\noindent \textbf{Proof of Theorem \nameref{thm_sat_pr}.} 
\phantomsection \label{proof_sat_pr}
For $\SatOne{pr}$, given $\taam$, 
the verification of all the 3 cases for every $s \in Stmt[G]$ is in co-NP by Lemma \ref{lem_co_np_problems} 
and by the fact that there are only $|Stmt[G]|$ members in $Stmt[G]$.  
%With the restriction on $\taam$ that it be the power-set abstraction of itself, 
%the verification is in P by Lemma \ref{lem_reduce_conp}. 
%$\SatOne{pr}$ is therefore in co-NP. 
\hfill$\Box$ \\%and $\SatOneP{pr}$ in P.   
%Since every subset of $\pmb{Themes}$ is uniquely identified by 
%a (polynomially-bounded) $|\pmb{Themes}|$-bit sequence $\{0,1\}^{|\pmb{Themes}|}$ and 
%since the verification must be done for each of the subsets, 
%$\Sat{pr}$ is then in co-NP(NP) and $\SatP{pr}$ is in co-NP
%by Lemma \ref{lem_subsumption_complexity_classes}. \hfill$\Box$ 

%%\noindent \textbf{Proof of Theorem \ref{thm_sat_ss}.}
%
%For $\SatOne{ss}$, given  $\taam$, it needs to be shown that the verification of 
%the following for every $T \subseteq \pmb{Themes}$ and every $s \in Stmt[G]$ 
%is in NP: 
%$T \subseteq \Pi(s)$ and $\mathcal{I}(T, s) \not= \emptyset$ 
%materially imply both $\pmb{0} \not= \bigwedge \mathcal{I}(T, s)$ 
%and $\pmb{1} \not= \bigwedge \mathcal{I}(T, s)$. 
%%Suppose also its sub-problem SUBSS$^{\mathfrak{p}}$ 
%%which restricts the first parameter $\taam$ to the power-set abstraction of itself.
%Lemma \ref{lem_np_problems}
%%and \ref{lem_reduce_np} 
%and the 
%straightforward observation that $T \subseteq \Pi(s)$ and $\mathcal{I}(T, s)$ are polynomial-time verifiable/computable for every $T \subseteq \pmb{Themes}$ and every $s \in Stmt[G]$ 
%prove the result.  \hfill$\Box$ \\
%
\noindent \textbf{Proof of Theorem \nameref{thm_sat_mat}.}
\phantomsection \label{proof_sat_mat}
%suppose a decision problem SUBMAT which takes 
% $\taam$ 
We need to verify that, for every $T_1, T_2 \subseteq \pmb{Themes}$, 
$T_1 \subseteq T_2$ materially implies 
$\mathcal{I}(T_1, \omega) \subseteq \mathcal{I}(T_2, \omega)$. 
%Suppose also its sub-problem SUBMAT$^{\mathfrak{p}}$ which restricts 
%the first parameter $\taam$ to the power-set abstraction of itself. 
This is in co-NP by Lemma
\ref{lem_co_np_problems} and by the fact that $T_1 \subseteq T_2$ is polynomial-time verifiable. \hfill$\Box$ \\
%and SUBMAT$^{\mathfrak{p}}$ is in P by Lemma \ref{lem_reduce_conp}.  
%$\SatOne{mat}$ is therefore in co-NP and $\SatOneP{mat}$ in P. 
%Now, let us assume $\pmb{Themes} = \{t_1, \ldots, t_n\}$ with no loss of 
%generality. Every pair of subsets of $\pmb{Themes}$ 
%is identified uniquely by 
%a (polynomially-bounded) $2n$-bit sequence $\{0, 1\}^{2n}$.  
%Specifically, one of them is identified by the first half 
%of the $2n$ bits, while the other is identified by 
%the second half of the $2n$ bits. Since the verification needs to be conducted for 
%every pair, $\Sat{mat}$ is in co-NP(SUBMAT) = co-NP(NP)  
%and $\SatP{mat}$ is in co-NP(SUBMAT$^{\mathfrak{p}}$) = \mbox{co-NP} 
%by Lemma \ref{lem_subsumption_complexity_classes}. \hfill$\Box$ 

\noindent \textbf{Proof of Theorem \nameref{thm_sat_ss}.}
\phantomsection \label{proof_sat_ss}
For $\SatOne{ss}$, given  $\taam$, it needs to be shown that the verification of 
the following for every $T \subseteq \pmb{Themes}$ and every $s \in Stmt[G]$ 
is in NP: 
$T \subseteq \Pi(s)$ and $\mathcal{I}(T, s) \not= \emptyset$ 
materially imply both $\pmb{0} \not= \bigwedge \mathcal{I}(T, s)$ 
and $\pmb{1} \not= \bigwedge \mathcal{I}(T, s)$. 
%Suppose also its sub-problem SUBSS$^{\mathfrak{p}}$ 
%which restricts the first parameter $\taam$ to the power-set abstraction of itself.
Lemma \ref{lem_np_problems}
%and \ref{lem_reduce_np} 
and the 
straightforward observation that $T \subseteq \Pi(s)$ and $\mathcal{I}(T, s)$ are polynomial-time verifiable/computable for every $T \subseteq \pmb{Themes}$ and every $s \in Stmt[G]$ 
prove the result.  \hfill$\Box$ \\
%See Appendix:  proofs. \hfill$\Box$\\ 

%Now, for any $s \in Stmt[G]$, the upper bound of $\Pi(s)$ is $\pmb{Themes}$. Thus,   
%similarly to the preceding proofs, $\Sat{ss}$ is in co-NP(SUBSS) = co-NP(NP) 
%and $\SatP{ss}$ is in co-NP. \hfill$\Box$ 

%the verification of $\taam$'s satisfiability of \pmb{ss} is PSPACE. 
%For a partial verification of \pmb{ss}, for a given $s \in Stmt[G]$ and a given $T \subseteq \Pi(s)$, 
%the verifications of $\mathcal{I}(T, s) \not= \emptyset$, of $\pmb{0} \not= \bigwedge \mathcal{I}(T, s)$, 
%and of $\pmb{1} \not= \bigwedge \mathcal{I}(T, s)$ are all NP. Consequently,   
%the verification of the three conditions for every $s \in Stmt[G]$ and 
%for every polynomially-bounded number of subsets   
%of $\Pi(s)$ is P(NP). \hfill$\Box$ 

\noindent \textbf{Proof of Theorem \nameref{thm_sat_esr}.} 
\phantomsection \label{proof_sat_esr}
Suppose the following 3 decision problems. 
\begin{itemize} 
	\item SUBESR with inputs (1) $\taam$, (2) 
		some subsets $T_1, T_2$ of $\pmb{Themes}$, 
		and (3) some $t.\mathcal{C} \in PStmt[G]$ 
		for verifying $\mathcal{I}(T_1, t.\mathcal{C}) 
		\cap Common = \mathcal{I}(T_2, t.\mathcal{C}) \cap 
		Common$. 
	\item SUBENSR with inputs (1) $\taam$, (2) 
		some subsets $T_1, T_2$ of $\pmb{Themes}$, 
		and (3) some $t.a \in PStmt[G]$ 
		for verifying $\mathcal{I}(T_1, t.a) 
		\cap Common = \mathcal{I}(T_2, t.a) \cap 
		Common$. 
	\item SUBEOS with inputs (1) $\taam$, (2) 
		some subsets $T_1, T_2$ of $\pmb{Themes}$, 
		and (3) some $a \in OStmt[G]$ 
		for verifying $\mathcal{I}(T_1, a) 
		\cap Common = \mathcal{I}(T_2, a) \cap 
		Common$. 
\end{itemize} 
$Common \equiv \mathcal{I}(T_1, \omega) \cap \mathcal{I}(T_2, \omega)$ 
in each of them. 
%\mbox{SUBESR$^{\mathfrak{p}}$}, \mbox{SUBENSR$^{\mathfrak{p}}$} and \mbox{SUBEOS$^{\mathfrak{p}}$}  
%are the corresponding sub-problems restricting the first parameter $\taam$ to the power-set abstraction 
%of itself. 

Let us consider SUBESR. %and SUBESR$^{\mathfrak{p}}$, 
For each $i \in \{1,2\}$, 
it takes fewer than 
$|\mathcal{I}(T_i, t.\mathcal{C})| \cdot 
|\mathcal{I}(T_1, \omega)| \cdot |\mathcal{I}(T_2, \omega)|$ 
tautology checks for enumerating all members of 
$\mathcal{I}(T_i, t.\mathcal{C}) \cap Common$.  
%With no loss of generality, we assume that only the members of $\mathcal{I}(T_1, \omega)$ 
%are getting enumerated. This assumption is for eliminating the need of checking any more logical equivalence. 
This enumeration 
is in P(co-NP) = 
P(NP) by Lemma \ref{lem_subsumption_complexity_classes}. 
%For SUBESR$^{\mathfrak{p}}$, it is in P(P) = P by Lemma \ref{lem_subsumption_complexity_classes} again.  
Now, if $|\mathcal{I}(T_1, t.\mathcal{C}) \cap 
Common| \not= |\mathcal{I}(T_2, t.\mathcal{C}) \cap Common|$, 
then $\mathcal{I}(T_1, t.\mathcal{C}) \cap Common = 
\mathcal{I}(T_2, t.\mathcal{C}) \cap Common$ is not satisfiable. 
Otherwise, the verification of $\mathcal{I}(T_1, t.\mathcal{C}) \cap Common = 
\mathcal{I}(T_2, t.\mathcal{C}) \cap Common$  
requires fewer than 
$|\mathcal{I}(T_1, t.\mathcal{C}) \cap Common|^2$ tautology checks.  
This part is also in P(co-NP) = P(NP). 
%of members of $\mathcal{I}(T_1, \omega)$.  
%Due to the above assumption, there is no need of redoing logical equivalence check. 
Hence, SUBESR is in P(NP).

Suppose a more general decision problem SUBESRALLS 
with inputs (1) $\taam$ and (2) some subsets $T_1, T_2 
\subseteq \pmb{Themes}$ for verifying $\mathcal{I}(T_1, t.\mathcal{C})
\cap Common = \mathcal{I}(T_2, t.\mathcal{C}) \cap Common$ 
for every $t.\mathcal{C} \in PStmt[G]$. % as well as its sub-problem SUBESRALLS$^{\mathfrak{p}}$ 
%which restricts the first parameter $\taam$ to the power-set abstraction of itself. 
Since there are fewer than or equal to $|Stmt[G]|$ 
members in $PStmt[G]$, SUBESRALLS is polynomial-time Turing 
reducible to SUBESR and is in P(P(NP)) = P(NP) by Lemma \ref{lem_subsumption_complexity_classes}. 
%while SUBSERALLS$^{\mathfrak{p}}$ is polynomial-time Turing reducible to SUBESR$^{\mathfrak{p}}$ 
%which is thus in P(P) = P by Lemma \ref{lem_subsumption_complexity_classes} again. 
Hence, $\SatOne{esr}$ is in P(SUBESRALLS) which is in P(NP) by Lemma \ref{lem_subsumption_complexity_classes}. %and $\SatOneP{esr}$ is in SUBESRALLS$^{\mathfrak{p}}$ 
%= P. 
%Then, similarly to the preceding proofs, $\Sat{esr}$ 
%is in co-NP(SUBESRALLS) = co-NP(NP) and  
%$\SatP{esr}$ is in co-NP(SUBESRALLS$^{\mathfrak{p}}$) = co-NP 
%by Lemma \ref{lem_subsumption_complexity_classes}.  
The complexities of $\SatOne{ensr}$ 
and $\SatOne{eos}$ are similarly derived from 
SUBENSR and SUBEOS. \hfill$\Box$ \\

\noindent \textbf{Proof of Theorem \nameref{thm_sat_das}.} 
\phantomsection \label{proof_sat_das}
Suppose a decision problem SUBDAS 
with inputs 
(1) $\taam$, (2) some $(s', s) \in Rel[G]$ and (3) 
some subset $T$ of $\pmb{Themes}$ 
for verifying that 
$\emptyset \subset T \subseteq \pmb{Themes}$ 
and $T \cup \{`attack$'$\} \subseteq \Pi((s', s))$ 
materially imply $\bigwedge \mathcal{I}(T, s') \not\in 
{\downarrow \{\bigwedge \mathcal{I}(T, s)\}} 
\cup {\uparrow \{\bigwedge \mathcal{I}(T, s)\}}$ 
and that $\emptyset \subset T \subseteq \pmb{Themes}$ 
and $T \cup \{`support$'$\} \subseteq \Pi((s', s))$ 
materially imply $\bigwedge \mathcal{I}(T, s') = 
\bigwedge \mathcal{I}(T, s)$. %SUBDAS$^{\mathfrak{p}}$ is  
%its sub-problem which restricts the first parameter $\taam$ to the power-set abstraction 
%of itself. 
SUBDAS is in P(NP) straightforwardly by Lemmas \ref{lem_np_problems} and 
\ref{lem_co_np_problems}. %while SUBDAS$^{\mathfrak{p}}$ is in P by Lemmas \ref{lem_reduce_np} and 
%\ref{lem_reduce_conp}. 
Suppose a more general decision problem 
SUBDASALLPAIRS with the inputs (1) and (3)  
for verifying the above conditions for every $(s', s) \in Rel[G]$.  
%SUBDASALLPAIRS$^{\mathfrak{p}}$ is its sub-problem which restricts (1) 
%to the power-set abstraction of itself. 
Since there are at most $|Stmt[G]|^2$ members of $Rel[G]$, 
SUBDASALLPAIRS is polynomial-time Turing reducible 
to SUBDAS. %and SUBDASALLPAIRS$^{\mathfrak{p}}$ polynomial-time Turing reducible 
%to SUBDAS$^{\mathfrak{p}}$. 
$\SatOne{das}$ is polynomial-time Turing reducible to SUBDASALLPAIRS which is then in P(P(P(NP))) = P(NP) %and
%$\SatOneP{das}$ is in SUBDASALLPAIRS$^{\mathfrak{p}}$ = P(P) = P 
by Lemma \ref{lem_subsumption_complexity_classes}. 
%Then, similarly to the preceding proofs, 
%$\Sat{das}$ is in \mbox{co-NP(NP)} and $\SatP{das}$ is in co-NP.  
\hfill$\Box$ \\

\noindent \textbf{Proof of Theorem \ref{thm_sat_core}.} 
\phantomsection \label{proof_inclusion_sat_all_f}
Suppose a decision problem PARTIAL  
which takes  (1) $\taam$ and (2) one subset $T_1$ of 
$\pmb{Themes}$ and (3) another subset $T_2$ of $\pmb{Themes}$ for verifying \mbox{1-Partial:\pmb{$x$}} for every 
constraint \pmb{$x$} in 
\pmb{All}$^{- \text{\pmb{F}}} \backslash \{$\pmb{vi}$\}$. 
%The first input is passed to 
%\mbox{1-Partial:\pmb{$vi$}}, 
The first two inputs are passed to 
\mbox{1-Partial:\pmb{$x$}} for \pmb{$x$} $\in \{$\pmb{aass}, 
\pmb{i}, \pmb{bat}, \pmb{pr}, \pmb{ss}, \pmb{das}, \pmb{nwci}$\}$, 
and all the three inputs are passed to 
\mbox{1-Partial:\pmb{$x$}} for \pmb{$x$} $\in \{$\pmb{mat}, 
\pmb{manss}, \pmb{esr}, \pmb{ensr}, \pmb{eos}$\}$.  
PARTIAL is in P($\bigcup_{\pmb{x} \in \text{(\pmb{All}$^{- \text{\pmb{F}}}
\backslash \{$\pmb{vi}$\}$)}}\SatOne{x}$) which is 
in P(NP) by Lemmas \ref{lem_subsumption_complexity_classes} 
and \ref{lem_partial}. 

In terms of PARTIAL, $\Sat{ \pmb{All}$^{- \text{\pmb{F}}}$}$ 
is a decision problem which takes $\taam$ 
for verifying $\SatOne{vi}$ and 
PARTIAL for every $T_1, T_2 \subseteq \pmb{Themes}$. 
Now, there is a polynomially-bounded bit sequence 
$\{0,1\}^{2\cdot|\pmb{Themes}|}$ which uniquely identifies every combination 
of two subsets of $\pmb{Themes}$. Specifically, the first 
$|\pmb{Themes}|$ bits identify a subset $T_1$ of $\pmb{Themes}$ 
and the remaining $|\pmb{Themes}|$ bits identify a subset $T_2$ 
of $\pmb{Themes}$. Since 
$\Sat{ \pmb{All}$^{- \text{\pmb{F}}}$}$ needs to call PARTIAL 
for every $T_1, T_2 \in \pmb{Themes}$, 
$\Sat{ \pmb{All}$^{- \text{\pmb{F}}}$}$ is in  
\mbox{co-NP(PARTIAL)} which is in 
\mbox{co-NP(NP)} by Lemma \ref{lem_subsumption_complexity_classes}. 
\hfill$\Box$ \\

\noindent \textbf{Proof of Lemma \ref{lem_hardness_aass}.}
\phantomsection \label{proof_hardness_aass}
We divide $\Sat{aass}$ into two parts  
by supposing 
a decision problem AASSF which takes 
$\taam$ for verifying  
the following: for every $\emptyset \subset T \subseteq \pmb{Themes}$ 
and every $(s', s) \in Rel[G]$, $\{`attack$'$\} \cup T \subseteq \Pi((s', s))$ 
materially implies $\emptyset \not= \bigwedge \mathcal{I}(T, s') \not= \bigwedge \mathcal{I}(T, s) \not= \emptyset$, 
and another decision problem AASSL which takes 
$\taam$ for verifying the following: 
for every $\emptyset \subset T \subseteq \pmb{Themes}$ 
and every $(s', s) \in Rel[G]$, $\{`support$'$\} \cup T \subseteq \Pi((s', s))$ 
materially implies $\bigwedge \mathcal{I}(T, s') \in {\uparrow \{\bigwedge \mathcal{I}(T, s) \}} 
		\cup {\downarrow \{\bigwedge \mathcal{I}(T, s)\}}$.   
For any 3-CNF Boolean formula $F$ with unquantified variables $x_1, \ldots, x_n, y_1, \ldots, y_m$ 
with polynomially-bounded $n$ and $m$, %\footnote{We assume readers 
%are familiar with the terminologies of classical logic.} 
 we show that 
$\exists x_1 \cdots \exists x_n \forall y_1 \cdots 
\forall y_m.F(x_1, \ldots, x_n, y_1, \ldots, y_m)$  
is polynomial-time many-one reducible to co-AASSF. 
Here, $F(x_1, \ldots, x_n, y_1, \ldots, y_m)$ denotes 
a propositional Boolean formula in which 
$\{x_1, \ldots, x_n, y_1, \ldots, y_m\}$ 
is the set of all propositional variables occurring in $F$. Assume with no loss of generality that  
$x_1, \ldots, x_n, y_1, \ldots, y_m$ all range over $\{0,1\}$.\footnote{There is then 
a corresponding Boolean formula $\exists x\forall y.F'(x, y)$ 
such that $x$ and $y$ range over $\{0,1\}^{n+m}$ and 
that $\exists x\forall y.F'(x,y)$ is a true formula  
iff $\exists x_1 \cdots \exists x_n \forall y_1 \cdots \forall 
y_m.F(x_1, \ldots, x_n, y_1, \ldots, y_m)$ is.} 
We construct the corresponding representation of $\taam$  
as follows.  
\begin{itemize}    
		\begin{multicols}{2} 
	\item $\pmb{Themes} = \{x_1, \ldots, x_n\}$.  
	\item $\mathcal{D}$ is represented by 
		$[y_1, \ldots, y_m]$.   
	\item $Stmt[G] = \{s', s\}$. 
	\item $Rel[G] = \{(s', s)\}$. 
	\item $\Pi((s', s)) = \{`attack$'$\} \cup \pmb{Themes}$.  
		$\Pi(s') = \Pi(s) = \pmb{Themes}$.  
		\end{multicols} 
		\vspace{-0.4cm} 
	\item $\mathcal{I}$ is such that, 
		for every $\emptyset \subset T \subseteq \pmb{Themes}$,  
		$\mathcal{I}(T, s') = \{\pmb{1}\}$ 
		and $\mathcal{I}(T, s) = \{F_{\star}\}$ where $F_{\star}$ 
		is almost exactly $F$ except for the following difference: 
		for every $1 \leq i \leq n$, 
		if $x_i \in T$, then $x_i$ is replaced by \pmb{1}, 
		and if $x_i \not\in T$, then $x_i$ is replaced by \pmb{0}. 
\end{itemize}  
$\kappa$ is chosen so that $k \equiv (|Stmt[G]| + |Rel[G]| + |\pmb{Themes}| + \log (|\pmb{Atom}(\mathcal{D})|) + \kappa)$ 
is just as long as $k$ is no smaller than the length of $\exists x_1 \cdots \exists x_n \forall y_1 \cdots \forall y_m.F(x_1, \ldots, 
x_n, y_1, \ldots, y_m)$. The length of the above representations 
is then bounded by a polynomial $p(k)$. $\exists x_1 \cdots \exists x_n \forall y_1 \cdots y_m
F(x_1, \ldots, x_n, y_1, \ldots, y_m)$ is a true formula 
iff co-AASSF accepts the input described above. 
Hence, \mbox{co-A}ASSF is \mbox{NP(N}P)-hard or AASSF is \mbox{co-NP(N}P)-hard, as required. Since 
AASSF is called in $\Sat{aass}$, $\Sat{aass}$ is \mbox{co-NP(N}P)-hard. \hfill$\Box$ \\

\noindent \textbf{Proof of Theorem \ref{thm_model_existence_model_checking}.}
\phantomsection \label{proof_thm_model_existence_model_checking}
By Theorem \ref{thm_core}, the existence 
is guaranteed as long as $\modell$ satisfies \pmb{tr}, \pmb{nnp}, \pmb{nsa} and \pmb{kos}. 
By Theorems \nameref{thm_satisfaction_tr}, \nameref{thm_satisfaction_nnp}, \nameref{thm_satisfaction_nsa}
and \nameref{thm_satisfaction_kos}, the satisfiability of these constraints is polynomial-time verifiable. This concludes 
the first part. For the second part, by Theorem \ref{thm_sat_core_complete}, 
%Theorems \ref{thm_sat_aass} 
%$\sim$ \ref{thm_sat_i} and 
%\ref{thm_sat_bat} $\sim$ \ref{thm_sat_ss},  
\pmb{Core} is not polynomial-time verifiable 
unless P = NP. \hfill$\Box$  \\

\noindent \textbf{Proof of Lemma \ref{lem_computing_width_statements_sets}.}
\phantomsection \label{proof_computing_width_statements_sets}
$WidS[s,t,trel]$ includes every set $S$ that satisfies all the following 
conditions. %By Proposition \ref{prop_uniqueness}, if $t \in \Pi(s)$, 
%then $WidS[s,t,rel]$ always 
%exists for the given inputs and is a singleton set. If $t \not\in \Pi(s)$, 
%then $WidS[s,t,rel]$ does not exist. 
\begin{enumerate} 
	\item $s \in S$. 
	\item for every $s' \in S$,  $s \not= s'$ materially implies 
		$(s', s) \in Rel[G]$. 
	\item for every $s' \in S$, $(s', s) \in Rel[G]$ 
		materially implies $\{trel, t\} \subseteq \Pi((s', s))$. 
	\item for every $s' \in S$, $t \in \Pi(s')$. 
	\item $S$ is a maximal set that satisfies 
		the above conditions 1. $\sim$ 4.. 
\end{enumerate}    
By Proposition \ref{prop_uniqueness}, $|WidS[s,t,trel]|$ is at most 1. 
%If $t \not\in \Pi(s)$, then WSS outputs $\emptyset$. Otherwise, 
Let $S_{\star}$ be a dynamic working set which initially 
is $\{s\} \cup \{s' \in Stmt[G] \mid (s', s) \in Rel[G]\}$. 
$S_{\star}$ is polynomial-time enumerable, 
and by definition already satisfies the first and the second conditions.  
For every $s' \in S_{\star}$, if $(s', s) \in Rel[G]$, then 
the 3rd and the 4th conditions are polynomial-time verifiable. If 
either of them does not hold, then WSS removes $s'$ off $S_{\star}$. 
The set of the remaining elements of $S_{\star}$ is 
the member of $WidS[s,t,trel]$. It may be empty. WSS 
outputs $\{S_{\star}\}$. \hfill$\Box$ \\

\noindent \textbf{Proof of Lemma \ref{lem_polynomial_time_transformation_dag_labelled_dag}.} 
\phantomsection \label{proof_polynomial_time_transformation_dag_labelled_dag}  
Since we only need to show 
polynomial-time computability, we shall prove it in smaller steps.  
We presume the following order which is trivially a partial order. 
%The labelled DAG of a DAG is polynomial-time obtainable. 
%To see that it is so, start by observing in 
%Example \ref{ex_maximal_reverse_path}  
%that, if we had another DAG which is almost the same DAG save 
%it does not have the edge $(v_1, v_4)$, then the labelled DAG 
%of the new DAG would still have the same $label$ as for 
%the DAG in Example \ref{ex_maximal_reverse_path}. This implies 
%that $label$ for the labelled DAG of a DAG can be obtained from a smaller DAG 
%with respect to the following order. 
\begin{innerdefinition}[Order $\leq_{\mathsf{DAG}}$] \rm \label{def_partial_order}
    We define $\leq_{\mathsf{DAG}}$ to be an order 
	on DAGs defined as follows.     
	Let $(V_1, E_1)$ and $(V_2, E_2)$ denote 
	a DAG, 
	let $((V_1, E_1), label_1)$ denote 
	the labelled DAG of $(V_1, E_1)$ and 
	let $((V_2, E_2), label_2)$ denote 
	the labelled DAG of $(V_2, E_2)$, then 
	$(V_1, E_1) \leq_{\mathsf{DAG}} (V_2, E_2)$  
	iff all the following hold:  
	(1) $V_1 = V_2$; (2) $E_1 \subseteq E_2$; (3)  
	$label_1(v, v') = label_2(v, v')$ for every $v, v' \in V_1$.   
	\hfill$\spadesuit$ 
\end{innerdefinition} 

\noindent The following result holds. 
\begin{innertheorem}[Unique traversal of an edge] \label{thm_unique_traversal_of_edge} 
    Given a DAG $(V, E) \in (\mathcal{G}^+\backslash \mathcal{G})$, 
	if $(V, E)$ is a minimal DAG with respect to $\leq_{\mathsf{DAG}}$, 
	then the following hold for every $(v_1, v_2) \in E$:  
	(1) for any node $v_3$, if there is a reverse path 
	from $v_3$ to $v_2$, there is a maximal reverse path 
	from $v_3$ to a terminal node which traverses both $v_2$ and $v_1$; 
	(2) there is one and only one reverse path from $v_2$ to $v_1$. 
\end{innertheorem}  
\textbf{Inner-proof.} Let $v_3$ be a node from which there is a reverse path to 
$v_2$. 

For (1), we need to show that at least one reverse path from $v_3$ to a terminal node 
which traverses both $v_1$ and $v_2$ must be a maximal reverse path 
from $v_3$ to the terminal node. Suppose otherwise, then 
no member of the set of all maximal reverse paths from $v_3$ to the terminal node 
traverse both $v_1$ and $v_2$. Thus, none of them could subsume 
any reverse path from $v_3$ to the terminal node that traverses 
both $v_1$ and $v_2$, contradicting the supposition that the presumed 
set includes all the maximal reverse paths from $v_3$ to the terminal node. 

For (2), we need to show that there cannot be a reverse path from $v_2$ to $v_1$ 
that traverses any other nodes than $v_2$ and $v_1$ alone.  
Suppose otherwise, then for any reverse path from $v_2$ 
to $v_1$, the path that traverses $v_1$ right after $v_2$ 
is subsumed by another reverse path from $v_2$ to $v_1$.  
Hence, for any maximal reverse path from $v_3$ to a terminal node, 
if it traverses both $v_2$ and $v_1$, then  
there is some $v' (\not= v_1)$ that is traversed right after $v_2$. 
Then, $(V, E \backslash \{(v_1, v_2)\}) \leq_{\mathsf{DAG}}
(V, E)$, contradicting the assumption that $(V, E)$ is a minimal DAG.  
\hfill {\it Q.E.D.} \\

By Inner-Theorem \ref{thm_unique_traversal_of_edge}, 
the labelled DAG of a $\leq_{\mathsf{DAG}}$-minimal DAG is such that 
$label(v, v')$ is the set of all previous nodes of $v'$ in some reverse 
path from $v$ to a terminal node that traverses $v'$. Hence: 
%The labelled DAG of any $\leq_{\mathsf{DAG}}$-minimal DAG is 
%thus polynomial-time computable. 
\begin{innerlemma}[Polynomial-time transformation of a $\leq_{\mathsf{DAG}}$-minimal DAG into a labelled DAG] \label{lem_polynomial_time_transformation_minimal_dag_labelled_dag}
      Given a DAG $(V, E) \in (\mathcal{G}^+ \backslash \mathcal{G})$,   
	if $(V, E)$ is minimal with respect to $\leq_{\mathsf{DAG}}$, 
	then the labelled DAG of $(V, E)$ is polynomial-time computable. 
\end{innerlemma} 
\textbf{Inner-proof.}  
Algorithm \ref{algo_labelled_dag} 
generates the labelled DAG.  
%\incmargin{1em} 
%\restylealgo{boxed} 
{\small 
\begin{algorithm}[h]
	\SetAlgoLined
	\SetKwInOut{Input}{input} 
	\SetKwInOut{Output}{output}  
	\Input{a $\leq_{\mathsf{DAG}}$-minimal DAG $(V, E)$} 
	\Output{a labelled DAG $((V, E), label)$} 
	\BlankLine   
%	$sortedV \gets \mathsf{topologicalSort}((V, E))$; \\ 
	\For{$v \in V$}{  
	    $\mathsf{init}(sortedV, v)$; $\mathsf{init}(label, v)$; 
	    $\mathsf{init}(index, 1)$;    \\
%	    $\mathsf{init}(lenDict, sortedV)$;\\   
		\While{$index \leq |sortedV|$}{ 
			$v_{this} \gets sortedV.get(index)$; \\
			%$\mathsf{init}(minLen,  lenDict.get(v_{this}))$; \\ 
			\If{$v_{this} \not= v$}{ 
			   $label(v, v_{this}) \gets \mathsf{prevList}(sortedV, v_{this})$; 
			   } 
			   	$index$++; \\ 
			%\ForEach{$v_{prev} \in \mathsf{prevList}(sortedV, v_{this})$}{  
			%   \If{$minLen==1 \text{ or } lenDict.get(v_{prev}) \leq minLen$}{ 
			%	$minLen \gets lenDict.get(v_{prev}) - 1$;  
			%	$lenDict.set(v_{this}, minLen)$; 
%				$label(v, v_{this}) \gets \{v_{prev}\}$; 
			   }
%			   \ElseIf{$lenDict.get(v_{prev}) == minLen + 1$}{ 
%				$label(v, v_{this}) \gets label(v, v_{this}) \cup \{v_{prev}\}$; 
%			   }
%			}
		%	}
				%}
	}
	\Return{$((V, E), label)$}; 
	\caption{Algorithm to obtain the labelled DAG of a 
	$\leq_{\mathsf{DAG}}$-minimal DAG. $\mathsf{init}$ initialises.}
	\label{algo_labelled_dag}
\end{algorithm}
}
For each $v \in V$ (\mbox{Line 1}), 
\mbox{Line 2} initialises its local data. $\mathsf{init}(sortedV, v)$ generates 
$sortedV$ which is a list of members of $V$ such that 
\begin{itemize} 
	\item the members of $sortedV$ are precisely those of $\{v' \in V \mid (v', v) \in E^*\}$ 
(where $E^*$ is the reflexive-transitive 
closure of $E$). 
	\item $sortedV$ is topologically sorted so that, for any $v', v'' \in sortedV$, $(v'', v') \in E$ (necessarily $v'' \not= v'$) materially implies that $v'$ appears at an earlier index than $v''$. 
\end{itemize} 
Next, 
$\mathsf{init}(label, v)$ sets $label(v, v')$ to $\emptyset$ for every $v' \in V$. 
Next, $index$ that is used to refer to an index of $sortedV$ is initialised to 1. Here, 
it is assumed that the first index of $sortedV$ is 1. %Finally, $\mathsf{init}(lenDict, sortedV)$
%initialises a hash map $lenDict$ with the set of its keys being exactly the set of all members of $sortedV$ 
%and with $lenDict(v')$ being 0 if $v = v'$; 1, otherwise. In this algorithm, the traversal cost of 
%$-1$ is presumed on all the edges, thus the value 1 is considered as an 
%extremely large value for a length of a path. 
Topological sorting 
is known to be done in polynomial-time, so \mbox{Line 2} is straightforwardly polynomial-time computable. 

We start to traverse $(V, E)$ from $v$ in the reverse direction (\mbox{Line 3}). 
\mbox{Line 4} sets the current node $v_{this}$ to the node stored in $sortedV$ at the current index $index$. 
%\mbox{Line 5} initialises $minLen$ to the length value of $v_{this}$ as stored in $lenDict$ which is: 0 if 
%$v_{this} = v$; 1, otherwise. 
Since $v$ does not have any preceding nodes in the reverse paths that start at $v$, 
if $v_{this}$ is $v$, we do nothing (\mbox{Line 5}). Otherwise, we obtain 
the list of all the immediately preceding nodes of $v_{this}$ in $(V, E)$ that appear in $sortedV$. 
Note that if $v_x$ is one of them, then $(v_{this}, v_x) \in E$ holds. That is, 
$v_x$ precedes $v_{this}$ in a reverse path. 
This computation is assumed to be done in $\mathsf{prevList}(sortedV, v_{this})$ 
on \mbox{Line 6} which is straightforwardly polynomial-time computable.    
We then assign $\mathsf{prevList}(sortedV, v_{this})$ to $label(v, v_{this})$. 
%For each immediately preceding node $v_{prev} \in \mathsf{prevList}(sortedV, v_{this})$, %we look up the 
%length value of $v_{prev}$ in $lenDist$. 
%Since $lenDict.get(v)$ is 0 and since we assume the weight of $- 1$ on each edge, 
%$lenDict.get(v_{prev}) \leq minLen$ (\mbox{Line 8}) means that the reverse path to arrive at $v_{this}$ from $v_{prev}$ 
%has a strictly lower traversal cost than $minLen$. Hence, we update $minLen$, update the value for the key $v_{this}$ 
%in $lenDict$ to the new $minLen$, and set $label(v, v_{this})$ to $\{v_{prev}\}$ (\mbox{Line 9}). 
%In case $label(v, v_{this})$ already had some set of nodes, it is replaced by $\{v_{prev}\}$. 
%If, instead, $lenDict.get(v_{prev})$ is $minLen + 1$ (\mbox{Line 11}), then the reverse path to arrive at $v_{this}$ from $v_{prev}$ 
%requires just as low cost as from some other preceding nodes. Hence, we update $label(v, v_{this})$ 
%by adding $v_{prev}$ to the current $label(v, v_{this})$. 
The index value is updated (\mbox{Line 8}) and 
this process is repeated until we obtain the required result (\mbox{Line 11}). It is then straightforward
to see that this algorithm runs in polynomial time. \hfill{\it Q.E.D.} \\

\noindent Further, InnerTheorem \ref{thm_unique_traversal_of_edge} implies that transformation of a DAG 
into a $\leq_{\mathsf{DAG}}$-minimal DAG  is efficient. 
\begin{innerlemma}[Polynomial-time transformation of a DAG into a $\leq_{\mathsf{DAG}}$-minimal DAG] \label{lem_polynomial_time_transformation_dag_minimal_dag}
      Given a DAG $(V, E) \in (\mathcal{G}^+ \backslash \mathcal{G})$, 
	let $(V, E')$ be a DAG satisfying $(V, E') \leq_{\mathsf{DAG}} (V, E)$, 
	then $(V, E')$ is polynomial-time computable.  
\end{innerlemma} 
\textbf{Inner-proof.} 
By InnerTheorem \ref{thm_unique_traversal_of_edge}, it suffices to 
remove from $E$ any edge $(v_1, v_2) \in E$ when 
there are multiple paths from $v_1$ to $v_2$.  
The presence of an additional path 
from $v_1$ to $v_2$ is identified by the breadth-first search 
from $v_1$ which runs in polynomial-time. There are only 
$|E|$ edges in $(V, E)$ to test. \hfill{\it Q.E.D.} \\

\noindent By InnerLemma \ref{lem_polynomial_time_transformation_dag_minimal_dag}, 
a $\leq_{\mathsf{DAG}}$-minimal DAG $(V, E')$ of $(V, E)$ 
is polynomial-time obtainable. By InnerLemma \ref{lem_polynomial_time_transformation_minimal_dag_labelled_dag}, the labelled DAG $((V, E'), label)$ of $(V, E')$ is polynomial-time obtainable. By InnerDefinition 
\ref{def_partial_order} and Theorem \ref{thm_unique_traversal_of_edge}, 
the labelled DAG of $(V, E)$ is $((V, E), label)$. \hfill$\Box$ \\

\noindent \textbf{Proof of Proposition \ref{prop_decide_membership_deps}.}
\phantomsection \label{proof_decide_membership_deps}
$((V, E), expand, label)$ with $((V, E), expand)$ 
being $dag(G_{t, trel})$ and \linebreak $((V, E), label)$ being the labelled DAG
of $(V, E)$ is polynomial-time constructed; see  
Lemmas \ref{lem_polynomialtime_transformation_g_dagg} and \ref{lem_polynomial_time_transformation_dag_labelled_dag}.  
%constructing the labelled DAG $((V, E)for $t$ and $trel$ is in P. 
%Let us assume a dynamic set $S_{\star}$ which contains all the members of $S$ and nothing else. 
%If $S_{\star}$ does not contain any terminal node of the labeleld DAG, then there is no $s \in Stmt[G]$ 
%such that $S$ is a member of $DepS[s,t,trel]$. Otherwise, we remove 

The given decision problem is answered in Algorithm \ref{algo_deps}. 
$G_{t, trel}$, $S$, and the $((V, E), expand, label)$ 
form the input to it. 
{\small 
\begin{algorithm}[h]
	\SetAlgoLined
	\SetKwInOut{Input}{input} 
	\SetKwInOut{Output}{output}  
	\Input{$G_{t, trel}$, $S$ and $((V, E), expand, label)$} 
	\Output{true or false} 
	\BlankLine    
	\If{$S = \emptyset$ or $S \not\subseteq Stmt[G_{t, trel}]$}{ \Return false;} 
			$\mathsf{init}(V', \bigcup_{s \in S}condense_{G_{t,trel}}(s))$;
		%\tcp*{condensation is over $G_{\{t\},trel}$}   
		$\mathsf{init}(V_{termi}, \text{the set of all the terminal nodes 
		in } V)$; \\
	\If{$S$ \text{ is not closed under reverse path traversal in } $G_{t, trel}$}{ 
	      \Return{false}; 
	   }
%	\If{$V' \cap V_{termi} == \emptyset$}{ \Return{false}; }
	%     } 
%	$v'' \gets \text{a member of } V' \cap V_{termi}$; \\
	%$\mathsf{init}(v'', V' \cap V_{termi})$; \\
	\While{$V' \not= \emptyset$}{ 
		\If{$|V' \cap V_{termi}| \not= 1$}{
			\Return{false}; 
		}
		$v'' \gets \text{a member of } V' \cap V_{termi}$; \\
		$V_{termi} \gets \bigcup_{v \in V'} label(v, v'')$;     
	     $V' \gets V' \backslash \{v''\}$; \\  
	     	}
	\Return{true}; 
	\caption{Algorithm to decide that $S$ is 
	a depth-statements-set with respect to $t$ for $trel$. $\mathsf{init}$ initialises.} 
	%$\mathsf{terminal}$ obtains terminal nodes.}
	\label{algo_deps}
\end{algorithm}
}
To explain the algorithm in detail,  
since a depth-statements-set of $Stmt[G_{t, trel}]$ 
is by definition a non-empty subset of $Stmt[G_{t, trel}]$, 
it first checks if 
$S$ is non-empty and is in $Stmt[G_{t,trel}]$ (Line 1). If not, $S$ cannot be a depth-statements-set with respect to $t$ 
for $trel$, so the algorithm rejects the input (Line 2).  
%it copies $\bigcup_{s \in S} condense(s)$ to $V'$ and the set of all the terminal nodes of $V$ to $V_{termi}$ 
%(Line 1), and 
It then tests if $S$ is closed under reverse path traversal in $G_{t, trel}$  
(Line 5). For this test, it suffices to obtain $V' = \bigcup_{s \in S} condense_{G_{t,trel}}(s)$  (Line 4) and then 
check the following: for each $v' \in V'$, if $v'$ is in $V$, then $expand(v')$ is a subset of  $S$. 
This is polynomial-time checkable. If $S$ fails this check, $S$ cannot be a member of $DepS[s,t,trel]$, 
so the algorithm rejects the input (Line 6).  
Otherwise, %assume a dynamic set $V'_{\star}$ containing all the members of $V'$ and nothing else.
it checks whether $V'$ has one and just one member that is a terminal node in $(V, E)$ (Line 9). The set of terminal nodes 
of $V$, $V_{termi}$, is already available (Line 4). This check is polynomial-time 
doable. 

*** If there is no such member in $V'$ or if there are more than one such member in $V'$, then 
$S$ cannot be a member of $DepS[s,t,trel]$ and the input is rejected (Line 10). 
Otherwise, it calls the solo member of $V' \cap V_{termi}$ as $v''$ (Line 12).  
This search is polynomial-time doable. 
It then obtains $\bigcup_{v \in V'} label(v, v'')$ (which contains all the nodes that precede $v''$ in $V'$), 
assigning it to $V_{termi}$ and remove $v''$ off $V'$ (Line 13). 
At this point, $V'$ being empty means that $\bigcup_{s \in S}condense_{G_{t,trel}}(s)$ is one of the reverse paths from 
$v''$ to a terminal node in $(V, E)$ 
and so the input is accepted (Line 15). 
If $V'$ is not empty, then $V'$ needs to have a member that is in $V_{termi}$ (Line 9). 

We go to *** above and repeat the process until the input is rejected or accepted. As $V'$ decreases 
in size by 1 at each iteration, the number of iterations is bounded polynomially by $|\bigcup_{s \in S}condense_{G_{t,trel}}(s)|$. 
\hfill$\Box$\\ 

\noindent \textbf{Proof of Proposition \ref{lem_deciding_subtest_and_subtestall}.}
\phantomsection \label{proof_deciding_subtest_and_subtestall}
Checking whether $D_1$ is non-empty and whether $s_1 \not= s_2$ holds is polynomial-time doable. 
Once this check is done, we need to show 
$(\bigwedge D_1 \not= \bigwedge \mathcal{I}(\{t\}, s_1)) \vee   
(\bigwedge \mathcal{I}(\{t\}, s_1) \not= \bigwedge \mathcal{I}(\{t\}, s_2))
\vee (D_1 \not\subseteq \mathcal{I}(\{t\}, s_2))$.  By Lemmas
\ref{lem_np_problems} and \ref{lem_co_np_problems} 
(\ref{lem_co_np_problems} for $D_1 \not\subseteq \mathcal{I}(\{t\}, s_2)$), 
\mbox{SUBTEST} = \mbox{NP}. 
Choosing a non-empty subset $D_2$ of $D_1$ is polynomial-time 
doable. Checking whether 
$(\bigwedge D_2 = \bigwedge \mathcal{I}(\{t\}, s_1)) 
\wedge (\mathcal{I}(\{t\}, s_1) = \bigwedge \mathcal{I}(\{t\}, s_2)) 
\wedge (D_2 \subseteq \mathcal{I}(\{t\}, s_2))$  
is a tautology %is to check whether 
%$\neg ((\bigwedge D_2 = \bigwedge \mathcal{I}(\{t\}, s_1)) 
%\wedge (\mathcal{I}(\{t\}, s_1) = \bigwedge \mathcal{I}(\{t\}, s_2)) 
%\wedge (D_2 \subseteq \mathcal{I}(\{t\}, s_2)))$ is satisfiable 
is \mbox{co-N}P by Lemma \ref{lem_co_np_problems}. 
Now, suppose a bit sequence $\{0,1\}^m$ where $m$ is an integer bounded by a polynomial 
of the length of the representation of $\taam$ but 
is otherwise no smaller than any $|\mathcal{I}(\{t_x\}, s)|$ for any $t_x \in \pmb{Themes}$ and 
$s \in Stmt[G]$. Then, every subset of $D_1$ is identified 
uniquely by the first $|D_1|$ bits 
of the bit sequence. Since we only need to find 
just one non-empty subset $D_2$ of $D_1$ satisfying 
the required conditions, 
\mbox{SUBTESTSUCSOME} is in NP(co-NP) = NP(NP) by Lemma 
\ref{lem_subsumption_complexity_classes}. \hfill$\Box$\\

\pagebreak
\section*{{\hyperref[text_implementation_detail]{Appendix: implementation, gathered data and 
run-time evaluation}}} \label{app_implementation_detail}
The theme aspect argumentation model and all the constraints on the typed 
argumentation graph as well as on the theme aspect argumentation model 
are implemented, available in \citep{Nakai22}. 
The code was written in \mbox{Python 3.10.0}. 
%, which are, together with all the collected data, 
%available in 
We acknowledge the use of dill \citep{McKerns10,McKerns12} in our code.  
The following data were collected: theme aspect argumentation 
models for Examples \ref{ex_role_tr} through \ref{ex_role_fad}; 
and 118,073 randomly generated theme aspect argumentation models. 
The data were collected on %two different MacBook Pros.   
%\begin{itemize} 
%	\item A 16-inch MacBook Pro (2021), Apple M1 Max, 
%64 GB Unified Memory. We refer to this MacBook Pro by Mac1. 
	%\item 
		a 14-inch MacBook Pro (2021), Apple M1 Pro, 16 GB Unified Memory.  
		We refer to this MacBook Pro by Mac. 

%In this section, %we consolidate theoretical 
%results in the p
%We obtain the mean and the maximum run times. 
%for the constraints, confirming that 
%the actual run times align decently to the theoretical expectations. 
%The implementation and the collected data 
%are all available in \citep{Nakai22}. %All the examples in 
%Sections 
%\ref{sec_core_constraints_of_theme_aspect_argumentation_model} and 
%\ref{sec_more_constraints} 
%are also in \citep{Nakai22}, as we mentioned earlier. 
%and at hackmd \mbox{\url{https://hackmd.io/tws_mNTAR9eEzi7_YIs3Sg}}. 
%We acknowledge the use of dill in our implementation \citep{McKerns10,McKerns12}.
%Of course, we cannot implement an infinite structure in practice. Our implementation 
%enforces the following conditions. 
%\begin{itemize} 
%	\item \ryuta{And the conditions.} 
%\end{itemize}  
%\subsection*{Theme aspect argumentation models for Examples  
%for the constraints} 
%Concrete theme aspect argumentation models for 
%Examples \ref{ex_role_tr} through \ref{ex_role_fad} 
%are in \citep{Nakai22}.  
\subsection*{On the random generation of theme aspect argumentation models} 
%for  } 
Each of the theme aspect argumentation models was generated 
randomly but in a way that 
all the conditions in Table \ref{tbl_run_conditions} are met. 
%\end{itemize} 
Theme aspect argumentation models $\taam$ used in the runs are generated randomly but 
in a way that all the conditions 
in Table \ref{tbl_run_conditions} are satisfied. $f$ was determined empirically so  
$G$ would not be too dense most of the time. 
%$|\bigcup_{s \in Stmt[G]} condense_G(s)|$, {\it i.e.} the number of 
%strongly connected components in $G$, will not end up in too small a number (1, in particular) most of the times.  \\
%On Mac1 316,466 $\taam$s were, and on Mac2 235,556 $\taam$s were, 
In total, 118,073 theme aspect argumentation models were 
generated. For every theme aspect argumentation model $\taam$ so 
generated and for every constraint $\pmb{x}$, (un)satisfaction of $\pmb{x}$ 
and the run time were recorded.  

\begin{table*}[!t]
	\resizebox{\textwidth}{!}{
\begin{tabular}{@{}rccccccccc@{}}\toprule
	& $|\pmb{Themes}|$ & $|Stmt[G]|$ & $|Rel[G]|$ & $|OStmt[G]|$ & $|PStmt[G]|$ & $\log(|\pmb{Atom}(\mathcal{D})|)$ & 
	$|\Pi(s)|$ & $|\mathcal{I}(T, s)|$ & $\mathcal{I}(T, \omega)$ \\ \midrule   
%	Mac1 & $1 \sim 4$ & $1 \sim 10$ & $f(|Stmt[G]|)$ & $y$ & $|Stmt[G]| - y$ & $1 \sim 2$ & $1 \sim 4$ & $0 \sim 4$ & same for $\mathcal{I}(T, s)$
%	($\approx$ 50\%) or\\  
%	&&&&&&&&& a bit pattern ($\approx$ 50\%) \\
	Mac & $1 \sim 4$ & $1 \sim 10$ & $f(|Stmt[G]|)$ & $y$ & $|Stmt[G]| - y$ & $1 \sim 3$ & $1 \sim 4$ & $0 \sim 4$ & same for $\mathcal{I}(T, s)$
	($\approx$ 50\%) or\\
	&&&&&&&&& a bit pattern ($\approx$ 50\%) \\
%	\midrule
%	$\mathfrak{p}$-Sat: & $\mathsf{\Pi}_1$ & $\mathsf{\Pi}_1$ & P & $\mathsf{\Pi}_1$ & $\mathsf{\Pi}_1$ & $\mathsf{\Pi}_1$ 
%	& $\mathsf{\Pi}_1$ & $\mathsf{\Pi}_1$ & $\mathsf{\Pi}_1$ & 
%	$\mathsf{\Pi}_1$ & $\mathsf{\Pi}_1$ & $\mathsf{\Pi}_1$ & $\mathsf{\Pi}_1$ & $\mathsf{\Pi}_2$-hard & $\mathsf{\Pi}_1$-hard\\
%	1-$\mathfrak{p}$-Sat: &  P  & P & P &
%	P & P & P & P & P & P  
%	& P & P & P & P & $\mathsf{\Pi}_2$-hard & $\mathsf{\Pi}_1$-hard \\
\end{tabular}  
	}
	\caption{The conditions satisfied by $\taam$s used in the experiment for Mac. %1 and Mac2. 
	``$n \sim m$'' means some integer selected from the range. 
	$f(n) = \lfloor n^2 \cdot r_1 \cdot r_2^{0.3} \rfloor$ where each of $r_1$ and $r_2$ is a randomly selected
	number between 0 and 1 inclusive. 
	$y$ is a randomly selected integer between 0 and $|Stmt[G]|$ inclusive.   
%	$|\pmb{Atom}(\mathcal{D})|$ is a randomly selected integer from $\{2,4\}$ for Mac1 and from $\{2,4,8\}$ for Mac2.
	For each $s \in Stmt[G]$, $|\Pi(s)|$ is an integer between 1 and 4 inclusive. 
	For each $T \subseteq \pmb{Themes}$ and each $s \in Stmt[G]$, $|\mathcal{I}(T, s)|$ is an 
	integer between 0 and 4 inclusive. For each 
	$T \subseteq \pmb{Themes}$, there is 
	a 50\% chance that $\mathcal{I}(T, \omega)$   
	gets assigned $x \in \{0,1,2,3,4\}$ Boolean formulas 
	and a 50\% chance that it is a bit pattern. 
	 }
	\label{tbl_run_conditions}  
\end{table*}

\subsubsection{The run time}   
\begin{table*} 
\resizebox{5.1in}{!}{ 
\begin{tabular}{lllrrrrrrrrrr}
\toprule
  &   &   & \multicolumn{2}{l}{tr\_time} & \multicolumn{2}{l}{nnp\_time} & \multicolumn{2}{l}{nsa\_time} & \multicolumn{2}{l}{kos\_time} & \multicolumn{2}{l}{nss\_time} \\
  &   &   &     mean &      max &     mean &      max &     mean &      max &     mean &      max &     mean &      max \\
	$|Stmt[G]|$ & $|\pmb{Themes}|$ & $\log(|\pmb{Atom}(\mathcal{D})|)$ &          &          &          &          &          &          &          &          &          &          \\
\midrule
3 & 1 & 1 & 9.41e-06 & 2.60e-05 & 5.16e-06 & 1.69e-05 & 2.39e-06 & 7.87e-06 & 2.90e-06 & 8.82e-06 & 4.42e-06 & 3.00e-05 \\
  &   & 2 & 9.43e-06 & 2.12e-05 & 5.19e-06 & 1.79e-05 & 2.46e-06 & 1.00e-05 & 2.90e-06 & 8.82e-06 & 4.38e-06 & 2.98e-05 \\
  &   & 3 & 9.81e-06 & 2.91e-05 & 5.27e-06 & 1.12e-05 & 2.48e-06 & 6.20e-06 & 2.98e-06 & 1.10e-05 & 4.58e-06 & 2.69e-05 \\
  & 2 & 1 & 6.23e-06 & 1.62e-05 & 4.31e-06 & 1.10e-05 & 2.40e-06 & 4.05e-06 & 2.32e-06 & 2.48e-05 & 4.53e-06 & 3.10e-05 \\
  &   & 2 & 6.36e-06 & 2.00e-05 & 4.39e-06 & 9.06e-06 & 2.45e-06 & 5.01e-06 & 2.32e-06 & 5.96e-06 & 3.91e-06 & 2.81e-05 \\
  &   & 3 & 7.39e-06 & 2.29e-05 & 4.85e-06 & 4.12e-05 & 2.63e-06 & 5.96e-06 & 2.47e-06 & 6.91e-06 & 4.26e-06 & 2.41e-05 \\
  & 3 & 1 & 6.55e-06 & 2.10e-05 & 5.07e-06 & 1.31e-05 & 2.38e-06 & 5.96e-06 & 1.74e-06 & 1.79e-05 & 4.23e-06 & 4.41e-05 \\
  &   & 2 & 7.08e-06 & 2.19e-05 & 5.31e-06 & 2.00e-05 & 2.49e-06 & 7.87e-06 & 1.80e-06 & 1.00e-05 & 4.48e-06 & 3.19e-05 \\
  &   & 3 & 8.13e-06 & 4.32e-05 & 5.81e-06 & 1.31e-05 & 2.64e-06 & 1.10e-05 & 1.85e-06 & 1.22e-05 & 4.44e-06 & 3.50e-05 \\
6 & 1 & 1 & 1.84e-05 & 5.08e-05 & 5.57e-06 & 1.00e-05 & 2.68e-06 & 1.79e-05 & 3.36e-06 & 9.06e-06 & 6.41e-06 & 4.60e-05 \\
  &   & 2 & 1.93e-05 & 5.41e-05 & 5.89e-06 & 9.30e-06 & 2.71e-06 & 5.25e-06 & 3.48e-06 & 7.87e-06 & 5.53e-06 & 4.77e-05 \\
  &   & 3 & 1.92e-05 & 4.70e-05 & 5.87e-06 & 1.22e-05 & 2.76e-06 & 8.11e-06 & 3.49e-06 & 7.15e-06 & 6.66e-06 & 4.72e-05 \\
  & 2 & 1 & 8.48e-06 & 3.67e-05 & 6.22e-06 & 1.31e-05 & 2.67e-06 & 5.01e-06 & 3.41e-06 & 8.11e-06 & 5.79e-06 & 5.48e-05 \\
  &   & 2 & 8.33e-06 & 3.91e-05 & 6.37e-06 & 1.31e-05 & 2.78e-06 & 5.96e-06 & 3.44e-06 & 7.15e-06 & 5.92e-06 & 3.79e-05 \\
  &   & 3 & 1.00e-05 & 3.60e-05 & 6.96e-06 & 1.53e-05 & 2.96e-06 & 7.15e-06 & 3.63e-06 & 1.62e-05 & 6.42e-06 & 6.10e-05 \\
  & 3 & 1 & 7.42e-06 & 3.58e-05 & 6.83e-06 & 2.19e-05 & 2.72e-06 & 6.91e-06 & 3.40e-06 & 9.06e-06 & 5.67e-06 & 5.89e-05 \\
  &   & 2 & 8.27e-06 & 3.60e-05 & 7.20e-06 & 1.69e-05 & 2.86e-06 & 5.96e-06 & 3.48e-06 & 7.15e-06 & 6.84e-06 & 4.98e-05 \\
  &   & 3 & 9.61e-06 & 3.91e-05 & 8.10e-06 & 1.60e-05 & 3.03e-06 & 1.62e-05 & 3.85e-06 & 1.10e-05 & 6.25e-06 & 6.29e-05 \\
9 & 1 & 1 & 3.50e-05 & 9.11e-05 & 8.05e-06 & 1.38e-05 & 2.86e-06 & 5.25e-06 & 4.94e-06 & 9.06e-06 & 7.11e-06 & 6.70e-05 \\
  &   & 2 & 3.63e-05 & 1.02e-04 & 8.12e-06 & 1.69e-05 & 2.86e-06 & 5.25e-06 & 4.92e-06 & 1.29e-05 & 6.90e-06 & 5.67e-05 \\
  &   & 3 & 3.94e-05 & 1.07e-04 & 8.55e-06 & 1.57e-05 & 2.92e-06 & 5.96e-06 & 5.06e-06 & 9.06e-06 & 6.93e-06 & 6.51e-05 \\
  & 2 & 1 & 8.55e-06 & 5.89e-05 & 7.66e-06 & 1.41e-05 & 2.91e-06 & 7.15e-06 & 4.05e-06 & 1.91e-05 & 7.46e-06 & 7.30e-05 \\
  &   & 2 & 9.59e-06 & 5.82e-05 & 7.88e-06 & 1.69e-05 & 2.98e-06 & 7.15e-06 & 4.18e-06 & 9.06e-06 & 7.50e-06 & 8.18e-05 \\
  &   & 3 & 1.09e-05 & 4.20e-05 & 9.07e-06 & 1.88e-05 & 3.22e-06 & 7.15e-06 & 4.42e-06 & 1.22e-05 & 7.10e-06 & 7.30e-05 \\
  & 3 & 1 & 7.82e-06 & 3.41e-05 & 7.87e-06 & 1.62e-05 & 2.90e-06 & 6.20e-06 & 3.92e-06 & 8.82e-06 & 7.06e-06 & 6.60e-05 \\
  &   & 2 & 8.91e-06 & 5.60e-05 & 8.83e-06 & 2.03e-05 & 3.08e-06 & 9.78e-06 & 4.09e-06 & 1.03e-05 & 8.09e-06 & 7.99e-05 \\
  &   & 3 & 1.02e-05 & 4.41e-05 & 9.99e-06 & 2.79e-05 & 3.36e-06 & 8.11e-06 & 4.26e-06 & 9.06e-06 & 8.74e-06 & 7.30e-05 \\
\bottomrule
\end{tabular}
}{\ }\\
\resizebox{6.3in}{!}{
\begin{tabular}{lllrrrrrrrrrrrrrrrr}
\toprule
  &   &   & \multicolumn{2}{l}{aass\_time} & \multicolumn{2}{l}{i\_time} & \multicolumn{2}{l}{vi\_time} & \multicolumn{2}{l}{bat\_time} & \multicolumn{2}{l}{pr\_time} & \multicolumn{2}{l}{mat\_time} & \multicolumn{2}{l}{manss\_time} & \multicolumn{2}{l}{ss\_time} \\
  &   &   &      mean &      max &     mean &      max &     mean &      max &     mean &      max &     mean &      max &     mean &      max &       mean &      max &     mean &      max \\
	$|Stmt[G]|$ & $|\pmb{Themes}|$ & $\log(|\pmb{Atom}(\mathcal{D})|)$ &           &          &          &          &          &          &          &          &          &          &          &          &            &          &          &          \\
\midrule
3 & 1 & 1 &  2.19e-04 & 3.28e-03 & 1.30e-04 & 1.25e-03 & 2.06e-06 & 8.11e-06 & 2.00e-04 & 9.46e-03 & 2.64e-04 & 2.24e-03 & 7.98e-05 & 1.23e-03 &   2.59e-04 & 1.92e-03 & 2.97e-04 & 1.95e-03 \\
  &   & 2 &  9.29e-04 & 1.68e-02 & 5.87e-04 & 4.49e-03 & 2.19e-06 & 1.26e-05 & 8.03e-04 & 8.06e-03 & 1.79e-03 & 1.46e-02 & 5.16e-04 & 8.64e-03 &   1.77e-03 & 1.11e-02 & 2.72e-03 & 1.24e-02 \\
  &   & 3 &  3.15e-03 & 8.08e-02 & 2.04e-03 & 1.40e-02 & 2.22e-06 & 5.25e-06 & 2.85e-03 & 1.44e-02 & 7.48e-03 & 7.19e-02 & 1.91e-03 & 2.69e-02 &   7.33e-03 & 4.61e-02 & 1.46e-02 & 6.06e-02 \\
  & 2 & 1 &  1.84e-04 & 2.86e-03 & 1.39e-04 & 7.15e-04 & 2.11e-06 & 7.87e-06 & 2.49e-04 & 8.89e-03 & 3.45e-04 & 3.28e-03 & 1.28e-04 & 1.80e-03 &   2.73e-04 & 1.65e-03 & 2.88e-04 & 2.70e-03 \\
  &   & 2 &  7.82e-04 & 1.36e-02 & 5.84e-04 & 4.88e-03 & 2.18e-06 & 5.25e-06 & 1.09e-03 & 6.60e-03 & 2.67e-03 & 2.29e-02 & 6.04e-04 & 7.74e-03 &   2.16e-03 & 1.08e-02 & 3.06e-03 & 1.23e-02 \\
  &   & 3 &  2.32e-03 & 4.52e-02 & 2.30e-03 & 1.35e-02 & 2.26e-06 & 6.91e-06 & 4.01e-03 & 1.37e-02 & 1.14e-02 & 9.97e-02 & 2.70e-03 & 2.65e-02 &   8.65e-03 & 4.35e-02 & 1.73e-02 & 6.43e-02 \\
  & 3 & 1 &  1.77e-04 & 2.84e-03 & 1.36e-04 & 9.41e-04 & 2.10e-06 & 1.41e-05 & 3.22e-04 & 1.07e-02 & 4.82e-04 & 4.69e-03 & 1.28e-04 & 2.14e-03 &   1.27e-04 & 1.86e-03 & 3.10e-04 & 2.49e-03 \\
  &   & 2 &  7.44e-04 & 1.80e-02 & 5.59e-04 & 5.54e-03 & 2.18e-06 & 5.25e-06 & 1.30e-03 & 4.88e-02 & 3.23e-03 & 3.73e-02 & 6.45e-04 & 8.56e-03 &   8.85e-04 & 1.55e-02 & 3.40e-03 & 2.04e-02 \\
  &   & 3 &  2.42e-03 & 5.79e-02 & 2.14e-03 & 1.31e-02 & 2.28e-06 & 7.15e-06 & 4.55e-03 & 1.57e-02 & 1.42e-02 & 1.82e-01 & 2.54e-03 & 4.40e-02 &   3.16e-03 & 4.98e-02 & 2.03e-02 & 1.15e-01 \\
6 & 1 & 1 &  3.01e-04 & 6.05e-03 & 1.29e-04 & 1.75e-03 & 2.06e-06 & 5.01e-06 & 1.82e-04 & 8.39e-03 & 4.88e-04 & 2.46e-03 & 7.84e-05 & 1.26e-03 &   4.08e-04 & 2.13e-03 & 2.89e-04 & 1.70e-03 \\
  &   & 2 &  1.54e-03 & 2.62e-02 & 5.66e-04 & 4.95e-03 & 2.20e-06 & 5.25e-06 & 7.97e-04 & 4.23e-03 & 3.77e-03 & 1.96e-02 & 5.23e-04 & 6.33e-03 &   3.29e-03 & 1.41e-02 & 3.58e-03 & 1.70e-02 \\
  &   & 3 &  4.83e-03 & 7.03e-02 & 2.17e-03 & 2.14e-02 & 2.25e-06 & 5.01e-06 & 2.77e-03 & 1.31e-02 & 1.60e-02 & 9.21e-02 & 1.88e-03 & 2.35e-02 &   1.43e-02 & 6.05e-02 & 2.29e-02 & 7.41e-02 \\
  & 2 & 1 &  2.46e-04 & 3.74e-03 & 1.51e-04 & 1.06e-03 & 2.13e-06 & 9.78e-06 & 2.94e-04 & 1.03e-02 & 4.23e-04 & 3.74e-03 & 1.14e-04 & 1.33e-03 &   4.00e-04 & 2.00e-03 & 3.15e-04 & 2.38e-03 \\
  &   & 2 &  1.15e-03 & 2.87e-02 & 6.14e-04 & 5.81e-03 & 2.19e-06 & 4.05e-06 & 1.09e-03 & 6.27e-03 & 3.11e-03 & 2.94e-02 & 6.18e-04 & 7.78e-03 &   3.32e-03 & 2.09e-02 & 4.10e-03 & 2.36e-02 \\
  &   & 3 &  3.46e-03 & 6.82e-02 & 2.25e-03 & 1.34e-02 & 2.29e-06 & 6.91e-06 & 3.86e-03 & 1.46e-02 & 1.41e-02 & 1.59e-01 & 2.27e-03 & 2.58e-02 &   1.38e-02 & 7.11e-02 & 2.69e-02 & 9.23e-02 \\
  & 3 & 1 &  2.33e-04 & 2.98e-03 & 1.38e-04 & 1.18e-03 & 2.10e-06 & 5.25e-06 & 3.18e-04 & 9.72e-03 & 3.83e-04 & 4.09e-03 & 1.32e-04 & 2.63e-03 &   4.18e-04 & 2.28e-03 & 3.17e-04 & 2.14e-03 \\
  &   & 2 &  1.16e-03 & 1.78e-02 & 6.01e-04 & 6.03e-03 & 2.21e-06 & 1.41e-05 & 1.27e-03 & 7.04e-03 & 2.51e-03 & 2.12e-02 & 6.45e-04 & 7.25e-03 &   2.99e-03 & 1.77e-02 & 3.97e-03 & 2.20e-02 \\
  &   & 3 &  3.24e-03 & 6.38e-02 & 2.09e-03 & 1.44e-02 & 2.29e-06 & 6.91e-06 & 4.53e-03 & 1.35e-02 & 1.05e-02 & 7.66e-02 & 2.45e-03 & 2.80e-02 &   1.27e-02 & 6.92e-02 & 3.12e-02 & 1.43e-01 \\
9 & 1 & 1 &  3.53e-04 & 8.93e-03 & 1.51e-04 & 2.38e-03 & 2.06e-06 & 5.01e-06 & 2.31e-04 & 9.09e-03 & 5.90e-04 & 4.47e-03 & 8.84e-05 & 1.49e-03 &   5.83e-04 & 2.74e-03 & 3.24e-04 & 1.85e-03 \\
  &   & 2 &  1.95e-03 & 2.98e-02 & 5.54e-04 & 3.60e-03 & 2.20e-06 & 7.15e-06 & 7.68e-04 & 6.47e-03 & 4.83e-03 & 3.08e-02 & 4.75e-04 & 6.01e-03 &   4.78e-03 & 1.93e-02 & 3.87e-03 & 1.70e-02 \\
  &   & 3 &  5.31e-03 & 8.80e-02 & 2.20e-03 & 1.32e-02 & 2.28e-06 & 5.01e-06 & 2.94e-03 & 2.27e-02 & 1.98e-02 & 1.48e-01 & 2.29e-03 & 2.45e-02 &   2.02e-02 & 9.66e-02 & 2.78e-02 & 1.03e-01 \\
  & 2 & 1 &  2.86e-04 & 4.70e-03 & 1.50e-04 & 1.80e-03 & 2.08e-06 & 5.96e-06 & 3.47e-04 & 9.03e-03 & 5.58e-04 & 3.85e-03 & 1.38e-04 & 2.11e-03 &   6.13e-04 & 2.57e-03 & 3.01e-04 & 2.47e-03 \\
  &   & 2 &  1.37e-03 & 2.42e-02 & 5.70e-04 & 3.28e-03 & 2.23e-06 & 5.01e-06 & 1.06e-03 & 6.74e-03 & 3.94e-03 & 2.48e-02 & 5.66e-04 & 6.71e-03 &   4.79e-03 & 1.99e-02 & 4.08e-03 & 2.20e-02 \\
  &   & 3 &  4.15e-03 & 8.10e-02 & 2.06e-03 & 1.20e-02 & 2.28e-06 & 7.15e-06 & 3.79e-03 & 1.42e-02 & 1.62e-02 & 9.90e-02 & 2.18e-03 & 2.39e-02 &   1.92e-02 & 8.29e-02 & 3.40e-02 & 1.36e-01 \\
  & 3 & 1 &  2.46e-04 & 2.82e-03 & 1.33e-04 & 1.23e-03 & 2.05e-06 & 1.62e-05 & 3.18e-04 & 1.06e-02 & 4.91e-04 & 3.38e-03 & 1.27e-04 & 3.37e-03 &   5.69e-04 & 2.47e-03 & 3.10e-04 & 3.38e-03 \\
  &   & 2 &  1.26e-03 & 2.85e-02 & 6.08e-04 & 5.70e-03 & 2.22e-06 & 5.96e-06 & 1.25e-03 & 4.48e-03 & 3.49e-03 & 3.59e-02 & 6.60e-04 & 9.13e-03 &   4.60e-03 & 2.06e-02 & 4.23e-03 & 2.61e-02 \\
  &   & 3 &  3.50e-03 & 6.48e-02 & 1.99e-03 & 1.22e-02 & 2.29e-06 & 7.15e-06 & 4.59e-03 & 3.70e-02 & 1.51e-02 & 1.09e-01 & 2.35e-03 & 2.85e-02 &   1.92e-02 & 9.06e-02 & 3.61e-02 & 1.52e-01 \\
\bottomrule
\end{tabular}
}
{\ }\\
\resizebox{6.3in}{!}{
\begin{tabular}{lllrrrrrrrrrrrrrr}
\toprule
  &   &   & \multicolumn{2}{l}{esr\_time} & \multicolumn{2}{l}{ensr\_time} & \multicolumn{2}{l}{eos\_time} & \multicolumn{2}{l}{das\_time} & \multicolumn{2}{l}{nwci\_time} & \multicolumn{2}{l}{faD\_time} & \multicolumn{2}{l}{faW\_time} \\
  &   &   &     mean &      max &      mean &      max &     mean &      max &     mean &      max &      mean &      max &     mean &      max &     mean &      max \\
	$|Stmt[G]|$ & $|\pmb{Themes}|$ & $\log(|\pmb{Atom}(\mathcal{D})|)$  &          &          &           &          &          &          &          &          &           &          &          &          &          &          \\
\midrule
3 & 1 & 1 & 4.73e-04 & 6.34e-03 &  4.49e-04 & 6.04e-03 & 7.20e-04 & 1.17e-02 & 1.67e-04 & 1.40e-03 &  5.19e-04 & 8.77e-03 & 6.30e-03 & 1.13e-01 & 5.48e-03 & 9.30e-02 \\
  &   & 2 & 3.74e-03 & 5.69e-02 &  4.00e-03 & 5.12e-02 & 7.02e-03 & 9.80e-02 & 7.64e-04 & 9.95e-03 &  3.03e-03 & 4.65e-02 & 1.25e-01 & 2.48e+00 & 1.09e-01 & 2.47e+00 \\
  &   & 3 & 1.64e-02 & 2.15e-01 &  1.84e-02 & 2.31e-01 & 3.15e-02 & 4.15e-01 & 3.43e-03 & 3.97e-02 &  1.31e-02 & 1.81e-01 & 6.58e-01 & 8.90e+00 & 5.74e-01 & 7.83e+00 \\
  & 2 & 1 & 4.24e-04 & 9.84e-03 &  4.35e-04 & 7.92e-03 & 1.21e-03 & 1.04e-02 & 1.91e-04 & 1.17e-03 &  3.10e-04 & 5.04e-03 & 3.43e-03 & 9.57e-02 & 2.83e-03 & 4.54e-02 \\
  &   & 2 & 5.43e-03 & 1.94e-01 &  6.90e-03 & 1.81e-01 & 1.84e-02 & 2.06e-01 & 8.37e-04 & 9.18e-03 &  2.23e-03 & 4.99e-02 & 1.08e-01 & 6.14e+00 & 8.74e-02 & 2.82e+00 \\
  &   & 3 & 3.76e-02 & 5.67e-01 &  4.48e-02 & 7.22e-01 & 1.74e-01 & 1.70e+00 & 3.72e-03 & 2.97e-02 &  9.57e-03 & 1.57e-01 & 4.54e-01 & 1.57e+01 & 3.63e-01 & 8.73e+00 \\
  & 3 & 1 & 1.07e-03 & 1.71e-02 &  2.12e-04 & 1.12e-02 & 4.74e-04 & 1.08e-02 & 2.00e-04 & 1.49e-03 &  4.01e-04 & 7.85e-03 & 4.31e-03 & 1.65e-01 & 3.86e-03 & 1.23e-01 \\
  &   & 2 & 2.06e-02 & 4.51e-01 &  5.02e-03 & 3.66e-01 & 9.90e-03 & 3.04e-01 & 1.03e-03 & 1.39e-02 &  2.61e-03 & 5.61e-02 & 1.78e-01 & 6.31e+00 & 1.57e-01 & 5.90e+00 \\
  &   & 3 & 4.34e-01 & 4.29e+00 &  5.71e-02 & 3.02e+00 & 1.82e-01 & 5.90e+00 & 4.50e-03 & 3.94e-02 &  1.05e-02 & 1.93e-01 & 1.23e+00 & 3.65e+01 & 1.00e+00 & 2.80e+01 \\
6 & 1 & 1 & 2.01e-04 & 3.71e-03 &  5.48e-04 & 6.47e-03 & 1.64e-03 & 1.89e-02 & 2.04e-04 & 1.47e-03 &  6.45e-04 & 1.09e-02 & 7.43e-03 & 2.11e-01 & 6.10e-03 & 6.52e-02 \\
  &   & 2 & 2.05e-03 & 4.57e-02 &  5.83e-03 & 1.22e-01 & 1.89e-02 & 1.98e-01 & 1.18e-03 & 1.01e-02 &  4.80e-03 & 9.91e-02 & 7.61e-01 & 1.13e+01 & 4.13e-01 & 6.36e+00 \\
  &   & 3 & 6.59e-03 & 1.52e-01 &  2.54e-02 & 3.20e-01 & 9.97e-02 & 9.52e-01 & 5.41e-03 & 4.36e-02 &  1.60e-02 & 3.22e-01 & 5.83e+00 & 7.65e+01 & 2.52e+00 & 5.76e+01 \\
  & 2 & 1 & 4.18e-04 & 5.97e-03 &  9.29e-04 & 1.33e-02 & 1.81e-03 & 1.61e-02 & 2.24e-04 & 1.49e-03 &  4.74e-04 & 1.35e-02 & 6.04e-03 & 9.79e-02 & 5.45e-03 & 7.64e-02 \\
  &   & 2 & 6.55e-03 & 1.29e-01 &  1.62e-02 & 2.75e-01 & 3.03e-02 & 3.70e-01 & 1.25e-03 & 8.77e-03 &  3.45e-03 & 1.20e-01 & 5.05e-01 & 2.16e+01 & 3.39e-01 & 7.89e+00 \\
  &   & 3 & 4.14e-02 & 7.34e-01 &  1.09e-01 & 1.82e+00 & 3.07e-01 & 2.96e+00 & 6.56e-03 & 9.67e-02 &  1.27e-02 & 1.77e-01 & 6.14e+00 & 1.36e+02 & 2.94e+00 & 5.52e+01 \\
  & 3 & 1 & 6.58e-04 & 1.09e-02 &  1.06e-03 & 1.03e-02 & 1.61e-03 & 1.37e-02 & 2.40e-04 & 1.66e-03 &  4.66e-04 & 7.73e-03 & 5.09e-03 & 7.86e-02 & 4.76e-03 & 7.70e-02 \\
  &   & 2 & 1.12e-02 & 2.70e-01 &  1.98e-02 & 4.30e-01 & 2.91e-02 & 5.48e-01 & 1.46e-03 & 4.16e-02 &  2.89e-03 & 6.08e-02 & 4.09e-01 & 2.64e+01 & 3.22e-01 & 1.29e+01 \\
  &   & 3 & 1.82e-01 & 3.27e+00 &  3.93e-01 & 4.35e+00 & 6.35e-01 & 7.11e+00 & 6.40e-03 & 5.07e-02 &  1.26e-02 & 1.41e-01 & 8.40e+00 & 2.00e+02 & 4.31e+00 & 9.40e+01 \\
9 & 1 & 1 & 3.08e-04 & 4.97e-03 &  9.51e-04 & 7.39e-03 & 2.25e-03 & 1.99e-02 & 2.09e-04 & 1.55e-03 &  7.81e-04 & 2.08e-02 & 6.11e-03 & 6.45e-02 & 5.95e-03 & 7.37e-02 \\
  &   & 2 & 2.82e-03 & 5.78e-02 &  1.05e-02 & 1.25e-01 & 2.44e-02 & 2.27e-01 & 1.29e-03 & 8.54e-03 &  6.37e-03 & 1.78e-01 & 1.40e+00 & 3.83e+01 & 6.61e-01 & 1.78e+01 \\
  &   & 3 & 1.26e-02 & 2.57e-01 &  5.69e-02 & 7.19e-01 & 1.49e-01 & 1.13e+00 & 5.73e-03 & 3.95e-02 &  2.00e-02 & 7.34e-01 & 2.92e+01 & 2.55e+02 & 9.36e+00 & 1.34e+02 \\
  & 2 & 1 & 4.83e-04 & 7.69e-03 &  1.37e-03 & 2.04e-02 & 2.82e-03 & 3.77e-02 & 2.57e-04 & 1.48e-03 &  5.67e-04 & 1.99e-02 & 6.43e-03 & 9.33e-02 & 6.09e-03 & 8.69e-02 \\
  &   & 2 & 6.89e-03 & 1.31e-01 &  1.89e-02 & 3.08e-01 & 3.56e-02 & 6.92e-01 & 1.37e-03 & 9.20e-03 &  3.54e-03 & 4.82e-02 & 7.98e-01 & 4.86e+01 & 5.57e-01 & 1.66e+01 \\
  &   & 3 & 4.70e-02 & 1.07e+00 &  1.99e-01 & 2.37e+00 & 4.04e-01 & 3.97e+00 & 6.91e-03 & 4.67e-02 &  1.35e-02 & 2.10e-01 & 2.36e+01 & 5.58e+02 & 7.52e+00 & 1.57e+02 \\
  & 3 & 1 & 6.01e-04 & 1.13e-02 &  1.43e-03 & 1.80e-02 & 2.53e-03 & 3.11e-02 & 2.62e-04 & 1.63e-03 &  4.86e-04 & 1.20e-02 & 5.85e-03 & 6.91e-02 & 5.14e-03 & 7.44e-02 \\
  &   & 2 & 1.24e-02 & 4.37e-01 &  2.94e-02 & 5.79e-01 & 3.69e-02 & 5.80e-01 & 1.39e-03 & 1.01e-02 &  3.46e-03 & 6.22e-02 & 4.59e-01 & 1.17e+01 & 4.42e-01 & 1.05e+01 \\
  &   & 3 & 2.04e-01 & 3.38e+00 &  5.21e-01 & 5.40e+00 & 9.28e-01 & 1.25e+01 & 7.07e-03 & 5.88e-02 &  1.21e-02 & 1.71e-01 & 3.40e+01 & 8.31e+02 & 1.14e+01 & 2.60e+02 \\
\bottomrule
\end{tabular}
}
\caption{(\pmb{Top}) The mean and the maximum run time (in seconds) for the constraints on the typed graph. \\ 
(\pmb{Middle}) The mean and the maximum run time (in seconds) for \pmb{Core} constraints. \\
(\pmb{Bottom}) The mean and the maximum run time (in seconds) for the remaining constraints. } 
\label{tbl_run_time_first} 
\end{table*} 
%The specifications of Mac1 and Mac2 are different. 
%For the run time, therefore, we use only the samples collected on Mac2 for 
%fair comparisons. 
Table \ref{tbl_run_time_first} shows the mean and the maximum run times (in seconds) 
for satisfiability checks on 
the constraints. % \ref{tbl_run_time_second} and \ref{tbl_run_time_third}, 
%each of which is further divided into 3 cases: the top sub-table 
%lists the run time of each constraint across all the samples, 
%the middle sub-table lists the run time of each constraint 
%across all the samples that satisfy it, and the bottom sub-table lists the run time 
%of each constraint across all the samples that do not satisfy it. NaN 
%signifies that no occurrence of a sample meeting the condition was obtained during the runs. 
While the table %$\sim$ \ref{tbl_run_time_third} 
shows the data for limited combinations of 
$|Stmt[G]|$, $|\pmb{Themes}|$ and $\log(|\pmb{Atom}(\mathcal{D})|)$, 
the csv file containing the full data is available in \citep{Nakai22}.

As is expected, the satisfiability of the constraints on the typed argumentation graph 
can be checked considerably more efficiently than that of the constraints on the theme aspect argumentation model. 
Indeed, when $(|Stmt[G]|, |\pmb{Themes}|, \log(|\pmb{Atom}(\mathcal{D})|))$ 
is increased from $(3,1,1)$ to $(9,3,3)$, {\it i.e.} 
when it is tripled, the run times for \pmb{tr}, \pmb{nnp}, \pmb{nsa}, \pmb{kos} and \pmb{nss} 
only marginally increase in general. %The run time for \pmb{nss} increases 
%slightly faster but is still kept within the $3t$ bound where $t$ is the run time 
%for $(3,1,1)$. 

By contrast, the only constraint on the theme aspect argumentation model 
for which satisfiability can be checked as efficiently is \pmb{vi}. 
The data shows the following for the others by going to $(9,3,3)$ from $(3,1,1)$, exhibiting a decent alignment to the theoretical expectation obtained in Section \ref{sec_computational_complexities}. 
%even though the run times for \pmb{E} constraints can be suggesting some room for improvement 
%of the implementation. 
% the data shows the following.
\begin{itemize} 
	\item The maximum run time for \pmb{Core} constraints, 
		\pmb{das} and \pmb{nwci} becomes 3.91 times (\pmb{bat}) to 77.9 times (\pmb{ss}) longer and 
		the mean run time for them becomes 15.3 times (\pmb{i}) to 74.1 times (\pmb{manss}) longer. 
	\item For the \pmb{E} constraints,  the maximum run time becomes 533 times (\pmb{esr}) to 1,070 times (\pmb{eos}) 
		longer and the mean run time becomes 431 times (\pmb{esr}) to 1,290 times (\pmb{eos}) longer.    
%		\pmb{E} constraints, \pmb{das} and \pmb{nwci} becomes 
%		9 times (\pmb{i}) to 107 times (\pmb{ss}) longer and, with the exception of \pmb{bat}, the mean run time 
%		becomes 13 times (\pmb{esr}) to 130 times (\pmb{ss}) longer. 
%		at least 
%		9 times longer. 
		%by up to the multiple of $2.086^3 (< 9.08)$ - generally with the exception of \pmb{bat}'s maximum run time.\footnote{The rate of satisfaction of 
%\pmb{bat} is very low which likely led to the lower-biased result.} 
%	\item Among these constraints, \pmb{ss} saw the biggest run time 
%		increase at least by the multiple of $4.747^3 (< 107)$. 
	\item The \pmb{F} constraints fair worse: the maximum run time 
		becomes 2,800 times (\pmb{faW}) to 7,350 times (\pmb{faD}) longer 
		and the mean run time becomes 
		2,080 times (\pmb{faW}) to 5,400 times (\pmb{faD}) longer.  
%		at least by the multiple of $1.380^{3^3} (< 6060)$.  
%	\item For \pmb{Core}, the sum of the maximum run times of 
%	     \pmb{Core} constraints then becomes 24 times longer and 
%		the sum of the mean run times becomes around 17 times longer.  
%	\item For \pmb{All}$^{- \text{\pmb{F}}}$, the sum of the maximum run times of 
%	     \pmb{Core} constraints then becomes 24 times longer and 
%		the sum of the mean run times becomes around 17 times longer.   
%	\item For \pmb{All}, the sum of the maximum run times of 
%	     \pmb{Core} constraints then becomes 24 times longer and 
%		the sum of the mean run times becomes around 17 times longer.   
\end{itemize}  
%While the \pmb{E} constaints appears closer to the \pmb{F} constraints, when we go to $(9,3,3)$ from $(3,1,3)$ instead, 
%that is, when we fix $\log(|\pmb{Atom}(\mathcal{D})|)$ and vary the other parameters, the maximum run time 
%for \pmb{E} constraints becomes 
Summarising the data, we obtain the following conclusions on the run time increases by going to (9,3,3) from (3,1,1). 
\begin{itemize} 
	\item The sum of the maximum run times and that of the mean run times for \pmb{tr}, \pmb{nnp}, \pmb{nsa} and \pmb{kos}  
		become 1.50 times and respectively 1.40 times longer (and 1.81 times and respectively 1.51 times 
		longer for all the 5 constraints on typed argumentation graphs). 
	\item The sum of the maximum run times and that of the mean run times for 
	     \pmb{Core} constraints become 23.2 times and respectively 57.2 times longer. 
	\item The sum of the maximum run times and that of the mean run times for 
	     \pmb{All}$^{- \text{\pmb{F}}}$ constraints become 396 times and respectively 611 times longer. 
	\item The sum of the maximum run times and that of the mean run times for 
	     \pmb{All} constraints become 4,250 times and respectively  3,070 times longer. 
\end{itemize} 
%}

\bibliography{references}  
\section*{Q \& A}   
\begin{enumerate}[label=Q\arabic*]     
    \item In Remark 1, should it not be that Alice's statement is also fallacious 
	    because there perhaps is an unexpressed premise: ``{\it I am an expert 
		on this.}"?  
		\begin{itemize} 
			\item[A:] No. 
				%No unexpressed information may be 
				%attached to expressed information 
				%without changing it.  
				Refer to Remark \ref{note_2}. 
				Further, opinions 
				are opinions and no such premise 
				is stated to exist. 
				Cf. footnote  
				\footref{fnlabel}.
				%They cannot be regarded 
				%as fallacious 
				%for the reason that some  
				%extended version of it may be. 
				If a statement could be judged
				fallacious by identifying 
				one extended and modified version of it that 
				may be fallacious, 
				every statement would be fallacious 
				since it can be extended into 
				a self-contradictory statement. 
%				When I say ``We shall go to the mountain.", 
%				my intention is to go to the mountain. 
%				You may interpret it as ``We shall go to the mountain
%				because it is sunny", ``We shall go to the mountain 
%				because there lives our parents there" 
%				and so on, with any reason 
%				that you may see is plausible. 
%				All these alternative statements, 
%				however, are your interpretations and none 
%				of them are expressed by myself. Indeed, 
%				if one's statement is fallacious 
%				so long as a modified version of 
%				the statement is, any statement by anybody 
%				is fallacious for the same reason.  
%				An interesting question to ask is: 
%				is ``The Eiffel Tower is" a fallacious statement? 
%				Well, it is not even a sentence, but you worry not, 
%				because it is 
%				``The Eiffel Tower is a wrought-iron lattice tower located in Paris, France. It is one of the most recognizable landmarks in the world." 
%				when I passed it to ChatGPT. Now, 
%				this sentence is just as much fallacious 
%				as ``We shall go to the mountain," for  
%				 
%				 
%				
%				 
%				 
%				 
% 
%				Remark 1 does not presuppose the presence 
%				of any statement that has not not expressed, 
%				or the need of a premise. 
%				If Alice states {\it We should buy fire insurance.}, 
%				we may gather that it is Alice's position 
%				that they should buy fire insurance but we 
%				cannot attach any other information if 
%				we are to consider the statement as it is expressed. 
%				Hence, it is clear that Alice's statement  
%				is not fallacious because of an unexpressed 
%				statement. 
		\end{itemize}  
    \item How many formal constraints are needed for identifying   
	    a class ({\it ad hominem} {\it etc.}) of 
	    informally characterised fallacies? 
%	    all the fallacies in a class ({\it ad hominem} {\it etc.})  
%	    within the informally classified fallacy repertoire? 
		\begin{itemize} 
			\item[A:] This question is not entirely well-conceived. 
				Refer to Section \ref{sec_introduction}. 
				 Informal concepts  
				 are innately ambiguous, and 
				the boundary of every class 
				of informally characterised fallacies 
				fluctuates without any rigid agreement. 
				Since it fluctuates,   
				no class of informally characterised 
				fallacies can be precisely known. 
				%It then follows that  
				No one consistent 
				 set 
				of formal constraints may identify
				that which cannot be known. %though 
%				this question seems to assume otherwise.  
				This applies to all 
				informal 
				concepts such as informal truth, 
				informal knowledge, informal belief 
				and informal acceptability. 
				Cf. \cite{Kant08,Tarski56}.   
				What we do in this paper is to 
				identify a fallacy with a failure 
				to satisfy formal constraint(s). 
				It is this that is the idea of formal 
				fallacy identification, 
				not the futile attempt to 
				assign a set of constraints to 
				an unknowable class of 
				informally characterised fallacies 
				as its equivalent. 
				%Formal investigation 
				%of truth, knowledge, belief, 
				%acceptability is not about 
				%finding formal constraints 
				%that cover informal `truth', 
				%`knowledge', `belief', or 
				%%`acceptability', but to identify 
				%the phenomena 
				%But I wish to ask the same question 
				%to you. {\it How many classes 
				%of informally characterised fallacies 
				%are needed for identifying 
				%fallacies? Is a fallacy a fallacy 
				%insofar as it is classified as 
				%one?} 
				%It appears that this question 
				%contains within it 
				%an assumption that cannot hold. 
				%is futile 
				%since the boundary itself 
				%is undefined. 
%				No formal logic endeavours 
%				to delimit the informal `truth', 
%				the informal `knowledge' 
%				or 
%				the informal `belief'. The attempt 
%				is bound to be unsuccessful.
			%	Cf. 
			%	\cite{Kant08,Tarski56}.  
				%This paper does not aim 
				%to model any class of informally
				%classified fallacies. 
		\end{itemize}  
%	\item Is it possible to eventually identify every  
%		or most of 
%		informally classified fallacies 
%		by adding formal constraints? 
%		\begin{itemize} 
%			\item[A:] Implausible. See the previous  
%			     response.  
%			Moreover, this question 
%				contains within it 
%				the assumption 
%			  that the set of every or most 
%				of informally 
%				classified fallacies 
%				is known, which 
%				cannot be the case. 
%				At least, 
%				in formal fallacy 
%				identification, 
%``the process of fallacy judgement is explainable in detail: 
%from how an argumentation subjected to fallacy judgement is being modelled; 
%to under which context(s) the judgement is being made; and, if it is judged a fallacy, what formal constraints 
%are unsatisfied. Moreover, formal analysis on the interactions 
%				among the constraints is possible.'' See section \ref{subsec_solution_high_level_research_problem}. 
%		\end{itemize} 
	\item Is it possible to identify 
		a fallacy that occurs within 
		a single statement? 
		\begin{itemize} 
			\item[A:] Yes. For example, 
				refer to Examples 
				\ref{ex_role_nsa}, 
				\ref{ex_role_pr},
				\ref{ex_role_kos} and 
				\ref{ex_role_ss}. 
		\end{itemize}  
	\item To identify specific statements causing 
		violation of a constraint, do we need to
	      keep removing statements off the typed argumentation graph 
	      and keep recomputing until they are identified? 
	      \begin{itemize} 
		      \item[A:] There is no need to remove any 
			      statement. 
			    Satisfaction of a constraint 
			    is mostly judged for statements and edges 
			    linking them. Unsatisfaction, when induced 
			    by them, identifies them 
			      at that instant.  
			    Refer to the definitions of the constraints. 
	      \end{itemize} 
      	\item Suppose we have a theme aspect argumentation model 
		$\taam$ and suppose we like to append more statements 
		in $G$ for capturing the dynamic development 
		of argumentation, 
		do we need to recompute satisfaction of constraints 
	      each time such addition occurs? 
	      \begin{itemize} 
		      \item[A:] A simple answer is that investigation into 
			      the properties 
			      of the dynamics is out of the scope 
			      of this paper.  
			      Let us, however, discuss 
			      this matter under the following 
			      assumption: the semantic granularity  
			      (see Example \ref{ex_illustration_semantic_granularity}) 
			      does not change before and after 
			      the statement is added.  
%			      The rationale behind 
%			      this assumption is as follows: 
%			      since up to the current 
%			      \pmb{Assumption 2}: 
%			      granularity (see Example \ref{ex_illustration_semantic_granularity}) and that ,  
			      Then, recomputing 
			      is not necessary 
			      for already computed parts of $G$ 
			      unless types change for 
			      the statements and the edges 
			      in the parts. Such changes to 
			      $\Pi$, however, 
			      are manipulations and are hence 
			      discouraged: 
			      discussion themes  
			      and utterers of statements 
			      should not be permitted to alter in the middle
			      of the argumentation for the themes. 
			      As is clear from the definitions 
			      of the constraints, 
			      except for \pmb{F} constraints,  
			      no constraints require 
			      any more than the semantics of 
			      the added statement and those 
			      statements it acts on. 
			      As for \pmb{F} constraints, 
			      if $\taam$ does not satisfy \pmb{F}, 
			      then the new theme aspect argumentation 
			      model also does not. 
			      If $\taam$ does satisfy \pmb{F}, 
			      it is possible that 
			      the new theme aspect argumentation
			      model does not. However, 
			      redundancy with respect 
			      to a theme (see Definition 
			      \ref{def_redundancy}) is judged  
			      only for a pair of statements. 
			      So, it suffices to check 
			      every required pair 
			      in which one of them is the added 
			      statement.  
	      \end{itemize} 
      \item Is it not the case that an argumentation 
	      that is fallacious 
	      can be corrected by prompting people 
	      to clarify their arguments? 
	      \begin{itemize} 
		      \item[A:] That may well be the case. 
			      However, this does not alter  
			      the fact that the argumentation 
			      was fallacious before the clarification. 
			      Also, refer to section 
			      \ref{subsec_logic_based_argumentation} 
			      for the impracticality associated with assuming 
			      every participant in society is honest 
			      and well-intentioned. Such clarifications 
			      can be exploited as a way of 
			      prevarication by self-interested 
			      and malicious participants. 
	      \end{itemize} 
%	\item Suppose some statement is added, 
%		will the satisfaction of 
%		the formal constraints have to 
%		be redone all over again?  
%		\begin{itemize} 
%			\item[A:] Negative for \pmb{Core} 
%			constraints, 
%			and also negative for 
%				most of the other 
%				formal constraints.  
%		\end{itemize}   
%	\item Do we need to recompute? 
%		\begin{itemize} 
%			\item[A:] No need to recopmute 
%				most of the cases. 
%				Just check 
%				for all the constraints. 
%				The time complexity 
%				is the same. 
%		\end{itemize} 
\end{enumerate}
%\noindent \textbf{Q1} In Remark 1, is not the Alice statement be itself fallacious, since 
%		she is not providing a reason for it? We could 
%		 provide ``I am an expert on this 
%		subject" as a justification of her claim and 
%		then Alice's statement may be an argument from authority. \\
%\noindent \textbf{A} 
%If an unstated information
%as is suggested in \textbf{Q1}, 
%can be freely added to a statement and the statement can be 
%called a fallacy on the basis that the extended statement appears fallacious, 
%every statement may be turned into a fallacy by adding unexpressed information 
%on top of it.  
		
\end{document}